\documentclass[11pt, twoside, openright]{book}
\usepackage{microtype}
\usepackage[b5paper,
            layout=b5paper,
            headheight=15pt,
            paperheight=24truecm,
            paperwidth=17truecm,
            bindingoffset=8mm,
            marginratio={4:6,5:7}
            ]{geometry} 

\usepackage[english,dutch]{babel}
\usepackage[labelfont={bf,footnotesize},
            textfont={footnotesize}]{caption}
\usepackage{ragged2e}

\usepackage{xcolor}
\usepackage{lettrine}

\usepackage[outline]{contour}
\contourlength{1pt}

\usepackage[]{natbib}

\usepackage[noindentafter]{titlesec}
\titleformat{\section}
{\large\scshape}
{\textbf\thesection}{1em}{\MakeLowercase}
\titlespacing{\section}
{0pt}{3.5ex plus .1ex minus .2ex}{1.5ex minus .1ex}

\titleformat{\subsection}
{\normalsize\itshape}
{\scshape\large\textbf\thesubsection}{1em}{}
\titlespacing{\subsection}
{0pt}{3.5ex plus .1ex minus .2ex}{1.5ex minus .1ex}

\titleformat{\paragraph}[runin]
{\normalfont\normalsize\scshape}{\theparagraph}{1em}{\MakeLowercase}

\RequirePackage[palatino]{quotchap}
\renewcommand*{\chapterheadstartvskip}{\vspace*{-3.0\baselineskip}}
\renewcommand*{\chapterheadendvskip}{\vspace{1.3\baselineskip}}

\makeatletter
\renewcommand{\chapnumfont}{%
 \usefont{\f@encoding}{\@defaultcnfont}{b}{n}\fontsize{60}{130}\selectfont%
 \color{chaptergrey}}
\let\size@chapter\LARGE
\renewcommand\chapter{%
  \if@openright\cleardoublepage\else\clearpage\fi
  \thispagestyle{empty}%
  \global\@topnum\z@
  \null\hfill\@printcites\par
  \@afterindentfalse
  \secdef\@chapter\@schapter
}

\renewcommand{\@makechapterhead}[1]{%
  \chapterheadstartvskip%
  {\size@chapter{\sectfont\centering
    {\chapnumfont
      \ifnum \c@secnumdepth >\m@ne%
      \if@mainmatter\thechapter%
      \fi\fi
      \par\nobreak }%
    {\scshape\centering\advance\leftmargin10em\interlinepenalty\@M \MakeLowercase{#1}\par}}
  \nobreak\chapterheadendvskip}}
\makeatother

\usepackage{adjustbox}
\usepackage{todonotes}

\usepackage{amssymb}
\usepackage{rotating}
\usepackage{bibentry}
\usepackage{comment}
\usepackage{arydshln}

\usepackage{graphics}
\usepackage{graphicx}
\usepackage{mathptmx}
\usepackage[tbtags]{amsmath}
\usepackage{latexsym}
\usepackage{quotchap}
\usepackage{booktabs}
\usepackage{colortbl}
\usepackage{multirow}
\usepackage{tabularx}

\usepackage{fontawesome}
\usepackage{standalone}
\usepackage[misc]{ifsym}
\usepackage{soul}
\usepackage{float}
\usepackage{multirow}
\usepackage{multibib}
\usepackage{graphicx}

\usepackage{ccicons}
\usepackage{pdfpages}

\usepackage{algorithm2e}

\usepackage{enumitem}

\usepackage{tikz}
\usepackage{pgfplots}
\pgfplotsset{compat=newest}
\usepgfplotslibrary{colorbrewer}
\usepgfplotslibrary[groupplots,dateplot]
\usetikzlibrary[patterns,shapes.arrows]
\unexpanded{\def\startgroupplot{\groupplot}}
\unexpanded{\def\stopgroupplot{\endgroupplot}}
\usetikzlibrary{arrows,%
                petri,%
                topaths}%
\usepackage{tkz-berge}

\usepackage{fancyhdr}
\usepackage{titling}

\usepackage{etoolbox}
\apptocmd{\sloppy}{\hbadness 10000\relax}{}{}

\usepackage[T1]{fontenc}

\usepackage[osf,sc]{mathpazo}
\usepackage[euler-digits]{eulervm} 

\usepackage[scaled=0.90]{helvet}

\usepackage[scaled=0.85]{beramono}

\newcommand{\xvar}[1]{\text{#1}}

\newcommand{\xvbox}[2]{\makebox[#1][l]{#2}}

\usepackage[section]{placeins}

\usepackage[marginal]{footmisc}

\usepackage{tocloft}
\usepackage{fmtcount}

\DeclareMathOperator*{\argmax}{arg\,max}

\definecolor{nc}{HTML}{ff6f69}
\definecolor{pc}{HTML}{88d8b0}
\definecolor{gr}{gray}{0.8}
\definecolor{bl}{HTML}{666666}

\definecolor{PaleYellow}{RGB}{255,230,204}
\definecolor{PaleRed}{RGB}{248,206,204}
\definecolor{PalePurple}{RGB}{225,213,231}
\definecolor{PaleGreen}{RGB}{213,232,212}
\definecolor{PaleBlue}{RGB}{218,232,252}

\definecolor{PartYellow}{RGB}{215,155,0}
\definecolor{PartRed}{RGB}{184,84,80}
\definecolor{PartPurple}{RGB}{150,115,166}
\definecolor{PartGreen}{RGB}{130,179,102}
\definecolor{PartBlue}{RGB}{108,142,191}

\definecolor{AquaBlue}{RGB}{176,227,230}
\definecolor{AquaMarine}{RGB}{14,128,136}
\definecolor{Lavender}{RGB}{86,81,126}

\definecolor{DrawPaleGreen}{cmyk}{0.08,0.00,0.09,0.09}
\definecolor{DrawPaleOrange}{cmyk}{0,0.14,0.31,0.02}
\definecolor{DrawOrange}{cmyk}{0,0.44,0.98,0.29}
\definecolor{DrawPaleRed}{cmyk}{0,0.13,0.15,0.02}
\definecolor{DrawRed}{cmyk}{0,0.63,0.71,0.32}
\definecolor{DrawPaleAqua}{cmyk}{0.23,0.01,0.00,0.10}
\definecolor{DrawAqua}{cmyk}{0.90,0.06,0.00,0.47}
\definecolor{DrawPaleBlue}{cmyk}{0.26,0.08,0.00,0.06}
\definecolor{DrawBlue}{cmyk}{0.90,0.27,0.00,0.38}
\definecolor{DrawPalePurple}{cmyk}{0.08,0.08,0.00,0.11}
\definecolor{DrawPurple}{cmyk}{0.32,0.36,0.00,0.51}
\definecolor{DrawPaleMarine}{cmyk}{0.12,0.05,0.00,0.17}
\definecolor{DrawMarine}{cmyk}{0.62,0.27,0.00,0.64}
\definecolor{DrawBaby}{cmyk}{0.43,0.26,0.00,0.25}
\definecolor{DrawPaleBaby}{cmyk}{0.13,0.08,0.00,0.01}
\definecolor{DrawPaleLavender}{cmyk}{0.03,0.08,0.00,0.09}
\definecolor{DrawLavender}{cmyk}{0.10,0.31,0.00,0.35}
\definecolor{DrawPaleGold}{cmyk}{0.00,0.05,0.20,0.00}
\definecolor{DrawGold}{cmyk}{0.00,0.15,0.60,0.16}

\newcommand\pale[1]{\cellcolor{DrawPaleOrange}{#1}}
\newcommand\purp[1]{\cellcolor{DrawPaleLavender}{#1}}

\newcommand{\hlfancy}[2]{\colorlet{hlcolor}{#1}\sethlcolor{hlcolor}\hl{#2}}

\SetAlFnt{\footnotesize}
\SetAlCapNameFnt{\footnotesize}
\SetAlCapFnt{\scriptsize}

\SetAlCapSty{xAlCapSty}

\SetCommentSty{xCommentSty}
\LinesNumbered               
\SetSideCommentRight        
\DontPrintSemicolon         
\RestyleAlgo{algoruled}     

\nobibliography*

\PassOptionsToPackage{hyphens}{url}\usepackage[
    colorlinks=true,
    linkcolor=gray,
    filecolor=gray,
    urlcolor=gray,
    citecolor=gray,
    backref=page]{hyperref}
\usepackage{cleveref}

\renewcommand*{\backref}[1]{}
\renewcommand*{\backrefalt}[4]{\scriptsize
    \ifcase #1 (Not cited)%
    \or        (p.~#2)%
    \else      (pp.~#2)%
    \fi}

\begin{document}
\pagenumbering{gobble}
\includepdf{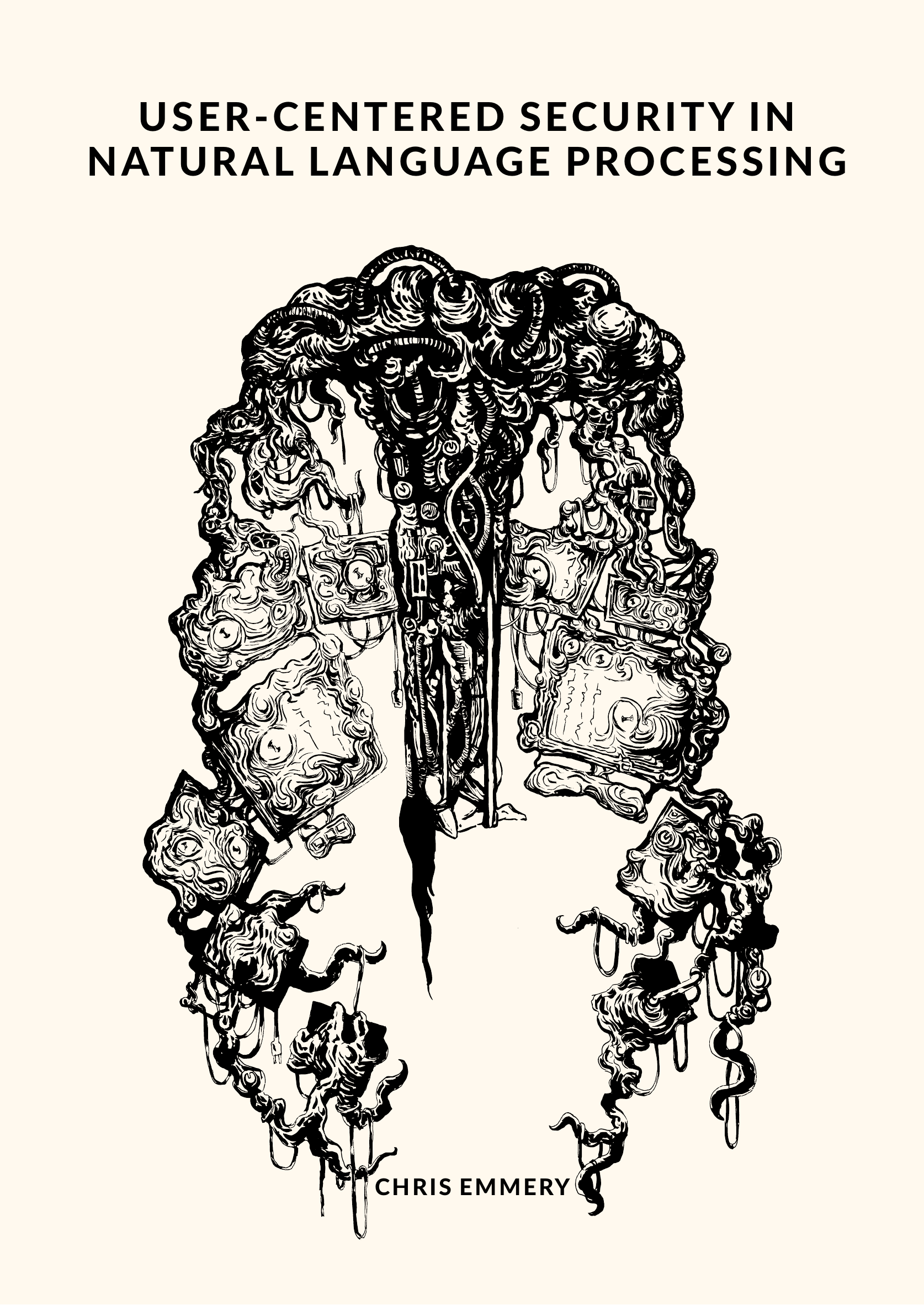}
\cleardoublepage
\selectlanguage{english}
\renewcommand{\chaptermark}[1]{\markboth{#1}{}}
\renewcommand{\headrulewidth}{0pt}
\raggedbottom
\frontmatter
\pagenumbering{gobble}
\def \fulltitle{User-Centered Security in Natural Language Processing}
\def \title{User-Centered Security in Natural Language Processing}
\def \subtitle{}
\def \author{Chris Emmery}

\def \coadvisorOne{Prof. Dr. Walter Daelemans}
\def \coadvisorTwo{Prof. Dr. Eric Postma}
\def \coadvisorThree{Dr. Grzegorz Chrupała}

\def \promotorOne{prof. dr. E.O. Postma (Tilburg University)}
\def \promotorTwo{prof. dr. ing. W. Daelemans (University of Antwerp)}
\def \promotorThree{dr. G.A. Chrupała (Tilburg University)}

\def \committeeOne{Dr. Albert Gatt}
\def \committeeTwo{Prof. Dr. Willem-Jan van den Heuvel}
\def \committeeThree{Prof. Dr. Marie-Francine (Sien) Moens}
\def \committeeFour{Prof. Dr. Malvina Nissim}
\def \committeeFive{Juniorprof. Dr. Martin Potthast}

\def \commissieOne{prof. dr. M.F. Moens (KU Leuven)}
\def \commissieTwo{prof. dr. W. van den Heuvel (JADS)}
\def \commissieThree{prof. dr. M. Nissim (Rijksuniversiteit Groningen)}
\def \commissieFour{dr. A. Gatt (Utrecht University)}
\def \commissieFive{dr. M. Potthast (Leipzig University)}

\def \degree{Doctor of Philosophy}
\def \field{Computational Linguistics}
\def \degreeyear{2023}
\def \degreemonth{January}
\def \department{Cognitive Science and Artificial Intelligence}

\def \university{Tilburg University}
\def \universitycity{Tilburg}
\def \universitystate{The Netherlands}

\thispagestyle{empty}
\begin{center}
{\huge \title \bigbreak}
    {\Large \subtitle \bigbreak}

\bigbreak

{ \large \subtitle }

\vfill

ter verkrijging van de graad van doctor aan Tilburg University op \par
gezag van de rector magnificus, Prof. Dr. W.B.H.J. van de Donk, \par
in het openbaar te verdedigen ten overstaan van een door \par
het college voor promoties aangewezen commissie \par
in de aula van de Universiteit op \par vrijdag 13 januari 2023 \par om 13:30 uur \par \bigbreak
door Christian David Emmery \par geboren te Eindhoven

\vfill

$\infty$
\end{center}
\clearpage
\thispagestyle{empty}

\noindent \textsc{promotores} \par
\noindent \promotorOne \par
\noindent \promotorTwo \par
\bigbreak
\noindent \textsc{co-promotores} \par
\noindent \promotorThree \par
\bigbreak
\noindent \textsc{leden promotiecommissie} \par
\noindent \commissieOne \par
\noindent \commissieTwo \par
\noindent \commissieThree \par 
\noindent \commissieFour \par
\noindent \commissieFive \par

\vspace*{\fill}
\justify{
\noindent \title \par
\noindent \subtitle \par
\noindent \author \par
\noindent PhD Thesis \par
\noindent \university, \degreeyear
\bigbreak
\noindent Cover: Ambika Kirkland (\href{https://instagram.com/sendrinon}{\texttt{@sendrinon}}) \par
\noindent Design: Chris Emmery (\href{https://mastodon.social/@cmry}{\texttt{@cmry}}) \par
\noindent Print: Ridderprint (\href{https://ridderprint.nl}{\texttt{ridderprint.nl}}) \par
\bigbreak
\noindent Chapter~\ref{ch:bul} was part of AMiCA (IWT SBO-project 120007), funded by the government agency for Innovation by Science and Technology.
\bigbreak
\noindent \ccbyncsa{} \emph{\title} by \author{} is licensed under a \href{http://creativecommons.org/licenses/by-nc-sa/4.0/}{Creative Commons Attribution-Non Commercial-ShareAlike 4.0 International License}.
}
\clearpage

\thispagestyle{empty}

\strut

\vfill

\begin{center}
\large{\textit{For Owen}}
\end{center}

\vfill

\clearpage

\thispagestyle{empty}
\strut
\newpage

\cleardoublepage
\clearpage
\tableofcontents

\mainmatter
\justify
{ \let\cleardoublepage }
{ \let\clearpage }
\pagestyle{fancy}
\lhead{}
\fancyhf{}
\fancyhead[LE,RO]{\footnotesize\scshape\thepage}
\fancyhead[RE]{\scshape\MakeLowercase{\leftmark}}
\fancyhead[LO]{\scshape\MakeLowercase{\chaptername~\thechapter}}
\setcounter{chapter}{-1}  

\begin{figure}
    \centering
    \includegraphics[width=0.75\textwidth,angle=-90]{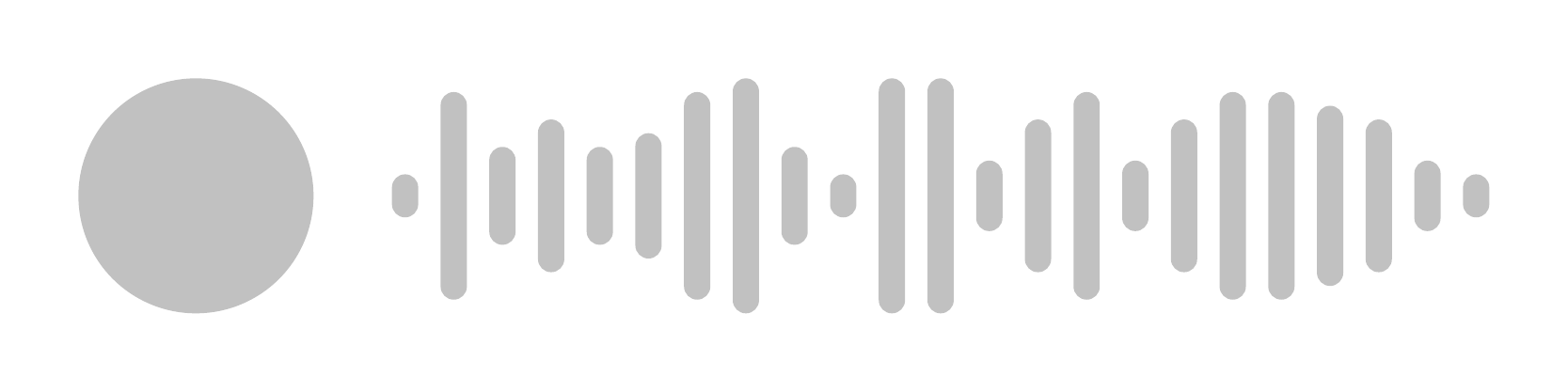}
\end{figure}

\thispagestyle{empty}
\strut
\newpage
\thispagestyle{empty}
\strut
\vfill

\noindent This chapter is based on the following work:

\begin{description}
    \item \bibentry{emmery-etal-2017-simple}
    \item \bibentry{emmery-2019-towards}
\end{description}   

\noindent Parts of \cite{emmery-etal-2017-simple} were adapted, or at the basis of the ideas presented in \Cref{sec:priv}. Part of \cite{emmery-2019-towards} is included verbatim in \Cref{sec:repro}. 

\newpage 
\chapter{Introduction}
\label{ch:introduction}

\lettrine[lines=4, lraise=0, nindent=0em, slope=0em]{T}{he internet} has become so ubiquitous that one could argue an introduction is quite the frivolous endeavor. After all, it plays an essential role in many human lives. Particularly younger generations (our `digital natives') center their personal and professional lives around it, and---according to long-standing popular belief---should therefore have become highly technologically proficient. It seems as though this rather optimistic view requires sober updating, however. In reality, technological proficiency remains varied across generations and is strongly correlated with sociodemographic factors \citep{https://doi.org/10.1111/j.1475-682X.2009.00317.x,doi:10.1177/0002764218787026,doi:10.1177/0963721420915872}. Additionally, its modern-day requirements have increased with system complexity: unprecedented ransomware attacks, spread of misinformation, and influence on elections via bots---they are all examples of sociotechnological security issues where human judgement of `the Internet' is the weakest link.

\pagestyle{fancy}

\parshape=0 Hence, it seems more meaningful to introduce the current state of the Internet, and what its users have made of it. The Internet as a free\footnote{As in ``free speech''.} platform, purposed solely for decentralized information sharing, has long passed \citep{DBLP:journals/nms/Morrison09}. The social web has put great emphasis on communication at a tremendous scale---particularly after the introduction of mobile devices. As content on the Internet has historically been predominantly free,\footnote{As in ``free beer''.} we pay with (quite literally) lives' worth of personal data; monetized to fuel the advertisement networks keeping it `free' and furthering its growth. This shift has given rise to the mechanism dubbed surveillance capitalism \citep{DBLP:journals/jitech/Zuboff15}: information pays for services, clicks generate revenue. We prove quick to relinquish control over our personal data in exchange for convenient (and effective) services \citep{DBLP:journals/isj/KehrKWF15}. This has created a rich-get-richer dynamic in which platforms that offer many (fast) data-driven services, that do not require technical expertise, attract most users. The Internet's hardware, software, and services are increasingly centralized under these parties \citep{evans2017dynamics,DBLP:conf/imc/MouraCHWH20}, who as a result may have become too big to regulate in handling our data \citep{DBLP:journals/ais/Arogyaswamy20}.

The current extensive dependence on this handful of large entities entails the line between what personal information we initially deemed private, and what we believed reasonably concerned institutions and corporations, has blurred---if not faded \citep{DBLP:journals/jitech/ConstantiouK15}. With it, the use of our data beyond the services we initially intended it for has inadvertently been obscured. If one requires a single illustratory example: Cambridge Analytica using a database of 87 million Facebook profiles worth of personal information for targeted voter manipulation (acquired through an intermediary app developer operating under the veil of scientific research) is certainly a fitting one \citep{DBLP:journals/computer/IsaakH18}.

As will become apparent throughout this dissertation, Artificial Intelligence (AI), or rather its subfield Machine Learning (ML --- i.e., predictive algorithms trained on data),  plays a significant role in obscuring how our data is repurposed \citep{Wachter2018ART,manheim2019artificial}. With the dominance of Deep Learning \citep[DL,][]{SCHMIDHUBER201585} currently still in full swing, and mobile consumer devices approaching desktop hardware computation levels \citep{DBLP:journals/cem/LemleyBC17,DBLP:conf/iccvw/IgnatovTKYWBWXG19}, complex tasks such as (among many others) indexing and categorizing photos by objects \citep{DBLP:journals/tsmc/HaralickSD73,DBLP:conf/iccv/Lowe99}, unlocking devices with facial recognition \citep{10.1162/jocn.1991.3.1.71}, tracking sleep \citep{10.1093/sleep/19.1.26}, activities \citep{DBLP:journals/vr/AbowdDOB98}, and controlling energy consumption \citep{DBLP:journals/tvlsi/SrivastavaCB96}, have made their way into most of the world's pockets. All the aforementioned applications deploy ML on highly sensitive data (i.e., they process related user data to train a model to perform the task). Those models often ship with (and are sometimes embedded in, and actively run on) the mobile devices themselves---scarcely to the owner's knowledge. Not incidentally, practically all major mobile technology players have set up AI labs and attracted prolific researchers to aid in-house development.

ML, having therefore become of great commercial interest, follows the same centralization trend as other information technology domains. The coinciding fast-paced, industry-driven research and development has significantly advanced the field; not only in terms of research output, but size of the experiments, impact and accessibility of the output, amount of subjects, and computational scale in general. This cultural shift has raised scientific and societal concerns, particularly whether research will (to the extent that it currently does) continue to align with public interest \citep{DBLP:journals/corr/abs-2102-01648}. As a number of those concerns lie at the core of the motivation and research questions of this dissertation, they will be discussed in the following sections: we will discuss how ML can be employed to compromise online privacy, how the framework of user-centered security is used to address this, and how, in turn, that framework can improve the accessibility of related research. 

\section{ML and Digital Privacy} \label{sec:priv}

Algorithms\footnote{Systems that use ML to deliver predictive output to a user, more popularly---though by definition being debatable terminology---referred to as algorithms.} have fundamentally changed our view into what it means to venture, and share on the Internet. Before, identifying information (such as IP addresses, browser fingerprints, cookies, and personal details) one would hand over through actions such as purchasing products online, were mostly seen by developers. Now, we receive almost instant feedback of such actions, e.g., via targeted ads. This (among others) has significantly increased public concern regarding digital privacy. Paradoxically, however, the average Internet user does little to protect their data \citep{BARTH20171038}; the utility and centralized character of the services requiring sensitive information make their users underestimate the risks of information disclosure \citep{DBLP:journals/isj/KehrKWF15}. Additionally, the lack of control users have over what data is shared induces a sense of cynicism and resignation regarding their privacy \citep{doi:10.1080/1369118X.2017.1293129,DBLP:journals/nms/LutzHR20}.

While, indeed, it can be argued\footnote{For middle to high-income countries in particular.} that one would have to be quite a pariah not to have a phone, bank card, or e-mail account, we do still have a choice regarding what we (directly) share, under what identifier, and which spaces we share in. For example, information security offers tools and frameworks through which one can protect their remaining bits of personal cyberspace; via end-to-end encryption. Interactions with people and services under this protocol are private, or at the very least controlled. Nevertheless, it seems that, while the public has a good understanding of the privacy threat models in (e.g.) chat clients, it lacks a technical understanding of encryption, and does not believe encryption to be a technical solution to protect its data \citep{DBLP:conf/eurosp/DechandND019}. Moreover, encryption mainly protects direct communication and local information, whereas social activities on the Internet partly involve \emph{voluntarily} disclosing particular information to a wider audience. Ideally, however, we should be able to do so without giving up our privacy regarding that which we chose not to share, or our anonymity while sharing.

The advent of `big'\footnote{Adequately-sized (for a given goal or task) is arguably a more fitting term.} data analysis and DL has had an adverse effect on our privacy in such scenarios. These models require troves of data to reach state-of-the-art results, often featuring people---not all equally (and thus fairly)\footnote{While not in the scope of this dissertation, fairness is a sizeable part of current societal and academic concerns. Further reading on this topic can be found in, e.g., \cite{bender-friedman-2018-data,DBLP:conf/fat/MitchellWZBVHSR19,DBLP:conf/fat/BenderGMS21}.} represented, and by far not as well protected as encryption does. For example, as the models store (some abstract representation of) information, they are vulnerable to extraction attacks \citep{DBLP:journals/ijsn/AtenieseMSVVF15,DBLP:conf/sp/ShokriSSS17,DBLP:conf/ccs/SongRS17}, ranging from membership inference (i.e., is an instance part of the original training data), to generation of sensitive content \citep{DBLP:journals/corr/abs-2012-07805,DBLP:conf/sp/PanZJY20}. This has prompted security and privacy researchers to investigate how ML models might correctly handle sensitive information \citep{DBLP:journals/corr/EdwardsS15,DBLP:conf/ccs/AbadiCGMMT016}, implemented under a similar framework as that of encryption. Protecting user information \emph{encoded} in ML models only covers a subset of the potential issues, however; more importantly, ML might be \emph{employed} to infer information beyond what is consciously shared.

\subsection{Latent User Information}

Physical activity using Internet-connected devices is a clear example of behavior facilitating invasive inference; location might reveal preferences and social connections \citep{DBLP:journals/pervasive/AnthonyHK07}, and sensors extensive physical and physiological information. Haptic sensors might even capture finger tap positions on a phone screen \citep{DBLP:journals/network/LiangCYHL18}. However, this is arguably less obvious for some information sources shared \emph{via} the Internet. With ML becoming increasingly accessible to non-experts, anything that is accessible to the public might have much more far-reaching implications than we can realistically predict, even as `experts'. For example, ML can infer demographic information through one's search terms \citep{murray1999inferring}, network of friends \citep[especially if those friends share personal information:][]{DBLP:conf/wsdm/MisloveVGD10,DBLP:conf/kdd/DongYTYC14}, purchases \citep{DBLP:conf/wsdm/WangGLXC16}, and media preferences \citep{DBLP:conf/mswim/SunLZ17}. Hence, simply by using the Internet, we leave traces of \emph{latent information} about ourselves.

Let us define the term latent information in this dissertation, which seems niche both to digital privacy research and ML. In the latter, it can be found using varying terminology; for example, \cite{DBLP:journals/tvt/HeCY18} refer to it as ``latent data'', whereas \cite{volkova-etal-2014-inferring} use the term ``latent attributes''. These can be reduced to the same phenomenon, and we would thereby define latent information as: \emph{information not part of the intended purpose of the data, but which can be inferred as a byproduct thereof}. To illustrate: an online purchase of diapers might (rather trivially) reveal you are a parent, as might (less trivially) a sudden increase in sweet vegetables (such as parsnip) among your groceries. Similarly, having only a few, closely connected friends on social media is shown to be a good predictor of old age \citep{DBLP:conf/kdd/DongYTYC14}, and watching Casablanca that you are female \citep{DBLP:conf/mswim/SunLZ17}.\footnote{Unfortunately, predictions still often rely on (harmfully) stereotypical cues in the training data \citep{koolen-van-cranenburgh-2017-stereotypes}. These data and models are hence not without bias regarding various demographic attributes \citep{DBLP:conf/fat/JoG20}.}

\subsection{Stylometric Profiling and Dual-Use}

Language is a prime example of data revealing latent information, and it is hence not incidentally the subject of this dissertation. The field of Natural Language Processing (NLP)---focusing on making human language machine-interpretable---offers a variety of techniques to infer things beyond the literal text. One's unique writing style might, apart from identity, also encode sensitive sociodemographic features such as gender and age \citep{DBLP:conf/aaaiss/SchlerKAP06,https://doi.org/10.1111/josl.12080}, personality \citep{plank-hovy-2015-personality}, education and income \citep{DBLP:conf/cikm/RaoYSG10,volkova-etal-2014-inferring}, region of origin \citep{bamman-etal-2014-distributed,tulkens-etal-2016-evaluating}, political or religious affiliation \citep{DBLP:conf/isi/KoppelAAB09,DBLP:conf/kdd/PennacchiottiP11}, and mental health issues \citep{DBLP:conf/icwsm/ChoudhuryGCH13,coppersmith2015}. These attributes can potentially, and often accurately, be inferred through stylometric analysis of publicly shared writing. Prediction of such author attributes, related to the inference of author identities \citep[as a binary task for one author, or multi-class prediction for many, see e.g., ][]{stamatatos-etal-1999-automatic}, is referred to as author profiling (see also Chapters~\ref{chap:textobf}~\&~\ref{chap:advsty}).

While these efforts have been greatly beneficial to various research fields such as computational sociolinguistics \citep{DBLP:conf/cicling/Daelemans13},  detecting fraud, deception, and identity theft \citep{badaskar-etal-2008-identifying,ott-etal-2011-finding,banerjee-etal-2014-keystroke,fornaciari-poesio-2014-identifying}, they potentially expose Internet users to adversaries inferring these attributes---which can be abused unbeknownst to the user. This is particularly harmful to individuals in a vulnerable position regarding race, political affiliation, mental health, or any other personal information made explicitly unavailable. It is thereby a prototypical case of function creep \citep[specific to information technology:][]{DBLP:journals/ieeesp/Schneier10}, or dual-use research \citep[as more broadly addressed by, e.g., health and natural sciences:][]{Ehni2008,10.1093/scipol/sct038}; i.e., technology which, while intentionally beneficial, has unintentional harmful applications or consequences. Typically, researchers reflect on such ethical issues---although in ML \citep{wallach-2014-big}, and NLP specifically \citep{hovy-spruit-2016-social}, this has until recently (see e.g., ethics review processes implemented since the 2021 *ACL conferences) rarely been the case. \cite{10.1093/scipol/sct038} recommend that regulation of such research should be a joint effort between all involved parties (e.g., science, governance), to which we would add that agency over technology is not restricted to the parties developing it, which in turn significantly complicates such efforts.

The application of these profiling methods has long been reserved for scientific purposes; collecting high quality labels for data is a costly process (both in time and resources), for which annotators typically have to be recruited and trained. Moreover, collecting the data itself, and fine-tuning models, would require some expertise and infrastructure. In \cite{emmery-etal-2017-simple}, however, we showed that simply querying for self-reported gender on social media generates enough (distantly supervised, noisy) data for on-par performance with models trained on `clean' annotated corpora.\footnote{\url{https://github.com/cmry/simple-queries}} Most importantly, this work ran within 24 hours, on a simple 4-core laptop, with out-of-the-box models; no tuning required. The study was later repeated,\footnote{\url{https://cmry.github.io/toku}} yielding similar results on a new sample of data from a different period, and for multiple attributes, in line with the results from prior work by \cite{beller-etal-2014-im}. This implies that anyone can potentially set up such profilers, making regulation significantly more challenging. Hence, providing Internet users with tools to mitigate such harmful inferences is an important contribution to their online security.

This introduces the first problem the current work addresses: NLP allows for privacy-invasive profiling inferences, the mechanism of which is not intuitive to humans. Contrary to many other areas of NLP, the `black box' nature of current DL techniques \citep{DBLP:journals/tnn/BenitezCR97} is not the main underlying issue. Stylometric classifiers may use high-level features \citep[e.g., punctuation use, sentence and word length, syllables:][]{10.2307/2282956,10.2307/2344178,10.3138/j.ctt15jjcf0,DBLP:conf/cikm/Otterbacher10}, particular tokens \citep[words:][]{10.1093/llc/10.2.111} and symbols \citep[emoticons and emojis:][]{DBLP:conf/cikm/PeersmanDV11}, or syntactical features \citep{biber-1995-dimensions,10.1093/llc/11.3.121,feng-etal-2012-syntactic,DBLP:conf/tsd/Hollingsworth12}---all of which can potentially be associated with some personal attribute. Without detailed insight into, and access to, these models their predictions, there is no reasonable heuristic by which Internet users might defend themselves against such inferences. Hence, they require a security tool; specifically, one that is ML-driven.

\section{User-Centered Security in NLP} \label{sec:sec}

Security in NLP is typically framed in an information security, or forensic context, where the focus lies on systems and data; more specifically, an entity providing a service, wishing to keep its (user) data secure \citep{DBLP:conf/nspw/AtallahMRN00,raskin-etal-2002-nlp}. Examples (among others) of such work are finding sensitive information in large corpora  \citep{DBLP:conf/IEEEares/CanforaSEFV18,DBLP:journals/corr/abs-2208-13081}, detecting malware \citep{DBLP:journals/corr/abs-2107-03072}, and securing representations in models handling text data \citep{DBLP:journals/tifs/FengHLWC20,lyu-etal-2020-differentially}. A broader categorization of security research in NLP is recent \citep{DBLP:conf/wsdm/FeyisetanGT20}; thus far, it does not cover harms directed at individuals. This makes for a(n interestingly) gray area where corporate and consumer interests only partially overlap. While online toxicity \citep{DBLP:conf/sigir/GreevyS04}, misinformation \citep{vlachos-riedel-2014-fact}, impersonation \citep{shropshire-2018-natural}, and surveillance can prove tremendously harmful to the public, platforms often do not have a direct monetary incentive to protect individual users against such external threats.

This brings us to the second problem touched upon in this dissertation: security research in NLP is predominantly \emph{industry-centered}, rather than \emph{user-centered}. This implies that, while it may solve societally relevant problems, the resulting tools are often not provided to be used by the public, and, more importantly, nor is it methodologically approached from a typical user's point of view. Research might, for example, use computational methods that are inaccessible even to an average research lab, only conduct experiments for a specific domain, platform, or setting, or generally have a research aim that does not align with public interest (e.g., improving invasive techniques).
Implementing such a user-centered framing has been discussed in other domains of NLP research, e.g., dialogue \citep{rieser-lemon-2008-automatic,nothdurft-etal-2015-interplay} and question-answering systems \citep{kelly-etal-2006-user}, summarization \citep{passali-etal-2021-towards}, explainability \citep{schuff-etal-2020-f1}, the clinical domain \citep{10.1136/amiajnl-2011-000465}, evaluation tools for NLP \citep{madnani-loukina-2020-user}, and design of ML libraries \citep{DBLP:conf/vl/CaiG19}.

Note that there is a strong reciprocity between user-centered security and privacy; it requires security tools to be deployed in such a way that they can be run on a local machine, thereby decentralizing data processing, and subsequent models (unlike related security frameworks such as Distributed Inference, or Federated Learning). NLP still relies on many public resources that potentially reveal sensitive author information \citep[for an overview, see][]{leidner-plachouras-2017-ethical}. Therefore, considering the increasing popularity of direct, or community-based communication platforms \citep[e.g., WhatsApp, Telegram, Discord]{10.1093/jcmc/zmy022} developing user-centered research practices could benefit the field as a whole. The currently open access to the richest data sources used to train state-of-the-art models (such as Reddit) is not guaranteed to last if their users' (indirect) contributions provide little in return, or might prove harmful. Towards such user-centered practices, this dissertation addresses two security applications: that of cyberbullying detection, and adversarial attacks.

\subsection{Cyberbullying Detection}

Cyberbullying detection---which we briefly discuss here, as Chapter~\ref{ch:bul} goes into extensive detail on the task and related work---is a subcategory of online harm prevention (and related to toxicity detection); its public interest concerns developing software that is able to spot incidents of online bullying. Practical applications hereof\footnote{Such as the Automatic Monitoring for Cyberspace Applications (AMiCA) project: \url{https://amicaproject.be/}} aim to employ such classifications for monitoring; i.e., sending reports to a moderator, or a trusted person \citep{10.1371/journal.pone.0203794}. To the best of our knowledge, the task has not been framed as a decentralized issue; i.e., to block contacts, or content on the receiver's end (closer to spam filtering). To that end, the research should ideally satisfy several criteria to be user-centered: for example, as a given user might use multiple platforms of communication, tailoring to a single one severely decreases detection potential; this implies existing classifiers should ideally generalize across many domains. Furthermore, for a single user, these classifiers cannot leverage the full network structure of social media; they are restricted to conversations, or a social media profile, as the scope of data. Hence, the collection and maintenance of useful training data is a main requirement, more so as the language of toxicity evolves rapidly (see Chapter~\ref{ch:aug}). It is therefore worth investigating to what extent it affects current classifiers, and if so, whether smaller data sources can be augmented to mitigate this.

\subsection{Adversarial Attacks}

Finally, we investigate adversarial attacks; i.e., changing a given input (generating so-called adversarial samples) such that a given ML model does not function as intended. State-of-the-art (neural) models prove highly vulnerable to such controlled perturbation attacks \citep{DBLP:journals/corr/SzegedyZSBEGF13,DBLP:journals/corr/GoodfellowSS14,DBLP:conf/iclr/KurakinGB17a,DBLP:conf/iclr/KurakinGB17,DBLP:conf/eurosp/PapernotMJFCS16,DBLP:conf/iclr/MadryMSTV18}. Often, they can be effective with minimal changes that, particularly in Computer Vision, do not significantly change the perceived output. For example, \citet{DBLP:conf/icse/TianPJR18} slightly change lighting conditions in the input to automated-driving systems, and show that these changes go undetected while influencing the system in its behavior. However, this relative ease of launching inconspicuous attacks does not apply for discrete inputs in domains such as language; grammaticality and the intended semantics need to be preserved for its techniques to be usable and, more importantly, remain undetected. Approaches on generating textual adversaries \citep{DBLP:journals/corr/LiMJ16a,alvarez-melis-jaakkola-2017-causal} therefore typically perform manual, or heuristic-based deletions and replacements on word level. \citet{DBLP:conf/milcom/PapernotMSH16}, for example, replace words with the nearest neighbor of their embedded representation. While conservative, they also require significant effort on the user side \cite[e.g., manual correction:][]{jia-liang-2017-adversarial}; fully automated (i.e., end-to-end) adversarial systems are therefore an increasingly popular line of work (Chapters~\ref{chap:textobf}~\&~\ref{chap:advsty} go into more detail and related work).

In this dissertation, such attacks are a common thread between our user-centered security problems; they are highly relevant for privacy-preserving obfuscation methods \citep{10.1007/978-3-030-15719-7_39} against author profiling \citep[i.e., adversarial stylometry: ][]{DBLP:journals/tissec/BrennanAG12}, and adversarial samples might also prove useful to assess the influence of lexical variation and augmentation on cyberbullying detection. For this research to fit the proposed user-centered framework, it again has to meet several criteria: whereas most of the adversarial attacks assume full, or at least restricted access to a target model on which they fit adversarial samples (i.e., white- and black-box attacks, respectively), a user employing them has no knowledge of the target model whatsoever. Furthermore, we cannot assume a user to be competent enough to tweak the attacks, nor to have the infrastructure to run them for a long time; hence, they have to be provided out of the box and run in a reasonable time. A methodological trade-off to be investigated here, is therefore that of targeted attacks versus end-to-end attacks; targeted attacks allow for more control, are more conservative, and thereby might be more faithful to the original input. More importantly, they often require less technical effort on the user-side. However, their limited changes might prove much less effective than end-to-end rewrites of entire sentences. Conversely, these systems might prove not to satisfy our usability criterion. Finally, the user-centered framework requires all (usable) research to be accessible to the public.

\section{Reproducibility \& Accessibility} \label{sec:repro}

With the (arguably unprecedented) speed at which the field currently produces research, recent attention has turned toward the lack of rigor in, and robustness of, the models achieving these results, and how this affects the broader ML community  \citep{doi:10.1126/science.359.6377.725}. This is perhaps surprising, given that one major driving force for the growth and interest has been the proliferation of powerful open-source libraries and pre-trained models that lower the bar for entry to the field.
Furthermore, many of the properties of ML research are conducive to open, reproducible research: large standardized publicly available datasets, generally less noisy data collection, community-driven shared tasks and benchmarks, the ability to share complete research code, and a tendency towards open-access publication \citep{Munafo2017}.

Despite these inherent advantages, (applied) ML research is still considered predominantly irreproducible \citep{Gundersen_Kjensmo_2018}; i.e., given \emph{different data} and an \emph{alternative implementation}, it would not produce the same results, nor support the inferences and conclusions of the original work. Efforts to fully reproduce the results (not just reuse published models) of seminal papers have been met with a multiplicity of issues, including unclear experimental descriptions \citep{pineau2018iclr,DBLP:conf/icml/TianMGSCPZ19}, failures to generalize to out-of-domain datasets \cite[e.g.][]{DBLP:conf/icml/RechtRSS19,DBLP:conf/iclr/ZhangBHMS20}, and model evaluations that do not replicate \citep{DBLP:conf/iclr/MelisDB18}. Significant blame has been allocated to the ecosystem of research in ML for exacerbating replication issues \citep{DBLP:journals/corr/abs-1812-01074,doi:10.1126/science.359.6377.725,DBLP:journals/cacm/LiptonS19,DBLP:conf/fat/MitchellWZBVHSR19,DBLP:journals/corr/abs-1901-06246,sculley2018winner,DBLP:journals/cacm/GebruMVVWDC21}.

It follows by definition that user-centered research should aim\footnote{We frame these as aims, as full reproducibility is a highly involved task, often with many more criteria than accessibility \citep{emmery-2019-towards}.} to be made available to the community; i.e., be reproducible, but most importantly accessible: well-documented, providing examples of how the code (or software) might be used for the intended purposes, and extended if possible. For \cite{emmery-etal-2017-simple} we provided an initial template for exactly those desiderata, which we will follow throughout the presented research.

\section{Research Questions}

As discussed thus far, this thesis juxtaposes the field of NLP with its frameworks of security, and privacy, and ties those in with research on the broader challenge of user-centered ML. Accordingly, this dissertation focuses on two security domains within NLP with great public interest: that of author profiling, and cyberbullying detection, and aims to answer the following research questions:

\begin{itemize}
    \item[\textbf{RQ1}] To what extent can end-to-end and targeted attacks be employed for user-centered adversarial stylometry?
    \item[\textbf{RQ2}] Do cyberbullying classifiers prove robust enough to be deployed in a user-centered fashion?
\end{itemize}

In doing so, we provide the following (at publication time of the chapters) unique contributions: we (i) investigate a user-centered framing of security research in NLP---specifically for the tasks of adversarial stylometry, and cyberbullying detection. In doing so, we (ii) propose a style-neutral framing for adversarial stylometry (Chapter~\ref{chap:textobf}), (iii) explicitly measure transferability of adversarial stylometry (Chapter~\ref{chap:advsty}), (iv) assess generalization of cyberbullying classifiers (Chapter~\ref{ch:bul}), (v) their sensitivity to lexical variation, and (vi) the potential to augment its corpora (Chapter~\ref{ch:aug}).

\thispagestyle{empty}
\strut
\newpage

\begin{figure}
    \centering
    \includegraphics[width=0.75\textwidth,angle=-90]{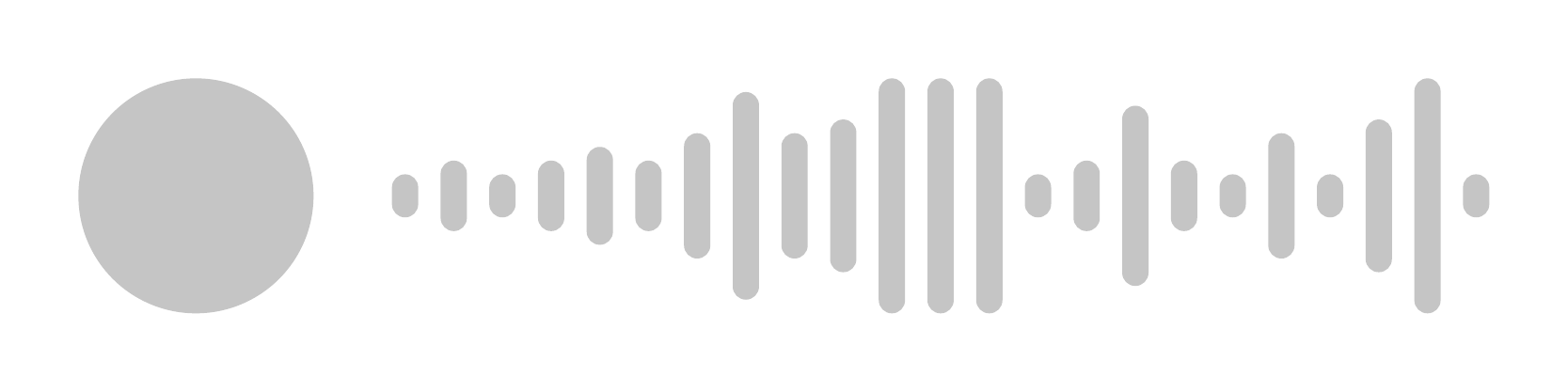}
\end{figure}

\thispagestyle{empty}
\strut
\newpage

\thispagestyle{empty}
\strut
\vfill

\noindent This chapter has been published as:

\begin{description}
    \item \bibentry{emmery-etal-2018-style}
\end{description}

\noindent Minor revisions have been made to the title, text, and figures. Figure~\ref{fig:grl}, and Tables~\ref{tab:ex1} \& \ref{tab:ex2} have been significantly improved.

\chapter{Textual Style Obfuscation by Invariance} \label{chap:textobf}

\lettrine[lines=4, lraise=0, nindent=0em, slope=0em]{T}{he task} of obfuscating writing style using sequence models has previously been investigated under the framework of obfuscation-by-transfer, where the input text is explicitly rewritten in another style. An adverse effect of this framework are the frequent major alterations to the semantic content of the input. In this chapter, we propose obfuscation-by-invariance, and investigate to what extent models trained to be explicitly style-invariant preserve semantics. We evaluate our architectures in parallel and non-parallel settings, and compare automatic and human evaluations on the obfuscated sentences. Our experiments show that the performance of a style classifier can be reduced to chance level, while the output is evaluated to be of equal quality to models applying style-transfer. Additionally, human evaluation indicates a trade-off between the level of obfuscation and the observed quality of the output in terms of meaning preservation and grammaticality.

\parshape=0
\section{Introduction}  \label{sec:int}

The fact that writing style uniquely characterizes a person, and can be leveraged for automatic author identification \citep{10.1093/llc/13.3.111,stamatatos-etal-2000-automatic}, has been well-studied in the field of (computational) stylometry \citep{DBLP:journals/csur/NealSFYXW17}. Similarly, work on author profiling \citep{DBLP:journals/lalc/KoppelAS02} has demonstrated that such stylometric features can be used to accurately infer an extensive set of personal information, such as age, gender, education, socio-economic status, and mental health issues \citep{eisenstein-etal-2011-discovering,DBLP:conf/icmla/AlowibdiBY13,volkova-etal-2014-inferring,plank-hovy-2015-personality,volkova-bachrach-2016-inferring}. Traditionally, these techniques relied on expensive human-labelled examples; however, more recent work has demonstrated near equal accuracy when only relying on self-reports as a distant supervision signal \citep{beller-etal-2014-im,emmery-etal-2017-simple,yates-etal-2017-depression}. While these efforts have been greatly beneficial to various research fields such as computational sociolinguistics \citep{DBLP:conf/cicling/Daelemans13}, they potentially expose users of such media to attacks where this information can be abused unbeknownst to them. This is particularly harmful to those in a vulnerable position regarding race, political affiliation, mental health, or any other personal information made explicitly unavailable by these individuals.

Adversarial stylometry, or style obfuscation, is one of the proposed methods  aimed at protecting users against such attacks. Its objective is to rewrite an input text such that a classifier (the adversary) trained on detecting a particular variable (such as a demographic attribute) is fooled---effectively protecting this variable. The main challenge is to preserve the original meaning of the input, whilst hiding the act of obfuscation \citep{DBLP:conf/clef/PotthastHS16}. Recent work on automatic obfuscation \citep{DBLP:conf/uss/ShettySF18,10.1007/978-3-319-65813-1_18} shows promising results in minimizing performance of the adversary; however, these models (as noted by the authors) struggle with maintaining the semantic content of the input. To illustrate, while rewriting \emph{school} to \emph{wedding}\footnote{Example taken from \cite{DBLP:conf/uss/ShettySF18}.} effectively fools an age classifier into thinking the text is written by an adult rather than a teen, the original meaning is not preserved in the output.

We propose that this observed shift in meaning is to some extent a by-product of the formulation of the obfuscation task. Content words that are strongly related to a particular attribute often play a significant role in the accuracy of a potential adversary. There is ample evidence for this phenomenon in age and gender classification work \citep[inter alia]{DBLP:journals/lalc/KoppelAS02,DBLP:conf/cikm/RaoYSG10,burger-etal-2011-discriminating,sap-etal-2014-developing,DBLP:conf/clef/PardoRVDPS16}. Taking examples from \cite{sap-etal-2014-developing} specifically, features with strong coefficient weights for gender include e.g., \emph{boxers, shaved, girlfriend, beard, fightin} for males, and \emph{purse, blueberry, pedicure, hubby, earrings} for females. It is therefore not a surprising result that models explicitly tasked to \emph{minimize} the performance of such classifiers perform what we will refer to as obfuscation-by-transfer. To illustrate, the adversary is easily fooled when a sentence looks strongly female even though it was written by a male. As such, the most efficient route to obfuscation from this perspective is a form of style-transfer: swapping a few strongly target-associated content words for their contrastive variant (\emph{wife} to \emph{husband}, \emph{school} to \emph{wedding}). When such variants are also close in semantic spaces that sequence models make use of, any reconstruction metrics---such as \textsc{bleu} \citep{papineni-etal-2002-bleu}, \textsc{meteor} \citep{banerjee-lavie-2005-meteor}, embedding distances, etc.---might prove to be inaccurate indicators of the true shift in meaning associated with these changes. 

\paragraph{Our contributions} In the current chapter, we propose a different approach to automatic obfuscation that we hypothesize partly overcomes the limitations to preserving meaning of the input: obfuscation-by-invariance. Here, the objective shifts towards maximizing the adversary's \emph{uncertainty}, implying its accuracy on the protected variable should be as close to chance level as possible. Fixing the adversary's performance around chance involves making the input text devoid of stylistic features that strongly correlate with any of the protected variables, thus producing language that is neutral with respect to these style differences. We test our hypothesis in several experimental conditions.\footnote{All code and data to fully replicate the experiments is available at \url{github.com/cmry/style-obfuscation}.} The main component in our encoder-decoder architecture to achieve a style-invariant encoding of the input is a Gradient Reversal Layer \citep[GRL, ][]{DBLP:conf/icml/GaninL15} inserted between the input sentence embedding and the style classifier. We investigate the effect of this module in isolation, as well as in a style-invariant soft transfer setting, by using a conditioned decoder. First, to gauge if this architecture can successfully decode the style-invariant encoding---and what the effect on an adversary's performance would be---we train sequence-to-sequence models on a parallel corpus of English Bible styles. Secondly, given that in a realistic obfuscation setting there is no access to such parallel sources, we drop the target pairs to create an autoencoder setting. In our experiments, we demonstrate a trade-off around chance-level performance: obfuscation-by-transfer in a parallel setting works well using a many-to-many translation model, but scores worse in the human evaluation than our style-invariant model. As such, we pose that there is potential in a style-invariant approach to obfuscation, and that it warrants further investigation.

\section{Related Work}

\subsection{Adversarial Stylometry}

The idea that computational stylometry might be used to compromise anonymity was first explored by \cite{DBLP:conf/uss/RaoR00}. They saw potential in concealing style information using machine translation (MT), but noted that it was not powerful enough at the time. \cite{kacmarcik-gamon-2006-obfuscating} continued the proposed line of work by informing users regarding characteristic features and deeper linguistic cues in their writing style. Recent related studies can be found in \citep{DBLP:conf/wistp/ThiSG15,caliskan2015coding}. \cite{DBLP:journals/tissec/BrennanAG12} explicitly frame obfuscation as an adversarial task and use MT (round-trip translation), similar to \citep{DBLP:conf/semco/CaliskanG12}. Rule-based perturbations \citep{DBLP:conf/ifip11-9/JuolaV11} and mixtures of both \citep{10.1007/978-3-319-65813-1_18} have also been applied for fully automatic obfuscation. Closest to our approach is recent work by \cite{DBLP:conf/uss/ShettySF18}, who pursue the task of learning obfuscation-by-transfer using a Generative Adversarial Network architecture. In their setup, a generator is trained to produce sentences that maximize the probability assigned by a discriminator that is, in turn, trained to distinguish real from generated sentences. While they incorporate different semantic losses, and demonstrate successful obfuscation on age and gender annotated microblog data and political speeches, their output suffers from the lack of semantic preservation we described before. Finally, a style-transfer approach with potential application to obfuscation is work by \cite{DBLP:journals/corr/abs-1711-04731}, who investigate textual zero-shot style-transfer using a sequence-to-sequence MT model inspired by zero-shot translation \citep{DBLP:journals/tacl/JohnsonSLKWCTVW17}. Their work demonstrates successful translation between different versions of the English Bible.

\subsection{Gradient Reversal}

The use of a GRL for learning domain-invariant feature representations was proposed by \cite{DBLP:conf/icml/GaninL15}, who demonstrated its application to lighting conditions in computer vision data. Since then, it has been applied to several language tasks: e.g., textual feature extraction \citep{DBLP:conf/sigir/PryzantCJ17}, POS tagging \citep{kim-etal-2017-cross,gui-etal-2017-part}, image captioning \citep{chen2017show}, and document classification \citep{liu-etal-2017-adversarial,xu-yang-2017-cross}.  Most importantly, \cite{DBLP:conf/nips/XieDDHN17} demonstrate a GRL can be used to implement an adversarial setting, and to improve performance for a number of language tasks, including generation. These results bode well for its application to obfuscation-by-invariance.

\section{Models}

\begin{figure}
\centering
\includegraphics[width=0.98\columnwidth]{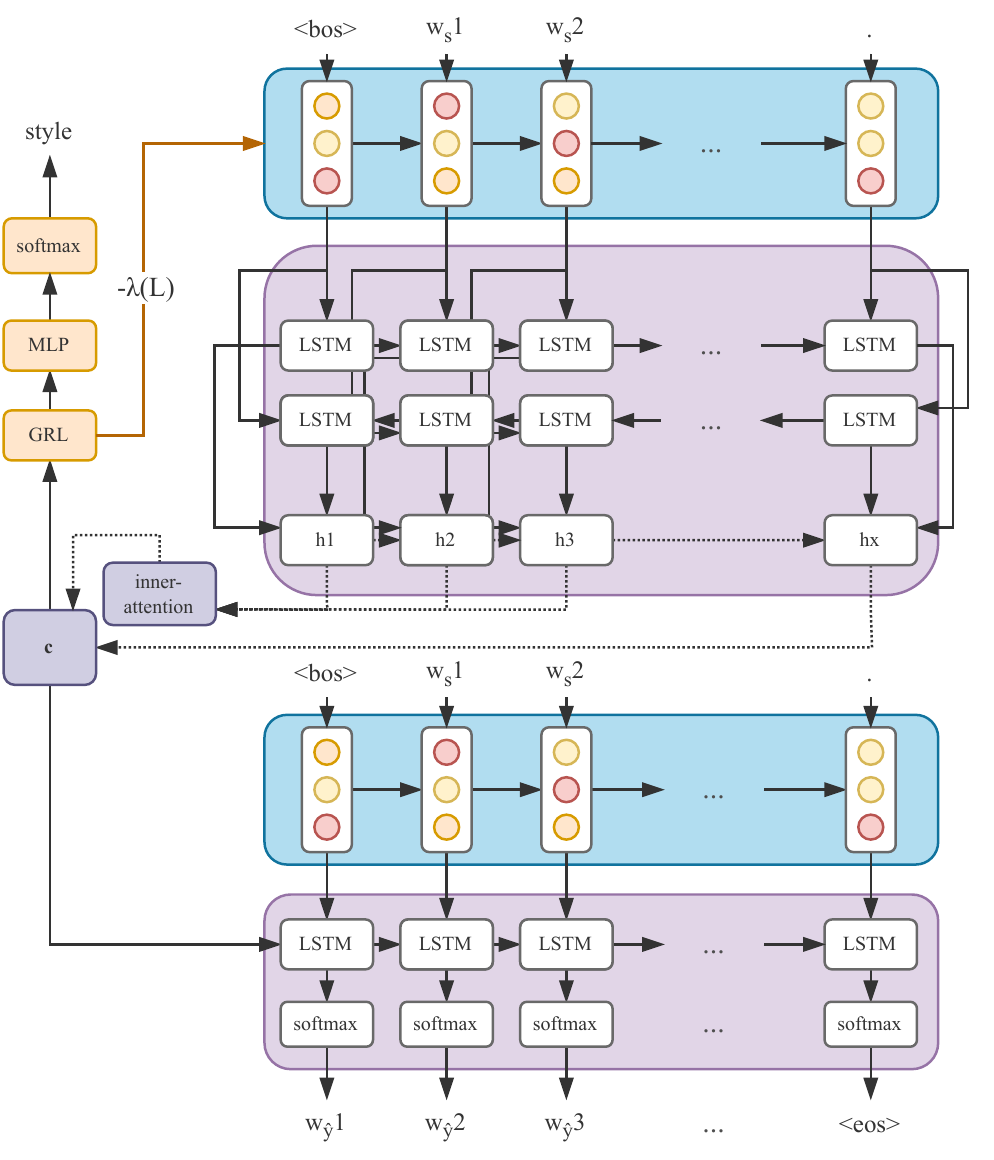}
\caption{Model architecture with a Gradient Reversal Layer (GRL) component: on the top the encoder part, to the left the GRL taking the input sentence embedding (or, the context vector) $\mathbf{c}$ as input, and on the bottom the decoder part of the architecture. Note that because the GRL inverses the loss ($-\lambda(L)$), the encoder both minimizes the style classification performance of the MLP, and has to produce a useful representation $\mathbf{c}$ for the decoder.}
\label{fig:grl}
\end{figure}

Our base architecture is a neural encoder-decoder \citep{DBLP:conf/nips/SutskeverVL14} model similar to that of \cite{DBLP:journals/corr/WuSCLNMKCGMKSJL16}, implemented in PyTorch \citep{DBLP:conf/nips/PaszkeGMLBCKLGA19}. Given an input sequence of one-hot encoded words, the encoder first embeds the words into dense vectors, which are then processed by one or more bidirectional \citep{DBLP:journals/tsp/SchusterP97} layers with LSTM cells \citep{DBLP:journals/neco/HochreiterS97}. The resulting sequence of processed vectors is then merged into a single representation using an inner-attention mechanism that will be described below. A neural language model is then trained to decode the output sequence conditioned on the encoded sentence embedding (a so-called context vector). Training is done by minimizing the locally-normalized per-word cross-entropy of the target sequence.

In an autoencoder setting, the goal of the network is simply to reconstruct the original sentence based on the encoded context vector \citep{DBLP:conf/nips/PLLKRRS14,DBLP:conf/acl/LiLJ15}. This set-up can be combined with the GRL to encourage the encoder to produce attribute-invariant context vectors. The target is the input itself in the case of an autoencoder (AE) architecture, or a paired sentence in the case of a sequence-to-sequence (S2S) architecture. See Figure~\ref{fig:grl} for a visual representation of the base architecture.

\subsection{Architecture Components}

\paragraph{Gradient Reversal Layer ({\textnormal{\textsc{grl}}})} The GRL \citep{DBLP:conf/icml/GaninL15} is applied on top of the context vector. During the forward pass, the GRL computes the identity function and feeds its input to a shallow Multi-Layer Perceptron (MLP) style classifier. However, during back-propagation, the gradient of the style classifier is flipped in sign. For the current work, its purpose is to optimize the encoder parameters such that they are style-invariant; i.e., encoding as little stylistic information of the input text as possible.

\paragraph{Conditional Decoder ({\textnormal{\textsc{c}}})} Previous research on neural language modeling has demonstrated the effectiveness of conditioning a models on sentential and contextual variables (such as tense, modality, or voice). In our experiments, we evaluate a conditional autoencoder in which the decoder is conditioned on the input style label. We implement conditioning following \cite{ficler2017controlling}; i.e., concatenating the corresponding attribute embedding vector (in our case, the corresponding style embedding) to each of the word embeddings input to the decoder. Each style is therefore associated with an embedding vector $\mathbf{c}$, which is fed into the architecture at each step. In contrast to the simple autoencoder, the conditional autoencoder can switch to a desired style at test time. We expect that, by conditioning the decoder on the original style label, the output retains its linguistic features, and consistency, without fully recovering the targeted style.

\paragraph{Token Transfer ({\textnormal{\textsc{tt}}})} 

We argue that an MT system relying on a parallel corpus of styles (be it attributes or authors) performs obfuscation-by-transfer, and---as translation is largely a meaning-preserving operation \citep{ide-etal-2002-sense,dyvik-2004-translations}---is  likely to have high semantic consistency. However, such parallel corpora are generally not available and have tremendous associated curation costs; it would require large amounts of identical (parallel) information (ideally on sentence-level), e.g., written by teens and adults. Textual style transfer by MT is therefore not a plausible use-case for obfuscation. However, it does provide a good indication of the performance of an obfuscation model under the framework of obfuscation-by-transfer. For this, we apply a sequence-to-sequence translation model trained on style, as discussed by \cite{DBLP:journals/corr/abs-1711-04731}. Following  \cite{DBLP:journals/tacl/JohnsonSLKWCTVW17}, we use a target style token, allowing for a model trained on many-to-many translation to rewrite an input sentence in a different style, simply by prepending a target token. This is the only configuration that uses an attention mechanism, as it is not tested in combination with the GRL.

\subsection{Architecture Details}

Our models use $300$-dimensional embeddings, and both the encoder and decoder a single layer of $1000$ LSTM cells. The sequence of hidden states from the encoder are merged into a single representation using a feature-wise inner-attention mechanism.
Let $w_t$ and $h_t$ denote respectively the input word embedding and hidden state of the LSTM for step $t$ for a total of $n$ total steps. The $i^{th}$ feature of the sentence embedding $c$ is computed by a weighted sum over the $i^{th}$ feature of the hidden states:
\begin{equation}
c_i = \sum_{t=1}^{n} a_i^t h_i^t
\end{equation}
where each weight $a_i^t$ is computed by:
\begin{equation}\label{eq:att}
a_i^t = \frac{\exp([W^Tz^t]_i)}{\sum_{s=1}^n \exp([W^Tz^s]_i)}
\end{equation}
In Equation \ref{eq:att}, $z_t$ is the concatenation of $w_t$ and $h_t$, $W \in \mathbb{R}^{(D+H)\times H}$ is an additional projection matrix to be learned, and $D$ and $H$ denote the dimensionality of the word embedding matrix and the LSTM layer respectively. In comparison to traditional merging models (such as max or mean pooling), the additional parameters help the model to learn what input words and what features are more important for the task. This is similar to conventional attention over hidden activation vectors, with the main difference being that weighting is done feature-wise and all information flow from the encoder to the decoder is passed through a single output encoding vector bottleneck. Note that a traditional alignment-style attention mechanism is not compatible with the application of the GRL, since the latter must have scope over all information being passed from the encoder to the decoder.

The decoder is conditioned on the context vector by concatenating the latter to the input of the decoder at each step. This facilitates learning by increasing the gradient signal to the encoder. All model parameters are optimized with Adam \citep{DBLP:journals/corr/KingmaB14} in mini-batches of $50$ examples and a learning rate of $0.001$ which is decreased by $0.75$ after each epoch. To avoid overfitting, dropout \citep{DBLP:journals/jmlr/SrivastavaHKSS14} is applied to the input of each LSTM layer with a dropping probability of $0.25$, and we stop training when loss stops decreasing for $3$ epochs. During test time, we approximate the global best output sequence, applying beam search with a beam of size $5$. The GRL is implemented in combination with a single-layer MLP using the same dimensionality as the encoder.

\section{Experimental Set-up}

Our goal is investigating the effectiveness of obfuscation-by-\ invariance, and more specifically to what extent style-invariant representations preserve sentential semantics of the input. To this end, we perform three experiments for different corpora and obfuscation settings, using the components described above. In each setting, there is an adversary trained to detect the variable to be obfuscated.

\subsection{Data}
\begin{figure}
  \begin{minipage}{.47\textwidth}
  \centering
  \begin{tikzpicture}
  \begin{axis}[enlargelimits=false,
  			   colorbar,
               tick pos=left,
               tick align=outside,
               major tick style = {black},
               point meta min=0,
               colormap={custom}{color(0)=(white) color(1)=(DrawPurple)},
               xticklabels={,\textsc{asv},\textsc{bbe},\textsc{dby},\textsc{web},\textsc{ylt}},
               yticklabels={,\textsc{asv},\textsc{bbe},\textsc{dby},\textsc{web},\textsc{ylt}},
               yticklabel style = {rotate=90},
               x grid style={white!69.019607843137251!black},
			   y grid style={white!69.019607843137251!black},]
    \addplot [matrix plot,point meta=explicit]
      coordinates {
        (0,0) [1.00] (1,0) [0.37] (2,0) [0.66] (3,0) [0.68] (4,0) [0.61]

        (0,1) [0.37] (1,1) [1.00] (2,1) [0.32] (3,1) [0.37] (4,1) [0.34]

        (0,2) [0.66] (1,2) [0.32] (2,2) [1.00] (3,2) [0.58] (4,2) [0.59]
        
        (0,3) [0.68] (1,3) [0.37] (2,3) [0.58] (3,3) [1.00] (4,3) [0.52]

        (0,4) [0.61] (1,4) [0.34] (2,4) [0.59] (3,4) [0.52] (4,4) [1.00]
      };
      \node[color=white] at (axis cs:0,0) {$1$};
      \node[color=white] at (axis cs:1,1) {$1$};
      \node[color=white] at (axis cs:2,2) {$1$};
      \node[color=white] at (axis cs:3,3) {$1$};
      \node[color=white] at (axis cs:4,4) {$1$};
      
      \node[color=black] at (axis cs:0,1) {$0.37$};
      \node[color=black] at (axis cs:0,2) {$0.66$};
      \node[color=black] at (axis cs:0,3) {$0.68$};
      \node[color=black] at (axis cs:0,4) {$0.61$};
      \node[color=black] at (axis cs:1,2) {$0.32$};
      \node[color=black] at (axis cs:1,3) {$0.37$};
      \node[color=black] at (axis cs:1,4) {$0.34$};
      \node[color=black] at (axis cs:2,3) {$0.58$};
      \node[color=black] at (axis cs:2,4) {$0.59$};
      \node[color=black] at (axis cs:3,4) {$0.52$};
      
      \node[color=black] at (axis cs:1,0) {$0.37$};
      \node[color=black] at (axis cs:2,0) {$0.66$};
      \node[color=black] at (axis cs:3,0) {$0.68$};
      \node[color=black] at (axis cs:4,0) {$0.61$};
      \node[color=black] at (axis cs:2,1) {$0.32$};
      \node[color=black] at (axis cs:3,1) {$0.37$};
      \node[color=black] at (axis cs:4,1) {$0.34$};
      \node[color=black] at (axis cs:3,2) {$0.58$};
      \node[color=black] at (axis cs:4,2) {$0.59$};
      \node[color=black] at (axis cs:4,3) {$0.52$};
      \clip (current bounding box.south west) rectangle (current bounding box.north east);
  \end{axis}
  
\end{tikzpicture}
  \caption{Jaccard index between vocabularies of the different Bible styles.} 
  \label{fig:jac}
  \vspace{1.11cm}
  \end{minipage} \hfill
  \begin{minipage}{.50\textwidth}
  	\centering
    \footnotesize
    \begin{tabular}{r|rrrr}
    	\toprule
        \textsc{style}			& \textsc{types} & \textsc{tok} & \textsc{uniq} & \textsc{punc} \\
        \midrule
        \uppercase{asv}	& $12.5$K & $3.7$M   &  $602 $  & $0.13$ \\
        \uppercase{bbe}	& $ 6.0$K & $3.9$M   &  $569 $  & $0.11$ \\
       	\uppercase{dby}	& $13.8$K & $3.7$M   &  $1729$  & $0.14$ \\
       	\uppercase{web}	& $12.8$K & $3.6$M   &  $1663$  & $0.14$ \\
       	\uppercase{ylt}	& $12.2$K & $3.9$M   &  $1682$  & $0.17$ \\
        \bottomrule
    \end{tabular}
    \captionof{table}{Corpus descriptives, showing type and token frequencies (tok), number of unique (uniq) types, and the punctuation (punc) ratio per Bible style.}
    \label{tab:desc}
  \end{minipage}
\end{figure}

We use a parallel corpus of five different versions of the English Bible (retrieved from GitHub).\footnote{\url{https://github.com/scrollmapper/bible_databases}, which was originally collected from: \url{http://www.openbible.info/labs/cross-references/}.} These versions are semantically consistent, but vary stylistically across different aspects; BBE is written in simplified English, YLT follows the syntactic structures of the original Greek and Hebrew, and other versions (DBY, ASV) correspond to older editions reflecting diachronic variations. The sentence-level verse coding (book + chapter + verse ID) is used for almost perfect pairing between the different versions (some missing pairs were removed), forming style quintuplets, which were tokenized using \texttt{spaCy}\footnote{\url{https://spacy.io/}} \citep{honnibal2020spacy}. The style tuples are divided amongst train ($80\%$), dev ($10\%$), and test ($10\%$) sets---ensuring all sentences in the development and test splits are unseen, regardless of the style combination that they occur in---and all style combinations (excluding a same-style combination) are used to form $620,752$ pairs. Further corpus descriptives for the different styles can be found in Figure~\ref{fig:jac} and Table~\ref{tab:desc}.

\subsection{Adversary} The adversary is a sentence-level \texttt{fastText}\footnote{\url{https://github.com/facebookresearch/fastText}} \citep{joulin-etal-2017-bag} classifier; a simple linear model with one hidden embedding layer that learns sentence representations using bag of words or $n$-gram input, producing a probability distribution over the given styles using the softmax function. The classifier is trained on the source side of the training split, as these are the instances we intend to obfuscate. It is run for $20$ epochs using $100$ dimensional embeddings, uni and bi-grams, a learning rate of $0.01$, and a bucket size of $1$M. It achieves an accuracy of $86.6\%$, and chance level performance for the adversary is $20\%$ given $5$ classes with an even distribution.

\subsection{Evaluation}

To automatically evaluate the reconstruction and semantic preservation of our generated sentences, we use MT metrics, as well as distance in semantic embedding space. Obfuscation is measured by the difference of the adversary's accuracy compared to chance.

\paragraph{MT Metrics (\textnormal{\textsc{bleu}}, \textnormal{\textsc{meteor}})} We calculate \textsc{bleu}-4 and \textsc{meteor} \citep{papineni-etal-2002-bleu,banerjee-lavie-2005-meteor} using \texttt{nlg-eval}.\footnote{\url{https://github.com/Maluuba/nlg-eval}} Given that this is not a standard MT task, we provide these scores between the generated sentence and the source sentence ($\leftarrow$), as well as the generated sentence and the target sentence ($\rightarrow$). For the sequence-to-sequence models, $\rightarrow$ is the primary indicator of successful obfuscation-by-transfer. However, $\leftarrow$ gives some indication how much the output is still related to the original. The GRL should decrease scores for both $\leftarrow$ and $\rightarrow$. For the autoencoder, evaluation can only be conducted on $\leftarrow$, for which scores should strongly decrease when adding the GRL.

\paragraph{Word Mover's Distance (WMD)} To measure the embedding distance of the obfuscated sentence to the original, we take the Word Mover's Distance (WMD) \citep{kusner2015word}, based on the English \texttt{fastText} embeddings for Wikipedia \citep{bojanowski2016enriching}. WMD takes the distance between two sentences in a weighted point cloud of embedded words as the minimum cumulative distance that this cloud of sentence $A$ needs to travel to align with that of sentence $B$---shown to capture some semantic relations.

\paragraph{Adversary Impact ($\Delta\ \textsc{acc}$)} We compare the accuracy of the adversary on the generated sentence to that of the original to assess obfuscation strength. However, as our goal is to keep the adversary's performance level close to chance, we define $\Delta\ \textnormal{accuracy} = \textnormal{accuracy} - p$ where $p$ is majority baseline. Therefore, if $\Delta\ \textnormal{accuracy}$ is negative, this means the adversary's performance has dropped below chance, and the task is closer to obfuscation-by-transfer rather than by-invariance. Subsequently, a positive score indicates the extent to which obfuscation fails to match both cases.

\paragraph{Gaussian Noise ($\sigma$)} To enforce a significant change in the decoded output, we add a Gaussian noise mask to the context vector during generation time. This mask is a random vector sampled from a Gaussian Distribution
$N(0,\sigma)$, where $\sigma = \{0.01, 0.05, 0.10,$ $0.15, 0.20\}$ and add this to the values of the context vector. This noise can be used to increase obfuscation (due to more random decoding behavior) at the cost of quality of the output.

\subsection{Experiments}

Using all components discussed above, we define three experimental settings to measure the effect of applying a GRL and a conditional decoder to achieve obfuscation-by-invariance. (\textbf{1}) We train our architecture on the style pairs from the English Bible corpus in a many-to-many sequence-to-sequence setting. By introducing a GRL here, words that are highly indicative of the target style are not captured by the encoder. To achieve an effective many-to-many MT system (and thus style-transfer) setting, we prepend the \texttt{<2\{stylename\}>} token. (\textbf{2}) We train an autoencoder on disconnected pairs. Here we introduce both the GRL, plus the conditioned decoder. We hypothesize that the conditional decoder allows for soft style-transfer from a neutral encoding, implying it would preserve semantic structure better than the MT model. (\textbf{3}) We use Gaussian noise to make the sequence-to-sequence and autoencoder models equivalent in obfuscation performance, and gauge their respective loss in quality at chance level performance.

\section{Results}

\begin{table}[t!]
 \centering
 \footnotesize
 \resizebox{0.95\textwidth}{!}{
 \begin{tabular}{l|ccc|rrrrrrrr}
 \toprule
\textsc{model} & \textsc{c} & \textsc{grl} & \textsc{tt}	&\textsc{ppl} & \textsc{bleu}  & \textsc{mtr} & \textsc{bleu} & \textsc{mtr} & \textsc{wmd} & \textsc{acc}  \\ 
 & & & & & $\leftarrow$ & $\leftarrow$ & $\rightarrow$ & $\rightarrow$ & & ${\Delta}$ \\
    \midrule
	\textsc{s2s} &    &   &   & $9.08$ &  $22.0$  &  \purp{$25.3$} & $17.8$ & $23.6$ & $1.50$ & \purp{$ 1.6$} \\
	\textsc{s2s} &    & $x$ &   & $9.27$ &  $21.8$  &  \purp{$25.2$} & $16.9$ & $22.9$ & $1.50$ & \purp{$  6.9$} \\ 
    \textsc{s2s}  &   &   & $x$ & $7.38$ &  $34.9$  &  \purp{$30.5$} & $39.2$ & $29.9$ & $1.24$ & \purp{$  -12.0$} \\     
    \midrule
    \textsc{ae}  &   &    &   & $1.51$ &  $79.5$  &  \purp{$52.9$} & $   -$ & $   -$ & $0.25$ & \purp{$64.4$} \\
    \textsc{ae}  &   & $x$  &   & $1.99$ &  $60.0$  &  \purp{$41.2$} & $   -$ & $   -$ & $0.65$ & \purp{$50.8$}  \\ 
    \textsc{ae}  & $x$ & $x$  &   & $1.87$ &  $51.9$  &  \purp{$38.1$} & $   -$ & $   -$ & $0.79$ & \purp{$18.3$} \\ 
    \midrule
    source       &   & 	  &   &     & $100.0$   & $100.0$ & $36.0$ & $34.5$ & $0.00$ & $66.6$ \\
    \bottomrule
  \end{tabular}}
  \caption{Results for the Bible experiments. The first column (model) indicates the setting of our base architecture: either sequence-to-sequence (\textsc{s2s}), or autoencoder (\textsc{ae}). The second column specifies which modules were incorporated: \textsc{c} for the conditional decoder, \textsc{grl} for the Gradient Reversal Layer, and \texttt{tt} for the prepended style token. The results show perplexity on the dev set (\textsc{ppl}), \textsc{bleu}-4 and \textsc{meteor} between source ($\leftarrow$) / target ($\rightarrow$) and the obfuscated sentence, the Word Mover's Distance score between source and the obfuscated sentence (\textsc{wmd}) and the extent to which the obfuscated sentence pushes the adversary to chance level performance ($\Delta$ \textsc{acc}). In the last row, we note a `source' baseline, that is achieved by copying the source. This shows the overall overlap between source and target, and the above-chance performance of the adversary.}
  \label{tab:scores}
\end{table}

\subsection{Experimental Results}

All results and automatic evaluations are shown in Table~\ref{tab:scores}. As we hypothesized, using style-transfer for obfuscation works well overall, performing either at, or below chance level. Looking at the target \textsc{bleu} and \textsc{meteor}, the sequence-to-sequence model without the target token generates sentences that are closer to source than they are to the target; and achieves low scores overall, with the sentences being quite far off based on \textsc{wmd}. However, note that this is many-to-many translation without any signal regarding the target, given languages with largely the same vocabulary. As such, \textsc{tt} is a more realistic reflection of style-transfer success. In terms of translation quality, it barely improves over the original baseline, but it does successfully perform obfuscation-by-transfer, as indicated by the 12\% accuracy below chance. Assessing the performance of the GRL in this setting, it does not seem to improve translation, nor obfuscation, as is largely in line with our expectations.

\begin{table}
 \centering
 \footnotesize
 \begin{tabular}{l|rrr|rrr|rrr}
 \multicolumn{2}{l}{}                & \multicolumn{2}{l}{\textsc{ae}} &  \multicolumn{3}{c}{\textsc{ae + grl}} & \multicolumn{3}{c}{\textsc{ae + c + grl}} \\
  \toprule
	\textsc{$\sigma$} & \textsc{bleu} & \textsc{mtr} & \textsc{acc} &  \textsc{bleu} & \textsc{mtr} & \textsc{acc} & \textsc{bleu} & \textsc{mtr} & \textsc{acc} \\
	 & $\leftarrow$ & $\leftarrow$ & ${\Delta}$ & $\leftarrow$ & $\leftarrow$ & ${\Delta}$ & $\leftarrow$ & $\leftarrow$ & ${\Delta}$ \\
    \midrule
  	$-   $  & $79.5$ & $52.9$ & $64.4$ & $60.0$ & $41.2$ & $50.8$ & $51.9$ & $38.1$ & $18.3$ \\
    $0.01$  & $78.7$ & $52.2$ & $64.1$ & $59.7$ & $41.4$ & $50.5$ & $52.0$ & $38.5$ & $18.5$ \\
  	$0.05$  & $52.9$ & $38.9$ & $57.2$ & $54.8$ & $39.1$ & $47.6$ & $49.1$ & $37.2$ & $16.5$ \\
    $0.10$  & $14.7$ & $21.6$ & $36.1$ & $32.8$ & $32.7$ & $36.9$ & $40.7$ & $33.6$ & $11.6$ \\
    $0.15$  & $ 4.2$ & $14.7$ & $23.3$ & $25.3$ & $26.2$ & $24.3$ & $30.0$ & $29.1$ & $ 4.6$ \\
    $0.20$  & $ 1.5$ & $11.3$ & $16.4$ & $15.3$ & $21.5$ & $15.5$ & $21.1$ & $24.9$ & $ 0.0$ \\
    \bottomrule
  \end{tabular}
  \caption{Effect of adding Gaussian Noise ($\sigma$) to the autoencoder.}
  \label{tab:noise}
\end{table}

The autoencoder setting provides a clearer look into the performance of the GRL, in particular in terms of obfuscation-by-invariance. The plain autoencoder to some extent successfully reproduces the target; looking at \textsc{bleu}, the adversary performance, and \textsc{wmd}, it is still closely related to the input. Introducing the GRL does impact the relation to the source sentence, but gains little in comparison in terms of obfuscation performance. However, when the conditioned decoder is added to the architecture, obfuscation performance is visibly impacted more than the decrease in \textsc{bleu} and \textsc{meteor}. Lastly, the effect of adding a Gaussian noise mask on the decoder can be found in Table~\ref{tab:noise}. Around $0.15$, the metrics seem to be comparable in terms of \textsc{bleu}, \textsc{meteor} and $\Delta\ \textnormal{accuracy}$. We further investigate the three most suitable models (\textsc{s2s + tt}, \textsc{ae + grl + c}, and \textsc{ae + grl + c + $\sigma$ = 0.15}) in a human evaluation.

\subsection{Human Evaluation} \label{subs:human}

For the human evaluation, $20$ pairs (original, obfuscated) were sampled from the output of the three models of interest (making $60$ pairs in total). Each pair was rated by four participants (all with a linguistics background) on three dimensions using a 3-point scale. These dimensions included semantic consistency between the original and the obfuscated sentence, the syntactical coherence of the latter, and the amount of changes in the output. The raters were made aware which sentence of the pairs was the original, and were explicitly asked to rate the dimensions with the original as reference. The raters were not aware that there were multiple models, and the pairs were shuffled so that comparing between pairs with the same original was impossible. To simplify the comparison to the original, we only sampled from BBE (Basic English Bible). See Table~\ref{tab:ins} for a sample of the instructions given to the participants, and Table~\ref{tab:hum} for the results. 

According to the evaluation results, \textsc{ae + grl + c} is preferred by raters in all three dimensions. Specifically, given that we are interested in semantic preservation, this model is evaluated better than a style-transfer model that has some access to semantic relations between source and target on the \textsc{semantics} dimension. Based on the \textsc{changes} dimension, the \textsc{ae + grl + c} is the most conservative, which is in line with the \textsc{bleu} and \textsc{meteor} scores in Table~\ref{tab:scores}, and likely propagates into the \textsc{grammaticality} dimension.

\begin{table}[t!]
\resizebox{\columnwidth}{!}{
\def\arraystretch{1.2}
\begin{tabular}{l|>{\raggedright\arraybackslash}p{4.2cm}>{\raggedright\arraybackslash}p{4.2cm}>{\raggedright\arraybackslash}p{4.2cm}}
\toprule
 &  \textsc{semantics}       &  \textsc{grammaticality}          &   \textsc{changes}    \\
\midrule
1 & Semantics are broken; sentence does not mean the same. & Part(s) of, or the complete sentence is garbled. & Change in special characters or flipping a single word. \\
2 & Slight semantic change, but not intrusive.              & Slight word order change that is ungrammatical.    & Multiple words were changed, but they align with the original. \\
\ 3 & Semantics are intact, changes do not alter the meaning. & Grammaticality has not been affected.
 & New parts have been introduced / rewritten in the sentence. \\
\bottomrule
\end{tabular}}
\caption{Explanations given to the raters in the human evaluation study, shown per rating for the three dimensions.}
\label{tab:ins}
\end{table}

\begin{table}[t!]
\centering
\footnotesize
\begin{tabular}{l|rrr}
\toprule
      		                            & \textsc{semantics} & \textsc{grammaticality} & \textsc{changes} \\
\midrule
\textsc{s2s + tt}                       &  $1.88$ & $2.21$  & $2.43$ \\
\textsc{ae + grl + c}                   &  \pale{$2.12$} & \pale{$2.32$}  & \pale{$1.99$} \\
\textsc{ae + grl + c + $\sigma$ = 0.15} &  $1.35$ & $1.66$  & $2.65$ \\
\bottomrule
\end{tabular}
\caption{Average human evaluation scores per model for the three dimensions. Higher is better, although \textsc{changes} is ideal around 2.}
\label{tab:hum}
\end{table}

\subsection{Qualitative Analysis}

Here we perform a comparison of the behavior, and the text generated by the three models that were evaluated by our raters in the previous section. Accordingly, we will identify the strengths and weaknesses of our experimental approach, and propose lines of future work.

One of the issues we identified with obfuscation-by-transfer was that of small, localized changes in the input, specifically focusing on words that are most relevant to the adversary. When looking at longer sentences such as Table~\ref{tab:ex1}, some (semi-)correct variants can be found in e.g., town $\rightarrow$ city, waste land $\rightarrow$ dry land, and in Table~\ref{tab:ex2}, beryl $\rightarrow$ onyx, stamp $\rightarrow$ seal, but incorrect ones also remain: rest $\rightarrow$ work. Overall, the longer the sequence, the more variation can be observed.

\begin{table}
\footnotesize
\centering
\def\arraystretch{1.4}
\begin{tabular}{p{1.75cm}|p{9cm}}
\textsc{original}                       & For the strong town is without men, an unpeopled living - place; and she has become a waste land: there the young ox will take his rest, and its branches will be food for him. \\
\textsc{s2s + tt}                       & For the strong \hlfancy{DrawPaleGold}{city} is \hlfancy{DrawPaleRed}{powerless}, an \hlfancy{DrawPaleGold}{astonishment} living \hlfancy{DrawPaleRed}{ } and she is become a \hlfancy{DrawPaleGold}{corrupt} land: \hlfancy{DrawPaleRed}{a} young ox \hlfancy{DrawPaleRed}{shall} rest, and its branches \hlfancy{DrawPaleGold}{shall} be \hlfancy{DrawPaleRed}{ } for him. \\
\textsc{ae + grl + c}                   & For the \hlfancy{DrawPaleRed}{ } \hlfancy{DrawPaleGold}{city} is without \hlfancy{DrawPaleGreen}{living} men, \hlfancy{DrawPaleRed}{there is without} a waste land; and she \hlfancy{DrawPaleGold}{makes an dry} land: \hlfancy{DrawPaleRed}{an} ox \hlfancy{DrawPaleGold}{shall} take his \hlfancy{DrawPaleGold}{work}, and \hlfancy{DrawPaleGold}{their} branches \hlfancy{DrawPaleGold}{shall} be food for him. \\
\textsc{ae + grl + c + $\sigma$ = 0.15} & \hlfancy{DrawPaleGreen}{A man dwelleth} without \hlfancy{DrawPaleGreen}{an dry land}; and \hlfancy{DrawPaleGreen}{, wandering she - }place; \hlfancy{DrawPaleGreen}{men shall there} become a waste a land: \hlfancy{DrawPaleRed}{an} ox \hlfancy{DrawPaleGreen}{- goat} \hlfancy{DrawPaleGold}{shall} take his \hlfancy{DrawPaleGold}{horses}, and \hlfancy{DrawPaleGold}{his} branches shall be \hlfancy{DrawPaleGold}{prepare} for him. \\
\end{tabular}
\caption{Example 1 --- Isaiah 27:10. 
\protect\hlfancy{DrawPaleRed}{Red} highlights indicate deletions or collapses, 
\protect\hlfancy{DrawPaleGold}{yellow} substitutes, and 
\protect\hlfancy{DrawPaleGreen}{green} insertions and unrelated substitutes.
}%
\label{tab:ex1}
\end{table}

A more interesting observation is that some parts of sentences are added to by the models, e.g. \emph{there is without a waste land ; and she makes an dry land}---while incorrectly inserting `\emph{without}', the rest can be considered as a correct expansion. The same holds for Table~\ref{tab:ex2}, where \emph{fixed in twisted frames of gold} is expanded to \emph{whereupon they bound in the skillfully woven red frames of gold }, partly erroneously, similar to \emph{living men} in Table~\ref{tab:ex1}. In contrast, the autoencoder + GRL  also seems to favor somewhat compressed phrases, removing adjectives such as \emph{young} in \emph{young ox}, \emph{strong} in \emph{strong city}, \emph{beryl} in \emph{beryl stones} and not incorporating \emph{an unpeopled living} altogether.

When directly comparing the sequence-to-sequence and autoencoder + GRL examples, it can be inferred that the transfer approach seems (at least in these examples) quite conservative, sticking to an almost exact alignment, and only making small changes. This however also causes the autoencoder to replace words with unrelated variants or insert not directly related ones; the same behavior can also be observed in the sequence-to-sequence model, however.

\begin{table}
\footnotesize
\centering
\def\arraystretch{1.4}
\begin{tabular}{p{1.75cm}|p{9cm}}
\textsc{original} & Then they made the beryl stones, fixed in twisted frames of gold and cut like the cutting of a stamp, with the names of the children of Israel.' \\
\textsc{s2s + tt} & Then they made the \hlfancy{DrawPaleGold}{onyx} stones \hlfancy{DrawPaleGreen}{as a hundred stones}, \hlfancy{DrawPaleGold}{burning} in \hlfancy{DrawPaleGold}{engraved stones} of gold\hlfancy{DrawPaleGreen}{,} and cut \hlfancy{DrawPaleGold}{as} the \hlfancy{DrawPaleGold}{marks} of a \hlfancy{DrawPaleGold}{seal}, with the names of the children of Israel.\hlfancy{DrawPaleRed}{ } \\
\textsc{ae + grl + c} & Then they \hlfancy{DrawPaleGold}{wrote} the \hlfancy{DrawPaleRed}{ } stones, \hlfancy{DrawPaleGreen}{whereupon they bound in the skillfully woven red} frames of gold and \hlfancy{DrawPaleGold}{made} like the cutting of a stamp, with the names of the children of Israel.' \\
\textsc{ae + grl + c + $\sigma$ = 0.15} & Then they \hlfancy{DrawPaleGold}{presented} the \hlfancy{DrawPaleGreen}{pillars that belonged in Henadad. The bottom of fine} gold and \hlfancy{DrawPaleGold}{made} like the \hlfancy{DrawPaleGold}{jewels} of a stamp \hlfancy{DrawPaleGreen}{of them}, \hlfancy{DrawPaleGreen}{at the dial} of the children of Israel.' \\

\end{tabular}
\caption{Example 2 --- Exodus 39:6}
\label{tab:ex2}
\end{table}

\begin{figure}
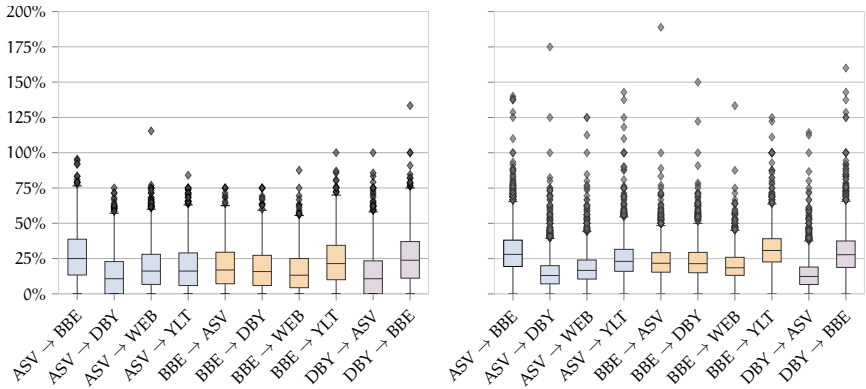

\centering
\begin{minipage}{0.48\columnwidth}
\includestandalone[scale=0.58, trim=0 0 0 4, clip]{chapters/1-style-obfuscation/gfx/box-left}
\end{minipage}
\begin{minipage}{0.48\columnwidth}
\includestandalone[scale=0.58]{chapters/1-style-obfuscation/gfx/box-right}
\end{minipage}
\caption{Distributions of the percentage of newly introduced tokens with respect to the original amount of tokens---shown for both the \textsc{ae + grl + c} obfuscation-by-invariance architecture (left), and the \textsc{s2s + tt} obfuscation-by-transfer model (right). Three out of five source styles are shown.}
\label{fig:box}
\end{figure}

The output of the autoencoder shows some evidence that minor \emph{rewrites} of the sentence are employed, which could potentially be an interesting path to further pursue. Including different variables in the conditional decoder such as sentence length, also demonstrated by \cite{ficler2017controlling}, would make experiments in this direction feasible. Rewrites are not restricted to the autoencoder only, as is demonstrated in Figure~\ref{fig:box}. We previously noted in Section~\ref{subs:human} that the \textsc{ae + grl + c} architecture seems more conservative in changing the input, which can also clearly be induced from the generally equal or higher amount of newly introduced tokens by the \textsc{s2 + tt} model, and its many more outliers (some even exceeding the original amount of tokens). This plot also reveals a tendency to add more tokens for BBE and YLT, which is in line with their larger amount of tokens shown in Table~\ref{tab:desc}. The extreme outliers are largely attributable to artifacts of recurrence (\emph{and Benaiah , the son of Zadok , and Eber , and Eliphelet}).

Finally, it must be noted that evidence of style-neutral rewrites is difficult to find in generated output concerning the Bible; not only due to possible archaic constructions (which we attempted to minimize by using BBE), but more so due to the fact that it requires a level of expertise to recognize style shifts. However, most importantly, applying the autoencoder to data that is non-parallel has some viable ground given the current results, and might therefore be extended to other domains in future work.

\section{Conclusion}

We presented an alternative framing of the task of automatic style obfuscation---obfuscation-by-invariance---and tested several components in a neural encoder-decoder architecture that were hypothesized to achieve style-invariant rewrites of the input text. We tested the effect of a Gradient Reversal Layer and a conditional decoder for obfuscation in parallel and non-parallel settings. Although strong evidence for style-neutral text was difficult to find for the Bible corpus, we demonstrated through human evaluation that our autoencoder architecture trained on non-parallel data obtained a better evaluation than a model trained on parallel data with partial access to semantic relations between source and target. In our qualitative analysis, we found evidence for semantically correct local changes of the input, as well as partial rewrites that fit the context of the verses. These results bode well for extending this architecture to other non-parallel corpora to test obfuscation in a practical use-case, e.g.\ author attributes such as age and gender.

\begin{figure}
    \centering
    \includegraphics[width=0.75\textwidth,angle=-90]{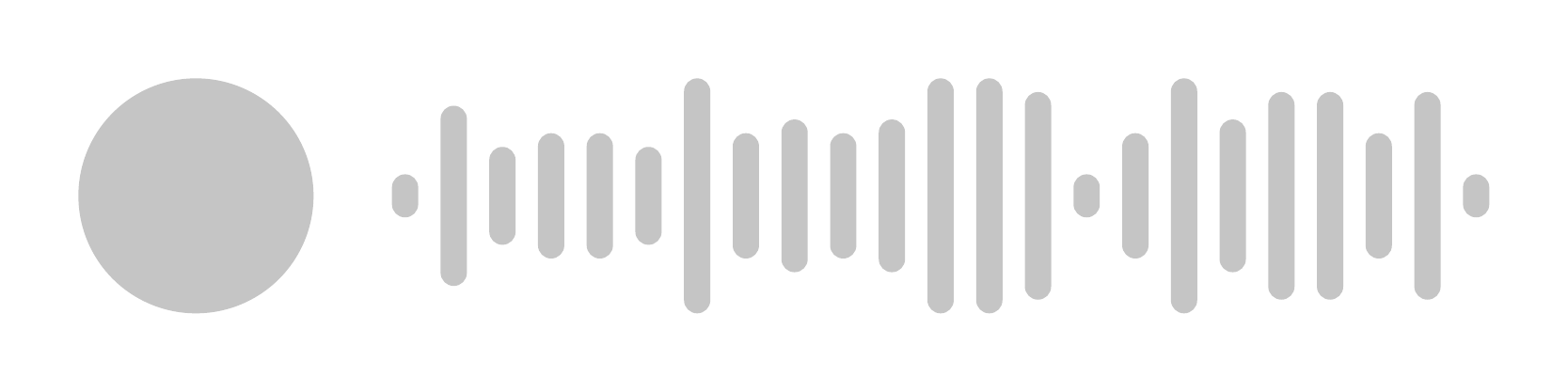}
\end{figure}

\thispagestyle{empty}
\strut
\newpage
\thispagestyle{empty}
\strut
\vfill

\noindent This chapter has been published as:

\begin{description}
    \item \bibentry{emmery-etal-2021-adversarial}
\end{description}

\noindent Minor formatting changes have been made to the text and figures (save for Figure~\ref{fig:obf} which has been significantly improved).

\chapter[Adversarial Stylometry in the Wild]{Adversarial Stylometry in the Wild: Transferable Lexical Substitution Attacks on Author Profiling} \label{chap:advsty}

\lettrine[lines=4, lraise=0, nindent=0em, slope=0em]{W}{ritten} language contains stylistic cues that can be exploited to automatically infer a variety of potentially sensitive author information. Adversarial stylometry intends to attack such models by rewriting an author's text. Our research proposes several components to facilitate deployment of these adversarial attacks in the wild, where neither data nor target models are accessible. We introduce a transformer-based extension of a lexical replacement attack, and show it achieves high transferability when trained on a weakly labeled corpus---decreasing target model performance below chance. While not completely inconspicuous, our more successful attacks also prove notably less detectable by humans. Our framework therefore provides a promising direction for future privacy-preserving adversarial attacks.

\newpage
\parshape=0
\section{Introduction}

The widespread use of machine learning on consumer devices and its application to their data has sparked investigation of security and privacy researchers alike in correctly handling sensitive information \citep{DBLP:journals/corr/EdwardsS15,DBLP:conf/ccs/AbadiCGMMT016}. Natural Language Processing (NLP) is no exception \citep{DBLP:conf/post/FernandesDM19,li-etal-2018-towards}; written text can contain a plethora of author information---either consciously shared or inferable through stylometric analysis \citep{DBLP:conf/uss/RaoR00,adams2006classification}. This characteristic is fundamental to
author profiling \citep{DBLP:journals/lalc/KoppelAS02}, and while the field's main interest pertains to the study of sociolinguistic and stylometric features that underpin our language use \citep{DBLP:conf/cicling/Daelemans13}, herein simultaneously lie its dual-use problems. Author profiling can, often with high accuracy, infer an extensive set of (sensitive) personal information, such as age, gender, education, socio-economic status, and mental health issues \citep{eisenstein-etal-2011-discovering,DBLP:conf/icmla/AlowibdiBY13,volkova-etal-2014-inferring,plank-hovy-2015-personality,volkova-bachrach-2016-inferring}. It therefore potentially exposes anyone sharing written online content to unauthorized information collection through their writing style. This can prove particularly harmful to individuals in a vulnerable position regarding e.g., race, political affiliation, or mental health. 

Privacy-preserving defenses against
such inferences can be found in the field of adversarial\footnote{These are adversarial attacks on models making stylometric predictions, not to be confused with adversarial learning.} stylometry. Our research\footnote{All code, data, and materials to fully reproduce the experiments are openly available at \url{https://github.com/cmry/reap}.} concerns the obfuscation subtask, where the aim is to rewrite an input text such that the style changes, and stylometric predictions fail. It is part of a growing body of research into adversarial attacks on NLP \citep{DBLP:journals/corr/abs-1207-0245}, which various modern models have proven vulnerable to; e.g., in neural machine translation \citep{ebrahimi-etal-2018-adversarial}, summarization \citep{DBLP:conf/aaai/ChengYCZH20}, and text classification \citep{DBLP:conf/ijcai/0002LSBLS18}.

Adversarial attacks on NLP are predominantly aimed at demonstrating vulnerabilities in existing algorithms or models, such that they might be fixed, or explicitly improved through adversarial training. Consequently, most related work focuses on white or black-box settings, where all or part of the target model is accessible (e.g., its predictions, data, parameters, gradients, or probability distribution) to fit an attack. The current research, however, does not intend to improve the targeted models; rather, we want to provide the attacks as tools to protect online privacy. This introduces several constraints over other NLP-based adversarial attacks, as it calls for a realistic, in-the-wild scenario of application. 

Firstly, authors seeking to protect themselves from stylometric analysis cannot be assumed to be knowledgeable about the target architecture, nor to have access to suitable training data (as the target could have been trained on any domain). Hence, we cannot optimally tailor attacks to the target, and need an accessible method of mimicking it to evaluate the obfuscation success. To facilitate this, we use a so-called substitute model, which for our purposes is an author profiling classifier trained in isolation (with its own data and architecture) that informs our attacks. Attacks fitted on substitute models have been shown to transfer their success when targeting models with different architectures, or trained on other data, in a variety of machine learning tasks \citep{DBLP:journals/corr/PapernotMG16}. The effectiveness of an attack fitted on a substitute model when targeting a `real' model is then referred to as \emph{transferability}, which we will measure for the obfuscation methods proposed in the current research. 

Secondly, for an obfuscation attack to work in practice (e.g., given a limited post history), it should suggest relevant changes --to-- the author's writing \emph{on a domain of their choice}. This implies the substitute models should be fitted locally, and therefore need to meet two criteria: reliable access to labeled data, and being relatively fast and easy to train. To meet the first criterion, the current research focuses on gender prediction, as: i) Twitter corpora annotated with this variable are by far the largest (and most common), ii) author profiling methods typically use similar architectures for different attributes; therefore, the generalization of attacks to other author attributes can be assumed to a large extent, and, most importantly, iii) \cite{beller-etal-2014-im} and \cite{emmery-etal-2017-simple} have shown that through distant labeling, a representative corpus for this task can be collected in under a day. This allows us to measure transferability of attacks fitted using realistically collected distant corpora to models using high-quality hand labeled corpora.

\begin{figure}[t!]
\centering
\includegraphics[width=\columnwidth]{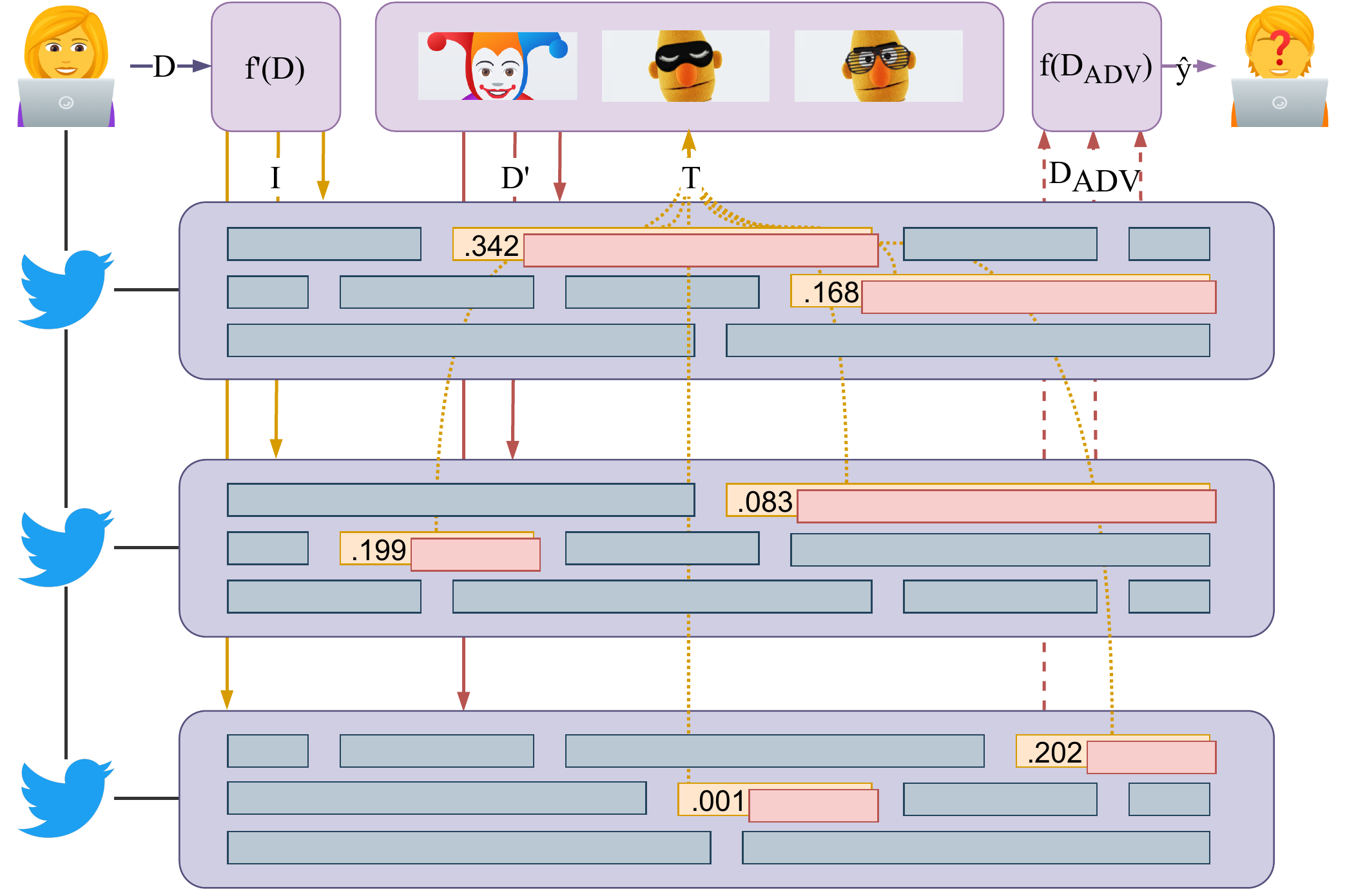}
\caption{Obfuscation scenario: model $f'$ trains on tweet batches, an omission score is used to determine and rank the words according to their classification contribution. These are then passed to either TextFooler, Masked BERT, or Dropout BERT to suggest top-$k$ replacement candidates. From these, a selection is made based on their class probability change on $f'(D)$. Finally, $f$ is evaluated on the perturbed tweets $D_{\textsc{adv}}$.}
\label{fig:obf}
\end{figure}

As for the attacks, we focus on lexical substitution of content words strongly related to a given label (i.e., attribute), as those have been shown to explain a significant portion of the accuracy of stylometric models \cite[see e.g.,][]{DBLP:conf/cikm/RaoYSG10,burger-etal-2011-discriminating,sap-etal-2014-developing,DBLP:conf/clef/PardoRVDPS16}. To that effect, we extend the substitution attack of \cite{DBLP:conf/aaai/JinJZS20} and apply it to author attribute obfuscation. Specifically, we explore the potential of training a simple (as to meet the speed criterion), non-neural substitute model $f'$ to indicate relevant words to perturb, during which retaining the original meaning is prioritized. Two transformer-based models are introduced to the framework to propose and rank lexical substitutions towards a change in the predictions of $f'$. We evaluate if the attacks on $f'$ transfer across corpora, architectures, and a separately trained target model $f$ (see Figure~\ref{fig:obf}). Finally, we measure the quality of changes using automatic evaluation metrics, and conduct a human evaluation that focuses on detection accuracy of the attacks.

\section{Related Work}

Stylometry, the study of (predominantly) writing style, dates back  several decades \citep{doi:10.1080/01621459.1963.10500849}, and has seen increased accessibility through the introduction of statistical models \cite[see surveys by ][]{10.1093/llc/13.3.111,DBLP:journals/csur/NealSFYXW17} and machine learning \cite[e.g.,][]{10.1093/llc/8.4.203,10.1093/llc/9.1.1}. Computational stylometry distinguishes several subtasks such as determining \cite[][]{baayen2002experiment} and verifying author identity \cite[][]{DBLP:conf/icml/KoppelS04}, and author profiling \citep{argamon2005lexical}; e.g., predicting demographic attributes. Adversarial stylometry (as conceptualized by \citeauthor{DBLP:journals/tissec/BrennanAG12}, \citeyear{DBLP:journals/tissec/BrennanAG12}) intends to subvert these inferences by changing an author's text through imitation, or, as pertains to our research, the obfuscation of writing style \citep{kacmarcik-gamon-2006-obfuscating,caliskan2015coding,DBLP:conf/wistp/ThiSG15,xu-etal-2019-alter}. 

These changes, or perturbations, can be produced in several ways, and the task is therefore often conflated with paraphrasing \citep{reddy-knight-2016-obfuscating}, style transfer \citep{kabbara-cheung-2016-stylistic}, and generating adversarial samples or triggers \citep{DBLP:journals/tist/ZhangSAL20}. Regardless of the employed method, the main challenge of obfuscation lies in retaining the original meaning of an input text; its written language medium limits any perturbations to discrete outputs, and unnatural discrepancies are significantly better discernible by humans than, say, a few pixel changes in an image. An additional, persistent limitation is the absence of evaluation metrics that guarantee complete preservation of the original meaning of the input whilst changes remain unnoticed \citep{DBLP:conf/clef/PotthastHS16}. This not only inhibits automatic evaluation of obfuscation, but all natural language generation (and processing, generally) research \citep{novikova-etal-2017-need}---placing an emphasis on human evaluation \citep{van-der-lee-etal-2019-best}.

It is perhaps for this reason that most obfuscation work uses heuristically-driven, controlled changes such as splitting or merging words or sentences, removing stop words, changing spelling, punctuation, or casing \cite[see e.g.,][]{10.1007/978-3-319-65813-1_18,eger-etal-2019-text}. These specific attacks are typically easier to mitigate through preprocessing \citep{DBLP:conf/ifip11-9/JuolaV11}.  Obfuscation through lexical substitution \citep{DBLP:conf/clef/MansoorizadehRA16,DBLP:conf/clef/JooH19,DBLP:journals/it/BevendorffWPH020} provides a middle ground of control, semantic preservation and attack effectiveness; however, they might prove less effective against models relying on deeper stylistic features (e.g.\ word order, part-of-speech (POS) tags, or reading complexity scores). End-to-end systems have been employed for similar purposes \citep{DBLP:conf/uss/ShettySF18,saedi-dras-2020-large}, or to rewrite entire phrases \citep[as in \Cref{chap:textobf}, and][]{bo-etal-2021-er} using (adversarially-driven) autoencoders. Such attacks seem less common, and provide less control over the perturbations and semantic consistency.

Our work does not assume the attacks to run end-to-end, but with a hypothetical human in the loop. We further opt for techniques that are more likely to find strong semantic mirrors to the original text while making minimal changes. A substitute model (the algorithm, hyper-parameters, and output of which an author can manipulate as desired) is employed to indicate candidate replacement words, and our attacks suggest and rank those against this substitute. Moreover, prior work typically attacks adversaries trained on the same data, whereas we add a transferability measure. Lastly, while author identification has been investigated in the wild \citep{DBLP:conf/ifip11-9/StolermanOAG14}, our work is, to our knowledge, the first to make a conscious effort towards realistic applicability of obfuscation techniques.

\section{Method}

Our framework extends TextFooler \cite[TF,][]{DBLP:conf/aaai/JinJZS20} in several ways. First, a substitute gender classifier is trained, from which the logit output given a document is used to rank words by their prediction importance through an omission score (Section~\ref{subs:twi}). For the top most important words, substitute candidates are proposed, for which we add two additional techniques (Section~\ref{subs:att}). These candidates can be checked and filtered on consistency with the original words (by their POS tags, for example), accepted as-is, or re-ranked (Section \ref{subs:checks}). For the latter, we add a scoring method. Finally, the remaining candidates are used for iterative substitution until TF's stopping criterion is met (i.e., the prediction changes, or candidates run out).

\subsection{Target Word Importance} \label{subs:twi}

We are given a target classifier $f$, substitute classifier $f'$, a document $D$ consisting of tokens $D_i$, and a target label $y$. Here, $f'$ is trained on some corpus $X$, and receives an author's new input text $D$, where the author provides label $y$. We denote a class label as $\bar{y}$ if $f'(D)$ predicts anything but $y$. Our perturbations form adversarial input $D_\textsc{adv}$, that intends to produce $f'(D_\textsc{adv}) = \bar y$, and thereby implicitly $f(D_\textsc{adv}) = \bar y$. Note that we only submit $D$ to $f$ for evaluating the attack effectiveness, and it is never used to fit the attack itself.

To create $D_\textsc{adv}$, a minimum number of edits is preferred, and thus we rank all words in $D$ by their omission score \cite[similar to e.g.,][]{kadar-etal-2017-representation} according to $f'$ (\texttt{omission\_score} in Algorithm~\ref{alg:obf}). Let $D_{\setminus i}$ denote the document after deleting $D_i$, and $o_y(D)$ the logit score by $f'$. The omission score is then given by $o_y(D) - o_y(D_{\setminus i})$, and used in an importance score $I$ of token $D_i$, as:
\begin{equation}
\label{eq:words_importance}
    I_{D_i}=
    \begin{cases}
      o_y(D) - o_y(D_{\setminus i}), \\ \quad \ \  \ \text{if}\ f'(D)=f'(D_{\setminus i}) = y. \\
      o_y(D) - o_y(D_{\setminus i}) + o_{\bar{y}}(D) - o_{\bar{y}}(D_{\setminus i}), \\ \quad \ \ \   
      \text{if}\ f'(D) = y, f'(D_{\setminus i})= \bar y, y\neq \bar y.
    \end{cases}
\end{equation}
With $I_{D_i}$ calculated for all words in $D$, the top $k$ ranked tokens are chosen as target words $T$.

\begin{algorithm}[t!]
\SetKwFunction{cumprod}{cumprod}
\SetKwFunction{length}{length}
\SetKwFunction{zeros}{zeros}
\SetKwFunction{ceil}{ceil}
\SetKwInOut{Input}{Input}
\SetKwInOut{Output}{Output}
\caption{Obfuscation by lexical replacement.\label{alg:obf}}
\Input{
		\hspace{2mm} \xvbox{2mm}{$f'$} -- substitute model \\
		\hspace{2mm} \xvbox{31mm}{$D = \{w_0, w_1, \ldots, w_n\}$} -- document \\
		\hspace{2mm} \xvbox{2mm}{$y$} -- target label \\
		\hspace{2mm} \xvbox{9mm}{\xvar{checks}} -- apply checks (bool) \\
		\hspace{2mm} \xvbox{2mm}{\xvar{k}} -- target max $k$-amount words \\
	  }
\Output{
		\hspace{2mm} \xvbox{7mm}{$D_\textsc{adv}$} -- obfuscated document
	   }

  \BlankLine
          
  \For{$D_i \in D$ }{
    $I_{D_i} \leftarrow$ \texttt{omission\_score}($f'$, $y$) $\quad$ \tcp{via Equation~\ref{eq:words_importance}}
	
  }
  \BlankLine
  $T \leftarrow$ \texttt{top\_k}(\texttt{argsort\_desc}($D, I_{D_i}$ scores), k)
  \BlankLine
  $D_\textsc{adv} = D$ \\
  \For{$t \in T$ }{
    \tcp{substitution attack on $t$}
    $C_t \leftarrow$\texttt{candidates}($t$) \\
    $A = ( D_{\textsc{adv}_{1:i-1}}, C_{t, j}, D_{\textsc{adv}_{i+1:n}} )_{1 \leq j \leq |C_t|}$ \\
    $\bar{A} =$ \texttt{filter/rank}($D$, $A;t$, checks) \\
    \tcp{test attack success on $f'$}
    \For{$D' \in \bar{A}$ }{
        \uIf{$\argmax o_y(D') \neq y$}{\Return $D_\textsc{adv} = D'$}
        \uElseIf{$o_y(D') < o_y(D_\textsc{adv})$}{
                $t$ in $D_\textsc{adv}$ is replaced with $c$ from $D'$  
        }
    }
}
\Return $D_\textsc{adv}$
\end{algorithm}

\subsection{Lexical Substitution Attacks} \label{subs:att}

Four approaches to determine how to perturb a target word $t \in T$ are considered in our experiments. These operations are referred to as \texttt{candidates} in Algorithm~\ref{alg:obf}.

\paragraph{Synonym Substitution (WS)}

This TF-based substitution embeds $t$ as $\boldsymbol{t}$ using a pre-trained embedding matrix $\boldsymbol{V}$. $C_t$ is selected by computing the cosine similarity between $\boldsymbol{t}$ and all available word-embeddings $\boldsymbol{w} \in \boldsymbol{V}$. We denote cosine similarity with $\Lambda(\boldsymbol{t}, \boldsymbol{w})$. A threshold $\delta$ is used to keep only reliable candidates $\Lambda(\boldsymbol{t}, \boldsymbol{w}) > \delta$.

\paragraph{Masked Substitution (MB)}

The embedding-based substitutions can be replaced by a language model predicting the contextually most likely token. BERT \citep{devlin-etal-2019-bert}---a bi-directional encoder \citep{DBLP:conf/nips/VaswaniSPUJGKP17} trained through masked language modeling and next-sentence prediction---makes this fairly trivial. By replacing $t$ with a mask, BERT produces a top-$k$ most likely $C_t$ for that position. Implementing this in TF does imply each previous substitution of $t$ might be included in the context of the current one. This method of contextual replacement has two drawbacks: i) semantic consistency with the \emph{original} word is not guaranteed (as the model has no knowledge of $t$), and ii) the replaced context means semantic drift can occur, as all subsequent substitutions follow the new, possibly (semantically) incorrect context. 

\paragraph{Dropout Substitution (DB)}

A method to circumvent the former (i.e., BERT's masked prediction limitations for lexical substitution), was presented by \cite{zhou-etal-2019-bert}. They apply dropout \citep{DBLP:journals/jmlr/SrivastavaHKSS14} to BERT's internal embedding of target word $t$ before it is passed to the transformer---zeroing part of the weights with some probability. The assumption is that $C_t$ (BERT's top-$k$) will contain candidates closer to the original $t$ than the masked suggestions.

\paragraph{Heuristic Substitution}

To evaluate the relative performance of the techniques we described before, we employ several heuristic attacks as baselines. In the order of Table~\ref{tab:results}: {\underline{1337}-speak}: converts characters to their leetspeak ($13375$P$34$K) variants, in a similar vein to e.g.\ diacritic conversion \citep{DBLP:conf/iclr/BelinkovB18}. {Character \underline{flip}}: inverts two characters in the middle of a word (wrod), which was shown to least affect readability \citep{Rayner2006-de}. {Random \underline{space}s}: splits a token into two at a random position (pos ition).

\subsection{Candidate Filtering and Re-ranking} \label{subs:checks}

Given $C_t$, either all, or only the highest ranked candidate can be accepted as-is. 
Alternatively, all $D'$ can be filtered by submitting them to checks, or re-ranked based on their semantic consistency with $D$. These operations are referred to as \texttt{rank/filter} in Algorithm~\ref{alg:obf}---both of which can be executed. 

\paragraph{Part-of-Speech and Document Encoding}

TF uses two checking components: first, it removes any $c$ that has a different POS tag than $t$. If multiple $D'$ exist such that $f'(D') = \bar y$, it selects the document $D'$ which has the highest cosine similarity to the Universal Sentence Encoder (USE) embedding \citep{cer-etal-2018-universal} of the original document $D$. If not, the $D'$ with the \emph{lowest} target word omission score is chosen (as per TF's method). 

\begin{table}
    \centering
    \footnotesize
    \begin{tabular}{l|rr|rr|rr}
    \toprule
                 & \textsc{authors}   
                           & \textsc{tweets}        
                                        & \textsc{female}    
                                                    & \textsc{male}      
                                                                & \textsc{train}     
                                                                            & \textsc{test}
                                                                            \\
    \midrule
    Huang        & $37,929$  & $47K$        & $26,758$    & $20,453$    & $30,602$    &  $7,651$ \\
    Emmmery      & $6,610$   & $16.789K$    & $61,736$    & $32,900$    & $75,918$    & $18,718$ \\ 
    Volkova      & $4,620$   & $12.227K$    & $32,376$    & $26,708$    & $47,298$    & $11,777$ \\
    \bottomrule
    \end{tabular}
    \caption{Corpus statistics indicating the number of authors, tweets, female and male labels, the size of the train and test splits, number of types (unique words) and tokens (total words), and average tokens per document (avg size). }
    \label{tab:data}
\end{table}

\paragraph{BERT Similarity} \label{p:bertsim}

\cite{zhou-etal-2019-bert} use the concatenation of the last four layers in BERT as a sentence's contextualized representation $\boldsymbol{h}$. We apply this in both Masked (MB) and Dropout (DB) BERT to re-rank all possible $D'$ by embedding them. Given document $D$, target $t$, and perturbation candidate document $D'$, $C_t$ would be ranked via an embedding similarity score:
\begin{equation}
    \begin{split}
    & \textsc{sim}\left(D, D^{\prime} ; t\right) =  \\ \sum_{i}^{n} w_{i, t} \times
    & \Lambda\left(\boldsymbol{h}\left(D_{i} | D \right), \boldsymbol{h}(D_{i}^{\prime} | D^{\prime}) \right)
    \end{split}
\end{equation}
where $\boldsymbol{h}\left(D_{i} | D\right)$ is BERT's contextualized representation of the $i^\text{th}$ token in $D$, and $w_{i, t}$
is the average self-attention score of all heads in all layers ranging from the $i^\text{th}$ token with respect to $t$ in $D$.\footnote{Zhou et al. (2019) additionally use a proposal score for finding $T$ that we replaced with the omission score.} 

\section{Experiment}

\subsection{Data} \label{subs:data-adv}

We use three author profiling sets (see Table~\ref{tab:data} for statistics) that are annotated for binary gender classification (male or female): first, that of \cite{DBLP:conf/aaai/VolkovaBAS15} which was crowdsourced via Mechanical Turk by annotating $5,000$\footnote{Profile counts in the current work differ due to collection limitations (e.g., removed accounts).} English Twitter profiles. This can be considered a `random' sample of Twitter profiles, and is therefore the most unbiased set of the three. Hence, we consider it the most representative of an author profiling set, and employ this as training split ($80\%$) for $f$, and test split for our attacks ($20\%$).

The second is the English portion of the Multilingual Hate Speech Fairness corpus of \cite{huang-etal-2020-multilingual}, which was collected with a different objective than author profiling. It was aggregated from existing hate speech corpora \cite[by][]{waseem-hovy-2016-hateful,waseem-2016-racist,DBLP:conf/icwsm/FountaDCLBSVSK18}---which were largely bootstrapped with look-up terms, selection of frequently abusive users, etc.---and annotated post-hoc with demographic information. The collection did not focus on profiles, and most authors are only associated with a single tweet. This can cause a significant domain shift compared to general author profiling. However, it can be seen as freely available (noisy) data.

Lastly, we include a weakly labeled author profiling corpus by \cite{emmery-etal-2017-simple}, collected through English keyword look-up for self-reports---similar to \cite{beller-etal-2014-im}. This corpus likely includes incorrect labels, but was collected in under a day, making it an ideal candidate for realistic access to (new) data to fit the substitute model.

\paragraph{Preprocessing \& Sampling} 

All corpora were tokenized with spaCy\footnote{\url{https://spacy.io}} \citep{honnibal2020spacy}. Other than lowercasing, allocating special tokens to user mentions and hashtags (\# and text were split), and URL removal, no preprocessing steps were applied. Every author timeline was divided into chunks for a maximum of $100$ tweets (i.e., some contain less), implying a maximum of $25$ instances per author (some contain one, $2,500$ is the API history limit). From the test set, the last\footnote{As the datasets are not shuffled to avoid overfitting on author-specific features, a few documents of the same author might spill from the train into the test split; this avoids incorporating those in our attack sample.} $200$ instances were sampled for the attack ($110$ male, $90$ female). While fairly small, this sample does reflect a realistic attack duration and timeline size, as they would be executed for a single profile.

\subsection{Attacks}

For the extension of TF, we re-implemented code\footnote{\url{https://github.com/jind11/TextFooler}} by \cite{DBLP:conf/aaai/JinJZS20} to work with Scikit-learn\footnote{\url{https://scikit-learn.org/}} \citep{DBLP:journals/jmlr/PedregosaVGMTGBPWDVPCBPD11}. For their synonym substitution component, we similarly used counter-fitted embeddings by \cite{mrksic-etal-2016-counter} trained on Simlex-999 \citep{hill-etal-2015-simlex}. The USE \citep{cer-etal-2018-universal} implementation uses TensorFlow\footnote{\url{https://tensorflow.org/}} \citep{DBLP:conf/osdi/AbadiBCCDDDGIIK16} as back-end, and all BERT-variants the Transformers\footnote{\url{https://huggingface.co/}} library \citep{wolf-etal-2020-transformers} with PyTorch\footnote{\url{https://pytorch.org/}} \citep{DBLP:conf/nips/PaszkeGMLBCKLGA19} as back-end.

We adopt the parameter settings from \cite{DBLP:conf/aaai/JinJZS20} throughout our TF experiments: they set $N$ (considered synonyms) and $\delta$ (cosine similarity minimum) empirically to $50$ and $0.7$ respectively. For MB and DB, we cap $T$ at $50$ and top-$k$ at $10$ (to improve speed). For DB, we follow \cite{zhou-etal-2019-bert} and set the dropout probability to $0.3$.

\subsection{Models}

For $f$ and $f'$ we require (preferably fast) pipelines that achieve high accuracy on author profiling tasks, and are sufficiently distinct to gauge how well our attacks transfer across architectures, rather than solely across corpora. As current state-of-the-art algorithms have not (yet) proven to be substantially effective for author profiling \citep{DBLP:conf/clef/JooH19}, we opt for $n$-gram features and linear models.

\paragraph{Logistic Regression} Logistic Regression (LR) trained on tf$\cdot$idf using uni and bi-gram features proved a strong baseline in author profiling in prior work. The simplicity of this classifier also makes it a substitute model that can realistically be run by an author. No tuning was performed: $C$ is set to $1$.

\paragraph{N-GrAM} Proposed by \cite{DBLP:conf/clef/BasileDMRHN17a} as a highly effective---yet simple---model, The New Groningen Author-profiling Model (N-GrAM) outperforms more complex (neural) alternatives on author profiling with little to no tuning. It uses tf$\cdot$idf-weighted uni and bi-gram token features, character hexa-grams, and sublinearly scaled term frequencies ($1+\log($tf$)$). These features are then passed to a Linear Support Vector Machine \citep{DBLP:journals/ml/CortesV95,DBLP:journals/jmlr/FanCHWL08}, where $C=1$.

\subsection{Experimental Setup}

\begin{table}
    \centering
    \footnotesize
    \begin{tabular}{l|p{8cm}}
        \toprule
        \textsc{data}          & Huang, Emmery, Volkova \\
        \textsc{importance}    & Omission score \\ 
        \textsc{attack}        & Heuristics, TextFooler, Masked BERT, Dropout BERT \\
        \textsc{model}         & Logistic Regression, N-GrAM \\
        \textsc{ranking}       & None, POS + USE, BERT Sim \\
        \bottomrule
    \end{tabular}
    \caption{Grid of possible experimental configurations.}
    \label{tab:exp}
\end{table}

To summarize (and see Table~\ref{tab:exp}), the experiment is conducted as follows: the substitute target model ($f'$)---LR for all experiments---is fit on a given corpus. The real target model ($f$, either LR or N-GrAM) is always fit on the corpus of \cite{DBLP:conf/aaai/VolkovaBAS15}. To evaluate the attacks, a $200$-instance sample is used. Target words are ranked via omission scores from $f'$, and fed to either Heuristics, TF, MB, or DB attacks. The heuristics directly change the target words, the rest outputs a ranked set of replacement candidates. The latter can either be evaluated against $f'$ through the TF pipeline, or the Top-1 candidate is returned. Filtering can be done through POS/USE for semantic similarity and POS compatibility checks (Check), or not (\st{Check}).

Note that we are predominantly interested in transferability, and would therefore like to test as many combinations of data and architecture access limitations as possible. If we assume an author does not have access to the data, the substitute classifier is trained on anything but the \citeauthor{DBLP:conf/aaai/VolkovaBAS15} corpus. If we assume the author does not know the target model architecture, the target model is N-GrAM (rather than LR). A full model transfer setting (both data and architecture) will therefore be, e.g.: data $f'$ = \citeauthor{emmery-etal-2017-simple}, data $f$ = \citeauthor{volkova-etal-2014-inferring}, $f'$ = LR, and $f$ = NGrAM.  Finally, for comparison to an optimal situation, we test a setting with access to the adversary's data.

\subsection{Evaluation}

\begin{table}[t!]
    \centering
    \footnotesize
    \begin{tabular}{@{\extracolsep{\fill}}r|r|cc|cc||cc}
        \toprule
        & & \multicolumn{6}{c}{\textbf{test} = \citeauthor{DBLP:conf/aaai/VolkovaBAS15} } \\
        \cmidrule(lr){3-8}
        & LR $\ f'$ $\rightarrow$ & \multicolumn{2}{c}{\citeauthor{huang-etal-2020-multilingual}} &
          \multicolumn{2}{c}{\citeauthor{emmery-etal-2017-simple}} &
          \multicolumn{2}{c}{\citeauthor{DBLP:conf/aaai/VolkovaBAS15}} \\
        \cmidrule(lr){3-4} \cmidrule(lr){5-6} \cmidrule(lr){7-8}
        & $f\ $ $\rightarrow$  & LR                 & NG            & LR              & NG       & LR            & NG  \\
        \midrule   
        
        & none                  &  $.885$              & $.940$          & $.885$           & $.940$         & $.885$          & $.940$    \\
        \cmidrule{1-8}
        \multirow{3}{*}{\rotatebox{90}{\scriptsize Heuristic}}
        & 1337                  &  $.770$              & $.850$          & $.775$           & $.835$         & $.715$          & $.860$    \\
        & flip                  &  $.900$              & $.950$          & $.885$           & $.905$         & $.840$          & $.905$    \\
        & space                 &  $.845$              & $.925$          & $.760$           & $.870$         & $.720$          & $.850$    \\
        \cmidrule{1-8}
        \multirow{3}{*}{\rotatebox{90}{\scriptsize Top-1}}
        & WS                    &  $.825$              & $.930$          & $.805$           & $.890$         & $.750$          & $.915$    \\
        & MB                    &  $.655$              & $.905$          & $.595$           & $.785$         & $.145$          & $.410$    \\
        & DB                    &  $.625$              & $.895$          & $.575$           & $.785$         & $.210$          & $.530$    \\
        \cmidrule{1-8}
        \multirow{3}{*}{\rotatebox{90}{\scriptsize\st{Check}}}
        & WS               &  $.540$              & $.855$          & $.355 $          & $.670$          & $\purp{.000}$  & $\purp{.009}$ \\          
        &  MB              &  $\purp{.415}$     & $.790$          & $\purp{.120}$  & $\purp{.420}$ & $\purp{.000}$  & $.085$    \\          
        &  DB              &  $.430$              & $\purp{.775}$ & $.175$           & $.430$          & $\purp{.000}$  & $.085$    \\          
        \cmidrule{1-8}
        \multirow{3}{*}{\rotatebox{90}{\scriptsize Check}}
        & TF                    &  $.705$              & $.920$           & $.780$          & $.910$          & $.375$          & $.700$    \\
        & TF + MB               &  $.640$              & $.880$           & $.760$          & $.890$          & $.380$          & $.725$    \\          
        & TF + DB               &  $.650$              & $.885$           & $.755$          & $.890$          & $.435$          & $.715$    \\          
        \bottomrule
    \end{tabular}
    \caption{Post-attack accuracy scores (below chance ($55\%$) = better) of $f$ on a test sample from the \citeauthor{DBLP:conf/aaai/VolkovaBAS15} corpus. Left, the attack conditions: heuristics, top-1 synonym, applying POS and USE similarity checks, or not applying those checks (\st{Check}). Splits per training corpus are noted for $f^\prime$ (always Logistic Regression (LR)). As target model, either LR, or N-GrAM (NG) was used. The substitution attacks are TextFooler (TF), Masked (MB) and Dropout BERT (DB). If TF's stopping criterion was used, TF + is noted. Word Similarity (WS), reflects the TF pipeline without checks.}
    \label{tab:results}
\end{table}

\paragraph{Metrics} 

The obfuscation success is measured as any accuracy score below chance level performance, which given our test sample is $55\%$. We would argue that random performance is preferred in scenarios where the prediction of the opposite label is undesired. For the current task, however, any accuracy drop to around or lower than chance level satisfies the conditions for successful obfuscation.\footnote{If an attack drops accuracy to $0\%$, this effectively flips (in case of a binary label) the label. This label might \emph{also} be undesired by the author (e.g., being classified as having polar opposite political views). This implies the target model being maximally unsure about the classification is desirable (see also Chapter~\ref{chap:textobf}).} To evaluate the semantic preservation of the attacked sentences, we calculate both \textsc{meteor} \citep{banerjee-lavie-2005-meteor,DBLP:journals/mt/LavieD09} using \texttt{nltk},\footnote{\url{https://www.nltk.org/_modules/nltk/translate/meteor_score.html}} and BERTScore \citep{zhang2019bertscore} between $D$ and $D_\textsc{adv}$. \textsc{meteor} captures flexible uni-gram token overlap including morphological variants, and BERTScore calculates similarities with respect to the sentence context.

\begin{table}
    \centering
    \footnotesize
    \begin{tabular}{l|l|rr}
    \toprule
    & \citeauthor{DBLP:conf/aaai/VolkovaBAS15} $\rightarrow$    &    \textsc{train} & \textsc{test} \\
    \midrule
    \multirow{2}{*}{\textsc{train}}
    & \citeauthor{huang-etal-2020-multilingual}                 &  $.640$            & $.620$         \\ [0.2cm]
    & \citeauthor{emmery-etal-2017-simple}                      &  $.725$            & $.890$         \\
    \bottomrule
    \end{tabular}
    \caption{Gender prediction accuracy of the substitute models $f^\prime$ on data splits of $f$.} 
    \label{tab:dom}
\end{table}

\paragraph{Human Evaluation} 

For the human evaluation, we sampled $20$ document pieces (one or more tweets) for each attack type in the best performing experimental configuration. A piece was chosen if it satisfied these criteria: i) contains changes for all three attacks, ii) consists of at least $15$ words (excluding emojis and tags), and iii) does not contain obvious profanity.\footnote{To avoid exposing the raters to overly toxic content, blatant examples were filtered using a keyword list. Some minor examples remained, for which we added a disclaimer.} All $60$ document pieces of the three models were shuffled, and the $20$ original versions were appended at the end (so that `correct' pieces were seen last). Each substitute model therefore has $80$ items for evaluation.

While in prior work it is common to rate semantic consistency, fluency, and label a text \cite[see e.g.,][]{DBLP:conf/clef/PotthastHS16,DBLP:conf/aaai/JinJZS20}, our Twitter data are too noisy (including many spelling and grammar errors in the originals), and document batches too long to make this a feasible task. Instead, our six participants (three per substitute) were asked to indicate if: a) a sentence was artificially changed, and if so, b) indicate one word that raised their suspicion. This way, we can evaluate which attack produces the most natural sentences, and the least obvious changes to the input. 

The items were rated individually; the human evaluators did not know beforehand that different versions of the same sentences were repeated, nor that the originals were shown at the end. All participants have a university-level education, a high English proficiency, and are familiar with the domain of the data. Several example ratings of the same sentence can be found in Table~\ref{tab:example}.

\section{Results}

\subsection{Domain Shift} As alluded to in Section~\ref{subs:data-adv}, both corpora used to train our substitute models were in fact not reference corpora for author profiling, and can therefore be considered as suboptimal, disjoint domains. The \citeauthor{huang-etal-2020-multilingual} corpus in particular shows a strong domain shift (see Table~\ref{tab:dom}) for both training and test sets. The distantly labeled \citeauthor{emmery-etal-2017-simple} corpus achieves $7.5\%$ more accuracy on the train split of \citeauthor{DBLP:conf/aaai/VolkovaBAS15}, and test performance is significantly higher ($27\%$). We might therefore expect better obfuscation performance from the latter.

\subsection{Baselines} The results for all attacks are shown in Table~\ref{tab:results}. Note that these are performances for $f$; therefore, when no attacks are applied (none), the performance for both substitute corpora stays the same (as those only influence the attacks). For the heuristic attacks, 1337 seems to make the more robust baseline; outperforming some of the other settings---even on transferability. A surface-level advantage is that this attack has a minor impact on readability (when applied conservatively) and does not change semantics; however, the heuristic attacks are fairly simple to mitigate in preprocessing \citep{DBLP:conf/ifip11-9/JuolaV11} and through character features (as shown by the performance of the N-GrAM model). For transferability, we evidently need to do more than trying to convert words to be out-of-vocabulary (OOV) with noise. While it can be argued the heuristics could change all words, shifting everything OOV would not be robust; the target model side could easily spot the anomalous input and act (e.g., reject) accordingly.

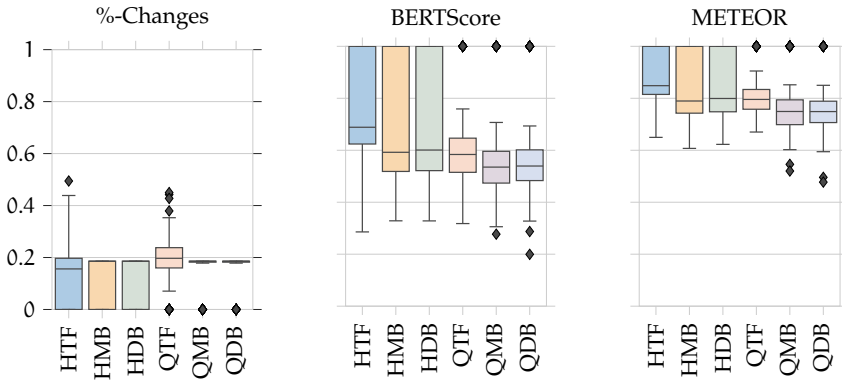
\begin{figure}
    \centering
    \begin{minipage}{.32\textwidth}
\begin{tikzpicture}

\definecolor{color2}{cmyk}{0.08,0.00,0.09,0.09}
\definecolor{color1}{cmyk}{0,0.14,0.31,0.02}
\definecolor{colorx}{cmyk}{0,0.44,0.98,0.29}
\definecolor{color3}{cmyk}{0,0.13,0.15,0.02}
\definecolor{DrawRed}{cmyk}{0,0.63,0.71,0.32}
\definecolor{colorx}{cmyk}{0.23,0.01,0.00,0.10}
\definecolor{DrawAqua}{cmyk}{0.90,0.06,0.00,0.47}
\definecolor{color0}{cmyk}{0.26,0.08,0.00,0.06}
\definecolor{DrawBlue}{cmyk}{0.90,0.27,0.00,0.38}
\definecolor{colorx}{cmyk}{0.08,0.08,0.00,0.11}
\definecolor{DrawPurple}{cmyk}{0.32,0.36,0.00,0.51}
\definecolor{colorx}{cmyk}{0.12,0.05,0.00,0.17}
\definecolor{DrawMarine}{cmyk}{0.62,0.27,0.00,0.64}
\definecolor{colorx}{cmyk}{0.43,0.26,0.00,0.25}
\definecolor{color5}{cmyk}{0.13,0.08,0.00,0.01}
\definecolor{color4}{cmyk}{0.03,0.08,0.00,0.09}
\definecolor{colorx}{cmyk}{0.10,0.31,0.00,0.35}
\definecolor{colorx}{cmyk}{0.00,0.05,0.20,0.00}
\definecolor{DrawGold}{cmyk}{0.00,0.15,0.60,0.16}

\begin{axis}[
title=\%-Changes,
axis line style={color=white!80!black},
tick align=outside,
x grid style={},
xmajorticks=true,
xmin=-0.5, xmax=5.5,
xtick style={color=white!80!black},
xtick={0,1,2,3,4,5},
xticklabels={HTF,HMB,HDB,QTF,QMB,QDB},
y grid style={color=white!80!black},
ymajorgrids,
ymajorticks=true,
ymin=0, ymax=1,
ytick style={color=white!15!black},
xticklabel style={rotate=90},
width=5cm,
height=6cm
]
\path [draw=white!28.6274509803922!black, fill=color0, semithick]
(axis cs:-0.4,0)
--(axis cs:0.4,0)
--(axis cs:0.4,0.197026022304833)
--(axis cs:-0.4,0.197026022304833)
--(axis cs:-0.4,0)
--cycle;
\path [draw=white!28.6274509803922!black, fill=color1, semithick]
(axis cs:0.6,0)
--(axis cs:1.4,0)
--(axis cs:1.4,0.185873605947955)
--(axis cs:0.6,0.185873605947955)
--(axis cs:0.6,0)
--cycle;
\path [draw=white!28.6274509803922!black, fill=color2, semithick]
(axis cs:1.6,0)
--(axis cs:2.4,0)
--(axis cs:2.4,0.185873605947955)
--(axis cs:1.6,0.185873605947955)
--(axis cs:1.6,0)
--cycle;
\path [draw=white!28.6274509803922!black, fill=color3, semithick]
(axis cs:2.6,0.159851301115242)
--(axis cs:3.4,0.159851301115242)
--(axis cs:3.4,0.237918215613383)
--(axis cs:2.6,0.237918215613383)
--(axis cs:2.6,0.159851301115242)
--cycle;
\path [draw=white!28.6274509803922!black, fill=color4, semithick]
(axis cs:3.6,0.182156133828996)
--(axis cs:4.4,0.182156133828996)
--(axis cs:4.4,0.185873605947955)
--(axis cs:3.6,0.185873605947955)
--(axis cs:3.6,0.182156133828996)
--cycle;
\path [draw=white!28.6274509803922!black, fill=color5, semithick]
(axis cs:4.6,0.182156133828996)
--(axis cs:5.4,0.182156133828996)
--(axis cs:5.4,0.185873605947955)
--(axis cs:4.6,0.185873605947955)
--(axis cs:4.6,0.182156133828996)
--cycle;
\addplot [semithick, white!28.6274509803922!black]
table {%
0 0
0 0
};
\addplot [semithick, white!28.6274509803922!black]
table {%
0 0.197026022304833
0 0.438661710037175
};
\addplot [semithick, white!28.6274509803922!black]
table {%
-0.2 0
0.2 0
};
\addplot [semithick, white!28.6274509803922!black]
table {%
-0.2 0.438661710037175
0.2 0.438661710037175
};
\addplot [black, mark=diamond*, mark size=2.5, mark options={solid,fill=white!28.6274509803922!black}, only marks]
table {%
0 0.494423791821561
};
\addplot [semithick, white!28.6274509803922!black]
table {%
1 0
1 0
};
\addplot [semithick, white!28.6274509803922!black]
table {%
1 0.185873605947955
1 0.185873605947955
};
\addplot [semithick, white!28.6274509803922!black]
table {%
0.8 0
1.2 0
};
\addplot [semithick, white!28.6274509803922!black]
table {%
0.8 0.185873605947955
1.2 0.185873605947955
};
\addplot [semithick, white!28.6274509803922!black]
table {%
2 0
2 0
};
\addplot [semithick, white!28.6274509803922!black]
table {%
2 0.185873605947955
2 0.185873605947955
};
\addplot [semithick, white!28.6274509803922!black]
table {%
1.8 0
2.2 0
};
\addplot [semithick, white!28.6274509803922!black]
table {%
1.8 0.185873605947955
2.2 0.185873605947955
};
\addplot [semithick, white!28.6274509803922!black]
table {%
3 0.159851301115242
3 0.0706319702602231
};
\addplot [semithick, white!28.6274509803922!black]
table {%
3 0.237918215613383
3 0.353159851301115
};
\addplot [semithick, white!28.6274509803922!black]
table {%
2.8 0.0706319702602231
3.2 0.0706319702602231
};
\addplot [semithick, white!28.6274509803922!black]
table {%
2.8 0.353159851301115
3.2 0.353159851301115
};
\addplot [black, mark=diamond*, mark size=2.5, mark options={solid,fill=white!28.6274509803922!black}, only marks]
table {%
3 0
3 0
3 0
3 0
3 0
3 0
3 0
3 0
3 0
3 0
3 0
3 0
3 0
3 0
3 0
3 0
3 0
3 0
3 0
3 0
3 0
3 0
3 0.442379182156134
3 0.449814126394052
3 0.379182156133829
3 0.427509293680297
};
\addplot [semithick, white!28.6274509803922!black]
table {%
4 0.182156133828996
4 0.178438661710037
};
\addplot [semithick, white!28.6274509803922!black]
table {%
4 0.185873605947955
4 0.185873605947955
};
\addplot [semithick, white!28.6274509803922!black]
table {%
3.8 0.178438661710037
4.2 0.178438661710037
};
\addplot [semithick, white!28.6274509803922!black]
table {%
3.8 0.185873605947955
4.2 0.185873605947955
};
\addplot [black, mark=diamond*, mark size=2.5, mark options={solid,fill=white!28.6274509803922!black}, only marks]
table {%
4 0
4 0
4 0
4 0
4 0
4 0
4 0
4 0
4 0
4 0
4 0
4 0
4 0
4 0
4 0
4 0
4 0
4 0
4 0
4 0
4 0
4 0
};
\addplot [semithick, white!28.6274509803922!black]
table {%
5 0.182156133828996
5 0.178438661710037
};
\addplot [semithick, white!28.6274509803922!black]
table {%
5 0.185873605947955
5 0.185873605947955
};
\addplot [semithick, white!28.6274509803922!black]
table {%
4.8 0.178438661710037
5.2 0.178438661710037
};
\addplot [semithick, white!28.6274509803922!black]
table {%
4.8 0.185873605947955
5.2 0.185873605947955
};
\addplot [black, mark=diamond*, mark size=2.5, mark options={solid,fill=white!28.6274509803922!black}, only marks]
table {%
5 0
5 0
5 0
5 0
5 0
5 0
5 0
5 0
5 0
5 0
5 0
5 0
5 0
5 0
5 0
5 0
5 0
5 0
5 0
5 0
5 0
5 0
};
\addplot [semithick, white!28.6274509803922!black]
table {%
-0.4 0.156133828996283
0.4 0.156133828996283
};
\addplot [semithick, white!28.6274509803922!black]
table {%
0.6 0.185873605947955
1.4 0.185873605947955
};
\addplot [semithick, white!28.6274509803922!black]
table {%
1.6 0.185873605947955
2.4 0.185873605947955
};
\addplot [semithick, white!28.6274509803922!black]
table {%
2.6 0.197026022304833
3.4 0.197026022304833
};
\addplot [semithick, white!28.6274509803922!black]
table {%
3.6 0.185873605947955
4.4 0.185873605947955
};
\addplot [semithick, white!28.6274509803922!black]
table {%
4.6 0.185873605947955
5.4 0.185873605947955
};
\end{axis}
\end{tikzpicture}
    \end{minipage}
    \begin{minipage}{.32\textwidth}
\begin{tikzpicture}

\definecolor{color2}{cmyk}{0.08,0.00,0.09,0.09}
\definecolor{color1}{cmyk}{0,0.14,0.31,0.02}
\definecolor{colorx}{cmyk}{0,0.44,0.98,0.29}
\definecolor{color3}{cmyk}{0,0.13,0.15,0.02}
\definecolor{DrawRed}{cmyk}{0,0.63,0.71,0.32}
\definecolor{colorx}{cmyk}{0.23,0.01,0.00,0.10}
\definecolor{DrawAqua}{cmyk}{0.90,0.06,0.00,0.47}
\definecolor{color0}{cmyk}{0.26,0.08,0.00,0.06}
\definecolor{DrawBlue}{cmyk}{0.90,0.27,0.00,0.38}
\definecolor{colorx}{cmyk}{0.08,0.08,0.00,0.11}
\definecolor{DrawPurple}{cmyk}{0.32,0.36,0.00,0.51}
\definecolor{colorx}{cmyk}{0.12,0.05,0.00,0.17}
\definecolor{DrawMarine}{cmyk}{0.62,0.27,0.00,0.64}
\definecolor{colorx}{cmyk}{0.43,0.26,0.00,0.25}
\definecolor{color5}{cmyk}{0.13,0.08,0.00,0.01}
\definecolor{color4}{cmyk}{0.03,0.08,0.00,0.09}
\definecolor{colorx}{cmyk}{0.10,0.31,0.00,0.35}
\definecolor{colorx}{cmyk}{0.00,0.05,0.20,0.00}
\definecolor{DrawGold}{cmyk}{0.00,0.15,0.60,0.16}


\begin{axis}[
title=BERTScore,
axis line style={color=white!80!black},
tick align=outside,
x grid style={white!80!black},
xmajorticks=true,
xmin=-0.5, xmax=5.5,
xtick style={color=white!80!black},
xtick={0,1,2,3,4,5},
xticklabels={HTF,HMB,HDB,QTF,QMB,QDB},
y grid style={white!80!black},
ymajorgrids,
ymajorticks=true,
ymin=0, ymax=1,
ytick style={color=white!80!black},
yticklabel style={color=white},
xticklabel style={rotate=90},
width=5cm,
height=6cm
]
\path [draw=white!28.6274509803922!black, fill=color0, semithick]
(axis cs:-0.4,0.624078780412674)
--(axis cs:0.4,0.624078780412674)
--(axis cs:0.4,0.999999642372131)
--(axis cs:-0.4,0.999999642372131)
--(axis cs:-0.4,0.624078780412674)
--cycle;
\path [draw=white!28.6274509803922!black, fill=color1, semithick]
(axis cs:0.6,0.518495783209801)
--(axis cs:1.4,0.518495783209801)
--(axis cs:1.4,0.999999642372131)
--(axis cs:0.6,0.999999642372131)
--(axis cs:0.6,0.518495783209801)
--cycle;
\path [draw=white!28.6274509803922!black, fill=color2, semithick]
(axis cs:1.6,0.52160918712616)
--(axis cs:2.4,0.52160918712616)
--(axis cs:2.4,0.999999642372131)
--(axis cs:1.6,0.999999642372131)
--(axis cs:1.6,0.52160918712616)
--cycle;
\path [draw=white!28.6274509803922!black, fill=color3, semithick]
(axis cs:2.6,0.515078783035278)
--(axis cs:3.4,0.515078783035278)
--(axis cs:3.4,0.646562561392784)
--(axis cs:2.6,0.646562561392784)
--(axis cs:2.6,0.515078783035278)
--cycle;
\path [draw=white!28.6274509803922!black, fill=color4, semithick]
(axis cs:3.6,0.473697461187839)
--(axis cs:4.4,0.473697461187839)
--(axis cs:4.4,0.595968931913376)
--(axis cs:3.6,0.595968931913376)
--(axis cs:3.6,0.473697461187839)
--cycle;
\path [draw=white!28.6274509803922!black, fill=color5, semithick]
(axis cs:4.6,0.483360603451729)
--(axis cs:5.4,0.483360603451729)
--(axis cs:5.4,0.601557895541191)
--(axis cs:4.6,0.601557895541191)
--(axis cs:4.6,0.483360603451729)
--cycle;
\addplot [semithick, white!28.6274509803922!black]
table {%
0 0.624078780412674
0 0.28577795624733
};
\addplot [semithick, white!28.6274509803922!black]
table {%
0 0.999999642372131
0 1.00000071525574
};
\addplot [semithick, white!28.6274509803922!black]
table {%
-0.2 0.28577795624733
0.2 0.28577795624733
};
\addplot [semithick, white!28.6274509803922!black]
table {%
-0.2 1.00000071525574
0.2 1.00000071525574
};
\addplot [semithick, white!28.6274509803922!black]
table {%
1 0.518495783209801
1 0.328468173742294
};
\addplot [semithick, white!28.6274509803922!black]
table {%
1 0.999999642372131
1 1.00000071525574
};
\addplot [semithick, white!28.6274509803922!black]
table {%
0.8 0.328468173742294
1.2 0.328468173742294
};
\addplot [semithick, white!28.6274509803922!black]
table {%
0.8 1.00000071525574
1.2 1.00000071525574
};
\addplot [semithick, white!28.6274509803922!black]
table {%
2 0.52160918712616
2 0.328165858983994
};
\addplot [semithick, white!28.6274509803922!black]
table {%
2 0.999999642372131
2 1.00000071525574
};
\addplot [semithick, white!28.6274509803922!black]
table {%
1.8 0.328165858983994
2.2 0.328165858983994
};
\addplot [semithick, white!28.6274509803922!black]
table {%
1.8 1.00000071525574
2.2 1.00000071525574
};
\addplot [semithick, white!28.6274509803922!black]
table {%
3 0.515078783035278
3 0.317894518375397
};
\addplot [semithick, white!28.6274509803922!black]
table {%
3 0.646562561392784
3 0.759037971496582
};
\addplot [semithick, white!28.6274509803922!black]
table {%
2.8 0.317894518375397
3.2 0.317894518375397
};
\addplot [semithick, white!28.6274509803922!black]
table {%
2.8 0.759037971496582
3.2 0.759037971496582
};
\addplot [black, mark=diamond*, mark size=2.5, mark options={solid,fill=white!28.6274509803922!black}, only marks]
table {%
3 1
3 0.999999642372131
3 1
3 1
3 1
3 1
3 0.999999284744263
3 1.00000071525574
3 1
3 0.999999642372131
3 0.999999642372131
3 0.999999642372131
3 1
3 1
3 1
3 0.999999642372131
3 1
3 1
3 1
3 1
3 1
3 0.999999642372131
};
\addplot [semithick, white!28.6274509803922!black]
table {%
4 0.473697461187839
4 0.305772572755814
};
\addplot [semithick, white!28.6274509803922!black]
table {%
4 0.595968931913376
4 0.707036972045898
};
\addplot [semithick, white!28.6274509803922!black]
table {%
3.8 0.305772572755814
4.2 0.305772572755814
};
\addplot [semithick, white!28.6274509803922!black]
table {%
3.8 0.707036972045898
4.2 0.707036972045898
};
\addplot [black, mark=diamond*, mark size=2.5, mark options={solid,fill=white!28.6274509803922!black}, only marks]
table {%
4 0.277099341154099
4 1
4 0.999999642372131
4 1
4 1
4 1
4 1
4 0.999999284744263
4 1.00000071525574
4 1
4 0.999999642372131
4 0.999999642372131
4 0.999999642372131
4 1
4 1
4 1
4 0.999999642372131
4 1
4 1
4 1
4 1
4 1
4 0.999999642372131
};
\addplot [semithick, white!28.6274509803922!black]
table {%
5 0.483360603451729
5 0.327439427375793
};
\addplot [semithick, white!28.6274509803922!black]
table {%
5 0.601557895541191
5 0.693508744239807
};
\addplot [semithick, white!28.6274509803922!black]
table {%
4.8 0.327439427375793
5.2 0.327439427375793
};
\addplot [semithick, white!28.6274509803922!black]
table {%
4.8 0.693508744239807
5.2 0.693508744239807
};
\addplot [black, mark=diamond*, mark size=2.5, mark options={solid,fill=white!28.6274509803922!black}, only marks]
table {%
5 0.287283480167389
5 0.199523091316223
5 1
5 0.999999642372131
5 1
5 1
5 1
5 1
5 0.999999284744263
5 1.00000071525574
5 1
5 0.999999642372131
5 0.999999642372131
5 0.999999642372131
5 1
5 1
5 1
5 0.999999642372131
5 1
5 1
5 1
5 1
5 1
5 0.999999642372131
};
\addplot [semithick, white!28.6274509803922!black]
table {%
-0.4 0.688551753759384
0.4 0.688551753759384
};
\addplot [semithick, white!28.6274509803922!black]
table {%
0.6 0.592202812433243
1.4 0.592202812433243
};
\addplot [semithick, white!28.6274509803922!black]
table {%
1.6 0.600841879844666
2.4 0.600841879844666
};
\addplot [semithick, white!28.6274509803922!black]
table {%
2.6 0.583801299333572
3.4 0.583801299333572
};
\addplot [semithick, white!28.6274509803922!black]
table {%
3.6 0.535189419984818
4.4 0.535189419984818
};
\addplot [semithick, white!28.6274509803922!black]
table {%
4.6 0.539352476596832
5.4 0.539352476596832
};
\end{axis}

\end{tikzpicture}
    \end{minipage}
    \begin{minipage}{.32\textwidth}
\begin{tikzpicture}

\definecolor{color2}{cmyk}{0.08,0.00,0.09,0.09}
\definecolor{color1}{cmyk}{0,0.14,0.31,0.02}
\definecolor{colorx}{cmyk}{0,0.44,0.98,0.29}
\definecolor{color3}{cmyk}{0,0.13,0.15,0.02}
\definecolor{DrawRed}{cmyk}{0,0.63,0.71,0.32}
\definecolor{colorx}{cmyk}{0.23,0.01,0.00,0.10}
\definecolor{DrawAqua}{cmyk}{0.90,0.06,0.00,0.47}
\definecolor{color0}{cmyk}{0.26,0.08,0.00,0.06}
\definecolor{DrawBlue}{cmyk}{0.90,0.27,0.00,0.38}
\definecolor{colorx}{cmyk}{0.08,0.08,0.00,0.11}
\definecolor{DrawPurple}{cmyk}{0.32,0.36,0.00,0.51}
\definecolor{colorx}{cmyk}{0.12,0.05,0.00,0.17}
\definecolor{DrawMarine}{cmyk}{0.62,0.27,0.00,0.64}
\definecolor{colorx}{cmyk}{0.43,0.26,0.00,0.25}
\definecolor{color5}{cmyk}{0.13,0.08,0.00,0.01}
\definecolor{color4}{cmyk}{0.03,0.08,0.00,0.09}
\definecolor{colorx}{cmyk}{0.10,0.31,0.00,0.35}
\definecolor{colorx}{cmyk}{0.00,0.05,0.20,0.00}
\definecolor{DrawGold}{cmyk}{0.00,0.15,0.60,0.16}


\begin{axis}[
title=METEOR,
axis line style={color=white!80!black},
tick align=outside,
x grid style={white!80!black},
xmajorticks=true,
xmin=-0.5, xmax=5.5,
xtick style={color=white!80!black},
xtick={0,1,2,3,4,5},
xticklabels={HTF,HMB,HDB,QTF,QMB,QDB},
y grid style={white!80!black},
ymajorgrids,
ymajorticks=true,
ymin=0, ymax=1,
ytick style={color=white!80!black},
yticklabel style={color=white},
xticklabel style={rotate=90},
width=5cm,
height=6cm
]
\path [draw=white!28.6274509803922!black, fill=color0, semithick]
(axis cs:-0.4,0.814920198658963)
--(axis cs:0.4,0.814920198658963)
--(axis cs:0.4,0.999999929898052)
--(axis cs:-0.4,0.999999929898052)
--(axis cs:-0.4,0.814920198658963)
--cycle;
\path [draw=white!28.6274509803922!black, fill=color1, semithick]
(axis cs:0.6,0.742744943434151)
--(axis cs:1.4,0.742744943434151)
--(axis cs:1.4,0.999999929898052)
--(axis cs:0.6,0.999999929898052)
--(axis cs:0.6,0.742744943434151)
--cycle;
\path [draw=white!28.6274509803922!black, fill=color2, semithick]
(axis cs:1.6,0.74807819469221)
--(axis cs:2.4,0.74807819469221)
--(axis cs:2.4,0.999999929898052)
--(axis cs:1.6,0.999999929898052)
--(axis cs:1.6,0.74807819469221)
--cycle;
\path [draw=white!28.6274509803922!black, fill=color3, semithick]
(axis cs:2.6,0.757864608546063)
--(axis cs:3.4,0.757864608546063)
--(axis cs:3.4,0.833986256308215)
--(axis cs:2.6,0.833986256308215)
--(axis cs:2.6,0.757864608546063)
--cycle;
\path [draw=white!28.6274509803922!black, fill=color4, semithick]
(axis cs:3.6,0.698603566345911)
--(axis cs:4.4,0.698603566345911)
--(axis cs:4.4,0.794058731404118)
--(axis cs:3.6,0.794058731404118)
--(axis cs:3.6,0.698603566345911)
--cycle;
\path [draw=white!28.6274509803922!black, fill=color5, semithick]
(axis cs:4.6,0.706606044771616)
--(axis cs:5.4,0.706606044771616)
--(axis cs:5.4,0.788814857462808)
--(axis cs:4.6,0.788814857462808)
--(axis cs:4.6,0.706606044771616)
--cycle;
\addplot [semithick, white!28.6274509803922!black]
table {%
0 0.814920198658963
0 0.649615041253703
};
\addplot [semithick, white!28.6274509803922!black]
table {%
0 0.999999929898052
0 0.999999987500527
};
\addplot [semithick, white!28.6274509803922!black]
table {%
-0.2 0.649615041253703
0.2 0.649615041253703
};
\addplot [semithick, white!28.6274509803922!black]
table {%
-0.2 0.999999987500527
0.2 0.999999987500527
};
\addplot [semithick, white!28.6274509803922!black]
table {%
1 0.742744943434151
1 0.607159365542696
};
\addplot [semithick, white!28.6274509803922!black]
table {%
1 0.999999929898052
1 0.999999987500527
};
\addplot [semithick, white!28.6274509803922!black]
table {%
0.8 0.607159365542696
1.2 0.607159365542696
};
\addplot [semithick, white!28.6274509803922!black]
table {%
0.8 0.999999987500527
1.2 0.999999987500527
};
\addplot [semithick, white!28.6274509803922!black]
table {%
2 0.74807819469221
2 0.622660473921631
};
\addplot [semithick, white!28.6274509803922!black]
table {%
2 0.999999929898052
2 0.999999987500527
};
\addplot [semithick, white!28.6274509803922!black]
table {%
1.8 0.622660473921631
2.2 0.622660473921631
};
\addplot [semithick, white!28.6274509803922!black]
table {%
1.8 0.999999987500527
2.2 0.999999987500527
};
\addplot [semithick, white!28.6274509803922!black]
table {%
3 0.757864608546063
3 0.670170181870418
};
\addplot [semithick, white!28.6274509803922!black]
table {%
3 0.833986256308215
3 0.905043792546014
};
\addplot [semithick, white!28.6274509803922!black]
table {%
2.8 0.670170181870418
3.2 0.670170181870418
};
\addplot [semithick, white!28.6274509803922!black]
table {%
2.8 0.905043792546014
3.2 0.905043792546014
};
\addplot [black, mark=diamond*, mark size=2.5, mark options={solid,fill=white!28.6274509803922!black}, only marks]
table {%
3 0.999999976218308
3 0.999999953042825
3 0.999999969488128
3 0.999999967612907
3 0.999999980324399
3 0.999999974597368
3 0.999999987165738
3 0.999999938428203
3 0.999999933594846
3 0.999999832551012
3 0.999999890693562
3 0.999999958905236
3 0.999999813823182
3 0.999999972198634
3 0.999999911343619
3 0.999999908286767
3 0.999999813823182
3 0.999999984741211
3 0.99999988664558
3 0.999999971552116
3 0.99999992824207
3 0.99999996097684
};
\addplot [semithick, white!28.6274509803922!black]
table {%
4 0.698603566345911
4 0.602273014128138
};
\addplot [semithick, white!28.6274509803922!black]
table {%
4 0.794058731404118
4 0.851927271883412
};
\addplot [semithick, white!28.6274509803922!black]
table {%
3.8 0.602273014128138
4.2 0.602273014128138
};
\addplot [semithick, white!28.6274509803922!black]
table {%
3.8 0.851927271883412
4.2 0.851927271883412
};
\addplot [black, mark=diamond*, mark size=2.5, mark options={solid,fill=white!28.6274509803922!black}, only marks]
table {%
4 0.54576043642859
4 0.520066938389673
4 0.999999976218308
4 0.999999953042825
4 0.999999969488128
4 0.999999967612907
4 0.999999980324399
4 0.999999974597368
4 0.999999987165738
4 0.999999938428203
4 0.999999933594846
4 0.999999832551012
4 0.999999890693562
4 0.999999958905236
4 0.999999813823182
4 0.999999972198634
4 0.999999911343619
4 0.999999908286767
4 0.999999813823182
4 0.999999984741211
4 0.99999988664558
4 0.999999971552116
4 0.99999992824207
4 0.99999996097684
};
\addplot [semithick, white!28.6274509803922!black]
table {%
5 0.706606044771616
5 0.594456658100074
};
\addplot [semithick, white!28.6274509803922!black]
table {%
5 0.788814857462808
5 0.849977852659941
};
\addplot [semithick, white!28.6274509803922!black]
table {%
4.8 0.594456658100074
5.2 0.594456658100074
};
\addplot [semithick, white!28.6274509803922!black]
table {%
4.8 0.849977852659941
5.2 0.849977852659941
};
\addplot [black, mark=diamond*, mark size=2.5, mark options={solid,fill=white!28.6274509803922!black}, only marks]
table {%
5 0.495800670976189
5 0.477415966386555
5 0.999999976218308
5 0.999999953042825
5 0.999999969488128
5 0.999999967612907
5 0.999999980324399
5 0.999999974597368
5 0.999999987165738
5 0.999999938428203
5 0.999999933594846
5 0.999999832551012
5 0.999999890693562
5 0.999999958905236
5 0.999999813823182
5 0.999999972198634
5 0.999999911343619
5 0.999999908286767
5 0.999999813823182
5 0.999999984741211
5 0.99999988664558
5 0.999999971552116
5 0.99999992824207
5 0.99999996097684
};
\addplot [semithick, white!28.6274509803922!black]
table {%
-0.4 0.848576299130433
0.4 0.848576299130433
};
\addplot [semithick, white!28.6274509803922!black]
table {%
0.6 0.789583062045209
1.4 0.789583062045209
};
\addplot [semithick, white!28.6274509803922!black]
table {%
1.6 0.79930494938492
2.4 0.79930494938492
};
\addplot [semithick, white!28.6274509803922!black]
table {%
2.6 0.795689115113054
3.4 0.795689115113054
};
\addplot [semithick, white!28.6274509803922!black]
table {%
3.6 0.749250945597902
4.4 0.749250945597902
};
\addplot [semithick, white!28.6274509803922!black]
table {%
4.6 0.748900913942554
5.4 0.748900913942554
};
\end{axis}

\end{tikzpicture}
    \end{minipage}
      \captionof{figure}{Rewrite metrics for the three different attacks: TextFooler (TF), Masked (MB) and Dropout BERT (DB)---split by corpora \citeauthor{huang-etal-2020-multilingual} (H), \citeauthor{emmery-etal-2017-simple} (Q). Shown are the relative number of changes (by document length), BERTScore ($F_1$), and \textsc{meteor} (MET) with respect to the original document. 
      }
    \label{fig:eval}
\end{figure}

\subsection{Attack Transferability} Transferability can be assessed by comparing the LR and N-GrAM (NG) columns. Globally it can be observed that the substitute models trained on the \citeauthor{emmery-etal-2017-simple} corpus systematically outperform those trained on \citeauthor{huang-etal-2020-multilingual}; both for the settings where the adversary's architecture is known (LR), and where it is unknown (NG). This matches our expectations from the observed domain shift. Our results also show that a noticeable decrease in obfuscation performance occurs ($10-30\%$ increased target model performance) when the attacks are transferred to different data and another model. In contrast, as can be observed from the last two columns in Table~\ref{tab:results}, in a practically unrealistic setting where the model and data are available, the obfuscation is specifically tailored to known weaknesses and therefore highly effective.\footnote{Jin et al. (2020) found similar drops to $0\%$ accuracy with a comparable percentage of changes for word-level models.}

\subsection{Transformer Performance} Looking at the Top-1, \st{Check} and Check brackets (Table~\ref{tab:results}), other than the BERT-based models having higher success of transferability than TF, they also retain obfuscation success; deteriorating the target model's performance to lower than chance level ($55\%$) for the settings not using additional checks. This comparison also demonstrates the synonym ranking to work (Top-1 vs. \st{Check} and Check), and the Check condition to be too restrictive; attaining lower attack power, and low transferability. This is further illustrated by the \%-changes shown in Figure~\ref{fig:eval}. Comparing the MB and DB variants, their performance seems almost identical, with masking having a slight advantage. As \cite{zhou-etal-2019-bert} argued, applying dropout should produce words that are closer to the original (compared to MB), which might affect obfuscation performance. Additionally, the BERT similarity ranking (described in Section~\ref{p:bertsim}) applied to the Masked substitution candidates could have some beneficial effect. This will have to be studied in more detail using the output evaluations.

\paragraph{Rewrite Metrics} 

\begin{table}
    \centering
    \footnotesize
    \begin{tabular}{l|r|rrr|rrr}
    \toprule
    & & \multicolumn{3}{c}{\citeauthor{huang-etal-2020-multilingual}} & \multicolumn{3}{c}{\citeauthor{emmery-etal-2017-simple}} \\
    \cmidrule(lr){3-5} \cmidrule(lr){6-8}
                     &  ORG & TF   & MB   & DB   & TF   & MB   & DB  \\
    \midrule
    \textsc{altered} & $.888$ & $.967$ & $.633$ & $.783$ & $.950$ & \purp{$.617$} & $.633$  \\
    \textsc{word}    & $  - $ & $.950$ & $.583$ & $.700$ & $.867$ & \purp{$.433$} & \purp{$.433$}  \\
    \bottomrule
    \end{tabular}
    \caption{Human accuracy scores of predicting if a text was altered, and guessing the attacked word (lower is better). All substitute models are those with the \st{Check} setting, trained on different corpora (i.e., different words are attacked per training corpus). ORG indicates correct prediction of the originals.}
    \label{tab:huval}
\end{table}

\begin{table}[t!]
    \def\arraystretch{1.4}
    \footnotesize
    \begin{tabular}{p{0.8cm}|p{10cm}}
        \textsc{org} & ready to go home already . a better relationship with god \includegraphics[width=0.7em]{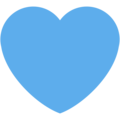} \includegraphics[width=0.7em]{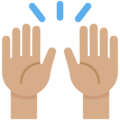} . i need another job asap . \\
        \textsc{htf} & \textbf{loan} to go \textbf{houses} already . a \textbf{improved} \textbf{relations} with \textbf{jesus} \includegraphics[width=0.7em]{chapters/2-adversarial-stylometry/gfx/emoji-heart.png} \includegraphics[width=0.7em]{chapters/2-adversarial-stylometry/gfx/emoji-hands.png} . i \textbf{should} another \textbf{labour} asap . \\
        \textsc{hmb} & ready to go \textbf{on} already . a better relationship with god \includegraphics[width=0.7em]{chapters/2-adversarial-stylometry/gfx/emoji-heart.png} \includegraphics[width=0.7em]{chapters/2-adversarial-stylometry/gfx/emoji-hands.png} . i need another \textbf{guy} \textbf{man} . \\
        \textsc{hdb} & ready to go \textbf{somewhere} already . a better relationship with god \includegraphics[width=0.7em]{chapters/2-adversarial-stylometry/gfx/emoji-heart.png} \includegraphics[width=0.7em]{chapters/2-adversarial-stylometry/gfx/emoji-hands.png} . i need another \textbf{position} \textbf{vs} . \\ [0.2cm]
        \textsc{org} & trump criticizes kim jong un after missile launch : ‘ does this guy have anything better to do ? ’ . \\
        \textsc{htf} & \textbf{tramp criticized kam yung jt} after \textbf{rocket start} : ‘ does this \textbf{boyfriend} have anything \textbf{best} to do ? ’ . \\
        \textsc{hmb} & trump criticizes \textbf{ha woman congressman} after \textbf{campaign} launch : ‘ does this \textbf{book} have anything \textbf{else} to do ? ’ . \\
        \textsc{hdb} & trump criticizes \textbf{in} \textbf{at} \textbf{sin} after \textbf{bomb} launch : ‘ does this \textbf{kid} have anything \textbf{less} to do ? ’ . \\
    \end{tabular}
    \caption{Example ratings of different attacks (not shown together to the human evaluators) on two sentences with varying semantic consitency and human detection accuracy. In the first example, \textsc{hmb} was marked unaltered by all raters, \textsc{hdb} by the majority, and \textsc{htf} by none. In the second, only \textsc{hdb} was marked unaltered, by only one rater. Attacked words are marked in bold, guessing any one of these would count as correctly identifying the attack.}
    \label{tab:example}
\end{table}

The metrics in Figure~\ref{fig:eval} show a common initial limitation in their application to this task: the more frequent an attack makes no changes, the higher the automatic evaluation metrics (BERTScore, \textsc{meteor}). Hence, to compare models, these scores need to be considered in light of the obfuscation performance, and related work. It can be observed that with consistently higher changes, MB and DB score lower on semantic consistency than TF. However, between MB and DB, and TF for the \citeauthor{emmery-etal-2017-simple} corpus, these differences are minor. Furthermore, despite being fit on a different domain, these scores are comparable to prior obfuscation work (e.g., \cite{DBLP:conf/uss/ShettySF18} show \textsc{meteor} scores between $0.69$ and $0.79$).

\paragraph{Human Evaluation}

The results in Table~\ref{tab:huval} reflect the same trend that can be observed in Table~\ref{tab:results}; high obfuscation success seems to result in higher human error when predicting if a sentence was obfuscated. Conversely, it seems that despite higher semantic consistency scores, the original TF pipeline is easier to detect. This can be attributed to the number of spelling and grammar errors the model makes without its additional checks. Furthermore, the $11\%$ error in identifying the original sentences also reflects some expected margin of error in this task, as our Twitter data is inherently noisy. Finally, while these results are in line with the obfuscation success, and are lower than detectability scores in related work \citep{mahmood-etal-2020-girl}, they also indicate that the models are still detectable above chance-level. Given three alternatives (including the original), performance should be $25\%$ or lower to indicate no intrusive changes are made to text (that are not semantically coherent or not inconspicuous enough---both metrics used by \citeauthor{DBLP:conf/clef/PotthastHS16}, \citeyear{DBLP:conf/clef/PotthastHS16}). Therefore, while the presented approaches are effective, and realistically transferable, there is room for improvement for practical applicability.

\section{Discussion and Future Work}

We have demonstrated the performance of author attribute obfuscation under a realistic setting. Using a simple Logistic Regression model for candidate suggestion, trained on a weakly labeled corpus collected in under a day, the attacks successfully transferred to different data and architectures. This is a promising result for future adversarial work on this task, and its practical implementation.

It remains challenging to automatically evaluate how invasive the required number of changes are for successful obfuscation---particularly to an author's message consistency as a whole. However, in practice such considerations could be left up to the author. In this human-in-the-loop scenario, a more extensive set of candidates could be suggested, and their effect on the substitute model shown interactively. This way, the attacks can be manually tuned to find a balance of effectiveness, inconspicuousness, and to guarantee semantic consistency. It would also show the author how their writing style affects potential future inferences.

Regarding the performance of the attacks: we demonstrated the general effectiveness of contextual language models in retrieving candidate suggestions. However, the quality of those candidates might be improved with more extensive rule-based checks; e.g., through deeper analyses using parsing. Nevertheless, such venues leave us with a core limitation of rewriting language, and therefore more broadly NLP: while the Masked attacks seemed more successful in our experiments, after manual inspection of the perturbations Dropout was found to often be semantically closer (see also Table~\ref{tab:example})---which was not reflected in the human evaluation. This begs the question if \emph{any} automated approach, evaluated under the current limitations of semantic consistency metrics, could realistically optimize for both obfuscation and inconspicuousness.

As such, we would argue that future work should focus on making as few perturbations as possible, retaining only the minimum amount of required obfuscation success. Given this, the other constraints become less relevant; one could generate short sentences (e.g., a single tweet) that might be semantically or contextually incorrect, but if this is inserted as a single message in a long post history, it will hardly be detectable or intrusive. This would likely require certain adversarial triggers (as demonstrated by \cite{wallace-etal-2019-universal} for example), and ascertaining how well they transfer. 

\section{Conclusion}

In this chapter, we argued realistic adversarial stylometry should be tested on transferability in settings where there is no access to the target model's data or architecture. We extended previous adversarial text classification work with two transformer-based models, and studied their obfuscation success in such a setting. We showed them to reliably drop target model performance below chance, though human detectability of the attacks remained above chance. Future work could focus on further minimizing this detection under our realistic constraints.
\thispagestyle{empty}
\strut
\newpage

\begin{figure}
    \centering
    \includegraphics[width=0.75\textwidth,angle=-90]{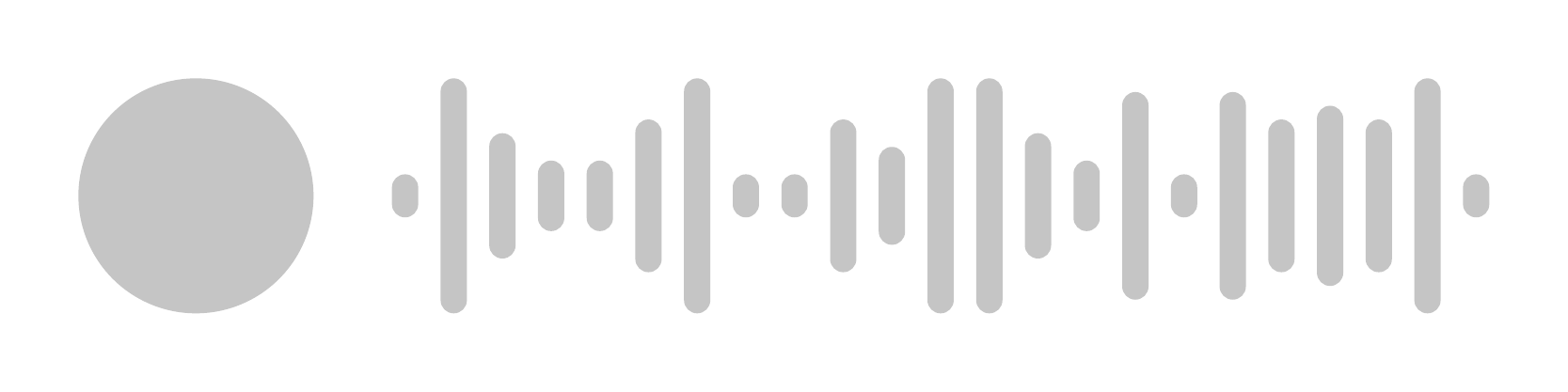}
\end{figure}

\thispagestyle{empty}
\strut
\newpage
\thispagestyle{empty}
\strut
\vfill

\noindent This chapter has been published as:

\begin{description}
    \item \bibentry{emmery-etal-2021-current}
\end{description}

\noindent Minor formatting changes have been made to the text and figures.

\chapter[Current Limitations in Cyberbullying Detection]{Current Limitations in Cyberbullying Detection: on Evaluation Criteria, Reproducibility, and Data Scarcity} \label{ch:bul}

\lettrine[lines=4, lraise=0, nindent=0em, slope=0em]{D}{etection} of online cyberbullying has seen an increase in societal importance, popularity in research, and available open data. Nevertheless, while computational power and affordability of resources continue to increase, the access restrictions on high-quality data limit the applicability of state-of-the-art techniques. Consequently, much of the recent research uses small, heterogeneous datasets, without a thorough evaluation of applicability. In this chapter, we further illustrate these issues, as we (i) evaluate many publicly available resources for this task and demonstrate difficulties with data collection. These predominantly yield small datasets that fail to capture the required complex social dynamics and impede direct comparison of progress. We (ii) conduct an extensive set of experiments that indicate a general lack of cross-domain generalization of classifiers trained on these sources, and openly provide this framework to replicate and extend our evaluation criteria. Finally, we (iii) present an effective crowdsourcing method: simulating real-life bullying scenarios in a lab setting generates plausible data that can be effectively used to enrich real data. This largely circumvents the restrictions on data that can be collected, and increases classifier performance. We believe these contributions can aid in improving the empirical practices of future research in the field.

\parshape=0
\section{Introduction}

Learning to accurately classify rare phenomena within large feeds of data poses challenges for numerous applications of machine learning. The volume of data required for representative instances to be included is often resource-consuming, and limited access to such instances can severely impact the reliability of predictions. These limitations are particularly prevalent in applications dealing with sensitive social phenomena such as those found in forensics: e.g., predicting acts of terrorism, detecting fraud, or uncovering sexually transgressive behavior. Their events are complex and require rich representations for effective detection. Conversely, online text, images, and meta-data capturing such interactions have commercial value for the platforms hosting them and are often off-limits to protect users' privacy.

An application affected by such limitations with increasing societal importance and growing interest over the last decade is that of cyberbullying detection. Not only is the data sensitive, but it is also inherently scarce in terms of public access. Most cyberbullying events are off-limits to the majority of researches, as they take place in private conversations. Fully capturing the social dynamics and complexity of these events requires much richer data than available to the research community up until now. Related to this, various issues with the operationalization of cyberbullying detection research were recently demonstrated by \cite{DBLP:journals/chb/RosaPRFCOCPST19}, who share much of the same concerns as we will discuss in this chapter. While their work focuses on methodological rigor in prior research, we will focus on the core limitations of the domain and complexity of cyberbullying detection. Through an evaluation of the current advances on the task, we illustrate how the mentioned issues affect current research, particularly cross-domain. Finally, we demonstrate crowdsourcing in an experimental setting to potentially alleviate the task's data scarcity. First, however, we introduce the theoretical framing of cyberbullying and the task of automatically detecting such events.

\subsection{Cyberbullying} \label{sec:bul}

Asynchrony and optional anonymity are characteristic of online communication as we know it today; it heavily relies on the ability to communicate with people who are not physically present, and stimulates interaction with people outside of one's group of close friends through social networks \citep{Madden2013}. The rise of these networks brought various advantages to adolescents: studies show positive relationships between online communication and social connectedness \citep{doi:10.1080/13691180701858851,ValkePeter2007fw}, and that self-disclosure on these networks benefits the quality of existing and newly developed relationships \citep{Steijn2013-hi}. The popularity of social networks and instant messaging among children has them connecting to the Internet from increasingly younger ages \citep{RePEc:ehl:lserod:50228}, with $95\%$ of teens\footnote{Survey conducted in 2011 among 799 American teens. Black and Latino families were oversampled.} ages 12--17 online, of which $80\%$ are on social media \citep{Lenhart2011}. For them, however, the transition from social interaction predominantly taking place on the playground to being mediated through mobile devices \citep{Livingstone2011} has also moved negative communication to a platform where indirect and anonymous interaction has a window into homes.

A range of studies conducted by the Pew Research Center\footnote{\url{www.pewinternet.org}}, most notably \cite{Lenhart2011}, provides detailed insight into these developments. While $78\%$ of teens report positive outcomes from their social media interactions, $41\%$ have experienced at least some adverse outcomes, ranging from arguments, trouble with school and parents, physical fights and ending friendships. From $19\%$ bullied in the 12 months prior to the study, $8\%$ of all teens reported this was some form of cyberbullying. These numbers are comparable ($7\%$ for Grades 6--12, and $15\%$ Grades 9--12 respectively) to other research \citep{Kann2014-mp,Morgan2015}. Bullying has for a while been regarded as a public health risk by numerous authorities \citep{xu-etal-2012-learning}, with depression, anxiety, low self-esteem, school absence, lower grades, and risk of self-medication as primary concerns.

The act of cyberbullying---other than being conducted online---shares the characteristics of traditional bullying: a power imbalance between the bully and victim \citep{https://doi.org/10.1002/cbm.123}, the harm is intentional, repeated over time, and has a negative psychological effect on the victim \citep{Dehue2008-rn}. With the Internet as a communication platform, however, some additional aspects arise: location, time, and physical presence have become an irrelevant factor in the act. Accordingly, several categories unique to this form of bullying are defined \citep{Willard2007-mm,Beran2008}: \emph{flaming} (sending rude or vulgar messages), \emph{outing} (posting private information or manipulated personal material of an individual without consent), \emph{harassment} (repeatedly sending offensive messages to a single person), \emph{exclusion} (from an online group), \emph{cyberstalking} (terrorizing through sending explicitly threatening and intimidating messages), \emph{denigration} (spreading online gossips), and \emph{impersonation}. Moreover, in addition to optional anonymity hiding the critical figures behind an act of cyberbullying, it could also obfuscate the number of actors (i.e., there might only be one even though it seems there are more). Cyberbullying acts can prove challenging to remove once published; messages or images might persist through sharing and be viewable by many (as is typical for hate pages), or available to a few (in group or direct conversations). Hence, it can be argued that any form of harassment has become more accessible and intrusive. This online nature has an advantage as well: in theory, platforms record these bullying instances. Therefore, an increasing number of researches are interested in the automatic detection (and prevention) of cyberbullying.

\subsection{Detection and Task Complexity} \label{subs:tskcomp}

The task of cyberbullying detection can be broadly defined as the use of machine learning techniques to automatically classify text in messages on bullying content, or infer characteristic features based on higher-order information, such as user features or social network attributes. Bullying is most apparent in younger age groups through direct verbal outings \citep{Vaez2004-sl}, and more subtle in older groups, mainly manifested in more complex social dynamics such as exclusion, sabotage, and gossip \citep{Privitera2009-iu}. Therefore, the majority of work on the topic focuses on younger age groups, be it deliberately or given that the primary source for data is social media---which will likely result in these being highly present for some media \citep{Duggan2015}. Apart from the well-established challenges that language use poses (e.g., ambiguity, sarcasm, dialects, slang, neologisms), two factors in the event add further linguistic complexity,  namely that of actor \emph{role} and associated \emph{context}. In contrast to tasks where adequate information is provided in the text of a single message, to completely map a cyberbullying event and pinpoint bully and victim implies some understanding of the dynamics between the involved actors and the textual interpretation of the \emph{register}.

\paragraph{Register} Firstly, to understand the task of cyberbullying detection as a specific domain of text classification, one should consider the full scope of the register that defines it. The bullying categories discussed in Section~\ref{sec:bul} include some  cues that can be inferred from text alone; flaming being the most overt through simple curse word usage, slurs, or other profanity. Similarly, threatening or intimidating messages that fall under cyberstalking are clearly denoted by particular word usage. The other categories are more subtle: outing could also be done textually, in the form of a phone number, or pieces of information that are personal or sensitive in nature. Denigration would include words that are not blatantly associated with abusive acts; however, misinformation about sensitive topics might for example be paired with a victim's name. One could further extend these cues based on the literature (as also captured in \citeauthor{VanHee2015guide}, \citeyear{VanHee2015guide}) to include bullying event cues, such as messages that serve to defend the victim, and those in support of the bully. The linguistic task could therefore be framed (partly based on \citeauthor{10.1371/journal.pone.0203794}, \citeyear{10.1371/journal.pone.0203794}) as \emph{identifying an online message context that includes aggressive or hurtful content against a victim}. Several additional communicative components in these contexts further change the interpretation of these cues, however.

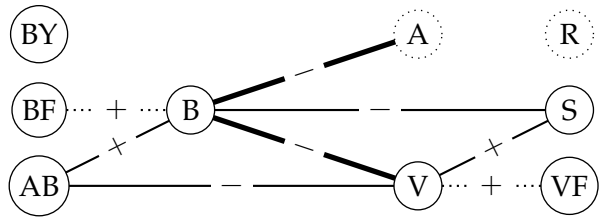
\begin{figure}[t]
    \centering
    \begin{tikzpicture}[scale=1,transform shape]
  \Vertex[x=0.0,y=2.0]{AB}
  \Vertex[x=2.0,y=3.0]{B}
  \Vertex[x=5.0,y=2.0]{V}
  \Vertex[x=7.0,y=3.0]{S}
  \Vertex[x=7.0,y=2.0]{VF}
  \Vertex[x=7.0,y=4.0, style=dotted]{R}
  \Vertex[x=0.0,y=4.0]{BY}
  \Vertex[x=5.0,y=4.0, style=dotted]{A}
  \Vertex[x=0.0,y=3.0]{BF}
  \tikzstyle{LabelStyle}=[fill=white,sloped]

  \Edge[label=$-$](AB)(V)
  \Edge[label=$-$, lw=2pt](B)(V)
  \Edge[label=$-$, lw=2pt](A)(B)
  \Edge[label=$+$](AB)(B)
  \Edge[label=$+$](S)(V)
  \Edge[label=$-$](S)(B)
  \Edge[label=$+$, style=dotted](BF)(B)
  \Edge[label=$+$, style=dotted](VF)(V)
\end{tikzpicture}
    \caption{Role graph of a bullying event. Each vertex represents an actor,
             labeled by their role in the event: bully ($B$), victim ($V$), bystander ($BY$), reinforcer ($AB$), assistant ($BF$), defender ($S$), reporter ($S$), accuser ($A$), and friend ($VF$). Each edge indicates a stream of
             communication, labeled by whether this is positive ($+$) or negative
             ($-$) in nature, and its strength indicating the frequency of interaction. Dotted edges indicate nonparticipation in the event, and vertices those added by \cite{xu-etal-2012-learning} to account for
             social-media-specific roles.}
    \label{fig:graph}
\end{figure}

\paragraph{Roles} \label{part:roles}
Secondly, there is a commonly made distinction between several actors within a cyberbullying event. A naive role allocation includes a bully $B$, a victim $V$ and bystander $BY$, the latter of whom may or may not approve of the act of bullying. More nuanced models such as that of \cite{xu-etal-2012-learning} include additional roles (see Figure~\ref{fig:graph} for a role interaction visualization), where different roles can be assigned to one person; for example, being bullied and reporting this. Most importantly, all shown roles can be present in the span of one single thread on social media, as demonstrated in Table~\ref{tab:bul}. While some roles clearly show from frequent interaction with either a positive or negative sentiment ($B$, $V$, $A$), others might not be observable through any form of conversation ($R$, $BY$), prove too subtle, or not distinguishable from other roles. 

\begin{table*}[t]
    \centering
    \footnotesize
    \resizebox{\textwidth}{!}{%
      \begin{tabular}{rlp{7cm}cr}
      \toprule
      Line & Role & Message & Bully & Type \\
      \midrule
      1 & V    & me and my friends hanging out tonight! :)                 &   & neutral \\
      2 & B    & @V lol b*tch, you dont have any friends.. ur fake as sh*t & \checkmark & curse, insult \\
      3 & AB   & @B haha word, shes so sad                                 & \checkmark & encouragement \\
      4 & VF   & @V you know it girl                                       &   & \\
      5 & S    & @V dont listen to @B, were gonna have fun for sure!        &   & defense \\
      6 & V    & @B shut up @B!! nobody asked your opinion!!!!              &   & defense \\
      7 & A    & @B you are a f*cking bully, go outside or smt             &   & insult \\
      8 & B    & @V @S haha you all so dumb, just kill yourself already!   & \checkmark & insult, curse \\
      9 & A, R & @B shut up or ill report you                                &   & \\
      10 & B    & @A u gonna cry? go ahead, see what happens tomorrow!      & \checkmark &  threat \\
      \bottomrule
     \end{tabular}
    }
     \caption{Fictional example of a cyberbullying conversation. Lines represent sequential turns. Roles are noted as described on Page~\pageref{part:roles} (under the eponymous paragraph), if the message can be considered bullying by \checkmark, and types according to the annotation guidelines from \cite{VanHee2015guide}.}
    \label{tab:bul}
\end{table*}

\paragraph{Context} Thirdly, the content of the messages has to be interpreted differently between these roles. While curse words can be a good indication of harassment, identification of a bully arguably requires more than these alone. Consider Table~\ref{tab:bul}: both $B$ and $A$ use insults (lines 7--8), the message of $V$ (line 6) might be considered as bullying in isolation, and having already determined $B$, the last sentence (line 10) can generally be regarded as a threat. In conclusion, the full scope of the task is complex; it could have a temporal-sequential character, would benefit from determining actors and their interactions, and then should have some sense of severity as well (e.g.\ distinguish bullying from teasing).

\subsection{Our Contributions}

Surprisingly, a significant amount of work on the task does not collect (or use) data that allows for the inference of such features (which we will further elaborate on in Section~\ref{sec:tco}). To confirm this, we reproduce part of the previous cyberbullying detection research on different sources. Predictions made by current automatic methods for cyberbullying classification are demonstrated not to reflect the above-described task complexity; we show performance drops across different training domains, and give insights into content feature importance and limitations. Additionally, we report on reproducibility issues in state-of-art work when subjected to our evaluation. To facilitate future reproduction, we will provide all code open-source, including dataset readers, experimental code, and qualitative analyses.\footnote{Available at \url{https://github.com/cmry/amica}.} Finally, we present a method to collect crowdsourced cyberbullying data in an experimental setting. It grants control over the size and richness of the data, does not invade privacy, nor rely on external parties to facilitate data access. Most importantly, we demonstrate that it successfully increases classifier performance. With this chapter, we provide suggestions on improving methodological rigor and hope to aid the community in a more realistic evaluation and implementation of this task.

\section{Related Work} \label{sec:prev}

The task of detecting cyberbullying content can be roughly divided into three categories. First, research with a focus on \emph{binary} classification, where it is only relevant if a message contains bullying or not. Second, more \emph{fine-grained} approaches where the task is to determine either the role of actors in a bullying scenario or the content type (i.e., different categories of bullying). Both binary and fine-grained approaches predominantly focus on text-based features. Lastly, \emph{meta-data} approaches that take more than just message content into account; these might include profile, network, or image information. Here, we will discuss efforts relevant to the task of cyberbullying classification within these three topics. We will predominantly focus on work conducted on openly available data, and those that report (positive) $F_1$-scores, to promote fair comparisons.\footnote{Unfortunately, numerous (recent) work on cyberbullying detection seems not to report such $F_1$-scores (in favor of accuracy), is limited to criticized datasets with high baseline scores (such as the \textsc{caw} datasets) or does not show enough methodological rigor---some are therefore not included in this overview.} For an extensive literature review and a detailed comparison of different studies, see \cite{DBLP:journals/chb/RosaPRFCOCPST19}. Finally, a significant portion of our research pertains to generalizability, and therefore the field of domain adaptation. We will discuss its prior observations related to text classification specifically, and their relevance to (future) research on cyberbullying detection.

\begin{table*}[t!]
   \resizebox{\textwidth}{!}{%
      \begin{tabular}{l>{\raggedright\arraybackslash}p{3.2cm}lllrr} 
    \toprule
    Author         & Other                                   & Name              & OS                                                          &  Platform   & Pos & Neg \\ 
    \midrule
    \citeauthor{yin2009detection} & \citeauthor{Nahar2013}                        & \textsc{caw\_kon} & \href{http://caw2.barcelonamedia.org/}{v}                   & Kongregate & $42$    & $4802$   
    \\ 
    
    \citeauthor{yin2009detection} & \citeauthor{Nahar2013}                        & \textsc{caw\_sls} & \href{http://caw2.barcelonamedia.org/}{v}                   & Slashdot   & $60$    & $4303$  
    \\ 
    
    \citeauthor{yin2009detection} & \citeauthor{Nahar2013}                        & \textsc{caw\_msp} & \href{http://caw2.barcelonamedia.org/}{v}                   & Myspace    & $65$    & $1946$  
    \\ 
    
    \citeauthor{6147681} & \citeauthor{DBLP:conf/websci/KontostathisRGE13,DBLP:conf/asunam/SquicciariniRLG15,DBLP:conf/fuzzIEEE/RosaCCMRC18,DBLP:conf/ijcnn/RosaM0CC18}  & \textsc{kon\_frm} & \href{http://www.chatcoder.com/DataDownload}{v}             & Formspring & $369$   & $3915$  \\ 

    \citeauthor{DBLP:conf/icwsm/DinakarRL11} & \citeauthor{}                             & \textsc{din\_ytb} & x                                                           & YouTube    & $2277$  & $4500$   
    \\ 
    
    \citeauthor{bayzick2011detecting} & \citeauthor{DBLP:journals/taffco/ZhaoM17,DBLP:conf/asunam/SquicciariniRLG15}          & \textsc{bay\_msp} & \href{http://www.chatcoder.com/DataDownload}{v}             & Myspace    & $415$   & $1647$  
    \\  
    
    \citeauthor{xu-etal-2012-learning} & \citeauthor{DBLP:journals/taffco/ZhaoM17}                          & \textsc{xu\_trec} & \href{https://research.cs.wisc.edu/bullying/data.html}{v}  & Twitter    & $684$   & $1762$  \\ 
    \citeauthor{Dadvar2014} & \citeauthor{}                              & \textsc{ddv\_msp} & x                                                           & Myspace    & $311$   & $8938$  \\ 
    
    \citeauthor{Dadvar2014} & \citeauthor{}                              & \textsc{ddv\_ytb} & x                                                           & YouTube    & $449$   & $4177$  \\ 
    
    \citeauthor{DBLP:conf/icis/BretschneiderWP14} & \citeauthor{}                                & \textsc{brt\_twi} & \href{http://www.ub-web.de/research/}{v}                    & Twitter    & $220$   & $5162$  \\ 
    
    \citeauthor{DBLP:conf/icis/BretschneiderWP14} & \citeauthor{}                                & \textsc{brt\_tw2} & \href{http://www.ub-web.de/research/}{v}                    & Twitter    & $194$   & $2599$  \\ 
    
    \citeauthor{van-hee-etal-2015-detection} & \citeauthor{10.1371/journal.pone.0203794}                    & \textsc{ami\_ask} & \href{https://osf.io/rgqw8/}{v}                                                  & Ask.fm     & $3787$  & $86419$ \\ 
    
    \citeauthor{DBLP:journals/corr/HosseinmardiMRH15a} & \citeauthor{DBLP:conf/sdm/ChengGSHL19}                       & \textsc{hos\_ins} & \href{https://sites.google.com/site/cucybersafety/home/
cyberbullying-detection-project/dataset}{v}                                                           & Instagram  & $567$   & $1387$  \\ 

    \citeauthor{Sui2015} & \citeauthor{DBLP:journals/taffco/ZhaoM17}                        & \textsc{sui\_twi} & \href{https://research.cs.wisc.edu/bullying/data.html.}{v}  & Twitter    & $2102$  & $5219$  \\ 
    
    \bottomrule
  \end{tabular}
    }
    \caption{Overview of datasets for cyberbullying detection. Lists the authors of the initial sets (Author), if the set was used by other work (Other), a reference name (Name), if the data is publicly available (OS, including a link to the source), which platform it was extracted from (Platform), the number of reported cyberbullying instances (Pos) and of non-cyberbullying (Neg). Please note that the instance numbers are as reported in the original work, and may have deviated through time (such as Twitter sets, and Formspring).}
    \label{tab:dat}
\end{table*}

\subsection{Binary Classification} \label{subs:bin}

One of the first traceable suggestions for applying text mining to the task of cyberbullying detection is made by \cite{doi:https://doi.org/10.1002/9780470689646.ch8}, who note that \cite{yin2009detection} previously tried to classify online harassment on the \textsc{caw} 2.0 dataset.\footnote{Data has been made available at \url{caw2.barcelonamedia.org}.} \citeauthor{yin2009detection} already state that the ratio of documents with harassing content to typical documents is challengingly small. Moreover, they foresee several other critical issues with regard to the task: a lack of positive instances will make detecting characteristic features a difficult task, and human labeling of such a dataset might have to face issues of ambiguity and sarcasm that are hard to assess when messages are taken out of conversation context. Even with very sparse datasets ($< 1\%$ positive instances), the harassment classifier outperforms the random baseline using tf$\cdot$idf, pronoun, curse word, and post similarity features. 

Following up \cite{yin2009detection}, \cite{6147681} note that the \textsc{caw} 2.0 dataset is generally unfit for cyberbullying classification: in addition to lacking bullying labels (it only provides harassment labels), the conversations are predominantly between adults. Their work, along with \cite{bayzick2011detecting}, is a first effort to create datasets for cyberbullying classification through scraping the question-answering website Formspring.me, as well as Myspace.\footnote{Data has been made available at \url{www.chatcoder.com/DataDownload}.} In contrast with similar research, they aim to use textual features while deliberately avoiding Bag-of-Words (BoW) features. Through a curse word dictionary and custom severity annotations, they construct several metrics for features related to these ``bad'' words. In more recent work, \cite{DBLP:conf/websci/KontostathisRGE13} redid analyses on the \textsc{kon\_frm} set, primarily focusing on the contribution curse words have in the classification of bullying messages. By forming queries from curse word dictionaries, they show that there is no one combination which retrieves all. However, by capturing them in an Essential Dimensions of Latent Semantic Indexing query vector averaged over known bullying content---classifying the top-$k$ (by cosine similarity) as positive---they show a significant Average Precision improvement over their baseline.

More recent efforts includes the work of \cite{DBLP:conf/icis/BretschneiderWP14}, who combined word normalization, Named Entity Recognition (to detect person-specific references), and multiple curse word dictionaries \citep{noswear,broadc,Millwood2000} in a rule-based pattern classifier, scoring well on Twitter data.\footnote{Data has been made available at \url{www.ub-web.de/research}.} Our own work \citep{van-hee-etal-2015-detection}, where we collected a large dataset with posts from Ask.fm, used standard BoW features. Later, these were extended in \cite{10.1371/journal.pone.0203794} with term lists, subjectivity lexicons, and topic model features. Recently popularized techniques of word embeddings and neural networks have been applied by \cite{DBLP:conf/icdcn/ZhaoZM16,DBLP:journals/taffco/ZhaoM17} on \textsc{xu\_trec}, \textsc{nay\_msp} and \textsc{sui\_twi}, both resulting in the highest performance for those sets. Convolutional Neural Networks (CNNs) on phonetic features were applied by \cite{DBLP:conf/icmla/ZhangTVWMKHLMD16}, and \cite{DBLP:conf/ijcnn/RosaM0CC18} investigate among others the same architecture on textual features in combination with Long Short-Term Memory Networks (LSTMs). Both \cite{DBLP:conf/ijcnn/RosaM0CC18} and \cite{DBLP:conf/ecir/AgrawalA18} investigate the C-LSTM \citep{DBLP:journals/corr/ZhouSLL15b}, the latter includes Synthetic Minority Over-sampling Technique (SMOTE). However, as we will show in the current chapter, both of these works suffer from reproducibility issues. Finally, fuzzified vectors of top-$k$ word lists for each class were used to conduct membership likelihood-based classification by \cite{DBLP:conf/fuzzIEEE/RosaCCMRC18} on \textsc{kon\_frm}, boosting recall over previously used methods.

\subsection{Fine-Grained Classification}

 A common thread in previously discussed research was a focus on detecting single messages with evidence of cyberbullying per instance. As argued in Section~\ref{subs:tskcomp}, however, there are more textual cues to infer than merely if a message might be interpreted as bullying. The work of \cite{xu-etal-2012-learning} proposed to expand this binary approach with fine-grained tasks by looking at \emph{bullying traces}; i.e., the responses to a bullying incident. They distinguished two tasks based on keyword-retrieved (\emph{bully}) Twitter data:\footnote{Data has been made available at \url{research.cs.wisc.edu/bullying/data.html}.} (1) a \emph{role} labeling task, where semantic role labeling was then used to distinguish person-mention roles, and (2) the incorporation of sentiment to determine \emph{teasing}, where despite high accuracy, $48\%$ of the positive instances were misclassified.
 
In our prior work, we extended this train of thought and demonstrated the difficulty of fine-grained classification of types of bullying (curse, defamation, defense, encouragement, insult, sexual, threat), and roles (harasser, bystander assistant, bystander defender, victim) with simple BoW and sentiment features --- especially in detecting types \citep{van-hee-etal-2015-detection,VanHee2015guide}. This was further addressed in \cite{10.1371/journal.pone.0203794} for both Dutch and English. Evaluated against a profanity (curse word lexicon) and word $n$-gram baseline, a multi-feature model (as discussed in Section~\ref{subs:bin}) achieved the lowest error rates over (almost) all labels, for both bullying type and role classification. Lastly, \cite{DBLP:conf/asunam/TomkinsGCZ18} also adapt fine-grained knowledge about bullying events in their sociolinguistic model; in addition to  bullying classification, they find latent text categories and roles, partly relying on social interactions on Twitter. It thereby ties in with the next category of work: leveraging meta-data from the network the data is collected from.

\subsection{Meta-data Features} \label{par:meta} A notable, yet less popular, aspect of this task is the utilization of a graph for visualizing potential bullies and their connections. This method was first adopted by \cite{Nahar2013}, who use this information in combination with a classifier trained on LDA and weighted tf$\cdot$idf features to detect bullies and victims on the \textsc{caw\_*} datasets. Work that more concretely implements techniques from graph theory is that of \cite{DBLP:conf/asunam/SquicciariniRLG15}, who used a wide range of features: network features to measure popularity (e.g., degree centrality, closeness centrality), content-based features (length, sentiment, offensive words, second-person pronouns), and incorporated age, gender, and number of comments. They achieved the highest performance on the \textsc{kon\_frm} and \textsc{bay\_msp} datasets. 

Work by \cite{DBLP:journals/corr/HosseinmardiMRH15a} focuses on Instagram posts and incorporates platform-specific features retrieved from images and its network. They are the first to adhere to the literature more closely and define cyberagression \citep{kowalski-2012-cyber} separately from cyberbullying, in that these are single negative posts rather than the repeated character of cyberbullying. They also show that certain LIWC (Linguistic Inquiry and Word Count) categories, such as death, appearance, religion, and sexuality, give a good indication of cyberbullying. While BoW features perform best, meta-data features (e.g., user properties and image content) and textual features from the top $15$ comments achieve a similar score. Cyberagression seems to be slightly easier to classify.

\subsection{Domain Adaptation}

As the majority of the work discussed above focuses on a single corpus, a serious omission seems to be gauging how this influences model generalization. Many applications in natural language processing (NLP) are often inherently limited by expensive high-quality annotations, whereas unlabeled data is plentiful. Idiosyncrasies between source and target domains often prove detrimental to the performance of techniques relying on those annotations \citep{mcclosky-etal-2006-reranking,chan-ng-2006-estimating,vilain-etal-2007-entity} when applied in the wild. The field of domain adaptation identifies tasks that suffer from such limitations, and aims to overcome them either in a supervised \citep{finkel-manning-2009-hierarchical,daume-iii-etal-2010-frustratingly} or unsupervised \citep{blitzer-etal-2007-biographies,DBLP:conf/cikm/JiangZ07,ma-2014-automatic,schnabel-schutze-2014-flors} way. For text classification, sentiment analysis is arguably closest to the task of cyberbullying classification \citep{DBLP:conf/icml/GlorotBB11,DBLP:conf/icml/ChenXWS12,DBLP:conf/www/PanNSYC10}. In particular, as imbalanced data exacerbates generalization \citep{li-etal-2012-active-learning}. However, while for sentiment analysis these issues are clearly identified and actively worked on, the same cannot be said for cyberbullying detection,\footnote{One very recent exception to the latter can be found in \cite{DBLP:conf/sdm/ChengGCL20}. Their work introduces a novel domain adaptation technique, and demonstrates it to increase performance on two text classification tasks, one being cyberbullying detection.} where concrete limitations have yet to be explored. We assume to find issues similar to those in sentiment analysis in the current task, as we will further discuss in the following section.

\section[Task Evaluation Importance and Hypotheses]{Task Evaluation Importance and \\ Hypotheses} \label{sec:tco}
\label{sec:oos}

The domain of cyberbullying detection is in its early stages, as can be seen in Table~\ref{tab:dat}. Most datasets are quite small, and only a few have seen repeated experiments. Given the substantial societal importance of improving the methods developed so far, pinpointing shortcomings in the current state of research should assist in creating a robust framework under which to conduct future experiments---particularly concerning evaluating (domain) generalization of the classifiers. The latter of which, to our knowledge, none of the current research seems involved with. This is therefore the main focus of our work. In this section, we define three motivations for assessing this, and pose three respective hypotheses through which we will further investigate current limitations in cyberbullying detection.

\subsection{Data Scarcity} \label{subs:scarcity}

Considering the complexity of the social dynamics underlying the target of classification, and the costly collection and annotation of training data, the issue of data scarcity can mostly be explained with respect to the aforementioned restrictions on data access: while on a small number of platforms most data is accessible without any internal access (commonly as a result of optional user anonymity), it can be assumed that a significant part of actual bullying takes place `behind closed doors'. To uncover this, one would require access to all known information within a social network (such as friends, connections, and private messages, including all meta-data). As this is unrealistic in practice, researchers rely on the small subset of publicly accessible data (predominantly text) streams. Consequently, most of the datasets used for cyberbullying detection are small and exhibit an extreme skew between positive and negative messages (as can be seen in Table~\ref{tab:dsc}). It is unlikely that these small sets accurately capture the language use on a given platform, and generalizable linguistic features of the bullying instances even less so. In line with domain adaptation research, we therefore anticipate that the samples are underpowered in terms of accurately representing the substantial language variation between platforms, both in normal language use and bullying-specific language use (Hypothesis \textbf{1}).

\subsection{Task Definition} \label{subs:task}

Furthermore, we argue that this scarcity introduces issues with adherence to the definition of the task of cyberbullying. The chances of capturing the underlying dynamics of \emph{cyberbullying} (as defined in the literature) are slim with the message-level (i.e., using single documents only) approaches that the majority of work in the field has used. The users in the collected sources have to be rash enough to bully in the open, and particular (curse) word usage that would explain the effectiveness of dictionary and BoW-based approaches in previous research. Hence, we also assume that the positive instances are biased; only reflecting a limited dimension of bullying (Hypothesis \textbf{2}). A more realistic scenario---where characteristics such as repetitiveness and power imbalance are taken into consideration---would require looking at the interaction between persons, or even profile instances rather than single messages, which, as we argued, is not generally available. The work found in the meta-data category (Section \ref{par:meta}) supports this argument, with improved results using this information.

This theory regarding the definition (or operationalization) of this task is shared by Rosa et al., who pose that ``\emph{the most representative studies on automatic cyberbullying detection, published from 2011 onward, have conducted isolated online aggression classification}'' \cite[p. 341]{ROSA2019333}. We will mainly focus on the shared notion that this framing is limited to verbal aggression; however, our focus will empirically assess its overlap with data framed to solely contain online toxicity data (i.e., online / cyberagression) to find concrete evidence.

\subsection{Domain Influence}

Enriching previous work with data such as network structure, interaction statistics, profile information, and time-based analyses might provide fruitful sources for classification and a correct operationalization of the task. However, they are also domain-specific, as not all social media have such a rich interaction structure. Moreover, it is arguably naive to assume that social networks such as Facebook (for which in an ideal case, all aforementioned information sources are available) will stay a dominant platform of communication. Recently, younger age groups have turned towards more direct forms of communication such as WhatsApp, Snapchat, or media-focused forms such as Instagram \citep{Smith2018}, and recently TikTok. This move implies more private and less affluent environments in which data can be accessed (resulting  in even more scarcity), and that further development in the field requires a critical evaluation of the current use of the available features, and ways to improve cross-domain generalization overall. This work, therefore, does not disregard textual features; they would still need to be considered as the primary source of information, while paying particular attention to the issues mentioned here. We further try to contribute towards this goal and argue that crowdsourcing bullying content potentially decreases the influence of domain-specific language use, allows for richer representations, and alleviates data scarcity (Hypothesis \textbf{3}).

\section{Data} \label{sec:data}

\begin{table}[t!]
      \resizebox{\textwidth}{!}{%
      \begin{tabular}{lrrrrrlrrr}
      \toprule
                   & \textsc{\lowercase{Pos}}     & \textsc{\lowercase{Neg}}   & \textsc{\lowercase{Types}}   & \textsc{\lowercase{Tokens}} & \multicolumn{2}{c}{\textsc{\lowercase{Avg Tok/Msg}}}  & \textsc{\lowercase{Emote}}  & \textsc{\lowercase{SweaN}}  & \textsc{\lowercase{SweaP}}                 \\ 
      \midrule
        $D_{twB}$     & $237$     & $5,258  $ & $12K $  & $78K   $  & $14 $ & ($\sigma = 8$)    & $961   $  & $277   $  & $867    $  \\
        $D_{frm}$     & $1,025$   & $11,742 $ & $21K $  & $348K  $  & $27 $ & ($\sigma = 29$)   & $3,322 $  & $1,228 $  & $2,871  $  \\
        $D_{msp}$     & $426$     & $1,627  $ & $13K $  & $803K  $  & $391$ & ($\sigma = 285$)  & $931   $  & $1,447 $  & $3,730  $  \\
        $D_{ytb}$     & $417$     & $3,045  $ & $52K $  & $827K  $  & $239$ & ($\sigma = 252$)  & $3,662 $  & $2,606 $  & $8,705  $  \\
        $D_{ask}$     & $5,001$   & $89,404 $ & $63K $  & $1,017K$  & $12 $ & ($\sigma = 23$)   & $17,362$  & $4,839 $  & $12,191 $  \\
        $D_{twX}$     & $281$     & $4,654  $ & $19K $  & $86K   $  & $18 $ & ($\sigma = 8$)    & $1,344 $  & $74    $  & $502    $  \\
        \midrule
        $D_{tox}$     & $15,279$  & $144,226$ & $220K$  & $12,924$K & $81 $ & ($\sigma = 121$)  & $11,876$  & $13,732$  & $22,404 $  \\
        \midrule
        $D_{ask\_nl}$ & $8,675$   & $70,557 $ & $58K $  & $776K  $  & $10 $ & ($\sigma = 15$)   & $16,905$  & $2,025 $  & $2,299  $  \\
        $D_{sim\_nl}$ & $2,330$   & $2,681  $ & $7K  $  & $55K   $  & $11 $ & ($\sigma = 16$)   & $434   $  & $682   $  & $194    $  \\ 
        $D_{don\_nl}$ & $152$     & $211    $ & $2K  $  & $7K    $  & $20 $ & ($\sigma = 24$)   & $33    $  & $47    $  & $19     $  \\        
        \bottomrule
    \end{tabular}
    }
    \caption{Corpus statistics for English and Dutch cyberbullying datasets, list number of positive (\textsc{\lowercase{Pos}}, bullying) and negative (\textsc{\lowercase{Neg}}, other) instances, \textsc{\lowercase{Types}} (unique words), \textsc{\lowercase{Tokens}} (total words), average number of tokens per message (\textsc{\lowercase{Avg Tok/Msg}}), number of emojis and emoticons (\textsc{\lowercase{Emote}}), and swear word occurrence per neutral (\textsc{\lowercase{SweaN}}), and positive (\textsc{\lowercase{SweaP}}) instance.\protect\footnotemark }
    \label{tab:dsc}
\end{table}

\footnotetext{Emojis were detected with \url{https://github.com/NeelShah18/emot}. Swears were detected with reference lists: for English these were taken from \url{www.noswearing.com} and the Dutch were manually composed.}

For the current research, we distinguish a large variety of datasets. For those provided through the AMiCA (Automatic  Monitoring in Cyberspace Applications)\footnote{\url{www.amicaproject.be}} project, the \emph{Ask.fm} corpus is partially available open-source,\footnote{\url{https://osf.io/rgqw8/}} and the \emph{Crowdsourced} corpus will be made available upon request. All other sources are publicly available datasets gathered from previous research\footnote{These were collected as complete as possible. Twitter, in particular, has low recall; approximately 60\% of the tweets were retrieved. Such numbers are expected given the classification problem; people tend to remove harassing messages as was shown before by \cite{xu-etal-2012-learning}.} as discussed in Section~\ref{sec:prev}. Corpus statistics of all data discussed below can be found in Table~\ref{tab:dsc}. The sets' abbreviations, language (\textsc{en} for English, \textsc{nl} for Dutch), and brief collection characteristics can be found below.

\subsection{AMiCA}

\paragraph{Ask.fm} ($D_{ask}$, $D_{ask\_nl}$, \textsc{en}, \textsc{nl}) were collected from the eponymous social network by \cite{van-hee-etal-2015-detection}. Ask.fm is a question answering-style network where users interact by (frequently anonymously) asking questions on other profiles, and answering questions on theirs. As such, a third party cannot react to these question-answer pairs directly. The anonymity and restrictive interactions make for a high amount of potential cyberbullying. Profiles were retrieved through profile seed list, used as a starting point for traversing to other profiles and collecting all existing question-answer pairs for those profiles---these are predominantly Dutch and English. Each message was annotated with fine-grained labels \cite[further details can be found in][]{VanHee2015guide}; however, for our experiments these were binarized, with any form of bullying being labeled positive.
    
\paragraph{Donated} ($D_{don\_nl}$, \textsc{nl}) contains instances of (Dutch) cyberbullying from a mixture of platforms such as Skype, Facebook, and Ask.fm. The set is quite small; however, it contains several hate pages that are valuable collections of cyberbullying directed towards one person. The data was donated for use in the AMiCA project by previously bullied teens, and it thereby provides a particularly reliable source of gold standard, real-life data.
    
\paragraph{Crowdsourced} ($D_{sim\_nl}$, \textsc{nl}) originates from a crowdsourcing experiment conducted by \cite{Broeck2014}, wherein $200$ adolescents aged 14 through 18 participated in a role-playing experiment on an isolated SocialEngine\footnote{\url{www.socialengine.com}} social network. Here, each respondent was given the account of a fictitious person and put in one of four roles in a group of six: a bully, a victim, two bystander-assistants, and two bystander-defenders. They were asked to read---and identify with---a character description and respond to an artificially generated initial post attributed to one of the group members. All were confronted with two initial posts containing either  low- or high-perceived severity of cyberbullying.

\subsection{Related Work}

\paragraph{Formspring} ($D_{frm}$, \textsc{en}) is taken from the research by \cite{6147681} and is composed of posts from Formspring.me, a question-answering platform similar to Ask.fm. As Formspring is mostly used by teenagers and young adults, and also provides the option to interact anonymously, it is notorious for hosting large amounts of bullying content \citep{uclan8378}. The data was annotated through Mechanical Turk, providing a single label by majority vote for a question-answer pair. For our experiments, the question and answer pairs were merged into one document instance.

\paragraph{Myspace} ($D_{msp}$, \textsc{en}) was collected by \cite{bayzick2011detecting}. Being an information retrieval task, the posts are labeled in batches of ten posts, and thus a single label applies to the entire batch (i.e., does it include cyberbullying). Each batch is considered an instance, and labeled accordingly. Due to this batching, the average tokens per instance are much higher than any of the other corpora.
    
\paragraph{Twitter} ($D_{twB}$, \textsc{en}) by \cite{DBLP:conf/icis/BretschneiderWP14} was collected from the stream between 20-10-2012 and 30-12-2012, and was labeled based on a majority vote between three annotators. Excluding re-tweets, the main dataset consists of $220$ positive and $5162$ negative examples, which adheres to the general expected occurrence rate of $4\%$. Their comparably-sized test set, consisting of $194$ positive and $2699$ negative examples, was collected by adding a filter to the stream for messages containing \texttt{school}, \texttt{class}, \texttt{college}, or \texttt{campus}. These sets are merged for the current experiments.
    
\paragraph{Twitter II} ($D_{twX}$, \textsc{en}) from \cite{xu-etal-2012-learning} focussed on \emph{bullying traces}, and was thus retrieved by keywords (\texttt{bully}, \texttt{bullying}), which generates a strong classification bias if left unmasked (both by word usage and being a mix of toxicity and victims). It does, however, allow for demonstrating the ability to detect bullying-associated topics, and (indirect) reports of bullying.

\subsection{Experiment-specific}

\paragraph{Ask.fm Context} ($C_{ask}$, $C_{ask}\_nl$, \textsc{en}, \textsc{nl}) --- the Ask.fm corpus was collected on profile level, but prior experiments have focused on single message instances \citep{10.1371/journal.pone.0203794}. Here, we aggregate all messages for a single profile, which is then labeled as positive when as few as a single bullying instance occurs on the profile. This aggregation shifts the task of cyberbullying message detection to victim detection on profile level, allowing for more access to context and profile-level severity (such as repeated harassment), and makes for a more balanced set ($1,763$ positive and $6,245$ negative instances). 

\paragraph{Formspring Context} ($C_{frm}$, \textsc{en}) --- similar to the Ask.fm corpus, was collected on profile level \citep{6147681}. However, the set only includes $49$ profiles, some of which only include a single message. Grouping on full profile level would result in very few instances; thus, we opted for creating small `context' in batches of five (of the same profile). Similar to the Ask.fm approach, if one of these messages contains bullying, it is labeled positive, balancing the dataset ($565$ positive and $756$ negative instances). 

\paragraph{Toxicity} ($D_{tox}$, \textsc{en}) --- based on the Detox set from Wikimedia \citep{Thain2017,DBLP:conf/www/WulczynTD17}, this set offers over 300k messages\footnote{Provided via Kaggle, more information here: \url{https://www.kaggle.com/c/jigsaw-toxic-comment-classification-challenge}.} of Wikipedia Talk comments with Crowdflower-annotated labels for toxicity (including subtypes).\footnote{See description at \url{https://meta.wikimedia.org/wiki/Research:Detox/Data_Release} for more information regarding operationalization of this dataset.} Noteworthy is how \emph{disjoint} both the task and the platform are from the rest of the corpora used in this research. While toxicity shares many properties with bullying, the focus here is on single instances of insults directed to anonymous people, who are most likely unknown to the harasser. Given Wikipedia as a source, the article- and moderation-focussed comments make it topically quite different from what one would expect on social media---the fundamental overlap being curse words, which is only one of many dimensions to be captured to detect cyberbullying (as opposed to toxicity). 

\subsection{Preprocessing} \label{subs:proc}

All texts were tokenized using spaCy\footnote{\url{https://spacy.io} (\texttt{v2.0.5})} \citep{honnibal2020spacy}. No preprocessing was conducted for the corpus statistics in Table~\ref{tab:dsc}. All models (Section~\ref{sec:exp}) applied lowercasing and special character removal only; alternative preprocessing steps proved to decreased performance (see Table~\ref{tab:archs}). 

\subsection{Descriptive Analysis} \label{subs:sum}

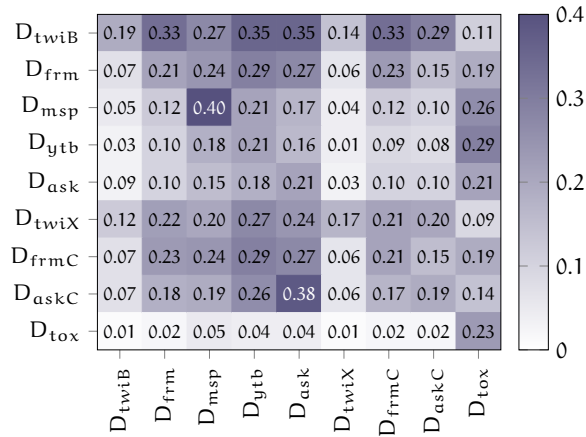
\begin{figure}[t!]
\centering
    \begin{tikzpicture}
  \begin{axis}[enlargelimits=false,
  			   colorbar,
               tick pos=left,
               tick align=outside,
               point meta min=0,
               colormap={custom}{color(0)=(white) color(1)=(Lavender)},
               try min ticks=8,
               xticklabels={,$D_{twiB}$,$D_{frm}$,$D_{msp}$,$D_{ytb}$,$D_{ask}$,$D_{twiX}$,$D_{frmC}$,$D_{askC}$,$D_{tox}$},
               yticklabels={,$D_{twiB}$,$D_{frm}$,$D_{msp}$,$D_{ytb}$,$D_{ask}$,$D_{twiX}$,$D_{frmC}$,$D_{askC}$,$D_{tox}$},
               xticklabel style = {rotate=90},
			   ]
    \addplot [matrix plot,point meta=explicit]
      coordinates {
        (0,0) [0.19] (1,0) [0.33] (2,0) [0.27] (3,0) [0.35] (4,0) [0.35] (5,0) [0.14] (6,0) [0.33] (7,0) [0.29] (8,0) [0.11]

        (0,1) [0.07] (1,1) [0.21] (2,1) [0.24] (3,1) [0.29] (4,1) [0.27] (5,1) [0.06] (6,1) [0.23] (7,1) [0.15] (8,1) [0.19]

        (0,2) [0.05] (1,2) [0.12] (2,2) [0.40] (3,2) [0.21] (4,2) [0.17] (5,2) [0.04] (6,2) [0.12] (7,2) [0.10] (8,2) [0.26]
        
        (0,3) [0.03] (1,3) [0.10] (2,3) [0.18] (3,3) [0.21] (4,3) [0.16] (5,3) [0.04] (6,3) [0.09] (7,3) [0.08] (8,3) [0.29]

        (0,4) [0.03] (1,4) [0.10] (2,4) [0.15] (3,4) [0.18] (4,4) [0.21] (5,4) [0.03] (6,4) [0.10] (7,4) [0.10] (8,4) [0.21]
        
        (0,5) [0.12] (1,5) [0.22] (2,5) [0.20] (3,5) [0.27] (4,5) [0.24] (5,5) [0.17] (6,5) [0.21] (7,5) [0.20] (8,5) [0.09]
        
        (0,6) [0.07] (1,6) [0.23] (2,6) [0.24] (3,6) [0.29] (4,6) [0.27] (5,6) [0.06] (6,6) [0.21] (7,6) [0.15] (8,6) [0.19]
        
        (0,7) [0.07] (1,7) [0.18] (2,7) [0.19] (3,7) [0.26] (4,7) [0.38] (5,7) [0.06] (6,7) [0.17] (7,7) [0.19] (8,7) [0.14]
        
        (0,8) [0.01] (1,8) [0.02] (2,8) [0.05] (3,8) [0.04] (4,8) [0.04] (5,8) [0.01] (6,8) [0.02] (7,8) [0.02] (8,8) [0.23]
      };
      \node[color=black] at (axis cs:0,0) {\footnotesize $0.19$};
      \node[color=black] at (axis cs:1,0) {\footnotesize $0.33$};
      \node[color=black] at (axis cs:2,0) {\footnotesize $0.27$};
      \node[color=black] at (axis cs:3,0) {\footnotesize $0.35$};
      \node[color=black] at (axis cs:4,0) {\footnotesize $0.35$};
      \node[color=black] at (axis cs:5,0) {\footnotesize $0.14$};
      \node[color=black] at (axis cs:6,0) {\footnotesize $0.33$};
      \node[color=black] at (axis cs:7,0) {\footnotesize $0.29$};
      \node[color=black] at (axis cs:8,0) {\footnotesize $0.11$};
      
      \node[color=black] at (axis cs:0,1) {\footnotesize $0.07$};
      \node[color=black] at (axis cs:1,1) {\footnotesize $0.21$};
      \node[color=black] at (axis cs:2,1) {\footnotesize $0.24$};
      \node[color=black] at (axis cs:3,1) {\footnotesize $0.29$};
      \node[color=black] at (axis cs:4,1) {\footnotesize $0.27$};
      \node[color=black] at (axis cs:5,1) {\footnotesize $0.06$};
      \node[color=black] at (axis cs:6,1) {\footnotesize $0.23$};
      \node[color=black] at (axis cs:7,1) {\footnotesize $0.15$};
      \node[color=black] at (axis cs:8,1) {\footnotesize $0.19$};
      
      \node[color=black] at (axis cs:0,2) {\footnotesize $0.05$};
      \node[color=black] at (axis cs:1,2) {\footnotesize $0.12$};
      \node[color=white] at (axis cs:2,2) {\footnotesize $0.40$};
      \node[color=black] at (axis cs:3,2) {\footnotesize $0.21$};
      \node[color=black] at (axis cs:4,2) {\footnotesize $0.17$};
      \node[color=black] at (axis cs:5,2) {\footnotesize $0.04$};
      \node[color=black] at (axis cs:6,2) {\footnotesize $0.12$};
      \node[color=black] at (axis cs:7,2) {\footnotesize $0.10$};
      \node[color=black] at (axis cs:8,2) {\footnotesize $0.26$};
      
      \node[color=black] at (axis cs:0,3) {\footnotesize $0.03$};
      \node[color=black] at (axis cs:1,3) {\footnotesize $0.10$};
      \node[color=black] at (axis cs:2,3) {\footnotesize $0.18$};
      \node[color=black] at (axis cs:3,3) {\footnotesize $0.21$};
      \node[color=black] at (axis cs:4,3) {\footnotesize $0.16$};
      \node[color=black] at (axis cs:5,3) {\footnotesize $0.01$};
      \node[color=black] at (axis cs:6,3) {\footnotesize $0.09$};
      \node[color=black] at (axis cs:7,3) {\footnotesize $0.08$};
      \node[color=black] at (axis cs:8,3) {\footnotesize $0.29$};
      
      \node[color=black] at (axis cs:0,4) {\footnotesize $0.09$};
      \node[color=black] at (axis cs:1,4) {\footnotesize $0.10$};
      \node[color=black] at (axis cs:2,4) {\footnotesize $0.15$};
      \node[color=black] at (axis cs:3,4) {\footnotesize $0.18$};
      \node[color=black] at (axis cs:4,4) {\footnotesize $0.21$};
      \node[color=black] at (axis cs:5,4) {\footnotesize $0.03$};
      \node[color=black] at (axis cs:6,4) {\footnotesize $0.10$};
      \node[color=black] at (axis cs:7,4) {\footnotesize $0.10$};
      \node[color=black] at (axis cs:8,4) {\footnotesize $0.21$};
      
      \node[color=black] at (axis cs:0,5) {\footnotesize $0.12$};
      \node[color=black] at (axis cs:1,5) {\footnotesize $0.22$};
      \node[color=black] at (axis cs:2,5) {\footnotesize $0.20$};
      \node[color=black] at (axis cs:3,5) {\footnotesize $0.27$};
      \node[color=black] at (axis cs:4,5) {\footnotesize $0.24$};
      \node[color=black] at (axis cs:5,5) {\footnotesize $0.17$};
      \node[color=black] at (axis cs:6,5) {\footnotesize $0.21$};
      \node[color=black] at (axis cs:7,5) {\footnotesize $0.20$};
      \node[color=black] at (axis cs:8,5) {\footnotesize $0.09$};
      
      \node[color=black] at (axis cs:0,6) {\footnotesize $0.07$};
      \node[color=black] at (axis cs:1,6) {\footnotesize $0.23$};
      \node[color=black] at (axis cs:2,6) {\footnotesize $0.24$};
      \node[color=black] at (axis cs:3,6) {\footnotesize $0.29$};
      \node[color=black] at (axis cs:4,6) {\footnotesize $0.27$};
      \node[color=black] at (axis cs:5,6) {\footnotesize $0.06$};
      \node[color=black] at (axis cs:6,6) {\footnotesize $0.21$};
      \node[color=black] at (axis cs:7,6) {\footnotesize $0.15$};
      \node[color=black] at (axis cs:8,6) {\footnotesize $0.19$};
      
      \node[color=black] at (axis cs:0,7) {\footnotesize $0.07$};
      \node[color=black] at (axis cs:1,7) {\footnotesize $0.18$};
      \node[color=black] at (axis cs:2,7) {\footnotesize $0.19$};
      \node[color=black] at (axis cs:3,7) {\footnotesize $0.26$};
      \node[color=white] at (axis cs:4,7) {\footnotesize $0.38$};
      \node[color=black] at (axis cs:5,7) {\footnotesize $0.06$};
      \node[color=black] at (axis cs:6,7) {\footnotesize $0.17$};
      \node[color=black] at (axis cs:7,7) {\footnotesize $0.19$};
      \node[color=black] at (axis cs:8,7) {\footnotesize $0.14$};
      
      \node[color=black] at (axis cs:0,8) {\footnotesize $0.01$};
      \node[color=black] at (axis cs:1,8) {\footnotesize $0.02$};
      \node[color=black] at (axis cs:2,8) {\footnotesize $0.05$};
      \node[color=black] at (axis cs:3,8) {\footnotesize $0.04$};
      \node[color=black] at (axis cs:4,8) {\footnotesize $0.04$};
      \node[color=black] at (axis cs:5,8) {\footnotesize $0.01$};
      \node[color=black] at (axis cs:6,8) {\footnotesize $0.02$};
      \node[color=black] at (axis cs:7,8) {\footnotesize $0.02$};
      \node[color=black] at (axis cs:8,8) {\footnotesize $0.23$};

      \clip (current bounding box.south west) rectangle (current bounding box.north east);
  \end{axis}
  
\end{tikzpicture}
    \caption{Jaccard similarity between training sets ($y$-axis) and test sets ($x$-axis).}
    \label{fig:jacs}
\end{figure}

Both Table~\ref{tab:dsc} and Figure~\ref{fig:jacs} illustrate stark differences; not only across domains but more importantly, between in-domain training and test sets. Most do not exceed a Jaccard similarity coefficient over 0.20 (Figure~\ref{fig:jacs}), implying a large part of their vocabularies do not overlap. This contrast is not necessarily problematic for classification; however, it does hamper learning a general representation for the negative class. It also clearly illustrates how even more disjoint $D_{twX}$ (collected by trace queries) and $D_{tox}$ are from the rest of the corpora and splits. Finally, the descriptives (Table~\ref{tab:dsc}) further show significant differences in size, message length, class balance, and type/token ratios (i.e., writing level). In conclusion, it can be assumed that the language use in both positive as negative instances will vary significantly, and that it will be challenging to model in-domain, and generalize out-of-domain.

\section{Experimental Setup} \label{sec:exp}

We attempt to address the following hypotheses posited in Section~\ref{sec:oos}:

\begin{itemize}
    \item Hypothesis \textbf{1}: As researchers can only rely on scarce data of public bullying, their samples are assumed to be underpowered in terms of accurately representing the substantial language variation between platforms, both in normal language use and bullying-specific language use.
     \item Hypothesis \textbf{2}: Given knowledge from the literature, it is unlikely that single messages capture the full complexity of bullying events. Cyberbullying instances in the considered corpora are therefore expected to be largely biased, only reflecting a limited dimension of bullying.
     \item Hypothesis \textbf{3}: With control over data generation and structure, crowdsourcing bullying content potentially decreases the influence of domain-specific language use, allows for richer representations, and alleviates data scarcity.
\end{itemize}
\noindent
Accordingly, we propose five main experiments. Experiments I (Hypothesis \textbf{1}) and III (Hypothesis \textbf{2}) deal with the problem of generalizability, whereas Experiment II (Hypothesis \textbf{1}) and V (Hypothesis \textbf{3}) will both propose a solution for restricted data collection. Experiment IV will reproduce a selection of state-of-the-art models for cyberbullying detection and subject them to our cross-domain evaluation, to be compared against our baselines.

\subsection{Experiment I: Cross-Domain Evaluation} \label{subs:crit}

In this experiment, we introduce the cross-domain evaluation framework, which will be extended in all other experiments. For this, we initially perform a many-to-many evaluation of a given model (baseline or otherwise) trained individually on all available data sources, split in train and test. In later experiments, we extend this with a one-to-many evaluation. This setup implies that (i) we fit our model on some given corpus' training portion and evaluate prediction performance on all available corpora their test portions (many-to-many) individually. Furthermore, we (ii) fit on all corpora their train portions combined, and evaluate on all their test portions individually (one-to-many). In sum, we report on `small' models trained on each corpus individually, as well as a `large' one trained on them combined, for each test set individually.

For every experiment, hyper-parameter tuning was conducted through an exhaustive grid search, using nested cross-validation (with ten inner and three outer folds) on the training set to find the optimal combination of the given parameters. Any model selection steps were based on the evaluation of the outer folds. The best performing model was then refitted on the full training set ($90\%$ of the data) and applied to the test set ($10\%$). All splits (also during cross-validation) were made in a stratified fashion, keeping the label distributions across splits similar to the whole set.\footnote{Indices (similar to any other random components) were fixed by the same seed for all experiments.} Henceforth, all experiments in this section can be assumed to follow this setup.

\begin{table}[t!]
      \centering
      \footnotesize
      \begin{tabular}{lll}
      \toprule
      \textsc{\lowercase{Part}} & \textsc{\lowercase{Params}} & \textsc{\lowercase{Values}}   \\ 
      \midrule
      BoW  & range & $(1, 1)$, $(1, 2)$, $(1, 3)$     \\
                    & level          & words                       \\
      \midrule
      SVM  & weight $y$   & default, balanced                    \\
                               & loss           & hinge, square hinge \\
                               & $C$            & $1\mathrm{e}{-3}$, $1\mathrm{e}{-2}$, $\ldots$, $1\mathrm{e}{2}$, $1\mathrm{e}{3}$ \\
      \bottomrule
    \end{tabular}
     \caption{SVM \texttt{baseline} and NBSVM grid used in hyper-parameter search.}
    \label{tab:hyp}
\end{table}

The many-to-many evaluation framework intends to test Hypothesis \textbf{1} (Section \ref{subs:scarcity}), relating to language variation and cross-domain performance of cyberbullying detection. To facilitate this, we employ an initial \texttt{baseline} model: Scikit-learn's \citep{DBLP:journals/jmlr/PedregosaVGMTGBPWDVPCBPD11} Linear Support Vector Machine (SVM) \citep{DBLP:journals/ml/CortesV95,DBLP:journals/jmlr/FanCHWL08} implementation trained on binary BoW features, tuned using the grid shown in Table~\ref{tab:hyp}, based on \cite{10.1371/journal.pone.0203794}. Given its use in previous research, it should prove a strong candidate. To ascertain out-of-domain performance compared to this baseline, we report test score averages across all test splits, excluding the set the model was trained on (in-domain).

Consequently, we add an evaluation criterion to that of related work: when introducing a novel approach to cyberbullying detection, it should not only perform best in-domain for the majority of available corpora, but should also achieve the highest out-of-domain performance on average to classify as a robust method. It should be noted that the selected corpora for this chapter are not all optimally representative. The tests in our experiments should, therefore, be seen as a proposal to improve the task evaluation.

\subsection{Experiment II: Gauging Domain Influence}

In an attempt to overcome domain restrictions on language use, and to further solidify our tests regarding Hypothesis \textbf{1}, we aim to improve the baselines' performance through changing our representations in three distinct ways: i) merging all available training sets (as to simulate a large, diverse corpus), ii) by aggregating instances on user-level, and iii) using state-of-the-art language representations over simple BoW features in all settings. We define these experiments as such:

\paragraph{Volume and Variety} Some corpora used for training are relatively small, and can thus be assumed insufficient to represent held-out data (such as the test sets). One could argue that this can be partially mitigated through simply collecting more data or training on multiple domains. To simulate such a scenario, we merge all available cyberbullying-related training splits (creating $D_{all}$), which then corresponds to the one-to-many setting of the evaluation framework. The hope is that corpora similar in size or content (the Twitter sets, Ask.fm and Formspring, YouTube and Myspace) would benefit from having more (related) data available. Additionally, training a large model on its entirety facilitates a catch-all setting for assessing the average cross-domain performance of the full task (i.e. across all test sets when trained on all available corpora). This particular evaluation will be used in Experiment IV (replication) for model comparison.

\paragraph{Context Change} Practically all corpora, save for MySpace and YouTube, have annotations based on short sentences, which is particularly noticeable in Table~\ref{tab:dsc}. This one-shot (i.e., based on a single message) method of classifying cyberbullying provides minimal content (and context) to work with. It does therefore not follow the definition of cyberbullying---as previously discussed in Section~\ref{subs:task}. As a preliminary simulation\footnote{Preferably, one would want to collect data on profile level by design. The corpora available were not specifically collected this way, making our set-up an approximation of such a setting.} of adding (richer) context, we merge the profiles of $D_{ask}$ and (batches of) $D_{frm}$ into single context instances (creating $C_{ask}$ and $C_{frm}$, see Section~\ref{sec:data}). This allows us to compare models trained on larger contexts directly to that of single messages, and evaluate how context restrictions affect performance on the task in general, as well as cross-domain.

\paragraph{Improving Representations}

Word embeddings as language representation have been demonstrated to yield significant performance gains for a multitude of NLP-related tasks \citep{DBLP:journals/jmlr/CollobertWBKKK11}. Given the general lack of training data---including negative instances for many corpora---word features (and weightings) trained on the available data tend to be a poor reflection of the language use on the platform itself, let alone other social media platforms. Therefore, pre-trained representations provide features that, in theory, should perform better in cross-domain settings. We consider two off-the-shelf embedding models per language that are suitable for the task at hand: for English, averaged 200-dimensional GloVe \citep{pennington-etal-2014-glove} vectors trained on Twitter,\footnote{\url{https://nlp.stanford.edu/projects/glove/} (\texttt{v1.2})} and DistilBERT \citep{DBLP:journals/corr/abs-1910-01108} sentence embeddings \citep{devlin-etal-2019-bert}.\footnote{\url{https://github.com/huggingface/transformers} (\texttt{1d646ba})} For Dutch, \texttt{fastText} embeddings \citep{bojanowski-etal-2017-enriching} trained on Wikipedia\footnote{\url{https://github.com/facebookresearch/fastText/blob/master/pretrained-vectors.md} (\texttt{2665eac})} and \texttt{word2vec} \citep{DBLP:journals/corr/abs-1301-3781,DBLP:conf/nips/MikolovSCCD13} embeddings\footnote{\url{https://github.com/clips/dutchembeddings} (\texttt{1e3d528})} \citep{tulkens-etal-2016-evaluating} trained on the COrpora from the Web (COW) corpus \citep{schafer-bildhauer-2012-building} embeddings. GLoVe, \texttt{fastText}, and \texttt{word2vec} embeddings were processed using Gensim\footnote{\url{https://radimrehurek.com/gensim/index.html} (\texttt{v3.4})} \citep{rehurek_lrec}.

As an additional baseline for this section, we include the Naive Bayes Support Vector Machine (NBSVM) from \cite{wang-manning-2012-baselines}, which should offer competitive performance on text classification tasks.\footnote{The implementations for these models can be found in our repository.} This model also served as a baseline for the Kaggle challenge related to $D_{tox}$.\footnote{\url{https://kaggle.com/jhoward/nb-svm-strong-linear-baseline}} NBSVM uses tf$\cdot$idf-weighted uni and bi-gram features as input, with a minimum document frequency of $3$, and corpus prevalence of $90\%$. The idf values are smoothed and tf scaled sublinearly ($1 + \log($tf$)$). These are then weighted by their log-count ratios derived from Multinomial Naive Bayes. 

Tuning of both embeddings and NB representation classifiers is done using the same grid as Table~\ref{tab:hyp}, however replacing $C$ with $[1, 2, 3, 4, 5, 10, 25, 50, 100, 200, 500]$. Lastly, we opted for Logistic Regression (LR), primarily as this was used in the NBSVM implementation mentioned above, as well as \texttt{fastText}. Moreover, we found SVM using our grid to perform marginally worse using these features, and fine-tuning DistilBERT using a fully connected layer \citep[similar to, e.g.,][]{10.1007/978-3-030-32381-3_16} to yield similar performance. The embeddings were not fine-tuned for the task. While this could potentially increase performance, it complicates direct comparison to our baselines---we leave this for Experiment IV.

\subsection{Experiment III: Aggression Overlap} 

In previous research using fine-grained labels for cyberbullying classification \cite[e.g.,][]{10.1371/journal.pone.0203794} it was observed that cyberbullying classifiers achieve the lowest error rates on blatant cases of aggression (cursing, sexual talk, and threats), an idea that was further adopted by \cite{ROSA2019333}. To empirically test Hypothesis \textbf{2} (see Section~\ref{subs:task})---related to the bias present in the available positive instances---we adapt the idea of running a profanity baseline from this previous work. However, rather than relying on look-up lists containing profane words, we expand this idea by training a separate classifier on toxicity detection ($D_{tox}$) and seeing how well this performs on our bullying corpora (and vice-versa). For the corpora with fine-grained labels, we can further inspect and compare the bullying classes captured by this model. 

We argue that high test set performance overlap of a toxicity detection model with models trained on cyberbullying detection gives strong evidence of nuanced aspects of cyberbullying not being captured by such models. Notably, in line with \cite{ROSA2019333}, that the current operationalization does not significantly differ from the detection of online aggression (or toxicity)---and therefore does not capture actual cyberbullying. Given enough evidence, both issues should be considered as crucial points of improvement for the development of classifiers in this domain. 

\subsection{Experiment IV: Replicating State-of-the-Art}

For this experiment, we include two architectures that achieved state-of-the-art results on cyberbullying detection. As a reference neural network model for language-based tasks, we used a Bidirectional \citep{DBLP:journals/tsp/SchusterP97,PMID:10743560} Long Short-Term Memory network \citep{DBLP:journals/neco/HochreiterS97,DBLP:journals/jmlr/GersSS02} (BiLSTM), partly reproducing the architecture from \cite{DBLP:conf/ecir/AgrawalA18}. We then attempt to reproduce the Convolutional Neural Network (CNN) \citep{kim-2014-convolutional} used in both \cite{DBLP:conf/ijcnn/RosaM0CC18} and \cite{DBLP:conf/ecir/AgrawalA18}, and the Convolutional LSTM (C-LSTM) \citep{DBLP:journals/corr/ZhouSLL15b} used in \cite{DBLP:conf/ijcnn/RosaM0CC18}. As \citeauthor{DBLP:conf/ijcnn/RosaM0CC18} do not report essential implementation details for these models (batch size, learning rate, number of epochs), there is no reliable way to reproduce their work. We will, therefore, take \cite{DBLP:conf/ecir/AgrawalA18} their implementation for the BiLSTM and CNN as the initial setup. Given that this work is available open-source, we run the exact architecture (including SMOTE) in our Experiment I and II evaluations. The architecture-specific details are as follows:

\paragraph{Reproduction} 

We initially adopt the basic implementation\footnote{\url{https://github.com/sweta20/Detecting-Cyberbullying-Across-SMPs/blob/master/DNNs.ipynb}}  by \cite{DBLP:conf/ecir/AgrawalA18}: randomly initialized embeddings with a dimension of $50$ (as the paper did not find significant effects of changing the dimension, nor initialization), run for $10$ epochs with a batch size of $128$, dropout probability of $0.25$, and a learning rate of 0.01. Further architecture details can be found in our repository.\footnote{\url{https://github.com/cmry/amica/blob/master/neural.py}} We also run a variant with SMOTE on, and one from the provided notebooks directly.\footnote{Note that this is for testing reproduction only, as it is not subjected to the same evaluation framework.} This and following neural models were run on an NVIDIA Titan X Pascal, using Keras \citep{chollet2015keras} with Tensorflow \citep{tensorflow2015whitepaper} as backend.

\paragraph{BiLSTM} For our own implementation of the BiLSTM, we minimally changed the architecture from \cite{DBLP:conf/ecir/AgrawalA18}, only tuning using a grid on batch size $[32,$ $64,$ $128,$ $256]$, embedding size $[50,$ $100,$ $200,$ $300]$, and learning rate $[0.1,$ $0.01,$ $0.05,$ $0.001,$ $0.005]$. Rather than running for ten epochs, we use a validation split (10\% of the train set) and initiate early stopping when the validation loss does not go down after three epochs. Hence---and in contrast to earlier experiments---we do not run the neural models in 10-fold cross-validation, but a straightforward 2-fold train and test split where the latter is $10\%$  of a given corpus. Again, we are predominantly interested in confirming statements made in earlier work; namely, that for this particular setting, tuning of the parameters does not meaningfully affect performance. 

\paragraph{CNN} 

We use the same experimental setup as for the BiLSTM. The implementations of \cite{DBLP:conf/ecir/AgrawalA18} and \cite{DBLP:conf/ijcnn/RosaM0CC18} use filter window sizes of $3$, $4$, and $5$---max pooled at the end. Given that the same grid is used, the word embedding sizes are varied and weights trained (whereas \cite{DBLP:conf/ijcnn/RosaM0CC18} use 300-dimensional pre-trained embeddings). Therefore, for direct performance comparisons, \cite{DBLP:conf/ecir/AgrawalA18} their results will be used as a reference. As CNN-based architectures for text classification are often also trained on character level, we include a model variant with this input as well. 

\paragraph{C-LSTM} For this architecture, we take an open-source text classification survey implementation.\footnote{\url{https://github.com/bicepjai/Deep-Survey-Text-Classification/}} This uses filter windows of $[10, 20,$ $30, 40, 50]$, $64$-dimensional LSTM cells and a final $128$-dimensional dense layer. Please refer to our repository for additional implementation details---for this and previous architectures.

\subsection{Experiment V: Crowdsourced Data}

Following up on the proposed shortcomings of the currently available corpora in Hypotheses \textbf{1} and \textbf{2}, we propose the use of a crowdsourcing approach to data collection. In this experiment, we will repeat Experiment I and II with the best out-of-domain classifier from the above evaluations with three (Dutch)\footnote{On account of the synthetic data being available in Dutch only. Experiment III was not repeated, as there is no equivalent toxicity dataset available in this language.} datasets: $D_{ask\_nl}$; the Dutch part of the Ask.fm dataset used before, $D_{sim\_nl}$; our synthetic, crowdsourced cyberbullying data, and lastly $D_{don\_nl}$; a small donated cyberbullying test set with messages from various platforms (full overview and description of these three can be found in Section~\ref{sec:data}). The only notable difference to our setup for this experiment is that we never use $D_{don\_nl}$ as training data. Therefore, rather than $D_{all}$, the Ask.fm corpus is merged with the crowdsourced cyberbullying data to make up the $D_{comb}$ set. 

\section{Results and Discussion}

We will now cover results per experiment, and to what extent these provide support for the hypotheses posed in Section~\ref{sec:oos}. As most of these required backward evaluation (e.g., Experiment III was tested on sets from Experiment I), the results of Experiment I-III are compressed in Table~\ref{tab:base}.  Table~\ref{tab:archs} comprises the \emph{Improving Representations} part of Experiment II (under `word2vec' and `DistilBERT') along with the preprocessing results effect of our baselines. The results of Experiment V can be found in Table~\ref{tab:neural}. For brevity, the latter two only report on the in-domain scores, and include the out-of-domain \emph{averages} for the $D_{all}$ models for comparison, and $D_{tox}$ averages in Table~\ref{tab:neural}.

\subsection{Experiment I}

\begin{table}[t!]
        \resizebox{\textwidth}{!}{%
        \begin{tabular}{lrrrrrrrrrr}
            \toprule
            Train         &                             \multicolumn{6}{c}{T1}                         & Avg         & \multicolumn{2}{c}{T2} & T3 \\
                          \cmidrule{2-7} \cmidrule{8-8} \cmidrule{9-10} \cmidrule{11-11}
                          & $D_{twB}$ & $D_{frm}$        & $D_{msp}$     & $D_{ytb}$ & $D_{ask}$ & $D_{twX}$ &         & $C_{frm}$ & $C_{ask}$ & $D_{tox}$  \\ 
            \midrule
            $D_{twB}$    & $.417$          & $.308$          & $.000$          & $.122$      & $.298$      & $.051$      & $.153$      & $.131$      & $.158$      & $.349$ \\
            $D_{frm}$    & $.423$          & $.454$          & $.042$          & $.379$      & $.418$      & $.041$      & $.321$      & $.682$      & $.259$      & $.465$ \\
            $D_{msp}$    & $.120$          & $.176$          & {\bf.941}       & $.324$      & $.168$      & $.043$      & $.197$      & $.364$      & $.185$      & $.185$ \\
            $D_{ytb}$    & $.074$          & $.160$          & $.375$          & $.365$      & $.138$      & $.000$      & $.183$      & $.338$      & $.197$      & $.140$ \\
            $D_{ask}$    & $.493$          & $.444$          & $.211$          & \bf{.421} & \bf{.561}     & $.139$      & $.351$      & $.389$      & $.357$      & $.584$ \\ 
            $D_{twX}$    & $.049$          & $.131$          & $.184$          & $.175$      & $.077$      & {\bf.508} &   $.205$      & $.496$      & $.325$      & $.082$ \\ 
            \midrule
            $D_{all}$    & \purp{\bf.524}  & \purp{\bf.473}  & \purp{\bf.941}  & \purp{$.397$} & \purp{$.553$} & \purp{$.194$} & \purp{\bf.557} & \purp{\bf.780} & \purp{$.570$} & \purp{$.587$} \\
            \midrule
            $C_{frm}$    & $.152$          & $.253$          & $.143$          & $.286$      & $.136$      & $.126$      & $.214$      & $.758$      & $.400$      & $.372$ \\ 
            $C_{ask}$    & $.286$          & $.237$          & $.359$          & $.244$      & $.356$      & $.107$      & $.310$      & $.582$      & {\bf.579} & $.280$ \\
            $D_{tox}$    & $.343$          & $.373$          & $.449$          & $.335$      & $.443$      & $.149$      & $.389$      & $.628$      & $.539$ & \bf{.806} \\
            \bottomrule
        \end{tabular}
        }
        \caption{Cross-corpora positive class $F_1$ scores for Experiment I (T1), II (T2), and III (T3). Models are fitted on the training proportion of the corpora row-wise, and tested column-wise. The out-of-domain average (Avg) excludes test performance of the parent training corpus. The best overall test score is noted in bold, the best out-of-domain performance in gray.} \label{tab:base}
\end{table}

Looking at Table~\ref{tab:base}, the upper group of rows under T1 represents the results for Experiment I. We posed in Hypothesis \textbf{1} that samples are underpowered regarding their representation of the language variation between platforms, both for bullying and normal language use. The data analysis in Section~\ref{subs:sum} showed minimal overlap between domains in vocabulary and notable variances in numerous aspects of the available corpora. Consequently, we raised doubts regarding the ability of models trained on these individual corpora to generalize to other corpora (i.e., domains).

Firstly, we consider how well our \texttt{baseline} performed on the \emph{in-domain test sets}. For half of the corpora, it achieves the highest performance on these specific sets (i.e., the test set portion of the data the model was trained on). More importantly, this entails that for four of the other sets, models trained on other corpora perform equal or better. Particularly the effectiveness of $D_{ask}$ was in some cases surprising; the YouTube corpus by \cite{DBLP:conf/ai/DadvarTJ14} ($D_{ytb}$), for example, contains much longer instances (see Table~\ref{tab:dsc}). 

It must be noted, though, that the baseline was selected from work on the Ask.fm corpus \citep{10.1371/journal.pone.0203794}. This data is one of the more diverse datasets (and largest) with exclusively short messages; therefore, one could assume a model trained on this data would work well on both longer and shorter instances. It is however also likely that particularly this baseline (binary word features) therefore enforces the importance of more shallow features. This will be further explored in Experiments II and III. 

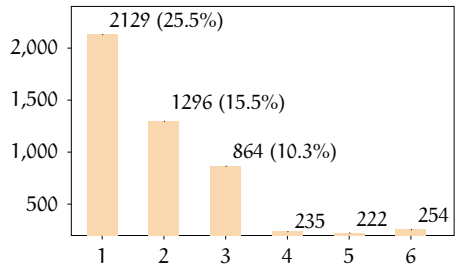
\begin{figure}[t!]
    \centering
    \begin{tikzpicture}
\pgfmathsetlengthmacro\MajorTickLength {
  \pgfkeysvalueof{/pgfplots/major tick length} * 0.4
}
\begin{axis}[
width=7.3cm,
height=5cm,
log basis y={10},
tick align=outside,
tick pos=left,
x grid style={white!69.01960784313725!black},
xmin=-0.5, xmax=5.7,
xtick style={color=black},
xtick={0,1,2,3,4,5},
xlabel={\strut},
xticklabel style = {rotate=0.0,font=\footnotesize},
xticklabels={$1$,$2$,$3$,$4$,$5$,$6$},
y grid style={white!69.01960784313725!black},
ymin=198.272206672272, ymax=2400.78342548652,
ytick style={color=black,font=\footnotesize},
major tick length=\MajorTickLength,
every tick label/.append style={font=\footnotesize},
]
\draw[fill=DrawPaleOrange,draw opacity=0] (axis cs:-0.25,0) rectangle (axis cs:0.25,2129);
\draw[fill=DrawPaleOrange,draw opacity=0] (axis cs:0.75,0) rectangle (axis cs:1.25,1296);
\draw[fill=DrawPaleOrange,draw opacity=0] (axis cs:1.75,0) rectangle (axis cs:2.25,864);
\draw[fill=DrawPaleOrange,draw opacity=0] (axis cs:2.75,0) rectangle (axis cs:3.25,235);
\draw[fill=DrawPaleOrange,draw opacity=0] (axis cs:3.75,0) rectangle (axis cs:4.25,222);
\draw[fill=DrawPaleOrange,draw opacity=0] (axis cs:4.75,0) rectangle (axis cs:5.25,254);
\draw[] (axis cs:0,2129) -- (axis cs:0,2129);
\node at (axis cs:0,2129)[
  scale=0.8,
  anchor=south west,
  text=black,
  rotate=0.0,
  yshift=-2pt
]{$2129$ ($25.5\%$)};
\draw[] (axis cs:1,1296) -- (axis cs:1,1296);
\node at (axis cs:1,1296)[
  scale=0.8,
  anchor=south west,
  text=black,
  rotate=0.0
]{$1296$ ($15.5\%$)};
\draw[] (axis cs:2,864) -- (axis cs:2,864);
\node at (axis cs:2,864)[
  scale=0.8,
  anchor=south west,
  text=black,
  rotate=0.0,
  yshift=-2pt
]{$864$ ($10.3\%$)};
\draw[] (axis cs:3,235) -- (axis cs:3,235);
\node at (axis cs:3,235)[
  scale=0.8,
  anchor=south west,
  text=black,
  rotate=0.0,
  yshift=-2pt
]{$235$};
\draw[] (axis cs:4,222) -- (axis cs:4,222);
\node at (axis cs:4,222)[
  scale=0.8,
  anchor=south west,
  text=black,
  rotate=0.0
]{$222$};
\draw[] (axis cs:5,254) -- (axis cs:5,254);
\node at (axis cs:5,254)[
  scale=0.8,
  anchor=south west,
  text=black,
  rotate=0.0
]{$254$};
\end{axis}

\end{tikzpicture}
    \caption{Test set occurrence frequencies (and percentages) of the top $5,000$ highest absolute feature coefficient values.}
    \label{fig:bar-top}
\end{figure}

For Experiment I, however, our goal was to assess the out-of-domain performance of these classifiers, not to maximize performance. For this, we turn to the \emph{Avg} column in Table~\ref{tab:base}. Between the \emph{top portion} of the Table, the $D_{ask}$ model performs best across all domains (achieving the highest scores on three of them, as mentioned above). The second-best model is trained on the Formspring data from \cite{6147681} ($D_{frm}$), akin to Ask.fm as a domain (question-answer style, option to post anonymously). It can be observed that almost all models perform the worst on the `bullying traces' Twitter corpus by \cite{xu-etal-2012-learning}, which was collected using queries. This result is relatively unsurprising, given the small vocabulary overlap with its test set shown in Figure~\ref{fig:jacs}. We also confirm in line with \cite{6147681} that the \textsc{caw} data from \cite{bayzick2011detecting} is unfit as a bullying corpus; achieving significant positive $F_1$-scores with a baseline, generalizing poorly and proving difficult as a test set.

Additionally, we observe that even the best performing models yield between $.1$ and $.2$ lower $F_1$ scores on other domains, or a $15-30\%$ drop from the original score. To explain this, we look at how well important features generalize across test sets. As our \texttt{baseline} is a Linear SVM, we can directly extract all grams with positive coefficients (i.e., related to bullying). Figure~\ref{fig:bar-top} shows the frequency of the top $5,000$ features with the highest coefficient values. These can be observed to follow a Zipfian-like distribution, where the important features most frequently occur in one test set ($25.5\%$) only, which quickly drops off with increasing frequency. Conversely, this implies that over $75\%$ of the top $5,000$ features seen during training do not occur in any test instance, and only $3\%$ generalize across all sets. This coverage decreases to roughly $60\%$ and $4\%$ respectively for the top $10,000$, providing further evidence of the strong variation in bullying-specific language use.

\begin{figure}[t!]
    \centering
    \begin{tikzpicture}
\pgfmathsetlengthmacro\MajorTickLength {
  \pgfkeysvalueof{/pgfplots/major tick length} * 0.4
}
\begin{axis}[
height=4cm,
width=17cm,
tick align=outside,
tick pos=left,
x grid style={white!69.01960784313725!black},
xmin=-0.5, xmax=14.9,
xtick style={color=black},
xtick={0,1,2,3,4,5,6,7,8,9,10,11,12,13,14,15,16,17,18,19},
xticklabel style = {rotate=45.0, xshift=-0.05cm, yshift=0.15cm},
xticklabels={b*tch,f*ck,f*gg*t,stfu,sl*t,g*y,c*nt,f*g,wh*r*,*sshole,n*gg*,hate,tw*t,d*ck,*ss}, 
y grid style={white!69.01960784313725!black},
ymin=0, ymax=1.0,
ytick style={color=black},
major tick length=\MajorTickLength,
every tick label/.append style={font=\footnotesize}
]
\draw[fill=DrawRed,draw opacity=0] (axis cs:-0.25,0) rectangle (axis cs:0.25,0.486159664958781);
\draw[fill=DrawRed,draw opacity=0] (axis cs:0.75,0) rectangle (axis cs:1.25,0.397253179557668);
\draw[fill=DrawRed,draw opacity=0] (axis cs:1.75,0) rectangle (axis cs:2.25,0.391205443140379);
\draw[fill=DrawRed,draw opacity=0] (axis cs:2.75,0) rectangle (axis cs:3.25,0.384622402673902);
\draw[fill=DrawRed,draw opacity=0] (axis cs:3.75,0) rectangle (axis cs:4.25,0.376668541047829);
\draw[fill=DrawRed,draw opacity=0] (axis cs:4.75,0) rectangle (axis cs:5.25,0.361748956873954);
\draw[fill=DrawRed,draw opacity=0] (axis cs:5.75,0) rectangle (axis cs:6.25,0.355548188760811);
\draw[fill=DrawRed,draw opacity=0] (axis cs:6.75,0) rectangle (axis cs:7.25,0.348264270625347);
\draw[fill=DrawRed,draw opacity=0] (axis cs:7.75,0) rectangle (axis cs:8.25,0.342418077994587);
\draw[fill=DrawRed,draw opacity=0] (axis cs:8.75,0) rectangle (axis cs:9.25,0.321320944748114);
\draw[fill=DrawRed,draw opacity=0] (axis cs:9.75,0) rectangle (axis cs:10.25,0.319181918889777);
\draw[fill=DrawRed,draw opacity=0] (axis cs:10.75,0) rectangle (axis cs:11.25,0.28227852361097);
\draw[fill=DrawRed,draw opacity=0] (axis cs:11.75,0) rectangle (axis cs:12.25,0.272251151873976);
\draw[fill=DrawRed,draw opacity=0] (axis cs:12.75,0) rectangle (axis cs:13.25,0.269157546152352);
\draw[fill=DrawRed,draw opacity=0] (axis cs:13.75,0) rectangle (axis cs:14.25,0.268011651014183);
\path [draw=black, semithick]
(axis cs:0,0.0560842165069208)
--(axis cs:0,0.916235113410641);

\path [draw=black, semithick]
(axis cs:1,0.0955653650260613)
--(axis cs:1,0.698940994089275);

\path [draw=black, semithick]
(axis cs:2,0.0220664643172784)
--(axis cs:2,0.760344421963479);

\path [draw=black, semithick]
(axis cs:3,0.000380341105266302)
--(axis cs:3,0.768864464242538);

\path [draw=black, semithick]
(axis cs:4,-0.0116548788890098)
--(axis cs:4,0.764991960984667);

\path [draw=black, semithick]
(axis cs:5,-0.0927967613667052)
--(axis cs:5,0.816294675114612);

\path [draw=black, semithick]
(axis cs:6,-0.0854321211301985)
--(axis cs:6,0.796528498651821);

\path [draw=black, semithick]
(axis cs:7,-0.0956444615248115)
--(axis cs:7,0.792173002775506);

\path [draw=black, semithick]
(axis cs:8,-0.0979864792153251)
--(axis cs:8,0.782822635204499);

\path [draw=black, semithick]
(axis cs:9,-0.0120080354445238)
--(axis cs:9,0.654649924940752);

\path [draw=black, semithick]
(axis cs:10,0.0176483956400034)
--(axis cs:10,0.620715442139551);

\path [draw=black, semithick]
(axis cs:11,-0.0181840890003955)
--(axis cs:11,0.582741136222336);

\path [draw=black, semithick]
(axis cs:12,-0.072987181681835)
--(axis cs:12,0.617489485429786);

\path [draw=black, semithick]
(axis cs:13,-0.120224980010959)
--(axis cs:13,0.658540072315663);

\path [draw=black, semithick]
(axis cs:14,-0.111389098729857)
--(axis cs:14,0.647412400758223);






\draw[] (axis cs:0.1,0.586159664958781) -- (axis cs:0.1,0.586159664958781);
\node at (axis cs:0.1,0.586159664958781)[
  scale=0.8,
  anchor= west,
  text=black,
  rotate=0.0
]{$1141$};
\draw[] (axis cs:1.1,0.497253179557668) -- (axis cs:1.1,0.497253179557668);
\node at (axis cs:1.1,0.497253179557668)[
  scale=0.8,
  anchor= west,
  text=black,
  rotate=0.0
]{$2685$};
\draw[] (axis cs:2.1,0.491205443140379) -- (axis cs:2.1,0.491205443140379);
\node at (axis cs:2.1,0.491205443140379)[
  scale=0.8,
  anchor= west,
  text=black,
  rotate=0.0
]{$11$};
\draw[] (axis cs:3.1,0.484622402673902) -- (axis cs:3.1,0.484622402673902);
\node at (axis cs:3.1,0.484622402673902)[
  scale=0.8,
  anchor= west,
  text=black,
  rotate=0.0
]{$247$};
\draw[] (axis cs:4.1,0.476668541047829) -- (axis cs:4.1,0.476668541047829);
\node at (axis cs:4.1,0.476668541047829)[
  scale=0.8,
  anchor= west,
  text=black,
  rotate=0.0
]{$119$};
\draw[] (axis cs:5.1,0.461748956873954) -- (axis cs:5.1,0.461748956873954);
\node at (axis cs:5.1,0.461748956873954)[
  scale=0.8,
  anchor= west,
  text=black,
  rotate=0.0
]{$11$};
\draw[] (axis cs:6.1,0.455548188760811) -- (axis cs:6.1,0.455548188760811);
\node at (axis cs:6.1,0.455548188760811)[
  scale=0.8,
  anchor= west,
  text=black,
  rotate=0.0
]{$130$};
\draw[] (axis cs:7.1,0.448264270625347) -- (axis cs:7.1,0.448264270625347);
\node at (axis cs:7.1,0.448264270625347)[
  scale=0.8,
  anchor= west,
  text=black,
  rotate=0.0
]{$409$};
\draw[] (axis cs:8.1,0.442418077994587) -- (axis cs:8.1,0.442418077994587);
\node at (axis cs:8.1,0.442418077994587)[
  scale=0.8,
  anchor= west,
  text=black,
  rotate=0.0
]{$279$};
\draw[] (axis cs:9.1,0.421320944748114) -- (axis cs:9.1,0.421320944748114);
\node at (axis cs:9.1,0.421320944748114)[
  scale=0.8,
  anchor= west,
  text=black,
  rotate=0.0
]{$202$};
\draw[] (axis cs:10.1,0.419181918889777) -- (axis cs:10.1,0.419181918889777);
\node at (axis cs:10.1,0.419181918889777)[
  scale=0.8,
  anchor= west,
  text=black,
  rotate=0.0
]{$621$};
\draw[] (axis cs:11.1,0.38227852361097) -- (axis cs:11.1,0.38227852361097);
\node at (axis cs:11.1,0.38227852361097)[
  scale=0.8,
  anchor= west,
  text=black,
  rotate=0.0
]{$1082$};
\draw[] (axis cs:12.1,0.372251151873976) -- (axis cs:12.1,0.372251151873976);
\node at (axis cs:12.1,0.372251151873976)[
  scale=0.8,
  anchor= west,
  text=black,
  rotate=0.0
]{$65$};
\draw[] (axis cs:13.1,0.369157546152352) -- (axis cs:13.1,0.369157546152352);
\node at (axis cs:13.1,0.369157546152352)[
  scale=0.8,
  anchor= west,
  text=black,
  rotate=0.0
]{$182$};
\draw[] (axis cs:14.1,0.368011651014183) -- (axis cs:14.1,0.368011651014183);
\node at (axis cs:14.1,0.368011651014183)[
  scale=0.8,
  anchor= west,
  text=black,
  rotate=0.0
]{$125$};
\draw[] (axis cs:15.1,0.348514820985584) -- (axis cs:15.1,0.348514820985584);
\end{axis}

\end{tikzpicture}
    \caption{Top 15 \emph{test set} words with the highest average coefficient values across all classifiers (minus the model trained on $D_{tox}$). Error bars represent standard deviation. Each coefficient value is only counted once per test set. The frequency of the words is listed in the annotation.}
    \label{fig:bar-error}
\end{figure}
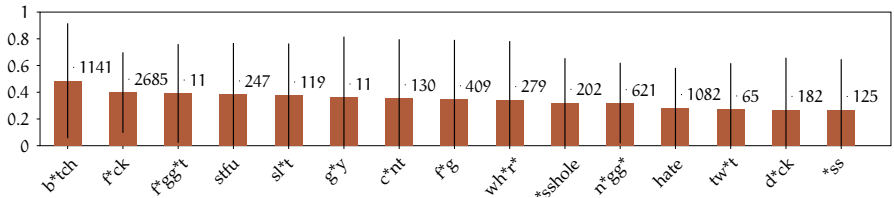

Figure~\ref{fig:bar-error} also indicates that the coefficient values are highly unstable across test sets, with most having roughly a $0.4$ standard deviation. Note that these coefficient values can also flip to negative for particular sets, so for some features, the range goes from associated with the other class to highly associated with bullying. Given the results of Table~\ref{tab:base}, and Figures~\ref{fig:bar-top} and \ref{fig:bar-error}, we can conclude that our \texttt{baseline} model shows not to generalize out-of-domain. Given the quantitative and qualitative results reported on in this Experiment, this particular setting partly supports Hypothesis \textbf{1}. 

\subsection{Experiment II}

The results for this experiment can be predominantly found in Table~\ref{tab:base} (middle and lower parts, and T2 in particular), and partly in Table~\ref{tab:archs} (word2vec, DistilBERT). In this experiment, we seek to further test Hypothesis \textbf{1} by employing three methods: merging all cyberbullying data to increase \emph{volume and variety}, aggregating on context level for a \emph{context change}, and \emph{improving representations} through pre-trained word embedding features. These are all reasonably straightforward methods that can be employed in an attempt to mitigate data scarcity.

\paragraph{Volume and Variety} The results for this part are listed under $D_{all}$ in Table~\ref{tab:base}. For all the following experiments, we now focus on the full results table (including that of Experiment I) and see which individual classifiers generalize best across all test sets (highlighted in gray). The \emph{Avg} column shows that our `big' model trained on all available corpora\footnote{This average excludes toxicity data from $D_{tox}$, which we found when added to substantially decrease performance on all domains, except for $D_{twX}$ and $C_{frm}$. Note that it also includes scopes from the \emph{context change} experiment.} achieves second-best performance on half of the test sets and best on the other half. More importantly, it has the highest average out-of-domain performance, without competition on any test set. These observations imply that for the \texttt{baseline} setting, an ensemble model of different smaller classifiers should not be preferred over the big model. Consequently, it can be concluded that collecting more data seems generally beneficial.

\begin{table}[t!]
    \caption{Examples of uni-gram weights according to the \texttt{baseline} SVM trained $D_{all}$, tested on $D_{twB}$ and $D_{ask}$. Words in red are associated with bullying, words in green with neutral content. The color intensity is derived from the strength of the SVM coefficients per feature (most are near zero). Black boxes indicate OOV words. Labels are divided between the gold standard ($y$) and predicted ($\hat{y}$) labels, \faThumbsODown{} for bullying content, \faThumbsOUp{} for neutral. }
    \footnotesize
    \centering
    \label{tab:qual}
    \begin{tabular}{llp{4.6cm}p{4.6cm}}
        \toprule
        $y$             & $\hat{y}$         &  $D_{twB}$     &  $D_{ask}$  \\
        \midrule
        \faThumbsOUp    & \faThumbsODown    & \colorbox{pc!1!}{about} \colorbox{pc!3!}{to} \colorbox{nc!42!}{leave} \colorbox{nc!4!}{this} \colorbox{pc!7!}{school} \colorbox{pc!1!}{library} \colorbox{pc!1!}{and} \colorbox{pc!3!}{take} \colorbox{nc!3!}{my} \colorbox{nc!62!}{*ss} \framebox{homeeee} & bigerrr ? \colorbox{pc!2!}{how} \colorbox{nc!3!}{much} ? \colorbox{pc!3!}{its} \colorbox{nc!4!}{gon} \colorbox{pc!4!}{na} \colorbox{nc!15!}{touch} \colorbox{pc!2!}{the} \colorbox{pc!5!}{sky} \colorbox{nc!0!}{?} a \colorbox{pc!5!}{wonder} \colorbox{nc!84!}{d*ck} \colorbox{nc!0!}{?}  \\
        \faThumbsODown   & \faThumbsODown   & \colorbox{nc!19!}{you} \colorbox{nc!63!}{p*ss} \colorbox{nc!8!}{me} \colorbox{nc!28!}{off} \colorbox{nc!1!}{so} \colorbox{nc!3!}{much} . & \colorbox{nc!0!}{r} \colorbox{nc!0!}{u} \colorbox{nc!0!}{a} \colorbox{nc!62!}{r*t*rd} \colorbox{nc!3!}{liam} \colorbox{nc!20!}{mate} \colorbox{nc!94!}{f*ck} \colorbox{nc!28!}{off}  \\
        \faThumbsODown   & \faThumbsOUp     & \framebox{@username} \colorbox{nc!0!}{i} \colorbox{nc!6!}{will} \colorbox{nc!11!}{skull} \colorbox{nc!2!}{drag} \colorbox{nc!19!}{you} \colorbox{pc!2!}{across} \colorbox{nc!1!}{campus} . &  \colorbox{nc!1!}{h*} \colorbox{nc!1!}{of} \colorbox{nc!1!}{me} \colorbox{nc!3!}{xoxoxoxoxoxoox} \\
        \bottomrule
    \end{tabular}
\end{table}

However, a qualitative analysis of the predictions made by this model clearly shows lingering limitations (see Table~\ref{tab:qual}). These three randomly-picked examples give a clear indication of the focus on blatant profanity (such as \texttt{d*ck}, \texttt{p*ss}, and \texttt{f*ck}). Especially combinations of words that in isolation might be associated with bullying content (\texttt{leave}, \texttt{touch}) tend to confuse the model. It also fails to capture more subtle threats (\texttt{skull drag}) and infrequent variations (\texttt{h*}). Both of these structural mistakes could be mitigated by providing more context that potentially includes either more toxicity or more examples of neutral content to decrease the impact of single curse words---hence, the next experiment.

\paragraph{Context Change} As for access to context scopes, we are restricted to the Ask.fm and Formspring data ($C_{frm}$ and $C_{ask}$ in Table~\ref{tab:base}). Nevertheless, in both cases, we see a noticeable increase for in-domain performance: a positive $F_1$ score of $.579$ for context scope versus $.561$ on Ask.fm, and $.758$ versus $.454$ on Formspring respectively. This increase implies that considering message-level detection for both individual sets should be preferred. On the other hand, however, these longer contexts do perform worse on out-of-domain sets. This can be partly explained due to the fact that including more data (therefore moving the data to profile, or conversation level) shifts the task to identifying bullying conversations, or profiles of victims. While variation will be higher, chances also increase that multiple single bullying messages will be captured in a one context. This would therefore allow learning the distinction between a profile or conversation with predominately neutral messages with a single toxic message (which might therefore be harmless), and one where there are multiple toxic messages, increasing the severity.

The change in scope clearly influences which features are deemed important.  On manual inspection, averaging feature importances of all \texttt{baseline} models on their in-domain test sets, the top $500$ most important features consist of $63\%$ profane words. For the models trained on Ask.fm and Formspring specifically ($D_{ask}$ and $D_{frm}$), this is an average of $42\%$. Strikingly, for the models trained on context scopes ($C_{frm}$ and $C_{ask}$), this percentage significantly reduced to $11\%$; many of their important bi-gram features include \texttt{you}, topics such as \texttt{dating}, \texttt{boys}, \texttt{girls}, and \texttt{girlfriend} occur, yet also positive words such as (\texttt{are}) \texttt{beautiful}---the latter of which could indicate messages from friends (defenders). This change is to an extent expected as by changing the scope, the task shifts to classifying profiles that are bullied, thus showing more diverse bullying characteristics.

These results provide evidence for extending classification to contexts to be a worthwhile platform-specific setting to pursue. However, we can conversely draw the same conclusions as Experiment I; that including direct context does not overcome the task its general domain limitations, therefore further supporting Hypothesis \textbf{1}. A solution to this could be improving upon the BoW features through more general representations of language, as found in word embeddings.

\paragraph{Improving Representations} 

\begin{table}[t!]
        \footnotesize
        \resizebox{\textwidth}{!}{%
        \begin{tabular}{lrrrrrrrrrr}
            \toprule
            Repr          &                             \multicolumn{6}{c}{T1}                             & Avg        & \multicolumn{2}{c}{T2} & T3 \\
                          \cmidrule{2-7} \cmidrule{8-8} \cmidrule{9-10} \cmidrule{11-11}
                              & $D_{twB}$ & $D_{frm}$    & $D_{msp}$   & $D_{ytb}$ & $D_{ask}$  & $D_{twX}$    &            & $C_{frm}$ & $C_{ask}$ & $D_{tox}$  \\ 
            \midrule
            {\tt baseline}    & \purp{$.417$}    & $.454$      & \purp{$.941$}  & $.365$      & $.561$      & $.508$       & $.557$      & $.758$      & $.579$      & $.806$ \\
            + clean           & $.408$       & \purp{$.477$}   & $.927$     & $.354$      & \purp{$.562$}   & \purp{$.517$}    & $.561$      & \purp{$.764$}   & $.592$      & \purp{$.807$} \\
            + preproc         & $.345$       & $.426$      & $.929$     & \purp{$.377$}   & $.506$      & $.293$       & $.512$      & $.600$      & $.582$      & $.734$ \\
            \midrule
            NBSVM             & $.364$       & $.462$      & $.929$     & $.231$      & $.508$      & $.469$       & $.542$      & $.635$      & $.592$      & $.779$\\
            + clean           & $.410$       & $.456$      & $.940$     & $.211$      & $.541$      & $.467$       & $.563$      & $.641$      & $.596$      & $.747$ \\
            + preproc         & $.318$       & $.466$      & $.907$     & $.320$      & $.480$      & $.305$       & \purp{$.566$}   & $.532$      & $.597$ & $.756$ \\
            \midrule
            word2vec          & $.368$       & $.394$      & $.860$     & $.338$      & $.304$      & $.323$       & $.366$      & $.698$      & $.572$      & $.634$ \\
            DistilBERT        & $.377$       & $.336$      & $.697$     & $.296$      & $.369$      & $.435$       & $.402$      & $.598$      & \purp{$.629$}      & $.642$ \\
                                          
            \bottomrule
        \end{tabular}
        }
        \caption{Overview of different feature representations (Repr) for Experiment I and II. The `+' parts show performance for preprocessing: removing all special characters (clean), and more sophisticated handling of social media tags and emojis (preproc). Their in-domain positive class $F_1$ scores for Experiment I (T1) and II (T2), and the out-of-domain average (Avg) for $D_{all}$. Baseline scores are from Table~\ref{tab:base}.}\label{tab:archs}
\end{table}

The aim of this experiment was to find (out-of-the-box) representations that would improve upon the simple BoW features used in our \texttt{baseline} model (i.e., achieving good in-domain performance as well as out-of-domain generalization). Table~\ref{tab:archs} lists both of our considered baselines, tested under different preprocessing methods. These are subsequently compared against the two different embedding representations. 

For preprocessing, several levels were used: the default for all models being 1) lowercasing only, then either 2) removal of special characters, or 3) lemmatization and more appropriate handling of special characters (e.g., splitting \texttt{\#word} to prepend a hashtag token) were added. The corresponding results in Table~\ref{tab:archs} do not reveal an unequivocal preprocessing method for either the BoW \texttt{baseline} or NBSVM. While the latter achieves highest out-of-domain generalization with thorough preprocessing (`+preproc', .566 positive $F_1$), the \texttt{baseline} model achieves best in-domain performance on five out of nine corpora, and an on-par out-of-domain average ($.566$ versus $.561$) with simple cleaning (`+clean').

According to our criterion proposed in Section~\ref{subs:crit}, a method that performs well both in- and out-of-domain should be preferred. The current consideration of preprocessing methods illustrates how the stricter evaluation criterion in this experiment potentially yields different overall results in contrast to evaluating in-domain only, or focusing on single corpora. Conversely, we opted for simple cleaning throughout the rest of our experiment (as mentioned in Section~\ref{subs:proc}), given its consistent performance for both baselines. 

The embeddings chosen for this experiment do not seem to provide representations that yield an overall improvement for the classification performance of our Logistic Regression model. Surprisingly, however, DistilBERT does yield significant gains over our baseline for the conversation-level corpus of Ask.fm ($.629$ positive $F_1$ over $.579$). This might imply that such representations would work well on more (balanced) data. While we did not see a significant effect on performance with shallowly fine-tuning DistilBERT, more elaborate fine-tuning would be a required point of further investigation before drawing strong conclusions. 

Moreover, given that we restricted our embeddings to averaged representations on document-level for word2vec, and the sentence representation token for BERT (following common practices), numerous settings remain unexplored. While both (i.e., fine-tuning and alternate input representations) of such potential improvements would certainly merit further exploration in future work focused on optimization, this is out of scope of the current research. Similarly, embeddings trained on a similar domain would be more ideal to represent our noisy data; we settled for strictly off-the-shelf ones that included web content, and a large vocabulary.

Therefore, we conclude that no alternative (out-of-the-box) baselines seem to clearly outperform our BoW baseline. We previously alluded to the effectiveness of binary BoW representations in previous work, and argued this being a result of capturing blatant profanity. We will further test this in the next experiment.

\subsection{Experiment III}

Here, we investigate Hypothesis \textbf{2}: the notion that positive instances across all cyberbullying corpora are biased, and only reflect a limited dimension of bullying. We found strong evidence for this in the previous Experiments I and II, Figures~\ref{fig:bar-error} and \ref{fig:bar-top}, Table~\ref{tab:qual}, and manual analyses of top features all indicated toxicity to be consistent top-ranking features. To add more empirical evidence to this, we trained models on toxicity, or cyber aggression, and tested them on bullying data (and vice-versa)---providing results on the overlap between the tasks. The results for this experiment can be found in the lower end of Table~\ref{tab:base}, under $D_{tox}$ and T3.

It can be noted that there is a substantial gap in performance between the cyberbullying classifiers (using $D_{all}$ as reference) performance on the $D_{tox}$ test set and that of the toxicity model (positive $F_1$ score of $.587$ and $.806$ respectively). More strikingly, however, the other way around, toxicity classifiers perform second-best on the out-of-domain averages (\emph{Avg} in Table~\ref{tab:base}). In the context scopes ($C_{frm}$ and $C_{ask}$) it is notably close, and for other sets relatively close, to the in-domain performance.

 Cyberbullying detection should include detection of toxic content, yet also perform on more complex social phenomena, likely not found in the Wikipedia comments of the toxicity corpus. It is therefore particularly surprising that it achieves higher out-of-domain performance on cyberbullying classification than all individual models using BoW features to capture bullying posts. Only when all corpora are combined, the $D_{all}$ classifier performs better than the toxicity model. This observation combined with previous results provides strong evidence that a large part of the available cyberbullying content is not complex, and current models to only generalize to a limited extent using predominately simple aggressive features, supporting Hypothesis \textbf{3}.

\subsection{Experiment IV}

\begin{table}[t!]
        \resizebox{\textwidth}{!}{%
        \begin{tabular}{lrrrrrrrrrr}
            \toprule
            Arch          &                             \multicolumn{6}{c}{T1}                         & Avg  &   \multicolumn{2}{c}{T2} & T3 \\
                          \cmidrule{2-7} \cmidrule{8-8} \cmidrule{9-10} \cmidrule{11-11}
                          & $D_{twB}$ & $D_{frm}$    & $D_{msp}$   & $D_{ytb}$ & $D_{ask}$  & $D_{twX}$  & $all$ \hspace{0.1cm} $tox$ & $C_{frm}$ & $C_{ask}$ & $D_{tox}$  \\ 
            \midrule
            {\tt baseline}  & $.417$         & $.454$        & $.941$       & $.365$      & \purp{$.561$} & \purp{$.508$}  & \purp{$.557$} \hspace{0.1cm} \purp{$.389$}   & \purp{$.758$} & .579      & \purp{$.806$} \\
            NBSVM           & $.383$         & \purp{$.486$}     & $.925$       & \purp{$.387$}   & $.476$      & $.396$       & $.551$ \hspace{0.1cm} $.385$             & $.703$      & .604      & .797 \\
            \midrule
            BiLSTM*         & $.171$         & $.363$        & $.938$       & $.152$      & $.504$      & $.400$       & $.440$ \hspace{0.1cm} $.349$             & $.609$      & $.507$      & $.762$ \\
            BiLSTM+         & $.188$         & $.396$        & \purp{$.951$}    & $.160$      & $.438$      & $.341$       & $.417$ \hspace{0.1cm} $.337$             & $.541$      & $.505$      & $.737$ \\
            BiLSTM          & $.182$         & $.341$        & $.905$       & $.148$      & $.463$      & $.246$       & $.479$ \hspace{0.1cm} $.356$             & $.608$      & $.522$      & $.774$ \\
            \midrule
            CNN*            & \purp{$.500$}      & $.276$        & $.790$       & $.133$      & $.462$      & $.438$       & $.364$ \hspace{0.1cm} $.350$             & $.000$      & $.306$      & $.753$ \\
            CNN             & $.444$         & $.416$        & $.816$       & $.000$      & $.498$      & $.438$       & $.464$ \hspace{0.1cm} $.342$             & $.000$      & $.610$      & $.754$ \\ 
            CNN$\star$      & $.444$         & $.419$        & $.816$       & $.000$      & $.499$      & $.375$       & $.460$ \hspace{0.1cm} $.362$             & $.000$      & \purp{$.647$}   & $.774$ \\ 
            C-LSTM*         & $.000$         & $.421$        & $.875$       & $.095$      & $.000$      & $.000$       & $.449$ \hspace{0.1cm} $.329$             & $.094$      & $.425$      & $.757$ \\
            C-LSTM          & $.000$         & $.019$        & $.829$       & $.000$      & $.066$      & $.000$       & $.463$ \hspace{0.1cm} $.355$             & $.095$      & $.518$      & $.761$ \\
            C-LSTM$\star$   & $.000$         & $.057$        & $.853$       & $.075$      & $.008$      & $.000$       & $.278$ \hspace{0.1cm} $.358$             & $.296$      & $.506$      & $.756$ \\
            \bottomrule
            
        \end{tabular} }
        \caption{Overview of different architectures (Arch) their in-domain positive class $F_1$ scores for Experiment I (T1) and II (T2), the out-of-domain average for $D_{all}$ ($all$), and $D_{tox}$ ($tox$). Baseline model (and scores) are from Table~\ref{tab:base}. Reproduction of \cite{DBLP:conf/ecir/AgrawalA18} is denoted by *, their oversampling method by +, character level models by $\star$, and all others are our tuned models.}
            \label{tab:neural}
\end{table}

So far, we have attempted to improve a straight-forward baseline that was trained on binary features with several distinct approaches. While changes in data (representations) seem to have a noticeable effect on performance (increasing the amount of messages per instance, merging all corpora), none of the experiments with different feature representations have had an impact. With the current experiment, we had hoped to leverage earlier state-of-the-art architectures by reproducing their methodology and subjecting the models to our evaluation framework.

As can be inferred from Table~\ref{tab:neural}, our baselines outperform these neural techniques on almost all in-domain tests, as well as the out-of-domain averages. Having strictly upheld the experimental set-up from \cite{DBLP:conf/ecir/AgrawalA18} and as close as possible that of \cite{DBLP:conf/ijcnn/RosaM0CC18}, we can conclude that---under stricter evaluation---there is sufficient evidence that these models do not provide state-of-the-art results on the task of cyberbullying.\footnote{Upon acquiring the results of the replication of \cite{DBLP:conf/ecir/AgrawalA18} (in particular failing to replicate the effect of the paper's oversampling) we investigated the provided code and notebooks. It is our understanding that oversampling before splitting the dataset into training and test sets causes the increase in performance; we measured overlap of positive instances in these splits and found \emph{no unique test instances}. Furthermore, after re-running the experiments directly from the notebooks with the oversampling conducted post-split, the effect was significantly decreased (similar to our results in Table~\ref{tab:neural}). The authors were contacted with our observations in March 2019, and have since confirmed our results. Our analyses can be found here: \url{https://github.com/cmry/amica/tree/master/reproduction}. The paper has not been retracted and has accumulated 271 citations at the time of writing (13th of December 2022). \label{foot:bul-rant}} Tuning these networks (at least in our set-up) does not seem to improve performance, rather decrease it. This indicates that the validation set on which early stopping is conducted is often not representative of the test set. Parameter tuning on this set is consequently sensitive to overfitting---unsurprisingly given the size of the corpora.

Some further noteworthy observations can be made related to the performance of the CNN architecture, achieving quite significant leaps on word level (for $D_{twB}$) and character level (for $C_{ask}$). Particularly, the conversation scopes ($C$, with a comparatively balanced class distribution) see much more competitive performance compared to the baselines. The same effect can be observed when more data is available; both averages test scores for $D_{all}$ and $D_{tox}$ are comparable to the baseline across almost all architectures. Additionally, the $D_{tox}$ scores indicate that all architectures show about the same overlap on toxicity detection, although interestingly, less so for the neural models than for the baselines. 

It can therefore be concluded that the current neural architectures do not provide a solution to the limitations of the task, rather, suffering more in performance. Our experiments do, however, once more illustrate that the proposed techniques of improving the representations of the corpora (by providing more data through merging all sources, and balancing by classifying batches of multiple messages, or conversations) allow the neural models to approach the baseline ballpark. As our goal here was not to completely optimize these architectures, but replication, the proposed techniques still could provide more avenues for further research. Finally, given its robust performance, we will continue to use the baseline model for the next experiment.

\subsection{Experiment V}

\begin{table}[t!]
    \centering
    \footnotesize
    \begin{tabular}{lrrrrr}
        \hline\noalign{\smallskip}
        train               & \multicolumn{3}{c}{T1}                      & \multicolumn{2}{c}{T2}         \\
                            \cmidrule{2-4} \cmidrule{5-6}
                            & $D_{ask\_nl}$ & $D_{sim\_nl}$ & $D_{don\_nl}$   &  $D_{askC\_nl}$    & $D_{simC\_nl}$ \\ 
        \noalign{\smallskip}\hline\noalign{\smallskip}
        $D_{ask\_nl}$       & $.598$          & $.516$          & $.495$          & $.264$             & $.533$             \\
        $D_{sim\_nl}$       & $.273$          & \bf{.708}     & \purp{\bf.667}      & $.501$             & $.800$             \\
        \noalign{\smallskip}\hline\noalign{\smallskip}
        $D_{comb}$          & \purp{\bf.608}    & $.681$          & $.516$          & \purp{\bf.801}       & $.808$             \\
        \noalign{\smallskip}\hline\noalign{\smallskip}
        $D_{askC\_nl}$      & $.165$          & $.361$          & $.182$          & $.505$             & $.750$             \\
        $D_{simC\_nl}$      & $.175$          & $.424$          & $.333$          & $.496$             & $.750$             \\
        \noalign{\smallskip}\hline\noalign{\smallskip}
        $D_{all\_nl}$       & $.577$          & $.677$          & $.516$          & $.379$             & \bf{.821}        \\
        \noalign{\smallskip}\hline
    \end{tabular}
        \caption{Positive class $F_1$ scores for Experiment IV on Dutch data. Models are fitted on the training proportion of the corpora row-wise and tested column-wise. The best overall test score is noted in bold. The scores of primary interest are highlighted.} \label{tab:basenl}
    
\end{table}

Due to the nature of its experimental set-up (which generates balanced data with simple language use, as shown in Table~\ref{tab:dat}), the crowdsourced data proves easy to classify. Therefore, we do not report out-of-domain averages, as this set would skew them too optimistically, and be uninformative. Regardless, we are primarily interested in performance when crowdsourced data is added, or used as a replacement for real data. In contrast to the other experiments, the focus will mostly be on the Ask.fm ($D_{ask\_nl}$) and donated ($D_{don\_nl}$) scores (see Table~\ref{tab:basenl}). The scores on the Dutch part of the Ask.fm corpus are quite similar to those on the English corpus ($.561$ vs $.598$ positive $F_1$ score), which is in line with earlier results \citep{VanHee2015guide}. Moreover, particularly for the small amount of data, the crowdsourced corpus performs surprisingly well on $D_{ask\_nl}$ ($.516$), and significantly better on the donated test data ($.667$ on $D_{don\_nl}$). This implies that a balanced, controlled bullying set, tailored to the task specifically, does not need a significant amount of data to achieve comparable (or even better) performance, which is a promising result.

Furthermore, in the settings that utilize context representations, training on conversation scopes initially does not seem to improve detection performance in any of the configurations (save for a  marginal gain on $D_{simC\_nl}$). However, it does simplify the task in a meaningful way at  test-time; whereas a slight gain is obtained for message-level $D_{ask\_nl}$ (from $.598$ $F_1$-score to $.608$), when merging both datasets a significant performance boost can be found when training on $D_{comb}$ and testing on $D_{askC\_nl}$ (from $.264$ and $.501$ to $.801$ on the combined). Hence, it can be further concluded that enriching the existing training set with crowdsourced data yields meaningful improvements.

Based on these results, we confirm the Experiment II results hold for Dutch: more diverse, larger datasets, and increasing context sizes contributes to better performance on the task. Most importantly, there is enough evidence to support Hypothesis \textbf{3}: the data generated by the crowdsourcing experiment helps detection rates for our in-the-wild test set, and its combination with externally collected data increases performance with and without additional context. These results underline the potential of this approach to collecting data.

\subsection{Suggestions for Future Work}

We hope our experiments have helped to shed light on, and raise further attention to, multiple issues with methodological rigor pertaining to the task of cyberbullying detection. It is our understanding that the disproportionate amount of work on the (oversimplified) classification task, versus the lack of focus on constructing rich, representative corpora reflecting the actual dynamics of bullying, has made critical assessment of the advances in this task difficult. We would therefore want to particularly stress the importance of simple baselines and the out-of-domain tests that we included in the evaluation criterion for this research. They would provide a fairer comparison for proposed novel classifiers, and a more unified method of evaluation. In line with this, the structural inclusion of domain adaptation techniques seems a logical next step to improve cross-domain performance, specifically those tailored to imbalanced data.

Furthermore, this should be paired with a critical view on the extent to which the full scope of the task is fulfilled. Novel research would benefit from explicitly finding evidence to support its assumptions that classifiers labeled `cyberbullying detection' do more than one-shot, message-level toxicity detection. We would argue that the current framing of the majority of work on the task is still too limited to be considered theoretically-defined cyberbullying classification. Here, we demonstrated several qualitative and quantitative methods that can facilitate such analyses. As popularity of the application of cyberbullying detection grows, this would avoid misrepresenting the conducted work, and that of in-the-wild applications in the future. 

We can imagine these conclusions to be relevant for more research within the computational forensics domain: detection of online pedophilia \citep{DBLP:journals/csl/BogdanovaRS14}, aggression and intimidation \citep{DBLP:journals/eswa/EscalanteVGLMP17}, terrorism and extremism \citep{DBLP:conf/icdm/KaatiOPS15,DBLP:journals/di/EbrahimiSO16}, and systematic deception \citep{feng-etal-2012-syntactic}--among others. These are all examples of heavily skewed phenomena residing within more complex networks, for which simplification could lead to serious misrepresentation of the task. As with cyberbullying research, a critical evaluation of multiple domains could potentially uncover problematic performance gaps.

While we demonstrated a method of collecting plausible cyberbullying with guaranteed consent, the more valuable sources of real-world data that allow for complex models of social interaction remain restricted. It is our expectation that future modeling will benefit from the construction of much larger (anonymized) corpora---as most fields dealing with language have, and we therefore hope to see future work heading this direction.

\section{Conclusion}

In this chapter, we identified several issues that affect the majority of the current research on cyberbullying detection. As it is difficult to collect accurate cyberbullying data in the wild, the field suffers from data scarcity. In an optimal scenario, rich representations capturing all required meta-data to model the complex social dynamics of what the literature defines as cyberbullying would likely prove fruitful. However, one can assume such access to remain restricted for the time being, and with current social media moving towards private communication, to not be generalizable in the first place. Thus, significant changes need to be made to the empirical practices in this field. To this end, we provided a cross-domain evaluation setup and tested several cyberbullying detection models, under a range of different representations to potentially overcome the limitations of the available data, and provide a fair, rigorous framework to facilitate direct model comparison for this task.

Additionally, we formed three hypotheses we would expect to find evidence for during these evaluations: \textbf{1}) the corpora are too small and heterogeneous to represent the strong variation in language use for both bullying and neutral content across platforms accurately, \textbf{2}) the positive instances are biased, predominantly capturing toxicity, and no other dimensions of bullying, and finally \textbf{3}) crowdsourcing poses a resource to generate plausible cyberbullying events, and that can help expand the available data and improve the current models. 

We found evidence for all three hypotheses: previous cyberbullying models generalize poorly across domains, simple BoW baselines prove difficult to improve upon, there is considerable overlap between toxicity classification and cyberbullying detection, and crowdsourced data yields well-performing cyberbullying detection models. We believe that the results of Hypotheses \textbf{1}) and \textbf{2} in particular are principal hurdles that need to be tackled to advance this field of research. Furthermore, we showed that both leveraging training data from all openly available corpora, and shifting representations to include context meaningfully improves performance on the overall task. 
Therefore, we believe both should be considered as an evaluation point in future work. More so given that we show that these do not solve the existing limitations of the currently available corpora, and could therefore provide avenues for future research focusing on collecting (richer) data. Lastly, we show reproducibility of models that previously demonstrated state-of-the-art performance on this task to fail. We hope that the observations and contributions made in this paper can aid to improve rigor in future cyberbullying detection work.

\begin{figure}
    \centering
    \includegraphics[width=0.75\textwidth,angle=-90]{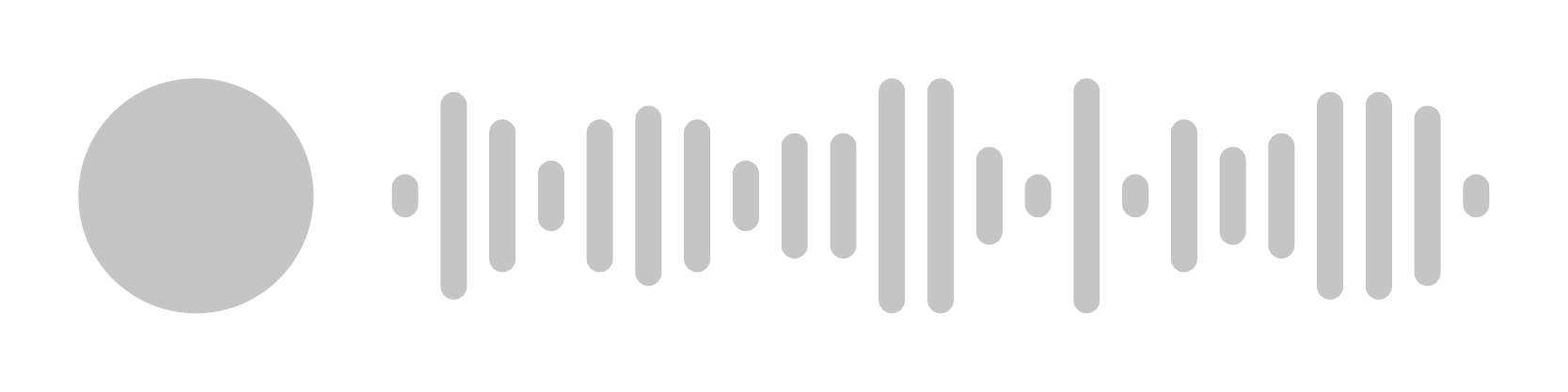}
\end{figure}

\thispagestyle{empty}
\strut
\newpage
\thispagestyle{empty}
\strut
\vfill

\noindent This chapter has been published as:

\begin{description}
    \item \bibentry{emmery-etal-2022-cyberbullying}
\end{description}

\noindent Minor formatting changes have been made to the text and figures.
\newpage

\chapter[Cyberbullying Classifiers are Sensitive to Perturbations]{Cyberbullying Classifiers are Sensitive to Model-Agnostic Perturbations} \label{ch:aug}

\lettrine[lines=4, lraise=0, nindent=0em, slope=0em]{A}{\ limited} amount of studies investigate the role of model-agnostic adversarial behavior in toxic content classification. As toxicity classifiers predominantly rely on lexical cues, (deliberately) creative and evolving language-use can be detrimental to the utility of current corpora and state-of-the-art models when they are deployed for content moderation. The less training data is available, the more vulnerable models might become. This study is, to our knowledge, the first to investigate the effect of adversarial behavior and augmentation for cyberbullying detection. We demonstrate that model-agnostic lexical substitutions significantly hurt classifier performance. Moreover, when these perturbed samples are used for augmentation, we show models become robust against word-level perturbations at a slight trade-off in overall task performance. Augmentations proposed in prior work on toxicity prove to be less effective. Our results underline the need for such evaluations in online harm areas with small corpora. The perturbed data, models, and code are available for reproduction at \url{https://github.com/cmry/augtox}.

\parshape=0
\section{Introduction}

Our online presence has simplified contact with our (in)direct network, and drastically changed how, and with whom  we interact. While online connections and self-disclosure are often socially beneficial \citep{ValkePeter2007fw}, the  absence of physical interaction has adverse effects: it reduces social accountability in (often anonymous) interactions, amplifies one's exposure to people with malicious intent, and through our frequent use of mobile devices, the invasiveness thereof \citep{https://doi.org/10.1002/pits.20301}. These factors accumulate to persistent online toxic  behavior---the scale of which online platforms continue to struggle with from a technical, legal, and ethical perspective.

Online harm \citep[provide a comprehensive taxonomy of this field]{banko-etal-2020-unified} and---particularly for Natural Language Processing (NLP)---abusive language, are highly complex phenomena. Their study spreads across several subfields (detection of hate speech, toxic comments, offensive and abusive language, aggression, and cyberbullying), all with their unique problem sets and (almost exclusively English) corpora \citep{10.1371/journal.pone.0243300}. Moreover, there are numerous open issues with these tasks, as highlighted in a range of critical studies \citep[Chapter~\ref{ch:bul},][]{ROSA2019333,swamy-etal-2019-studying,madukwe-etal-2020-data,DBLP:journals/corr/abs-2103-00153}. Those open issues primarily pertain to the contextual, historical, and multi-modal nature of toxicity, the specificity of the data, and poor generalization across domains.

The current chapter focuses on one of these subproblems: the continuously evolving nature of \emph{toxic content}. Apart from the disparate channels and media through which (young) users communicate, this development particularly applies to the related vocabulary: slang, hate speech, or general insults (e.g., \emph{karen}, \emph{simp}, \emph{coofer}, and \emph{covidiot}). Given the strong focus on lexical cues exhibited by state-of-the-art toxic content classifiers \citep{gehman-etal-2020-realtoxicityprompts}, the existing corpora would have to be continuously expanded for models to retain their performance. This puts costly requirements on any system moderating harmful content, while research in this domain still seems unconcerned with evaluating models in the wild \citep{DBLP:journals/corr/abs-2103-00153}.

\begin{figure}
\centering
\includegraphics[width=0.8\columnwidth]{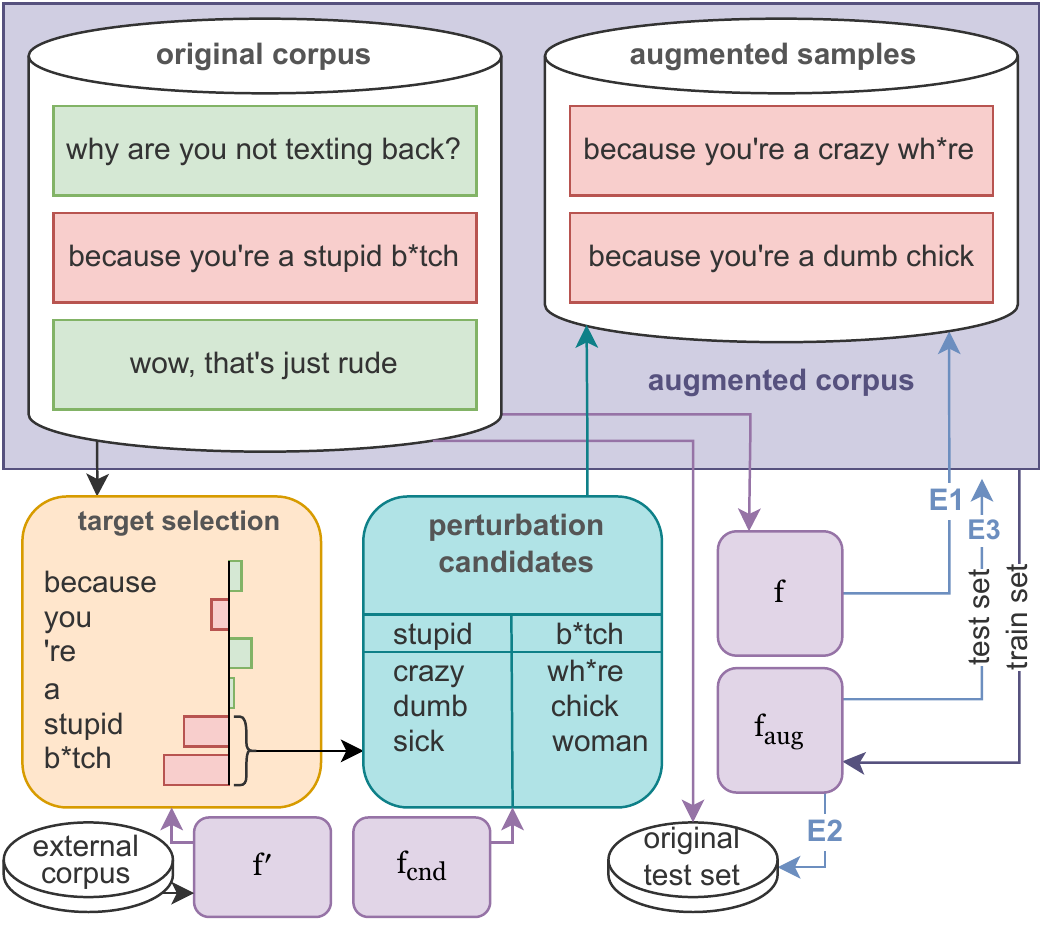}
\caption{Schematic overview of the presented experiments (E1-3) for data augmentation of cyberbullying content via model-agnostic lexical substitutions.}
\label{fig:paper}
\end{figure}

The adversarial nature of toxicity exacerbates these issues further; similar to any security application, it is safe to assume malicious actors will try to (actively) subvert any form of moderation they are subjected to. Yet, while ample work has investigated systems toward mimicking such behavior \citep[for example, feature attacks against Google's Perspective API]{DBLP:journals/corr/HosseiniKZP17,ebrahimi-etal-2018-hotflip,DBLP:conf/ndss/LiJDLW19}, work on toxic content detection rarely incorporates tests for robustness against adversarial attacks. More importantly, these attacks are commonly tailored to an existing toxicity classifier, whereas a human adversary would not have direct access to the models performing moderation. A realistic implementation of subversive human behavior would therefore require model-agnostic attacks.

Accordingly, the current research combines multiple ideas from previous work: we apply lexical substitution to an online harm subtask with small corpora (cyberbullying detection) to investigate how lexical variation (either natural, or adversarial) affects model performance and, by extension, evaluate the robustness of current state-of-the-art models. We do this in a model-agnostic fashion; an external classifier indicates which words might be relevant to substitute. Those words are perturbed through a variety of transformer-based models, after which we assess changes in the predictions of a target classifier (see Figure~\ref{fig:paper}). The perturbations are \emph{not} selected to be adversarial against the external or target classifier. We subsequently evaluate to what extent augmenting existing cyberbullying corpora improves classifier performance, robustness against word-level perturbations, and transferability across different substitution models. With this, we provide methods and language resources to test the robustness of cyberbullying classifiers against lexical variation in toxicity.

\begin{table*}
    \centering
    \footnotesize
    \setlength{\tabcolsep}{4.0pt}
    \begin{tabular}{l|lll>{\columncolor{DrawPaleAqua}}l>{\columncolor{DrawPaleAqua}}ll>{\columncolor{DrawPaleAqua}}llll}
        \toprule
         &   &       & &  \multicolumn{2}{c}{\textsc{targets}}          &       \\
            \cmidrule{5-6} \cmidrule{8-8}
        \textsc{prompt}  & You & are & a & \pale{r\hlfancy{black}{e}t\hlfancy{black}{a}rded} & \pale{dweeb}             & and & \pale{stupid}      & af & . \\
        \textsc{tokens}  & You & are & a & \pale{r\hlfancy{black}{e}\ \#\#t\hlfancy{black}{a}r\ \#\#d\hlfancy{black}{e}d} & \pale{d\ \#\#we\ \#\#eb} & and & \pale{stupid}      & a\ \#\#f & . \\
        \midrule
        \textsc{\#1} & You & are & a & silly     & baby                             & and & silly       & af & .   \\
        \textsc{\#2} & You & are & a & useless   & teenager                         & and & dumb        & af & .   \\
        \textsc{\#3} & You & are & a & sick      & b\hlfancy{black}{i}tch           & and & foolish     & af & .   \\
        \textsc{\#4} & You & are & a & crazy     & dog                              & and & useless     & af & .   \\
        \textsc{\#5} & You & are & a & dumb      & idiot                            & and & ignorant    & af & .   \\
        \bottomrule
    \end{tabular}
    \caption{Lexical substitution example using Dropout BERT. Shows the (bowdlerized) initial \textsc{prompt}, which words are targeted for substitution (highlighted), their word-piece encoding (\textsc{tokens}), and the generated samples.}
    \label{tab:lex-sub}
\end{table*}

\section{Lexical Substitution}

We employ word-level or token-level perturbations (i.e., substitutions, see Table~\ref{tab:lex-sub} for examples), which implies that for a given target word $w_t$ in document $D = (w_0, w_1, \ldots, w_t, \ldots, w_n)$, we find a set of perturbation candidates $C$ using substitute\footnote{Substitute does not refer to word substitution here, but a `replacement' classifier. Specifically, one with an architecture and training data distinct from any target classifier $f$ we use.} classifier $f'$ to exhaustively generate new samples $D'$, any of which \emph{potentially} produces an incorrect label for a target classifier $f$. However, the samples are not selected based on such label changes, which therefore does not make this an adversarial attack. We follow and improve upon the adversarial substitution framework\footnote{\url{https://github.com/cmry/reap} \texttt{(ba8ee44)}} from Chapter~\ref{chap:advsty}, which in turn extends that of TextFooler \citep{DBLP:conf/aaai/JinJZS20}\footnote{\url{https://github.com/jind11/TextFooler}} with transformer-based perturbations.

\subsection{Selecting Words to Perturb} Target words $T(D,\ f')$ are selected and ranked based on their contribution to the classification of a document. This importance, or omission score \citep[among others]{DBLP:journals/tnn/SamekBMLM17,DBLP:journals/coling/KadarCA17} is calculated by deleting a word at a given position $D_t$, denoted as $D_{\setminus t}$. The omission score is then $o_y(D) - o_y(D_{\setminus t})$, where $o_y$ is the logit score of a substitute classifier $f'$. Intuitively, this would provide us with highly toxic words, or text parts related to bullying, which can be perturbed in some way.

\subsection{Proposing Perturbation Candidates} As we intend to improve lexical variation, we focus on proposing synonyms as perturbation candidates. \cite{zhou-etal-2019-bert} condition BERT's masked language modeling on a given word by providing the original word its embedding to the masked position. They apply Dropout \citep{DBLP:journals/jmlr/SrivastavaHKSS14} as a surrogate mask, and show this to produce a top-$k$ of potential synonyms. The predicted words at the Dropout masked position by some separate transformer model $f_\text{cnd}$ are then our candidates $C(T, f_\text{cnd})$. To rank the candidates, they use a contextual similarity score:
\begin{equation}
\begin{split}
& \textsc{sim}\left(D, D^{\prime} ; t\right) =  \\ \sum_{i}^{n} \alpha_{i, t} \times
& \Lambda\left(\boldsymbol{h}\left(D_{i}\right), \boldsymbol{h}\left(D_{i}^{\prime} \right) \right)
\end{split}
\end{equation}
where: $\boldsymbol{h}\left(D_{i}\right)$ is the concatenation of $f_\text{cnd}$ its last four layers for a given $i^{th}$ token in document $D$, $D' = (w_0, \ldots, c_t, \ldots w_n)$ is the perturbed document $D$ where target word $w_t$ has been replaced with candidate $c$ at the index of $t$, $\Lambda$ is their cosine similarity, and $\alpha_{i, t}$ is the average self-attention score across all heads in all layers ranging from the $i^{th}$ token to the $t^{th}$ position in $D$. Finally, we sanitize the candidates: filtering single characters, plural and capitalized forms, sub-words, and sentence-level duplicates.

\subsection{Handling Out-of-vocabulary Words} BERT's associated tokenizers break down unknown words into word-pieces \citep{DBLP:journals/corr/WuSCLNMKCGMKSJL16}, meaning there is no single embedding to apply Dropout to. \cite{zhou-etal-2019-bert} do not mention how they handle such cases; however, they are problematically common for our task (see Table~\ref{tab:lex-sub}). We therefore extend their method with a back-off method: if $w_t$ is out-of-vocabulary (OOV),\footnote{Note that the vocabulary of $f'$ might include tokens not contained in the vocabulary of BERT.} we collapse the word-pieces into one, and zero the embedding at that position (which then acts as a mask).\footnote{Alternative approaches, such as averaging and summing the token embeddings, did not provide better representations.} Words other than $w_t$ that are OOV remain word-pieces.

\section{Experimental Set-up}

We employ and compare this lexical substitution method to produce new positive instances. We evaluate if these hold up as adversarial samples, and if they can be used for data augmentation.

\subsection{Data} \label{subs:data}

\begin{table}[t!]
      \footnotesize
      \centering
      \begin{tabular}{l|rrrrr}
      \toprule
                   & \includegraphics[width=9px]{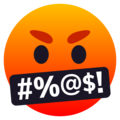}     & \includegraphics[width=9px]{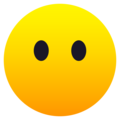}   & \textsc{ttr}   &  \textsc{avg tok/msg} \\  
      \midrule
        Ask.fm      & $5,001$   & $89,404$   & $.154$  & \ $12$ \hfill ($\sigma = 23$) \ \ \\ 
        MySpace     & $426$     & $1,627$   & $.016$  & $391$  \hfill ($\sigma = 285$) \\ 
        Twitter I   & $237$     & $5,258$   & $.154$  & \ $14$ \hfill  ($\sigma = 8$) \ \ \ \ \\ 
        Twitter II  & $281$     & $4,654$   & $.221$  & \ $18$ \hfill  ($\sigma = 8$)  \ \ \ \  \\ 
        YouTube     & $417$     & $3,045$   & $.063$  & $239$  \hfill ($\sigma = 252$) \\ 
        \midrule
        Formspring  & $1,025$   & $11,742$  & $.060$  & \ $27$ \hfill  ($\sigma = 29$) \ \ \\ 
        \bottomrule
    \end{tabular}
    \caption{Corpus statistics for cyberbullying data. Listed are the number of positive (angry emoji, bullying) and negative (cursing emoji) instances, Type-Token Ratio (\textsc{ttr}), and the (rounded) average number of tokens per message (\textsc{avg tok/msg}), and their standard deviation ($\sigma$).}
    \label{tab:data-dsc}
\end{table}

For our corpora (all are English), we use two question-answering-style social networks that allow for anonymous posting: \emph{Formspring} \citep{DBLP:conf/icmla/ReynoldsKE11} and \emph{Ask.fm} \citep{DBLP:conf/ranlp/HeeLVMDPDH15}. The latter features multi-label annotation, but is binarized (any indication of bullying\footnote{Includes self-defenses and assistants of the victim.} is labeled positive) to be compatible with other corpora. These corpora are significantly larger than the rest, as their platforms are typically used by young adults, and notorious for their bullying content \citep{uclan8378}. Two long-form platforms can be found in \emph{YouTube} \citep{DBLP:conf/icwsm/DinakarRL11} and \emph{MySpace} \citep{bayzick2011detecting}, the latter of which has instances of ten posts. The smallest two are from \emph{Twitter}, both collected using topical keywords \citep{xu-etal-2012-learning,DBLP:conf/icis/BretschneiderWP14}. The corpora's statistics can be found in Table~\ref{tab:data-dsc}.

\subsection{Augmentation Models} \label{subs:aug-mod}

The models were implemented using the \texttt{transformers} \citep{wolf-etal-2020-transformers} library.\footnote{\url{https://huggingface.co/transformers/}} Dependency versions can be found in our repository.

\paragraph{Target Word Selectors} All experiments follow the same model-agnostic approach: target words are determined through substitute classifier $f'$ (i.e., a distinct model trained on a different corpus than used in any other experiments). Generally, this is Gaussian Naive Bayes over tf$\cdot$idf-weighted vectors, trained on \emph{Formspring}.  We additionally investigate a pre-trained version of BERT fine-tuned on the Jigsaw dataset \citep[\texttt{unitary/toxic-bert}]{Detoxify} as a transformer-based alternative for $f'$ (denoted by a +). While the task it has been fine-tuned on is slightly different, we assume this model will have better representations and a larger vocabulary, which might make it more effective in choosing target words.

\paragraph{Substitutors}
We compare our own implementation of \cite{zhou-etal-2019-bert}'s substitution model (here: Dropout BERT) against other methods for masked word prediction: BERT \citep{DBLP:conf/naacl/DevlinCLT19} and BART \citep{lewis-etal-2020-bart}.  Table~\ref{tab:sub-comp} shows output examples of the various models. The probability of zeroing embedding dimensions in Dropout BERT is set to $0.2$, as we empirically found that values around $0.3$ often does not result in synonyms. We set the minimum required omission score to $0.005$ for tokens to be considered for substitution, which yields 1-3 target words per document on average. The substitutions do not incorporate prior substitutions; they are done simultaneously---best candidates  first---and exhaustively (i.e., while candidates are available) for maximally five samples.

For all BERT models, we use the pre-trained \texttt{bert-large-cased}, for BART we use \texttt{bart-large}.\footnote{We empirically found the \texttt{bert-base} models to perform significantly worse at finding good synonyms.} We also report experiments where we use a fine-tuned toxicity version of pre-trained BERT \citep[\texttt{GroNLP/hateBERT}]{caselli-etal-2021-hatebert} for $f_{\text{cnd}}$ in Dropout BERT (here referred to as Hate BERT, and when using a fine-tuned substitute classifier: Hate BERT+). This is similar to using Dropout BERT+; domain-specific vocabularies will likely result in better and more varied substitutions. Hate BERT(+) uses a different BERT-based toxicity model than the `+' model for $f'$, in order to keep these selections model-agnostic.

\paragraph{Baselines}

\begin{table}[t]
    \centering
    \footnotesize
    \begin{tabular}{l|p{5.7cm}}
        \toprule
        \textsc{prompt} & okay and stop calling me jaky you c\hlfancy{black}{o}ck \\
        \midrule
        \textsc{gpt-2} & s\hlfancy{black}{u}cker. you know it you f\hlfancy{black}{u}cking p\hlfancy{black}{u}ssy. you know you are an evil f\hlfancy{black}{u}cking b\hlfancy{black}{i}tch that only cares about getting her name in newspapers. i bet if you saw my face you wouldnt even believe i said ""oh yeah i think i can f\hlfancy{black}{u}ck you."" ... \\ [0.3em]
        \bottomrule
    \end{tabular}
    \caption{Output by GPT-2, receiving an original instance as prompt. The generated text (up to 70 tokens) is subsequently used as an augmented instance.}
    \label{tab:gpt-example}
\end{table}

The substitution models are compared with two baselines adapted from related work. Both have shown to improve toxicity detection \citep{gehman-etal-2020-realtoxicityprompts,quteineh-etal-2020-textual,DBLP:journals/corr/abs-2104-08826}, but have (to our knowledge) not been applied for augmenting cyberbullying content. Firstly, we employ the common data augmentation baseline: Easy Data Augmentation \citep[or EDA]{wei-zou-2019-eda}. EDA applies $n$ of the following operations to an input text: synonym replacement using WordNet \citep{DBLP:journals/cacm/Miller95}, random character insertions, swaps, and deletions. We set the number of augmentations made by EDA similar to that of our other models. Secondly, we employ fully unsupervised augmentation with GPT-2 \citep[implemented in the pre-trained \texttt{gpt2-large}]{radford2019language}. We use the positive instances (i.e., documents containing cyberbullying) of each dataset as prompt, with a maximum input length of $30$ tokens, and the generated output length to a maximum of $70$, as we found that toxicity is prevalent in the first part of the generation \citep[made similar observations]{gehman-etal-2020-realtoxicityprompts}. Table~\ref{tab:gpt-example} shows examples of the output, and the eventual divergence from toxicity.\footnote{GPT-2 in particular tends to descend into literary content after too many tokens are generated. We also experimented with GPT-3's \citep{DBLP:conf/nips/BrownMRSKDNSSAA20} \texttt{curie} from the OpenAI beta API (\url{https://beta.openai.com/}) but found systematically lower performance across all experiments compared to GPT-2. These results are therefore not included.}

\subsection{Classifiers} \label{subs:clf}
We follow recent state-of-the-art results \citep{DBLP:journals/access/ElsafouryKPR21} for our main classification model, and fine-tune all BERT-based models for 10 epochs with a batch size of 32 and a learning rate of $2e{-5}$, as suggested by \cite{DBLP:conf/naacl/DevlinCLT19}. Accordingly, we set the maximum sequence length to 128, and insert a single linear layer after the pooled output. For the transformer experiments, we fine-tune incrementally: first on the original set, then on the augmented training set (including the original instances)---both using the same configuration (learning rate, batch size, etc.), except for running it for 2 epochs. This should offer performance advantages \citep{DBLP:journals/corr/abs-1904-06652}, as well as increase model stability.\footnote{We found that fine-tuning on the mixed set renders augmentation ineffective for all models we tested.} 

We compare BERT against a previously tried-and-tested (Chapter~\ref{ch:bul}) `simple' linear baseline: the Scikit-learn \citep{DBLP:journals/jmlr/PedregosaVGMTGBPWDVPCBPD11} implementation of a Linear Support Vector Machine \citep[SVM ]{DBLP:journals/ml/CortesV95,DBLP:journals/jmlr/FanCHWL08} with binary Bag-of-Words (BoW) features, using hyperparameter ranges from \cite{10.1371/journal.pone.0203794}. Training of the SVM and BERT classifiers is done on a merged set of all the cyberbullying corpora in Table~\ref{tab:data-dsc}, except for \emph{Formspring} (reserved for substitute classifier $f'$)---always on the same 90\% split, augmented data or no. The SVM is tuned via grid search and nested, stratified cross-validation (with ten inner and three outer folds, no shuffling, using 10\% splits). The best settings (1-3-grams, class balancing, square hinge loss, and $C = 0.01$) are used in all experiments.

For both models, we also experiment with prepending a special token \citep[follow a similar approach]{daume-iii-2007-frustratingly,caswell-etal-2019-tagged} to the augmented instances (\texttt{<A>}). As per recommendations in \cite{DBLP:journals/corr/abs-2003-02245}, the token is not added to the vocabulary. These models are referred to as either $f$ or $f_\text{aug}$ in Figure~\ref{fig:paper}, depending on if they were trained on augmented data. If not, we skip the 2 fine-tuning epochs for BERT.

\begin{table}[t!]
      \footnotesize
      \centering
      \begin{tabular}{l|rr|rr}
      \toprule
                          & \multicolumn{2}{c}{\textsc{train}} &  \multicolumn{2}{c}{\textsc{test}} \\
                          \cmidrule{2-3} \cmidrule{4-5}
                          & \includegraphics[width=9px]{chapters/4-toxic-augmentation/gfx/angry.png}   & \includegraphics[width=9px]{chapters/4-toxic-augmentation/gfx/nomouth.png}  & \includegraphics[width=9px]{chapters/4-toxic-augmentation/gfx/angry.png}  &  \includegraphics[width=9px]{chapters/4-toxic-augmentation/gfx/nomouth.png} \\
      \midrule
        Merged            & $4,789 $        & $72,243$        & $561  $         &$ 8,001$   \\
        Augment Train     & $28,148$        & $72,243$        & $561  $         &$ 8,001$   \\
        Augment Test      & $4,789 $        & $72,243$        & $3,283$        & $8,001 $      \\
      \bottomrule
    \end{tabular}
    \caption{Instance counts for the different splits used in our experiments. Augment Test is used for Experiment 1 (gauging the adversarial nature of our samples), Augment Train in Experiment 2 (data augmentation).}
    \label{tab:set-freq}
\end{table}

\subsection{Evaluation}

To evaluate our classifiers on the main classification task, we use $F_1$-scores. The impact of the substitution models on classification performance is measured via a decrease in True Positive Ratio (TPR) between regular and substituted samples (i.e., how many previously positively classified samples classified as negative after perturbation).  Note that in these experiments, $f'$ is the same (either Naive Bayes or BERT); therefore, the substitute classifier always chooses the same target words to perturb. The amount of samples depends on the quality of the candidates the models propose.

TPR decrease by itself might also indicate an augmented instance is not toxic anymore; hence, to evaluate the semantic consistency of the samples produced by the various augmentation models, we calculate both \textsc{Meteor} \citep{banerjee-lavie-2005-meteor,denkowski-lavie-2011-meteor} using the implementation from \texttt{nltk}\footnote{\url{https://www.nltk.org/_modules/nltk/translate/meteor_score.html} (v3.5)}, and \textsc{BERTScore} \citep{sellam-etal-2020-bleurt} between the original sentences and their respective augmented samples. \textsc{Meteor} measures flexible uni-gram token overlap, and \textsc{BERTScore} transformer-based similarity with respect to the contextual sentence encoding.

\subsection{Experiments}

We run our substitution pipeline (visualized in Figure~\ref{fig:paper}) on the positive instances $X_{\text{pos}}$ of some given corpus (or the entire collection), using the different models discussed in Section~\ref{subs:aug-mod} for $f'$ and $f_{\text{cnd}}$. Per such configuration, this generates augmented samples $X_{\text{pos}}'$ (up to five per original instance). These can either be classified as is, or mixed in with the original corpus, producing the augmented corpus $X'$, with $X_{\text{train}}'$, and $X_{\text{test}}'$ splits. Using this configuration, we run our three experiments:

\paragraph{Experiment 1} We gauge the \emph{lexical variation} (and hence the `adversarial' character) in our augmented samples via $f(X_{\text{pos}}')$. $F_1$-scores and TPR changes close to $f(X_{\text{pos}})$ imply the substitutions are similar to the original words. We confirm this meaning preservation through  semantic consistency metrics for $X_{\text{pos}}'$.

\paragraph{Experiment 2} Here, we train via the \emph{data augmentation} scheme discussed in Section~\ref{subs:clf}; i.e., fine-tune for 2 epochs on $f(X_\text{train}')$. The resulting augmented classifier is referred to as $f_\text{aug}$, which we evaluate on the original $X_{\text{test}}$. An increase in $F_1$-score with respect to $f(X_{\text{test}})$ indicates the augmentation is a success.

\paragraph{Experiment 3} We measure \emph{robustness} against perturbations, and \emph{transferability} via $f_{\text{aug}}(X_{\text{test}}')$ by evaluating  $f_{\text{aug}}$ performance across different substitution models producing perturbed samples in $X_{\text{test}}'$; i.e., in a many-to-many evaluation. Any TPR increase implies augmentation improves robustness against perturbations. A total TPR higher than $f(X_{\text{test}})$ (Plain) does not necessarily increase the $F_1$-score (from Experiment 2). If this increase holds for multiple perturbation models, it implies the augmentations are transferable.

\begin{figure}
    \centering
\begin{tikzpicture}
\tikzstyle{every node}=[font=\scriptsize]
\definecolor{DrawPaleGreen}{cmyk}{0.08,0.00,0.09,0.09}
\definecolor{color0}{cmyk}{0,0.14,0.31,0.02}
\definecolor{DrawOrange}{cmyk}{0,0.44,0.98,0.29}
\definecolor{DrawPaleRed}{cmyk}{0,0.13,0.15,0.02}
\definecolor{DrawRed}{cmyk}{0,0.63,0.71,0.32}
\definecolor{color1}{cmyk}{0.23,0.01,0.00,0.10}
\definecolor{DrawAqua}{cmyk}{0.90,0.06,0.00,0.47}
\definecolor{colorx}{cmyk}{0.26,0.08,0.00,0.06}
\definecolor{DrawBlue}{cmyk}{0.90,0.27,0.00,0.38}
\definecolor{DrawPalePurple}{cmyk}{0.08,0.08,0.00,0.11}
\definecolor{DrawPurple}{cmyk}{0.32,0.36,0.00,0.51}
\definecolor{DrawPaleMarine}{cmyk}{0.12,0.05,0.00,0.17}
\definecolor{DrawMarine}{cmyk}{0.62,0.27,0.00,0.64}
\definecolor{DrawBaby}{cmyk}{0.43,0.26,0.00,0.25}
\definecolor{DrawPaleBaby}{cmyk}{0.13,0.08,0.00,0.01}
\definecolor{DrawPaleLavender}{cmyk}{0.03,0.08,0.00,0.09}
\definecolor{DrawLavender}{cmyk}{0.10,0.31,0.00,0.35}
\definecolor{DrawPaleGold}{cmyk}{0.00,0.05,0.20,0.00}
\definecolor{DrawGold}{cmyk}{0.00,0.15,0.60,0.16}

\begin{axis}[
axis line style={white!80!black},
legend cell align={left},
legend style={fill opacity=0.8, draw opacity=1, text opacity=1, draw=white!80!black},
legend pos=north west,
tick align=outside,
y=0.3cm,
y dir=reverse,
y grid style={white!80!black},
ymajorticks=true,
ymin=-0.5, ymax=7.5,
ytick style={color=white!15!black},
ytick={0,1,2,3,4,5,6,7},
ytick pos=upper,
yticklabels={EDA\phantom{p},BERT \phantom{p},Dropout BERT,Dropout BERT+,Hate BERT\phantom{p},Hate BERT+\phantom{p},BART\phantom{p},GPT-2\phantom{p}},
xtick pos=left,
x grid style={white!80!black},
xmajorgrids,
xmajorticks=true,
xmin=-1, xmax=0,
xtick style={color=white!15!black},
x=4cm
]
\addlegendimage{ybar,ybar legend,draw=white!18.8235294117647!black,fill=color0,semithick, legend image code/.code={\draw[] rectangle (0.5cm,0.1cm);}}
\addlegendentry{\ SVM}
\draw[draw=white!18.8235294117647!black,fill=color0,semithick] (axis cs:0,-0.4) rectangle (axis cs:-0.189968847352025,0);
\draw[draw=white!18.8235294117647!black,fill=color0,semithick] (axis cs:0,0.6) rectangle (axis cs:-0.483989722744181,1);
\draw[draw=white!18.8235294117647!black,fill=color0,semithick] (axis cs:0,1.6) rectangle (axis cs:-0.454219030520646,2);
\draw[draw=white!18.8235294117647!black,fill=color0,semithick] (axis cs:0,2.6) rectangle (axis cs:-0.543649579188982,3);
\draw[draw=white!18.8235294117647!black,fill=color0,semithick] (axis cs:0,3.6) rectangle (axis cs:-0.407928292829283,4);
\draw[draw=white!18.8235294117647!black,fill=color0,semithick] (axis cs:0,4.6) rectangle (axis cs:-0.47522303021216,5);
\draw[draw=white!18.8235294117647!black,fill=color0,semithick] (axis cs:0,5.6) rectangle (axis cs:-0.464097598760651,6);
\draw[draw=white!18.8235294117647!black,fill=color0,semithick] (axis cs:0,6.6) rectangle (axis cs:-0.721794392523365,7);

\addlegendimage{ybar,ybar legend,draw=white!18.8235294117647!black,fill=color1,semithick, legend image code/.code={\draw[] rectangle (0.5cm,0.1cm);}}
\addlegendentry{\ BERT}

\draw[draw=white!18.8235294117647!black,fill=color1,semithick] (axis cs:0,-2.77555756156289e-17) rectangle (axis cs:-0.119781931464174,0.4);
\draw[draw=white!18.8235294117647!black,fill=color1,semithick] (axis cs:0,1) rectangle (axis cs:-0.284541933504621,1.4);
\draw[draw=white!18.8235294117647!black,fill=color1,semithick] (axis cs:0,2) rectangle (axis cs:-0.230260896138126,2.4);
\draw[draw=white!18.8235294117647!black,fill=color1,semithick] (axis cs:0,3) rectangle (axis cs:-0.387605202754399,3.4);
\draw[draw=white!18.8235294117647!black,fill=color1,semithick] (axis cs:0,4) rectangle (axis cs:-0.19959495949595,4.4);
\draw[draw=white!18.8235294117647!black,fill=color1,semithick] (axis cs:0,5) rectangle (axis cs:-0.345265762051128,5.4);
\draw[draw=white!18.8235294117647!black,fill=color1,semithick] (axis cs:0,6) rectangle (axis cs:-0.310805577072037,6.4);
\draw[draw=white!18.8235294117647!black,fill=color1,semithick] (axis cs:0,7) rectangle (axis cs:-0.637121495327103,7.4);
\addplot [line width=1.08pt, white!26!black, forget plot]
table {%
0 -0.2
0 -0.2
};
\addplot [line width=1.08pt, white!26!black, forget plot]
table {%
0 0.8
0 0.8
};
\addplot [line width=1.08pt, white!26!black, forget plot]
table {%
0 1.8
0 1.8
};
\addplot [line width=1.08pt, white!26!black, forget plot]
table {%
0 2.8
0 2.8
};
\addplot [line width=1.08pt, white!26!black, forget plot]
table {%
0 3.8
0 3.8
};
\addplot [line width=1.08pt, white!26!black, forget plot]
table {%
0 4.8
0 4.8
};
\addplot [line width=1.08pt, white!26!black, forget plot]
table {%
0 5.8
0 5.8
};
\addplot [line width=1.08pt, white!26!black, forget plot]
table {%
0 6.8
0 6.8
};
\addplot [line width=1.08pt, white!26!black, forget plot]
table {%
0 0.2
0 0.2
};
\addplot [line width=1.08pt, white!26!black, forget plot]
table {%
0 1.2
0 1.2
};
\addplot [line width=1.08pt, white!26!black, forget plot]
table {%
0 2.2
0 2.2
};
\addplot [line width=1.08pt, white!26!black, forget plot]
table {%
0 3.2
0 3.2
};
\addplot [line width=1.08pt, white!26!black, forget plot]
table {%
0 4.2
0 4.2
};
\addplot [line width=1.08pt, white!26!black, forget plot]
table {%
0 5.2
0 5.2
};
\addplot [line width=1.08pt, white!26!black, forget plot]
table {%
0 6.2
0 6.2
};
\addplot [line width=1.08pt, white!26!black, forget plot]
table {%
0 7.2
0 7.2
};
\end{axis}

\end{tikzpicture}
    \caption{Decrease in True Positive Rate of the SVM and BERT classifiers after the respective substitution models have been applied (lower is more adversarial).}
    \label{fig:tpr-dec}
\end{figure}
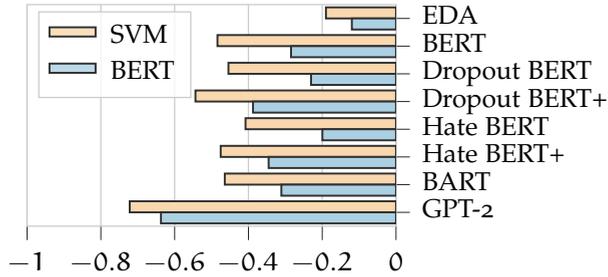

\section{Results and Discussion}

Here, we discuss the results of our three Experiments (Sections~\ref{subs:exp1} - \ref{subs:exp3}). The main results can be found in Table~\ref{tab:exp1-2} and \ref{tab:exp3}.

\subsection{The Effect of Lexical Variation} \label{subs:exp1}

The results can be found under the `Samples' row in Table~\ref{tab:exp1-2}.

\paragraph{Classifier Performance} It can be observed that unsupervised (prompt conditioning) samples (i.e., GPT-2) are the most difficult to classify. This is to be expected, as the generated output is not always toxic. However, it is arguably rather remarkable that a large amount of the generated sentences are labeled positive by the cyberbullying classifier. This confirms that (contextually more) harmful content is generated (as illustrated in Table~\ref{tab:gpt-example}), as also shown for toxicity detection by \cite{gehman-etal-2020-realtoxicityprompts} and \cite{DBLP:conf/acl/OusidhoumZFSY20}. Moving on, we can see the fine-tuned target selector models (`+') show most `adversarial' behavior, likely providing lexical diversity on words more important to the content classification. BART induces similar performance drops, but often inserts noise (see Table~\ref{tab:sub-comp}). The Dropout models seem to produce samples that are less diverse, but still show a solid $.1$ drop in $F_1$-score ($17.67\%$ avg). To emphasize, this decrease is based on \emph{untargeted} substitutions; i.e., without selecting the substituted words as to change the predictions of either $f$ or $f'$.

\paragraph{Adversarial Samples}

\begin{figure}
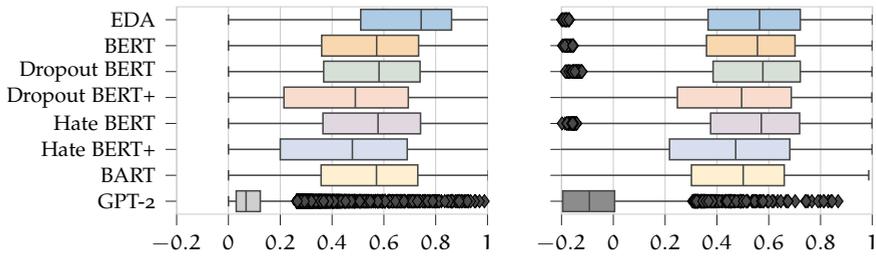

    \centering
    \includestandalone[width=\columnwidth]{chapters/4-toxic-augmentation/gfx/boxplot}
    \caption{Semantic consistency metrics (higher is better): \textsc{Meteor} (upper) and \textsc{BERTScore} (lower) scores per model---evaluating the augmented samples (positive only) with the original documents as a reference.}
    \label{fig:sem-con}
\end{figure}

Additional analyses can be found in Figure~\ref{fig:tpr-dec}. We observe the same patterns per substitution model\footnote{To equalize the length, we matched the amount of non-augmented test set instances for this experiment.} as in Table~\ref{tab:exp1-2}, with the BERT classifier showing to suffer around 20\% less in TPR compared to the SVM. This difference can partly be explained by the substitution models sampling from the same model as we fine-tuned for the classification task (\texttt{bert-large-cased}). As can be observed in this Figure, the difference is smaller when this is not the case (`+' models). This experiment not only underlines the strong focus on lexical cues\footnote{Which raises its own issues; see e.g., and \cite{zhou-etal-2021-challenges}, for work on bias and debiasing.} from linear classifiers, but also that transformer models are not immune to lexical variation---even when candidates are sampled from their own language model. This provides further evidence in line with research from \cite{DBLP:journals/access/ElsafouryKPR21}, and \cite{DBLP:conf/sigir/ElsafouryKWR21} (see Section~\ref{sec:rel-work}).

\begin{table*}[t!]
    \centering
    \footnotesize
    \setlength{\tabcolsep}{2pt}
    \begin{tabular}{l>{\raggedleft\arraybackslash}p{0.9cm}||>{\raggedleft\arraybackslash}p{0.9cm}>{\raggedleft\arraybackslash}p{1cm}>{\raggedleft\arraybackslash}p{1cm}>{\raggedleft\arraybackslash}p{1.05cm}>{\raggedleft\arraybackslash}p{1cm}>{\raggedleft\arraybackslash}p{1.05cm}>{\raggedleft\arraybackslash}p{1cm}>{\raggedleft\arraybackslash}p{1cm}}
      \toprule
                    &                                               &                                 &                          & D.out                          & D.out                 & Hate                              & Hate &        &  \\
        $X_{\text{train}}$  &    Plain                              & EDA                             & BERT                     &  BERT                            & BERT+                   &  BERT                             & BERT+                   & BART                   & GPT-2 \\
      \midrule 
                    &                                               & \multicolumn{8}{c}{$f(X_{\text{pos}}')$} \\
      \midrule
        Merged &                            $.614$ \scriptsize{$.009$}  &         $.598$  \scriptsize{$.012$} & $.458$ \scriptsize{$.002$}  & $.491$ \scriptsize{$.005$}            & $.439$ \scriptsize{$.002$}  &         $.520$ \scriptsize{$.005$}    & $.478$ \scriptsize{$.004$}  & $.436$ \scriptsize{$.007$} &          $.334$ \scriptsize{$.007$} \\
      \midrule \midrule
              & &    \multicolumn{8}{c}{$f_{\text{aug}}(X_{\text{test}})$} \\
      \midrule
        Merged      & \cellcolor{PaleRed}   $.563$ \scriptsize{$.014$}  &         $.553$  \scriptsize{$.007$} & $.538$ \scriptsize{$.014$}  & $.546$ \scriptsize{$.012$}            & $.523$ \scriptsize{$.004$}  & \textbf{$.562$} \scriptsize{$.007$}   & $.535$ \scriptsize{$.009$}  & $.536$ \scriptsize{$.013$} &          $.550$ \scriptsize{$.017$} \\ 
      \midrule
        Ask.fm      & \cellcolor{PaleRed}   $.621$ \scriptsize{$.011$}  &         $.591$  \scriptsize{$.011$} & $.581$ \scriptsize{$.016$}  & $.597$ \scriptsize{$.015$}            & $.574$ \scriptsize{$.003$}  & \textbf{$.611$} \scriptsize{$.007$}   & $.592$ \scriptsize{$.007$}  & $.587$ \scriptsize{$.015$} &          $.601$ \scriptsize{$.009$}\\ 
        Myspace     &                       $.436$ \scriptsize{$.093$}  &         $.476$  \scriptsize{$.021$} & $.403$ \scriptsize{$.058$}  & $.376$ \scriptsize{$.161$}            & $.351$ \scriptsize{$.040$}  &         $.414$  \scriptsize{$.042$}   & $.370$ \scriptsize{$.065$}  & $.387$ \scriptsize{$.043$} & \textbf{.$496$} \scriptsize{$.037$} \\
        Twitter I   &                       $.596$ \scriptsize{$.046$}  & \textbf{$.630$} \scriptsize{$.016$} & $.592$ \scriptsize{$.059$}  & $.531$ \scriptsize{$.048$}            & $.594$ \scriptsize{$.030$}  &         $.617$  \scriptsize{$.053$}   & $.533$ \scriptsize{$.051$}  & $.591$ \scriptsize{$.027$} &          $.583$ \scriptsize{$.097$} \\
        Twitter II  &                       $.308$ \scriptsize{$.059$}  &         $.290$  \scriptsize{$.038$} & $.297$ \scriptsize{$.043$}  & \textbf{$.329$} \scriptsize{$.034$}   & $.257$ \scriptsize{$.040$}  &         $.313$  \scriptsize{$.048$}   & $.268$ \scriptsize{$.027$}  & $.277$ \scriptsize{$.077$} &          $.295$ \scriptsize{$.048$} \\ 
        YouTube     &                       $.150$ \scriptsize{$.033$}  & \textbf{$.226$} \scriptsize{$.057$} & $.180$ \scriptsize{$.010$}  & $.201$ \scriptsize{$.066$}            & $.144$ \scriptsize{$.045$}  &         $.173$  \scriptsize{$.056$}   & $.152$ \scriptsize{$.009$}  & $.152$ \scriptsize{$.055$} &          $.207$ \scriptsize{$.061$} \\ 
      \bottomrule
    \end{tabular}
    \caption{BERT-based cyberbullying classification scores ($F_1$) for Experiments 1 (under $f(X_{\text{pos}}')$) and 2 (under $f_{\text{aug}}(X_{\text{test}})$). Classifiers are trained and tested on the indicated corpus (from Section~\ref{subs:data}), `Merged` is their combination. The other columns indicate, respectively: no substitutions (Plain), EDA, BERT-based models (where `+' indicates $f'$ uses BERT rather than an SVM, and `Hate' that $f_{\text{cnd}}$ is pre-trained), BART, and GPT-2. Highlighted cells indicate that non-augmented performance was highest, bold indicates the highest performance per augmentation model. Standard deviation (small script) is reported over $5$ runs with different seeds.}
    \label{tab:exp1-2}
\end{table*}

\paragraph{Semantic Consistency of Samples}

Here, we compared $X'_{\text{pos}}$ with $X_{\text{pos}}$ as a reference. The results for \textsc{Meteor} and \textsc{BERTScore} of these pairs can be found in Figure~\ref{fig:sem-con}. Generally, these confirm the trend from the previous two parts of the experiment: models that have higher semantic consistency have less effect on classification performance. A clear difference in \textsc{Meteor} can be observed between EDA and the other models. This is likely due to both the metric and model using WordNet, resulting in bias in favor of EDA. GPT-2 is a strong outlier, as it generates new data. 

The semantic consistency scores seem comparable, and at times slightly better, than previous lexical substitution work \citep[see Chapter~\ref{chap:textobf};][although these are all explicitly adversarial]{DBLP:conf/uss/ShettySF18,DBLP:journals/corr/abs-2011-03901}. We noticed that regarding the samples themselves, the transformer-based models often noticeably break down in terms of semantic preservation for the lower ranked candidates (see Table~\ref{tab:sub-comp}). For the models that do not use soft semantic constraints (such as Dropout), we already find antonyms, and generally ungrammatical and incoherent sentences within the top 5 candidates. Interestingly, at the same time, \textsc{BERTScore} assigns comparable scores to antonyms as it does for (intuitively) better substitution candidates. Given these observations, if mimicking adversarial behavior is to be given more weight, one should consider limiting the number of augmented samples, and tuning the omission score cut-off might prove to be worthwhile.

\begin{table}
    \setlength{\tabcolsep}{3.5pt}
    \centering
    \footnotesize
    \begin{tabular}{l|l|llll>{\columncolor{DrawPaleAqua}}l>{\columncolor{DrawPaleAqua}}lll|l}
        \toprule
        &                \# &            &   &       &     &  \multicolumn{2}{c}{\textsc{targets}}   & & & \textsc{bsc}  \\
                         \cmidrule{7-8}
                            &            & I & don't & get & why & \pale{people} & \pale{like} & u & .    &      \\
        \midrule
        \multirow{2}{*}{BE} & \textsc{1} & I & don't & get & why & they & want & u & .      & $.809$ \\
                            & \textsc{5} & I & don't & get & why & boys & hate & u & .      & $.807$ \\
        \midrule
        Dr                  & \textsc{1} & I & don't & get & why & everyone & want & u & .  & $.862$ \\
        BE                  & \textsc{5} & I & don't & get & why & you & love & u & .       & $.806$ \\
        \midrule
        \multirow{2}{*}{BA} & \textsc{1} & I & don't & get & why & you & are & u & .        & $.700$ \\
                            & \textsc{5} & I & don't & get & why & so & have & u & .        & $.634$ \\
        \bottomrule
    \end{tabular}
    \caption{Augmentations including the top 1 and 5 candidates from BERT (BE), Dropout BERT (Dr BE), and BART (BA), and the \textsc{BERTScore} (\textsc{bsc}) using the original text as reference, showing quality degradation (not well reflected in the metric) when sample size increases. BERT suggests antonyms, BART fails semantically.}
    \label{tab:sub-comp}
\end{table}

\subsection{Substitutions for Data Augmentation}

\begin{table*}[t!]
\footnotesize
\setlength{\tabcolsep}{3.6pt}
\centering
\begin{tabular}{l|>{\raggedleft\arraybackslash}p{0.7cm}||>{\raggedleft\arraybackslash}p{0.8cm}|>{\raggedleft\arraybackslash}p{0.8cm}|>{\raggedleft\arraybackslash}p{0.8cm}|>{\raggedleft\arraybackslash}p{0.8cm}|>{\raggedleft\arraybackslash}p{0.8cm}|>{\raggedleft\arraybackslash}p{0.8cm}|>{\raggedleft\arraybackslash}p{0.9cm}|>{\raggedleft\arraybackslash}p{0.98cm}}
\toprule
              & \textsc{init} & \multicolumn{8}{c}{\multirow{2}{*}{$f_{\text{aug}}$}} \\
              & \textsc{tpr}      \\
              \midrule
              &                     &                               &                               &         D.out               &         D.out                   &      Hate                     &      Hate                     &                               &        \\
Plain         &   $.537$              & EDA                           & BERT                          &         B                  &         B+                     &      B                     &      B+                    & BART                          & GPT-2  \\
\midrule
$X_{\text{test}}' \downarrow$    &  &  \multicolumn{8}{c}{\textsc{$\Delta$ tpr}} \\
\midrule
EDA           & $.498$                &  \purp{$.270$}                  &          $.106$                 &        $.104$                       &             $.108$                          & $ .101$                          &          $.107$                          &  \textbf{.114}                &  $.037$ \\
BERT          & $.390$                & $-.033$                         &  \purp{  $.195$}                &        $.144$                       &             $.110$                          & $ .128$                          &          $.092$                          &  \textbf{.149}                &  $.085$ \\
D.out B       & $.421$                & $-.017$                         &  \textbf{.183}                & \purp{ $.183$}                      &             $.143$                          & $ .143$                          &          $.111$                          &  $.152$                         &  $.086$ \\
D.out B+      & $.362$                & $ .015$                         &          $.228$                 &        $.236$                       &     \purp{  $.301$}                         & $ .177$                          &  \textbf{.238}                         &  $.191$                         &  $.088$ \\
Hate B        & $.444$                & $-.020$                         &  \textbf{.170}                &        $.148$                       &             $.104$                          & $ .162$                          &          $.123$                          &  $.143$                         &  $.073$ \\
Hate B+       & $.394$                & $ .034$                         &          $.188$                 &        $.174$                       &     \textbf{.238}                         & $ .215$                          &  \purp{  $.262$}                         &  $.183$                         &  $.065$ \\
BART          & $.378$                & $-.010$                         &  \textbf{.200}                &        $.157$                       &             $.127$                          & $ .142$                          &          $.115$                          &  \purp{ $.243$}  &  $.079$ \\
GPT-2         & $.303$                & $-.098$                         &          $.031$                 &        $.003$                       &             $-.020$                         & $.003$                           & $-.025$                                  &  \textbf{.048}                &  \purp{ $.597$} \\
\midrule
\textsc{mean} & $.399$                & $-.114$                         &  \textbf{.158}                &       $.138 $                       &              $.116$ &                          $.111 $ &                                  $.130$ &                           $.140$ &                       $.073$ \\
\bottomrule
\end{tabular}
\caption{Transferability and robustness of various $f_\text{aug}$ models on various augmented test samples $X_{\text{test}}'$. Shown is the original True Positive Rate (TPR, under \textsc{initial trp}) for $f$ (Plain), and the change ($\Delta$) in TPR of $f_{\text{aug}}(X_{\text{test}}')$ with respect to $f(X_{\text{test}}')$. Positive $\Delta$ \textsc{trp} shows robustness to perturbations after augmentation, negative the opposite. The classifiers most robust against same-model perturbations are highlighted, in bold the next-best. At the bottom, \textsc{mean} (per augmented classifier) TPR is shown, \emph{excluding} performance on the same model (to reflect transferability). Dropout is abbreviated to D.out, BERT to B.}
\label{tab:exp3}
\end{table*}

Given the strong performance effect the substitution models had on our classifiers, it seems plausible that they might prove to produce effective samples for augmentation purposes. Looking at the lower portion of Table~\ref{tab:exp1-2}; however, we can see that augmentation does not improve performance on the two biggest sets (Merged, and Ask.fm). Dropout BERT seems to improve performance for one of the smaller sets (Twitter II), but overall, EDA is generally a close contender with, if not more effective, than all of the more `advanced' models. Interestingly, the transformer-based models seem to yield sizable improvements on the Myspace set, with GPT-2 increasing it most. The latter might be attributed to the low Type-Token Ratio in this set (see Table~\ref{tab:data-dsc}).

\paragraph{Interpretation} Generally, none of these methods (baselines, or substitution-based augmentation) seem to yield the same performance improvement as observed in toxicity work \citep{DBLP:conf/icmla/IbrahimTE18,Jungiewicz_Smywinski-Pohl_2019}. However, note that, as we were interested in simulating adversarial behavior, we conducted model-agnostic augmentation (that is, given an unknown attacker, or noise). Hence, while we might employ these models in an explicit adversarial training scheme to directly improve model performance, this would require extensive transferability evaluations---typically requiring larger, higher quality datasets---and only satisfy one dimension (data). Given this, we argue an improvement in classifying the augmented sets, as in Experiment 1 (Section~\ref{subs:exp1}), is more significant.

\subsection{Augmentation for Robustness} \label{subs:exp3}

The results of Experiment 3 can be found in Table~\ref{tab:exp3}. We report TPR changes for $f(X_{\text{test}})$ (under initial TPR), and  $f_{\text{aug}}(X_{\text{test}}')$ per setting to create $f_{\text{aug}}$ and $X'$ respectively.

\paragraph{Robustness}

Generally, it can be observed that (unsurprisingly) the augmented classifiers increase TPR most when the substitutions come from the same type of model. Hence, the `second-best' TPR increases are more interesting. It can be observed that BERT and BART show strongest TPR improvements on three sets respectively, followed by the `+' models with one set respectively. This is quite a remarkable, contrasting result to Experiments 1 and 2, although it aligns with the observations from the semantic consistency scores that more conservative models are less effective augmenters. Hence, it seems that substitutions that are \emph{more} diverse, and distant from the original instances provide better robustness against perturbations. While their output might be less semantically consistent, this is generally not a relevant criterion when one is interested in improving task robustness.

\paragraph{Transferability} Systematic performance gain across all substitution models (i.e., transferability) is the final indicator of augmentation utility. First, it must be noted that the TPR differences between `same-model' and distinct model pairs are smaller for the transformer-based models ($.026$ on average) than EDA ($.156$). Lexical substitution using out-of-the-box BERT---in addition to high robustness---also achieves the highest transferability (mean $.158$) across substituted sets.

\paragraph{Performance Trade-Off}

The $f_{\text{aug}}$ results from the current experiment should be contextualized against the performance trade-off in $F_1$-score from Experiment 2. Using the information in Table~\ref{tab:exp3}, it can be inferred that the best performing model, BERT, actually improves absolute TPR on average; if we add its $.158$ mean TPR increase to the .399  average (= $.557$) this exceeds the $.537$ non-augmented TPR. However, as we showed in Experiment 2 (Table~\ref{tab:exp1-2}), this does not improve overall task performance; rather, it decreases performance. For BERT, the $F_1$-score slightly drops ($.025-.030$) on all sets. Hence, this is not a silver bullet, and such trade-offs should be considered when deploying these augmentation models to improve robustness against lexical variation.

\paragraph{Limitations and Future Work}

A substantial hurdle toward deploying the presented models for augmentation purposes is time. Upsampling the positive instances shown in Table~\ref{tab:set-freq} (5,350 total) with the transformer-based models takes 2-3 hours per model on a single NVIDIA Titan X (Pascal).\footnote{Training the BERT classifiers takes up to 31 hours, augmentation 40 minutes, predictions on the test set 5 minutes.} This impacts the amount of parameters that can be tweaked in reasonable time when using this architecture (such as omission score cut-offs, cosine similarity when ranking, dropout values, etc. which we all set empirically). Such computational demand is acceptable for smaller datasets like ours, and the augmentations can be run `offline' (i.e., one time only), but these limitations should certainly be taken into account when scaling is among one's desiderata.

Hence, recent work on decreasing the amount of queries for related models \citep{DBLP:journals/corr/abs-2106-07047} is particularly relevant for future work. Additionally, there is a myriad of components the base architecture we presented here could be improved with. Most are discussed in Chapter~\ref{chap:advsty}; however, some new work is specifically of interest to data augmentation, such as improving the substitutions using beam search \citep[as opposed to the simultaneous rollout we used in the current chapter]{DBLP:journals/corr/abs-2110-08036}. More broadly, adversarial training \citep{DBLP:conf/acl/SiZQLWLS21,DBLP:journals/corr/abs-2107-10137}, implementing more robust stylometric features \citep{markov-etal-2021-exploring}, or model-based weightings of the augmentation models could be explored; e.g., by selecting instances with a generation model in the loop \citep{DBLP:conf/aaai/Anaby-TavorCGKK20}. This could be a particularly worthwhile option when focusing on conversation scopes, rather than message-level cyberbullying content (Chapter~\ref{ch:bul}). 

Finally, this chapter focused specifically on augmenting cyberbullying corpora---a classification task which is generally (though equivocally) framed to extend beyond mere toxic content, and for which data is generally scarce. Although not in the scope of the presented work, our methods might be implemented in future work to further (critically) explicate the role of toxicity in this classification task, and thereby assist in curation of corpora, or contrast sets \citep{DBLP:journals/corr/abs-2004-02709,DBLP:conf/blackboxnlp/LiSLWZS20}, that are more representative of the theoretical underpinnings of cyberbullying.

\section{Related Work} \label{sec:rel-work}

Our work combines multiple sizeable---to the extent that they respectively produced several surveys \citep{DBLP:journals/csur/FortunaN18,gunasekara-nejadgholi-2018-review,DBLP:journals/corr/abs-1908-06024,banko-etal-2020-unified,madukwe-etal-2020-data,DBLP:journals/fi/MuneerF20,DBLP:journals/taffco/SalawuHL20,DBLP:journals/corr/abs-2106-00742,DBLP:journals/csur/MladenovicOS21}---areas of research; hence, we will provide a concise overview of the work directly related to our experimental setup.

For all tasks, the issue of generalization seems a particularly popular subject of study: for cyberbullying, in Chapter~\ref{ch:bul}, and \cite{DBLP:conf/asunam/LarochelleK20}, conclude there is little consensus in labeling practices, overlap between datasets, and that a combination of all datasets seems to transfer performance best. For hate speech, \cite{DBLP:journals/hcis/SalminenHCJAJ20}, and \cite{DBLP:journals/ipm/FortunaCW21}, draw similar conclusions, showing that general forms of harm (e.g., toxic, offensive) generalize better than specific ones, such as hate speech. Finally, \cite{nejadgholi-kiritchenko-2020-cross} provide unsupervised suggestions to address topic bias in data curation, potentially improving generalization. We draw from these works through cross-domain experiments on individual and combined corpora for cyberbullying, and pre-training on toxicity. 

Recent cyberbullying work \citep[e.g., are seminal work]{DBLP:conf/icmla/ReynoldsKE11,xu-etal-2012-learning,nitta-etal-2013-detecting,DBLP:conf/icis/BretschneiderWP14,DBLP:conf/ai/DadvarTJ14,van-hee-etal-2015-detection} has primarily focused on deploying Transformer-based models \citep{DBLP:conf/nips/VaswaniSPUJGKP17}; by and large fine-tuning \citep[e.g.]{swamy-etal-2019-studying,paul_cyberbert_2020,DBLP:journals/internet/Gencoglu21}, or re-training \citep{DBLP:journals/corr/abs-2010-12472} BERT. It is worth noting that \cite{DBLP:journals/access/ElsafouryKPR21,DBLP:conf/sigir/ElsafouryKWR21} show that although fine-tuning BERT achieves state-of-the-art performance in classification, its attention scores do not correlate with cyberbullying features, and they expect generalization of such models to be subpar. In our experiments, we employ similar domain-specific fine-tuned BERT models, and gauge generalization, sensitivity to perturbations, and the effects of augmentation to potentially improve the former.

Adversarial attacks on text \citep[e.g., provide broader surveys]{DBLP:journals/tist/ZhangSAL20,DBLP:journals/corr/abs-2103-00676} can roughly be divided in character-level and word-level. The former relates to purposefully misspelling or otherwise symbolically replacing text (e.g., \emph{fvk you}, \emph{@ssh*l3}) to subvert algorithms \citep{eger-etal-2019-text,DBLP:journals/corr/abs-1912-06872}. \cite{DBLP:conf/acl-alw/WuKS18} show such attacks on toxic content can be effectively deciphered. Word-level attacks are arguably straight-forward for humans, but significantly more challenging to automate---requiring preservation of toxicity; i.e, the semantics of the sentence. Previous work has investigated the effect of minimal edits on high-impact toxicity words, replacing them with harmless variants \citep{DBLP:journals/corr/HosseiniKZP17,brassard-gourdeau-khoury-2019-subversive}. Our current work is similar to that of \cite{DBLP:conf/micai/Tapia-TellezE20}, and closest to that of \cite{DBLP:conf/sepln/Guzman-Silverio20}, who apply simple synonym replacement using EDA, as well as adversarial token substitutions---the latter using TextFooler  on misclassified instances. We extend BERT-based lexical substitution \citep{zhou-etal-2019-bert} for model-agnostic perturbations, and data augmentation.

Finally, regarding data augmentation for online harms \citep[among others, provide more general-purpose overviews for various natural language data]{DBLP:journals/corr/abs-2107-03158,feng-etal-2021-survey}, toxicity work partly overlaps with work on adversarial attacks on text; for example, the synonym replacement from \cite{DBLP:conf/icmla/IbrahimTE18}, and \cite{Jungiewicz_Smywinski-Pohl_2019}, which are distinctly either unsupervised, or semi-supervised. Another such example can be found in \cite{rosenthal-etal-2021-solid} employed democratic co-training to collect a large corpus of toxic tweets, and \cite{gehman-etal-2020-realtoxicityprompts} find triggers that produce toxic content, querying GPT-like models \citep{radford2018improving}. Fully unsupervised augmentation has also been employed in \cite{quteineh-etal-2020-textual}, and \cite{DBLP:journals/corr/abs-2104-08826}. In our experiments, we use a pipeline of models for lexical substitution, and compare it to GPT generations \citep{radford2019language,DBLP:conf/nips/BrownMRSKDNSSAA20}.

\section{Conclusion}

In this chapter, we employed model-agnostic, transformer-based lexical substitutions to the task of cyberbullying classification. We show these perturbations significantly decrease classifier performance. Augmenting them using perturbed instances as new samples slightly trades off task performance with improved robustness against lexical variation. Future work should further investigate the use of these models to simulate and mitigate the effect of adversarial behavior in content moderation. 
\thispagestyle{empty}
\strut
\newpage

\begin{figure}
    \centering
    \includegraphics[width=0.75\textwidth,angle=-90]{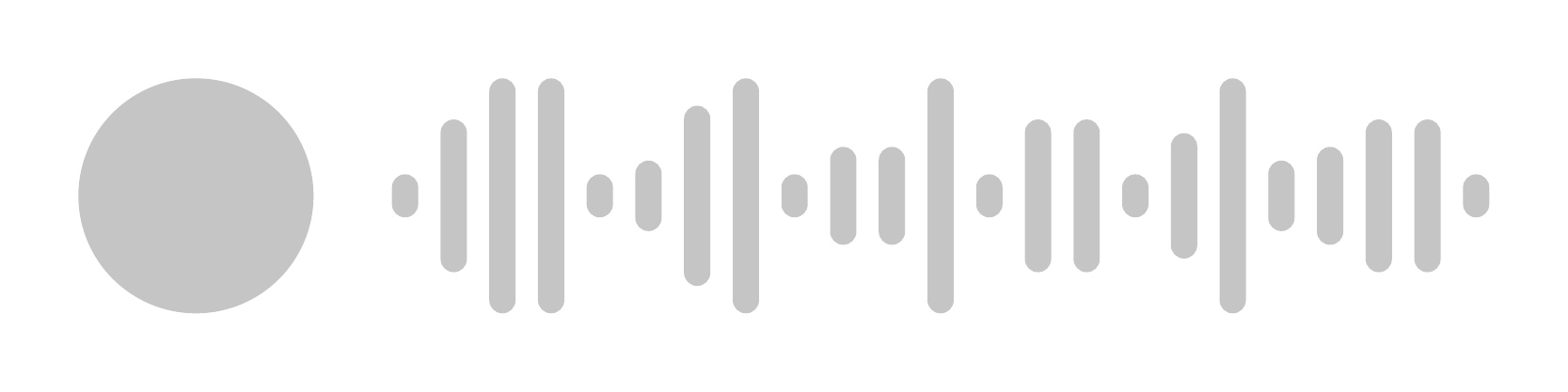}
\end{figure}

\thispagestyle{empty}
\strut
\newpage
\thispagestyle{empty}

\chapter{Discussion and Conclusion}
\label{ch:conclusion}

\lettrine[lines=4, lraise=0, nindent=0em, slope=0em]{I}{n this dissertation}, we have introduced the framework of \emph{user-centered security in Natural Language Processing}. We conducted a range of studies to demonstrate how such user-centered constraints can be implemented for two related domains: that of author profiling, and cyberbullying detection. Our experiments show these constraints are good indicators for task generalization. Not only from a Machine Learning perspective, but also a human one; i.e., does the current work on these tasks actually benefit humans? Can they employ these models in realistic scenarios? Toward these aims, we framed two research questions, which this chapter will further discuss while providing a summary of the conducted research, a discussion of its limitations, and future.   

\parshape=0
\section{Adversarial Stylometry}

For this task, we set out to answer the following research question: \newpage

\begin{itemize}
    \item[\textbf{RQ1}] To what extent can end-to-end and targeted attacks be employed for user-centered adversarial stylometry?
\end{itemize}

The initial part of this question, pertaining to end-to-end obfuscation, was covered in Chapter~\ref{chap:textobf}, where we explored the feasibility of an autoencoder-based obfuscator. Here, the envisaged scenario was for a user to train this model on their writing. An end-to-end model provided an obvious automated solution for this purpose. 

\subsection{End-to-end: Style Invariant Obfuscation}

After having identified limitations in the literature, we highlighted the importance of exploring the concept of `neutral' style.\footnote{Neutral with respect to the model. While framed as `formal' writing in style transfer research \citep[see e.g.,][]{madaan-etal-2020-politeness}, the concept of a genuinely neutral writing style is not a tangible one.} We initially argued this seemed much more faithful to the intended purpose of the obfuscation task; are we otherwise not just imitating another writing style? Moreover, though, and as indicated in this chapter, it seems an important limitation for most proposed end-to-end models: minimizing target model performance might require few alterations, but it is highly likely these are strongly correlated with a \emph{different} label. Not only does this encourage semantically inconsistent perturbations, it is also not a user-centered solution. One can imagine users of obfuscation models would not want to apply style \emph{transfer}, as being associated with an opposite style might be undesirable (adults writing teen slang, republicans talking about universal basic income, etc.).

Producing language that is invariant \cite[a more common term in Computer Vision, see e.g.,][]{DBLP:conf/nips/NamK18}, or neutral with respect to the different styles seems to be somewhat of an ambitious framing in hindsight, however. Ideally, we would want to produce language that is style neutral with respect to \emph{a target model}, or some approximation of that model. This distinction is particularly relevant---although an unlikely scenario in practice  due to user-specific resource requirements---if the target model party were to train a model to explicitly recover the attributes, which would likely still be able to recover style.\footnote{As demonstrated in, e.g., later work on attempting to debias models using Gradient Reversal by \cite{gonen-goldberg-2019-lipstick}.}

For the end-to-end work in Chapter~\ref{chap:advsty}, we were interested in implementing these ideas through several comparisons: i) what is the optimal case in which we obfuscate through \emph{style transfer} (i.e., unrealistically assuming we have access to parallel sentences), ii) how feasible is a style-invariant encoding, and iii) can we move away from parallel data by using a different architecture?\footnote{Similar efforts to move away from parallel data have later also been discussed for style transfer in \cite{john-etal-2019-disentangled}.} We compared sequence-to-sequence (i.e., parallel) models with autoencoders, and the effect of adding a Gradient Reversal Layer for style invariance (which worked most effectively via implicit style conditioning). To this effect, we formally defined what it means to optimize for obfuscation-by-transfer, and obfuscation-by-invariance: target model performance close to chance level implies \emph{style invariance}, and performance close to zero \emph{style transfer}. 

Formally, we achieved satisfactory results (i.e., close to chance performance, while still achieving high \textsc{bleu} and \textsc{meteor} scores). Our model's output was also rated of higher quality by human raters than a style transfer model with access to parallel data. Yet, the limitations of this initial study provide some sober context to these results. Generally, the set-up could have been improved with a more thorough evaluation (and justification) of the conditional autoencoder. In this work, we opted for \textsc{blue} and \textsc{meteor}; however, since we used the input sentence as reference sentence, the autoencoder had an unfair preservative advantage. Even WMD (which uses an embedding space) is likely to favor sentences that are `semantically close' to a (neural) model, and thus favor conservative models. For automatic evaluation, the combined source and target \textsc{blue} and \textsc{meteor} scores are interesting, but can only be employed when using parallel data. A fairer approach might be to use a Language Model (LM) to score the outputs based on perplexity, specifically for word changes.

Through manual inspection of the input, we also observed intrusive grammar changes for \emph{all} models. We alluded to this drawback of end-to-end models in Chapter~\ref{ch:introduction}; the lack of control over the output meant our models often replaced words with unrelated variants. This implies that the models do not seem to learn robust semantics from scratch, which in retrospect could have been partially avoided by using pre-trained embeddings.\footnote{Which is not surprising; the Bible is a very restricted language source.} Finally, despite being much more informative, and scores being in line with prior work, the human evaluation proved tedious and paired evaluation prevented testing inconspicuousness of the changes.

\subsection{Targeted: Lexical Substitution}

While building on ideas from Chapter~\ref{chap:textobf}, our work from Chapter~\ref{chap:advsty} primarily tried to address both the experimental and user-centered limitations from our initial research. Hence, we adapted the same style-neutral framing, but heavily focused on targeted obfuscation, transferability of the models, and a strict adherence to our realistic criteria for training and deployment. We improved the adversarial framework proposed by \cite{DBLP:conf/aaai/JinJZS20} with several transformer-based methods to find semantically similar perturbations, and showed these attacks fully transfer (that is, within the bounds of chance level performance). 
This implies that a given user can train their \emph{own architecture}, on their \emph{own data} (collectively: a substitute model), and assume it to effectively confuse an unknown target model. Moreover, a particularly promising result from this work is that the model with the highest transferability used a distantly collected dataset \cite{emmery-etal-2017-simple}. Hence, even the data collection was performed under user-centered restrictions, while the target model used gold standard corpora, from another domain.

These results are certainly promising, though some limitations remain, which are worthwhile to address in future research. Primarily, we only considered a single attribute (binary gender). As we argued, it can be assumed these techniques transfer to word-level models, as these architectures are generally similar in related work. However, this does not consider different stylometric features---nor does it take into account the potential complexity (future) transfer models might offer. An obvious patch would be `more of everything': different attributes, various domains (Reddit being particularly relevant), and various architectures. These, however, are subject to another limitation: forwarding through BERT is quite slow, and tailoring an attack to our substitute models can therefore make such a complex setup unfeasible with modest computational resources. Such complexity also does not benefit a user; however, with developments in distilled transformer variants \citep[see, e.g.,][]{DBLP:journals/corr/abs-1910-01108}, this might change in the future.

In this work, we opted for proposing an alternative human evaluation; mainly to decrease annotator effort, but more so as the format can be more conveniently integrated in an online annotation application in the future. A limitation of this approach is that it does not capture the quality of the replacements; it assumes that full inconspicuousness also correlates with perfect grammaticality. While it is a faithful simplification of metrics proposed by, e.g., \cite{DBLP:conf/clef/PotthastHS16} and \cite{DBLP:conf/clef/HagenPS17}, it is less useful from a user perspective; i.e., it says nothing about the \emph{intended} meaning, only if it is a plausible sequence. This is another instance where adding LM perplexity is a useful addition to the evaluation. Lastly, we did not explicitly incorporate recent work on detecting obfuscators \citep[see, e.g.,][]{mahmood-etal-2020-girl,mozes-etal-2021-frequency,DBLP:conf/iclr/DongLJ021}; however, we did measure human detectability.

\subsection{Future Work}

For the lines of work we set out in this dissertation in particular, there are some potentially fruitful avenues to explore. For Chapter~\ref{chap:textobf}, despite its discussed limitations, the invariant encoding still warrants further investigation. In particular, given the recent developments in pre-trained transformer models, the architecture would require much less domain-based fine-tuning. Furthermore, following \cite{gonen-goldberg-2019-lipstick}, it might be interesting to train an ensemble of invariant obfuscators against an ensemble of classifiers (i.e., boosting the obfuscator). Ensembles of obfuscators were recently employed by \cite{DBLP:journals/corr/abs-2109-07028}, to which this could prove a valuable addition.

For Chapter~\ref{chap:advsty}, in addition to the improvements we discussed in the previous section, there are numerous components that are worth expanding upon in the existing architecture. Currently, sampling of replacement candidates happens somewhat naively; every substitution candidate is evaluated against the target model in isolation, and whatever decreases classification probability the most is selected as a replacement. This does not consider other candidates, or combinations thereof. An interesting extension would be improving the attack's sampling method, as some related work has already investigated, through, for example, (improved) beam search, and combinatorial optimization \citep{yoo-etal-2020-searching,DBLP:journals/taslp/LiuLCT22,DBLP:journals/corr/abs-2110-08036}. 

The scope of the replacements, and framing of the task, might be somewhat of a higher-level point to address. We currently deliberately focus on surface-level (content) words, as they have the strongest direct influence; however, one might question if this a genuinely robust method of producing adversarial attacks. Are we just looking for a model's out-of-vocabulary words, or do we actually want to change the style? Current alternatives can be found in, e.g., orthographic changes \cite{DBLP:conf/iclr/BelinkovB18}, though they have proven not to be robust \citep{keller-etal-2021-bert}. Adversarial triggers \citep[see, e.g.,][]{wallace-etal-2019-universal,song-etal-2021-universal} remain a strong, non-invasive alternative, but they are unfortunately often tailored to one particular model, hence unlikely to transfer, and therefore do not adhere to a user-centered obfuscation approach. A meet-in-the-middle approach between end-to-end and targeted models might provide a solution here; in particular, allowing for some more attack creativity by taking larger spans of words, but constrained enough as not to change too much of the input.

Most of these suggestions require one key point of future work, though: finding an effective way to unify the individually flawed evaluation metrics for this task.
The classic Machine Translation and Natural Language Generation metrics are difficult to interpret with regard to \emph{changes}; ideally, those changes are very few, and as a consequence, substitutions should be close to perfect. Moreover, it might be worthwhile to investigate if these correlate with human judgement on this task specifically, as to provide some critical support for them. Currently, based on our observations from Chapter~\ref{chap:advsty}, obfuscation performance and the automatic evaluations do not seem to agree with human judgement. Alternative model-based semantic scorers, such as \textsc{Bleurt} \citep{sellam-etal-2020-bleurt} and \textsc{BERTScore} \citep{zhang2019bertscore} raise similar concerns; if one samples synonyms (or full semantic representations) from the same models, it seems rather evident these would score high, both on sentence and world-level.\footnote{We provide further evidence for this in Chapter~\ref{ch:aug}.} The same might be said for measures we did not try: \textsc{terp} \citep{snover-etal-2009-fluency}, \textsc{pinc} \citep{chen-dolan-2011-collecting}, LM-based perplexity---these will yield high sentence-level scores for controlled perturbations, giving the illusion models are performing well, and favoring more conservative models.

All things considered, the most valuable direction for future work for adversarial stylometry is therefore finding an appropriate balance between obfuscation performance and detectability. Not much progress can be found in this regard, with recent papers \citep[see, e.g.,][]{DBLP:journals/corr/abs-2001-07820} seemingly rediscovering points made in seminal work \citep{DBLP:conf/clef/PotthastHS16}. Future options, as we see them, are twofold: as we previously proposed, a unified metric (while not without limitations) could for example weight changes inversely against the number of perturbations, and the associated obfuscation power. We would, however, also argue it is somewhat difficult not to think of chasing such a metric as solely promoting competitive benchmarking. Alternatively, it seems more appropriate to improve the method through which humans evaluate this task. This should partially draw from existing work and related fields \citep[see, e.g.,][]{DBLP:conf/clef/PotthastSHS18,van-der-lee-etal-2019-best,ippolito-etal-2020-automatic,howcroft-etal-2020-twenty}, however, most importantly, take into account how this task differs. Particularly when using several target classifiers, obfuscators, and de-obfuscators, an interactive human-in-the-loop evaluation could yield high-quality evaluations, as well as a direct signal to improve the existing systems \citep[see e.g.,][]{mozes-etal-2021-contrasting}. Humans could even generate a set of 'perfect' edits interactively, providing a gold standard quality measure, and an upper bound on the typical evaluation scores at the same time.

\section{Cyberbullying Detection}

For this task, we set out to answer the following research question:

\begin{itemize}
    \item[\textbf{RQ2}] Do cyberbullying classifiers prove robust enough to be deployed in a user-centered fashion?
\end{itemize}

The biggest contribution to answering this question was made in Chapter~\ref{ch:bul}, where we conducted an elaborate set of experiments gauging how current baselines and state-of-the-art classifiers for cyberbullying detection generalize, how well they represent the task, and how current limitations might be addressed. As this work in itself is a discussion piece, we will mainly highlight how our findings relate to the user-centered framework we introduced.

\subsection{Task Robustness}

At the heart of executing this task in a user-centered fashion lies generalization. In the line of research conducted for \Cref{ch:bul}, we were interested in critically determining how the framing of the task of cyberbullying influences its ability to perform out of domain. In particular, we hypothesized that the current framing of ML research misrepresented the task through lack of theoretical and experimental rigor.\footnote{See \Cpageref{foot:bul-rant}.} Since our initial pre-print, several studies further delineated issues with broader online harms research \citep[see, e.g.,][]{10.1145/3476066}: all raising issues with task granularity, `toy' datasets, and the limited applicability of the models. Accordingly, generalization in particular remains a prevalent issue in cyberbullying detection research; the limited available data starkly contrasts the task complexity. Numerous research does state this in their limitations, but little to no work currently exists on addressing these issues. We believe that its core limitation therefore lies in cyberbullying detection being framed as an industry-centered task (providing platform-level classification) without having access to industry-scale data. The alternative framing of user-centered bullying detection is an interesting avenue that largely remains explored.

In \Cref{ch:bul}, we admittedly only scratched the surface. Mainly, we showed that---as argued in \Cref{ch:introduction}---if a user has multiple platforms of communication, the current systems will fail, as they do not generalize across domains. While we did investigate restrictions on the scope of the data through single messages (which is the most common format for this task), and conversation scopes (e.g., full profiles), these all assume access to many different users. The setting least fitting to our user-centered framework was profile-level classifications, which essentially turn the task into predicting if a person is being bullied, rather than (a collection of) messages classifying as bullying. For classification, however, it was greatly beneficial: with the datasets now more balanced, and larger contexts, it improved model performance significantly. Deployment of such models would need to be done in a centralized system, with integrated reporting to third parties, to be effective---exactly what we try to move away from. Further limitations apply to our efforts to collect more data by simulating bullying events; while not problematically centralized, such efforts provide arguably costly, and most likely static sources of data. The former limitation might be solved through collective effort; i.e., construction of much larger corpora, either anonymized or crowd-sourced. The latter limitation, however, would require such efforts to be frequent, as toxic language might change over time.

\subsection{Lexical Augmentation}

This brings us to another limitation of our, and a large part of, existing research: the sensitivity of these classifiers to lexical cues. We investigated lexical variety in line with Hypothesis 1 of \Cref{ch:bul}, the experiments of which show a strong focus on lexical cues for message-level classification---particularly, on blatant profanity. This has strong implications for existing work, as creativity, cross-cultural characteristics \citep{jbp:/content/journals/10.1075/jlac.1.1.05mat}, and bias in toxic language use \citep{sap-etal-2019-risk,wiegand-etal-2019-detection} all diminish the utility of the currently available corpora. 

Hence, in \Cref{ch:aug}, our goal was twofold: firstly, to demonstrate how this focus on lexical cues permeates (even)  state-of-the-art classifiers based on large language models. Secondly, we were interested in seeing if similar techniques used to generate creative language could also be used for data augmentation purposes for small(er) resources. The latter in particular is highly relevant for our user-centered framework. It would mean that we can generate local training data, as well as make existing sources more robust to lexical variety. In doing so, this research is the culmination of work investigated in this dissertation: creative language use might have a natural origin, but could also stem from adversarial behavior. Hence, while before we were interested in simulating detection \emph{models} and targeting them, here, we simulated \emph{users} providing adversarial input. There is one important deviation, though; the architecture we used in \Cref{chap:advsty} tailored substitutions to a target model, whereas here we did not. By simply accepting substitution candidates as-is, the method becomes semi-supervised. This aligns with our user-centered framework, as the core setup only requires a small annotated dataset (which could be historical data provided by a user, or similar to our setup: a freely available external corpus) and a simple classifier.

In this chapter, we already touched upon the core limitation of this work: time, and hence scalability. The different experiments we ran, where all parameters were set empirically beforehand, took several days to run on a(n expensive) GPU machine---the individual augmentation models up to a day. If we want to pursue user-sided classification, the lexical substitution models we proposed do not seem a realistic option. Fortunately, out-of-the-box BERT seemed to provide the most robust, transferable augmentations. This means that, similar to the proposed extensions for \Cref{chap:advsty}, more light-weight models \cite[][being another very recent example]{DBLP:journals/corr/abs-2110-08207} are a promising direction. In particular, this implies models not requiring custom embedding initialization (which the Dropout techniques required) can be employed, opening up many more possibilities. In line with this, we also made the observation that more noisy, and less lexically sound replacements produced higher quality augmentations. Hence, something  worthwhile exploring for this particular task---which we did not, given the computational cost of the experiments---is testing our semi-supervised framework using a wider range of augmentation models \cite[as can, e.g., be found in][]{DBLP:conf/emnlp/MorrisLYGJQ20}. As our augmentations proved ineffective to improving the classification task, we focused on in-depth analyses regarding the models, rather than cross-platform and cross-domain (e.g., transferability of these techniques to toxicity detection and hate speech) experiments. Future directions with less strong trade-offs between task performance and robustness, however, should.

\subsection{Future Work}

Since the start of research on cyberbullying detection in context of the AMiCA project, both this field and the more general variant of toxicity detection have become very popular topics of research. Here, we identify several current trends (context, fairness, augmentation, and LM prompting), and briefly discuss relevant recent work, and how these directions might advance the work presented in this dissertation.

As we discussed and demonstrated in \Cref{ch:bul}, conversational context might change both annotator labeling, and classification. \cite{DBLP:conf/icwsm/CartonMR20} combined these two ideas, by incorporating model-contextual explanations into annotator tasks, partly for decreasing annotation load. They found that among the annotators, false negative increase whereas false positives decrease, concluding their observations to be ambiguous with respect to utility. \cite{pavlopoulos-etal-2020-toxicity} find the same ambiguity for literal context in annotation; 5\% of the instances are annotated with a different label when context is not shown. Conversely, they found that conditioning models on context does not lead to model improvements. Despite a slightly different task of application, this provides an interesting contrast to what we found in \Cref{ch:bul}. Further observations of others provide different uses of context: e.g., \cite{DBLP:conf/icwsm/RadfarSC20}, who show social (rather than conversational) context to be of importance, as unconnected social media users are three times more likely to engage in toxic conversations. Additionally, \cite{DBLP:journals/corr/abs-2111-10223} find knowledge-distillation-enhanced systems to provide contextual change classification to toxicity, which they propose might be employed to improve toxicity detection (particularly moderation). The latter two proposed uses of context in particular can be deployed in an isolated fashion, not needing a bigger social media environments to operate on, and are therefore worthwhile to fit into a user-centered approach of detecting toxic content.

To that effect, the label representations we have used have been quite limited. While one can imagine different users do not deal with the same amount of toxicity (e.g., marginalized communities dealing with purely identity-focused hate), accounting for a wider variety can prove beneficial; binary classification of toxic and `dirty' content has previously been argued to reproduce society's harmful inequalities \citep{thylstrup2020detecting}. Algorithmic attempts to mitigate this are generally proposed through incorporating fairness constraints \citep{DBLP:journals/corr/abs-2011-06485,DBLP:journals/internet/Gencoglu21}; i.e., actively making toxicity classification invariant to, e.g., race, gender, and topical contexts, which additionally seems to boost classifier performance. While important in a wider context, such methods require extensive user information (at least for training), and by extension a larger social system to operate in. Hence, data sources providing better motivated, fine-grained approaches to classification might be more appropriate. \cite{DBLP:journals/corr/abs-2010-14952}, for example, propose better group representations, a taxonomy of subject matter, and only after applying an annotation scheme for properly contextualized severity. It remains to be seen, however, to what extent this can provide downstream accuracy. Some preliminary work shows ensemble models to be effective in such settings \citep{burtenshaw-kestemont-2021-dutch}, and further recommendations in working with small dataset are provided by \cite{DBLP:conf/www/ZhaoZH21}.

In addition to our results from \Cref{ch:aug}, augmentation seems a fruitful (and popular) direction. Recently, \cite{han-tsvetkov-2020-fortifying} have tried to augment their (toxicity) classifiers using a student-teacher setup where they actively uncover more subtle (`veiled') forms of toxicity such as misogyny, racist jokes, trans hate, etc. We would argue the diversity of datasets and how this propagates into models plays quite a big role here. In addition to \cite{gehman-etal-2020-realtoxicityprompts} showing toxic LM prompt generation behavior, and finding hateful Subreddits retained in the CommonCrawl dataset, cases can also be found outside of social media text; e.g., in \citet{DBLP:journals/corr/abs-2110-01963}, who found large bodies of toxic content and violent sexual material (image captioning) dataset. Hence, `normalization' of such content in these models might play a role in why querying for toxic content (similar to what we took advantage of in our work) is rather easy---partly demonstrated in \cite{welbl-etal-2021-challenges-detoxifying}. While this behavior of large LMs is damaging in deployment, it is worth further investigating to what extent unsupervised algorithms (particularly newer ones, such as GPT-3) could in the future be leveraged to generate, and identify toxic data instances. In particular, in order to better understand (we would not go as far as to say improve) harmful content in the language resources they are trained on, and if there is potential in leveraging them to ease computational burden on individual users, and their dependence on external parties providing content moderation.

\section{Conclusion}

This dissertation presented and motivated a user-centered framework for security research in Natural Language Processing. To investigate its application, we focused on two tasks: adversarial stylometry, focused on privacy, and cyberbullying detection, focused on security. In these respective tasks, we investigated the feasibility of several desiderata: generalization across different domains a user might operate in, individual classifiers that work on data accessible to a user, work out of the box, and in reasonable runtime. For adversarial stylometry, we investigated end-to-end, and fully model-agnostic targeted models. Lexical substitutions proved a good fit to the user-centered framework, with adequate obfuscation potential of sensitive author attributes, and high transferability. For cyberbullying detection, we investigated methods to gauge generalization, and improving the robustness of classifiers against lexical variation. Here, our model-agnostic setup also showed high transferability of defenses against potentially adversarial input. We believe the constraints provided by our framework have not only been demonstrated to provide more realistic evaluations when classifying natural language data, but, more importantly, it provides an avenue for individual users to reclaim consent over their data used in machine learning models and their utility. All research presented in this dissertation is provided open-source, open-access, with publicly available datasets, and documentation how to reproduce and re-use the presented work.

\thispagestyle{empty}
\fancyhead[RE]{\scshape acknowledgments}
\fancyhead[LO]{\scshape acknowledgments}
\begin{figure}
    \centering
    \includegraphics[width=0.75\textwidth,angle=-90]{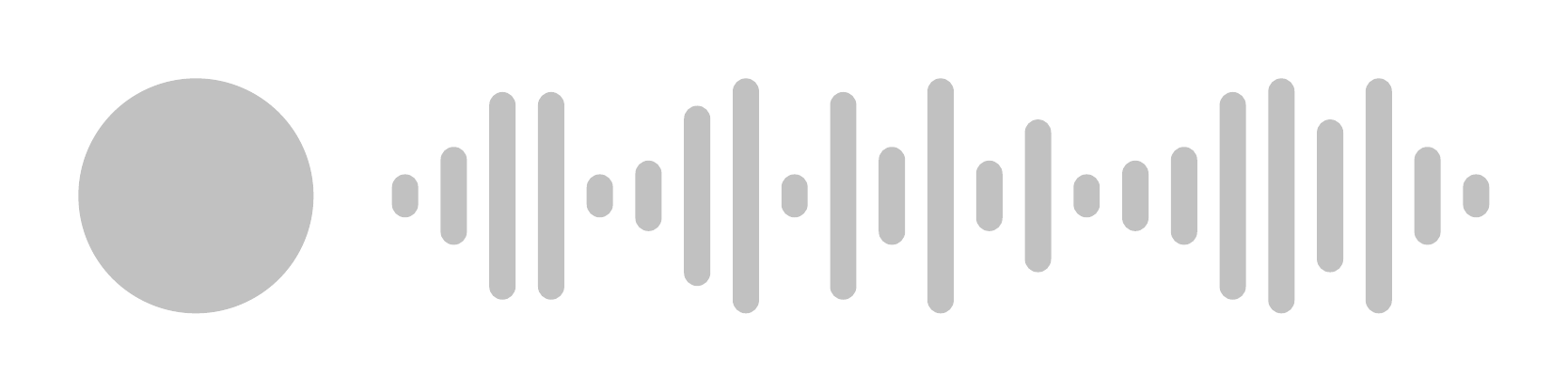}
\end{figure}
\vfill

\thispagestyle{empty}
\strut
\newpage
\thispagestyle{empty}
\strut
\vfill

\phantomsection
\addcontentsline{toc}{chapter}{Acknowledgments}

\chapter*{Acknowledgments}

\lettrine[lines=4, lraise=0, nindent=0em, slope=0em]{H}{ello}! If you're reading this, that means you cared enough to at least flip through some pages of this fancy bundling of papers---I appreciate you. I spent a long time producing it; not only in writing the papers, but also typesetting the document. I had about a year between submitting the first draft and defending, so I really went balls to the wall with \LaTeX{} for a bit. Truth be told, it was the only part that was exclusively fun. Hopefully that does not go unnoticed.

\parshape=0
That being said, in characteristic fashion, I have postponed writing this section until the last few days before committing it to print. If you spend seven years scrambling to find time between other obligations (first software development, later teaching) to work on the same project, crediting everyone being along for the journey becomes a bit trickier. I will do my best, though. \bigskip

If I have to point to one specific moment in time that kindled the core theme of my academic career, it would start with a conversation over coffee at my dad's place around the second year of my studies. Despite a seeming lack of academics in our family (that I know of), (techno-)skepticism was well-represented. Him having both a history working on low-level stuff as an embedded software engineer (e.g., ATMs, military comms), and levels of paranoia that seemed to increase with age, often made for some heated discussions. This particular one featured my early-20s brain downplaying his hypotheticals regarding data collection efforts by our government. Within a year of this conversation, Peter Vlemmix' (fellow Eindhovenaar) \emph{Panopticon} released, and Edward Snowden gave the world a peek into mass-surveillance activities. Quite the rude personal awakening.

Subsequently, while I quickly found Machine Learning (ML) a fascinating area of research, it did not take long into my PhD for my newly acquired skepticism patch to start applying itself there too. Many things were in stark contrast with my personal values. Sure, it was partly the rat race culture, but more so the numerous examples of impressive levels of disregard for experimental rigor, and in particular a trajectory of massive corpos disregarding basic ethical standards to drive the academic hype train for financial gain. So, I (quite ironically) found myself in a field actively (granted, often indirectly) contributing to monitoring of, and influence on, our daily lives. Luckily, ML proves rather effective at sabotaging itself, and hence there was fuel to write the dissertation you are reading through.

So yes, the work that lies before you is the result of skepticism, cynicism, maybe even some latent activism. While our relationship was far from great, it was some common ground I shared with my dad (who passed away during the last year of writing the dissertation), and I think he would've enjoyed reading it. For the most part, though, in due time I hope it will become an obsolete stack of papers. A booklet my son Owen will look at over a coffee with a similar 20-something brain I had before him; but, that for him---rather than the start of a trajectory of disenchantment---it will be a lesson not to spend his valuable time yelling at clouds.

This dissertation was written in turbulent times; personally, professionally, societally. Through it all, for the fifteen+ years we have been together, Monique continues to prove to be an invaluable member of team Life. There is no amount of words to express how much her and our son Owen enrich my existence. However long we still have together, I wouldn't trade it for the world. Without Monique's support, care, and understanding, there would not have been a dissertation. Thanks for everything.

For the academic part of life, first and foremost, many thanks to my supervisors. Prof. Dr. Walter Daelemans and his lab at CLiPS introduced me to the Computational Linguistics groups in the area, and his work and ideas were the starting point of most of the papers in this dissertation. Prof. Dr. Eric Postma trusted me enough to lecture in Tilburg’s Data Science master so that I could continue my dissertation, and was unwaveringly supportive of the line of research I tried to set out for this. His office was a place I felt I could always share my ideas and views without needing to put on my serious academic hat. Dr. Grzegorz Chrupa\l{}a took on the task of supervising this somewhat distant line of research. He’s proven a valuable critical filter through which I was allowed to push ideas, and even entire papers, often on extremely late notice. Most of all, I appreciate the patience the three of you had with my rather unconventional position and approach during all of this.

I’d also like to extend my gratitude to the committee: \committeeOne,
\committeeTwo, 
\committeeThree, 
\committeeFour, and 
\committeeFive; for their willingness to read, and---to my positive surprise---provide incredibly nice comments on the current work (I’m used to a particular part of the *ACL reviewer pool).

While I’ve been rather low-maintenance in supervision time (or so I’d like to think), two people have been particularly important members of our ``non-hierarchical coaching-oriented culture'': Dr. \'{A}kos K\'{a}d\'{a}r for research discussions, and Dr. Travis Wiltshire for everything else the university deems important. I’ve had the privilege to spend much time with both of you on things we enjoy outside of academia. It’s been a real pleasure working with both of you, and if I wasn’t overly contrarian, you would’ve certainly had to stand behind me for well over an hour.

I’d also like to thank the other members of the sad penguin group: my back-up clone Bram Willemsen, fellow academic dad Dr. Bertrand Higy (we miss both of you), and the tallest personality, Dr. Chris van der Lee. Our daily (digital) banter, (bi-)weekly beers (shoutout to Cat's Back), and ramen (same to Pig \& Rye), have been a real treat that kept work enjoyable, especially when it wasn't. You've all become very dear friends---may that last for life.

We have an awesome department, and I want to thank everyone for contributing to that (seriously, I'd list everyone if I wasn't on a tight schedule). In particular, however, I owe maintenance of my levelheadedness to a few people in particular: Dr. Henry Brighton, Dr. Elisabeth (Lisanne) Huis in 't Velt, Dr. Drew Hendrickson, and Dr. Emmanuel Keuleers. Thanks for helping me navigate the (partly, some of the less exciting) sides of academia.

To our current and former PhDs in particular: while slowly rolling out this work at two locations, I've seen many come and go. A lot of you were there for most of it, and I'm fairly confident all of you have helped get this work to where it is in some way or another. The Antwerp crew I was hands down the slowest of: Dr. Giovanni Cassani, Dr. Robert Grimm, Dr. Enrique Manjavacas, and Dr. St\'{e}phan Tulkens; the rogue PhD committee: George Aalbers, Lieke Gelderloos, and Dr. Paris Mavromoustakos Blom; and those of you grouped by lacking a convenient grouping: Dr. Thiago Castro Ferreira, Mariana Dias Da Silva, Dr. Mirjam de Haas, Dr. Nanne van Noord, Javad Pourmostafa, and Lisa Rombout. My thanks to you for being around on the Path.

To round off the academic credits, I'd like to thank Karin Berkhout and Eva Verschoor for excusing much of my bad organizational skills (they are improving, I promise) and subsequent last-minute requests, and Lars Biemans for keeping my battle station up.

On the slightly less academic side of life, I want to thank my other, older, but certainly dear friends: Dr. Suzanne Aussems, Thom van Iersel, Vincent Lichtenberg, Kristel Litjens, and Dr. Danny Merkx. Certainly the last few years, I've not been easy to get a hold of, and I'm very thankful that never seems to matter.

For your care and support in life outside of academia, I want to thank my family, particularly my mom Inge Koenen and stepdad Ignace Cichy for the many Fridays off the last two years, my parents-in-law: Bernadette, and Hans (who passed away the same year as my dad) being available any time their help was necessary, and my grandmother Trini Koenen (who passed away early 2020) who was Monique and my biggest fan. To close off, my younger---but without a doubt much brighter---brother Dani\"{e}l Emmery, who's currently also pursuing a PhD (doing actual science). It's a real privilege to have a sibling you connect with and can always rely on. I do not take that for granted. I hope we'll soon again find time to conveniently forget we're getting too old for some video games, and that we're able to produce some fun (but mostly useful) Emmery \& Emmery publications.

Last but not least, I'd like to thank my two trusty workhorses Onyx and Thurisaz (I sure hope somebody planted some trees to offset your carbon footprints), and Lubom\'{i}r Bene\v{s}, Vladim\'{i}r Jir\'{a}nek, Kees Prins, and Siem van Leeuwen for effectively distracting Owen during the last two years. \\ \strut \hfill A je to!


\thispagestyle{empty}
\fancyhead[RE]{\scshape summary}
\fancyhead[LO]{\scshape summary}
\clearpage

\phantomsection
\addcontentsline{toc}{chapter}{Summary}

\chapter*{Summary}

    

Services that collect, and process (sensitive) personal data through algorithms have become as ubiquitous as the Internet itself. Their increasingly centralized character, having now also permeated academia, raises societal and academic concern regarding its alignment with public interest. While related academic work may—by and large—address societally relevant problems, the resulting tools are often not provided to be used by the public, and, more importantly, nor are its methods designed from a typical user's point of view. This has become particularly true for the field of Natural Language Processing (NLP), where despite efforts to 'democratize' its state-of-the-art techniques (requiring enterprise-scale compute), there is a simultaneous trend toward monetizing access to such models. This is problematic from a security and privacy point of view, as it impedes inspection of the models, addressing dual-use issues, and control over data usage.

In an effort to address these concerns, this dissertation proposes a framework of user-centered security in NLP, and demonstrates how it can improve the accessibility of related research. Accordingly, it focuses on two security domains within NLP with great public interest. First, that of author profiling, which can be employed to compromise online privacy through invasive inferences. Without detailed insight into, and access to, these models their predictions, there is no reasonable heuristic by which Internet users might defend themselves from such inferences. Secondly, that of cyberbullying detection, which assumes a centralized approach; i.e., content moderation across social platforms. As access to appropriate data is restricted, and the nature of the task rapidly evolves (both through lexical variation, and cultural shifts), the effectiveness of its classifiers is greatly diminished and thereby often misrepresented.

Under the proposed framework, we investigate the use of adversarial attacks on language; i.e., changing a given input (generating adversarial samples) such that a given model does not function as intended. These attacks form a common thread between our user-centered security problems; they are highly relevant for privacy-preserving obfuscation methods against author profiling, and adversarial samples might also prove useful to assess the influence of lexical variation and augmentation on cyberbullying detection. We aim to answer the following research questions:

\begin{itemize}
    \item[\textbf{RQ1}] To what extent can end-to-end and targeted attacks be employed for user-centered adversarial stylometry?
    \item[\textbf{RQ2}] Do cyberbullying classifiers prove robust enough to be deployed in a user-centered fashion?
\end{itemize}

In doing so, we provide the following contributions: Chapter 1 proposes a style-neutral framing for adversarial stylometry, Chapter 2 explicitly measures the transferability of adversarial stylometry, Chapter 3 assesses the generalization of cyberbullying classifiers, and Chapter 4 cyberbullying classifiers their sensitivity to lexical variation, and the potential of augmenting its corpora.

\clearpage

\phantomsection
\selectlanguage{dutch}
\chapter*{Samenvatting}

Diensten die (gevoelige) persoonsgegevens verzamelen en verwerken middels algoritmen zijn net zo alomtegenwoordig als het Internet zelf. Hun steeds centralere karakter, en invloed op de academische wereld, baart maatschappelijke en academische zorgen over de waarborging van het publieke belang. Hoewel academisch werk over het algemeen maatschappelijk relevante problemen aan probeert te aanpakken, is de software die daar uit voortvloeit vaak niet afgestemd op publiek gebruik, en, wellicht nog belangrijker, zijn de methoden ook niet ontworpen met oog op de potentiële gebruikers daarvan. Dit is met name het geval in het gebied van natuurlijke taalverwerking (ook wel Natural Language Processing, of NLP). Ondanks pogingen om de state-of-the-art technieken (welke enorme hoeveelheden rekenkracht vereisen) te `democratiseren', ontwikkelde er gelijktijdig een trend om munt te slaan uit de toegang tot dergelijke modellen. Dit is uiterst zorgelijk vanuit het oogpunt van beveiliging en privacy; voornamelijk omdat het de transparantie van de modellen, het voorkomen van dubbelgebruik, en de controle over datagebruik in de weg staat.

In een poging om deze problemen aan te pakken introduceert dit proefschrift een raamwerk voor \emph{gebruikersgerichte beveiliging in NLP}, en laat het zien hoe dit de toegankelijkheid van gerelateerd onderzoek kan verbeteren. Het richt zich specifiek op twee veiligheidsdomeinen binnen NLP met een grote publieke belangstelling. Ten eerste, die van auteursprofilering, wat online privacy in gevaar kan brengen door middel van invasieve inferenties. Zonder gedetailleerd inzicht in---en toegang tot---deze modellen en hun voorspellingen, is er geen effectieve heuristiek waarmee internetgebruikers zich tegen dergelijke inferenties kunnen verdedigen. Ten tweede, die van de detectie van cyberpesten, waar het uitgaat van een gecentraliseerde aanpak; i.e., contentmoderatie op sociale platforms. Omdat de toegang tot geschikte gegevens beperkt is en de aard van de taak snel ontwikkelt (zowel door lexicale variatie als culturele verschuivingen), is de effectiviteit van de classificatiemodellen vaak significant minder dan wordt voorgedaan.

Binnen het voorgestelde raamwerk onderzoeken we tevens aanvallen op taal; i.e., taalveranderingen die een gegeven model minder effectief maken. Deze aanvallen vormen de rode draad tussen onze gebruikersgerichte beveiligingsproblemen; ze zijn zeer relevant voor privacybeschermende vertroebelingsmethoden tegen auteurprofilering, en kunnen ook nuttig zijn om de invloed van lexicale variatie en augmentatie op de detectie van cyberpesten te testen. Hiermee beantwoorden we de volgende onderzoeksvragen:

\begin{itemize}
    \item[\textbf{RQ1}] In hoeverre kunnen end-to-end en gerichte modellen worden gebruikt voor  gebruikersgerichte, stylometrische aanvallen?
    \item[\textbf{RQ2}] Zijn classificatiemodellen voor cyberpesten robuust genoeg om op een gebruikersgerichte manier te worden ingezet?
\end{itemize}

Dit proefschrift levert hiermee de volgende bijdragen: Hoofdstuk 1 stelt een stijlneutrale framing binnen stylometrische aanvallen voor, Hoofdstuk 2 meet de overdraagbaarheid van dergelijke stylometrische aanvallen, Hoofdstuk 3 beoordeelt generalisatie van cyberpestclassificatiemodellen en Hoofdstuk 4 hun gevoeligheid voor lexicale variatie en het de mogelijkheid om relevante corpora te vergroten.
\selectlanguage{english}

\thispagestyle{empty}
\phantomsection
\fancyhead[RE]{\scshape references}
\fancyhead[LO]{\scshape references}
\addcontentsline{toc}{chapter}{\itshape References}
\footnotesize{
\bibliography{references,the-anthology}

\begin{thebibliography}{427}
\expandafter\ifx\csname natexlab\endcsname\relax\def\natexlab#1{#1}\fi

\bibitem[{Abadi et~al.(2015)Abadi, Agarwal, Barham, Brevdo, Chen, Citro,
  Corrado, Davis, Dean, Devin, Ghemawat, Goodfellow, Harp, Irving, Isard, Jia,
  Jozefowicz, Kaiser, Kudlur, Levenberg, Man\'{e}, Monga, Moore, Murray, Olah,
  Schuster, Shlens, Steiner, Sutskever, Talwar, Tucker, Vanhoucke, Vasudevan,
  Vi\'{e}gas, Vinyals, Warden, Wattenberg, Wicke, Yu, and
  Zheng}]{tensorflow2015whitepaper}
Mart\'{\i}n Abadi, Ashish Agarwal, Paul Barham, Eugene Brevdo, Zhifeng Chen,
  Craig Citro, Greg~S. Corrado, Andy Davis, Jeffrey Dean, Matthieu Devin,
  Sanjay Ghemawat, Ian Goodfellow, Andrew Harp, Geoffrey Irving, Michael Isard,
  Yangqing Jia, Rafal Jozefowicz, Lukasz Kaiser, Manjunath Kudlur, Josh
  Levenberg, Dan Man\'{e}, Rajat Monga, Sherry Moore, Derek Murray, Chris Olah,
  Mike Schuster, Jonathon Shlens, Benoit Steiner, Ilya Sutskever, Kunal Talwar,
  Paul Tucker, Vincent Vanhoucke, Vijay Vasudevan, Fernanda Vi\'{e}gas, Oriol
  Vinyals, Pete Warden, Martin Wattenberg, Martin Wicke, Yuan Yu, and Xiaoqiang
  Zheng. 2015.
\newblock \href {http://tensorflow.org/} {{TensorFlow}: Large-scale machine
  learning on heterogeneous systems}.
\newblock Software available from tensorflow.org.

\bibitem[{Abadi et~al.(2016{\natexlab{a}})Abadi, Barham, Chen, Chen, Davis,
  Dean, Devin, Ghemawat, Irving, Isard, Kudlur, Levenberg, Monga, Moore,
  Murray, Steiner, Tucker, Vasudevan, Warden, Wicke, Yu, and
  Zheng}]{DBLP:conf/osdi/AbadiBCCDDDGIIK16}
Mart{\'{\i}}n Abadi, Paul Barham, Jianmin Chen, Zhifeng Chen, Andy Davis,
  Jeffrey Dean, Matthieu Devin, Sanjay Ghemawat, Geoffrey Irving, Michael
  Isard, Manjunath Kudlur, Josh Levenberg, Rajat Monga, Sherry Moore,
  Derek~Gordon Murray, Benoit Steiner, Paul~A. Tucker, Vijay Vasudevan, Pete
  Warden, Martin Wicke, Yuan Yu, and Xiaoqiang Zheng. 2016{\natexlab{a}}.
\newblock \href
  {https://www.usenix.org/conference/osdi16/technical-sessions/presentation/abadi}
  {Tensorflow: {A} system for large-scale machine learning}.
\newblock In \emph{12th {USENIX} Symposium on Operating Systems Design and
  Implementation, {OSDI} 2016, Savannah, GA, USA, November 2-4, 2016}, pages
  265--283. {USENIX} Association.

\bibitem[{Abadi et~al.(2016{\natexlab{b}})Abadi, Chu, Goodfellow, McMahan,
  Mironov, Talwar, and Zhang}]{DBLP:conf/ccs/AbadiCGMMT016}
Mart{\'{\i}}n Abadi, Andy Chu, Ian~J. Goodfellow, H.~Brendan McMahan, Ilya
  Mironov, Kunal Talwar, and Li~Zhang. 2016{\natexlab{b}}.
\newblock \href {https://doi.org/10.1145/2976749.2978318} {Deep learning with
  differential privacy}.
\newblock In \emph{Proceedings of the 2016 {ACM} {SIGSAC} Conference on
  Computer and Communications Security, Vienna, Austria, October 24-28, 2016},
  pages 308--318. {ACM}.

\bibitem[{Abowd et~al.(1998)Abowd, Dey, Orr, and
  Brotherton}]{DBLP:journals/vr/AbowdDOB98}
Gregory~D. Abowd, Anind~K. Dey, Robert~J. Orr, and Jason~A. Brotherton. 1998.
\newblock \href {https://doi.org/10.1007/BF01408562} {Context-awareness in
  wearable and ubiquitous computing}.
\newblock \emph{Virtual Real.}, 3(3):200--211.

\bibitem[{Adams(2006)}]{adams2006classification}
Carlisle~M. Adams. 2006.
\newblock \href {https://papers.ssrn.com/sol3/papers.cfm?abstract_id=999672} {A
  classification for privacy techniques}.
\newblock \emph{University of Ottawa Law \& Technology Journal}, 1.

\bibitem[{Adragna et~al.(2020)Adragna, Creager, Madras, and
  Zemel}]{DBLP:journals/corr/abs-2011-06485}
Robert Adragna, Elliot Creager, David Madras, and Richard~S. Zemel. 2020.
\newblock \href {http://arxiv.org/abs/2011.06485} {Fairness and robustness in
  invariant learning: {A} case study in toxicity classification}.
\newblock \emph{CoRR}, abs/2011.06485.

\bibitem[{Agrawal and Awekar(2018)}]{DBLP:conf/ecir/AgrawalA18}
Sweta Agrawal and Amit Awekar. 2018.
\newblock \href {https://doi.org/10.1007/978-3-319-76941-7\_11} {Deep learning
  for detecting cyberbullying across multiple social media platforms}.
\newblock In \emph{Advances in Information Retrieval - 40th European Conference
  on {IR} Research, {ECIR} 2018, Grenoble, France, March 26-29, 2018,
  Proceedings}, volume 10772 of \emph{Lecture Notes in Computer Science}, pages
  141--153. Springer.

\bibitem[{Alowibdi et~al.(2013)Alowibdi, Buy, and
  Yu}]{DBLP:conf/icmla/AlowibdiBY13}
Jalal~S. Alowibdi, Ugo~A. Buy, and Philip~S. Yu. 2013.
\newblock \href {https://doi.org/10.1109/ICMLA.2013.74} {Empirical evaluation
  of profile characteristics for gender classification on twitter}.
\newblock In \emph{12th International Conference on Machine Learning and
  Applications, {ICMLA} 2013, Miami, FL, USA, December 4-7, 2013, Volume 1},
  pages 365--369. {IEEE}.

\bibitem[{Alvarez-Melis and
  Jaakkola(2017)}]{alvarez-melis-jaakkola-2017-causal}
David Alvarez-Melis and Tommi Jaakkola. 2017.
\newblock \href {https://doi.org/10.18653/v1/D17-1042} {A causal framework for
  explaining the predictions of black-box sequence-to-sequence models}.
\newblock In \emph{Proceedings of the 2017 Conference on Empirical Methods in
  Natural Language Processing}, pages 412--421, Copenhagen, Denmark.
  Association for Computational Linguistics.

\bibitem[{Anaby{-}Tavor et~al.(2020)Anaby{-}Tavor, Carmeli, Goldbraich, Kantor,
  Kour, Shlomov, Tepper, and Zwerdling}]{DBLP:conf/aaai/Anaby-TavorCGKK20}
Ateret Anaby{-}Tavor, Boaz Carmeli, Esther Goldbraich, Amir Kantor, George
  Kour, Segev Shlomov, Naama Tepper, and Naama Zwerdling. 2020.
\newblock \href {https://aaai.org/ojs/index.php/AAAI/article/view/6233} {Do not
  have enough data? {D}eep learning to the rescue!}
\newblock In \emph{The Thirty-Fourth {AAAI} Conference on Artificial
  Intelligence, {AAAI} 2020, The Thirty-Second Innovative Applications of
  Artificial Intelligence Conference, {IAAI} 2020, The Tenth {AAAI} Symposium
  on Educational Advances in Artificial Intelligence, {EAAI} 2020, New York,
  NY, USA, February 7-12, 2020}, pages 7383--7390. {AAAI} Press.

\bibitem[{Anthony et~al.(2007)Anthony, Henderson, and
  Kotz}]{DBLP:journals/pervasive/AnthonyHK07}
Denise~L. Anthony, Tristan Henderson, and David Kotz. 2007.
\newblock \href {https://doi.org/10.1109/MPRV.2007.83} {Privacy in
  location-aware computing environments}.
\newblock \emph{{IEEE} Pervasive Comput.}, 6(4):64--72.

\bibitem[{Argamon et~al.(2005)Argamon, Dhawle, Koppel, and
  Pennebaker}]{argamon2005lexical}
Shlomo Argamon, Sushant Dhawle, Moshe Koppel, and James~W Pennebaker. 2005.
\newblock Lexical predictors of personality type.
\newblock In \emph{Proceedings of the 2005 Joint Annual Meeting of the
  Interface Foundation and the Classification Society of North America, St.
  Louis, MO, USA, June, 2005}, pages 1--16.

\bibitem[{Arogyaswamy(2020)}]{DBLP:journals/ais/Arogyaswamy20}
Bernard Arogyaswamy. 2020.
\newblock \href {https://doi.org/10.1007/s00146-020-00956-6} {Big tech and
  societal sustainability: an ethical framework}.
\newblock \emph{{AI} Soc.}, 35(4):829--840.

\bibitem[{Atallah et~al.(2000)Atallah, McDonough, Raskin, and
  Nirenburg}]{DBLP:conf/nspw/AtallahMRN00}
Mikhail~J. Atallah, Craig~J. McDonough, Victor Raskin, and Sergei Nirenburg.
  2000.
\newblock \href {https://doi.org/10.1145/366173.366190} {Natural language
  processing for information assurance and security: an overview and
  implementations}.
\newblock In \emph{Proceedings of the 2000 Workshop on New Security Paradigms,
  Ballycotton, Co. Cork, Ireland, September 18-21, 2000}, pages 51--65. {ACM}.

\bibitem[{Ateniese et~al.(2015)Ateniese, Mancini, Spognardi, Villani, Vitali,
  and Felici}]{DBLP:journals/ijsn/AtenieseMSVVF15}
Giuseppe Ateniese, Luigi~V. Mancini, Angelo Spognardi, Antonio Villani,
  Domenico Vitali, and Giovanni Felici. 2015.
\newblock \href {https://doi.org/10.1504/IJSN.2015.071829} {Hacking smart
  machines with smarter ones: How to extract meaningful data from machine
  learning classifiers}.
\newblock \emph{Int. J. Secur. Networks}, 10(3):137--150.

\bibitem[{Baayen et~al.(1996)Baayen, van Halteren, and
  Tweedie}]{10.1093/llc/11.3.121}
H~Baayen, H~van Halteren, and F~Tweedie. 1996.
\newblock \href {https://doi.org/10.1093/llc/11.3.121} {{Outside the cave of
  shadows: using syntactic annotation to enhance authorship attribution}}.
\newblock \emph{Literary and Linguistic Computing}, 11(3):121--132.

\bibitem[{Baayen et~al.(2002)Baayen, van Halteren, Neijt, and
  Tweedie}]{baayen2002experiment}
Harald Baayen, Hans van Halteren, Anneke Neijt, and Fiona Tweedie. 2002.
\newblock An experiment in authorship attribution.
\newblock In \emph{Proceedings of JADT 2002: Sixth International Conference on
  Textual Data Statistical Analysis}, volume~1, pages 29--37.

\bibitem[{Badaskar et~al.(2008)Badaskar, Agarwal, and
  Arora}]{badaskar-etal-2008-identifying}
Sameer Badaskar, Sachin Agarwal, and Shilpa Arora. 2008.
\newblock \href {https://aclanthology.org/I08-2115} {Identifying real or fake
  articles: Towards better language modeling}.
\newblock In \emph{Proceedings of the Third International Joint Conference on
  Natural Language Processing: Volume-{II}}.

\bibitem[{Baldi et~al.(1999)Baldi, Brunak, Frasconi, Soda, and
  Pollastri}]{PMID:10743560}
P~Baldi, S~Brunak, P~Frasconi, G~Soda, and G~Pollastri. 1999.
\newblock \href {https://doi.org/10.1093/bioinformatics/15.11.937} {Exploiting
  the past and the future in protein secondary structure prediction}.
\newblock \emph{Bioinformatics (Oxford, England)}, 15(11):937—946.

\bibitem[{Bamman et~al.(2014{\natexlab{a}})Bamman, Dyer, and
  Smith}]{bamman-etal-2014-distributed}
David Bamman, Chris Dyer, and Noah~A. Smith. 2014{\natexlab{a}}.
\newblock \href {https://doi.org/10.3115/v1/P14-2134} {Distributed
  representations of geographically situated language}.
\newblock In \emph{Proceedings of the 52nd Annual Meeting of the Association
  for Computational Linguistics (Volume 2: Short Papers)}, pages 828--834,
  Baltimore, Maryland. Association for Computational Linguistics.

\bibitem[{Bamman et~al.(2014{\natexlab{b}})Bamman, Eisenstein, and
  Schnoebelen}]{https://doi.org/10.1111/josl.12080}
David Bamman, Jacob Eisenstein, and Tyler Schnoebelen. 2014{\natexlab{b}}.
\newblock \href {https://doi.org/https://doi.org/10.1111/josl.12080} {Gender
  identity and lexical variation in social media}.
\newblock \emph{Journal of Sociolinguistics}, 18(2):135--160.

\bibitem[{Banerjee et~al.(2014)Banerjee, Feng, Kang, and
  Choi}]{banerjee-etal-2014-keystroke}
Ritwik Banerjee, Song Feng, Jun~Seok Kang, and Yejin Choi. 2014.
\newblock \href {https://doi.org/10.3115/v1/D14-1155} {Keystroke patterns as
  prosody in digital writings: A case study with deceptive reviews and essays}.
\newblock In \emph{Proceedings of the 2014 Conference on Empirical Methods in
  Natural Language Processing ({EMNLP})}, pages 1469--1473, Doha, Qatar.
  Association for Computational Linguistics.

\bibitem[{Banerjee and Lavie(2005)}]{banerjee-lavie-2005-meteor}
Satanjeev Banerjee and Alon Lavie. 2005.
\newblock \href {https://aclanthology.org/W05-0909} {{METEOR}: An automatic
  metric for {MT} evaluation with improved correlation with human judgments}.
\newblock In \emph{Proceedings of the {ACL} Workshop on Intrinsic and Extrinsic
  Evaluation Measures for Machine Translation and/or Summarization}, pages
  65--72, Ann Arbor, Michigan. Association for Computational Linguistics.

\bibitem[{Banko et~al.(2020)Banko, MacKeen, and Ray}]{banko-etal-2020-unified}
Michele Banko, Brendon MacKeen, and Laurie Ray. 2020.
\newblock \href {https://doi.org/10.18653/v1/2020.alw-1.16} {A unified taxonomy
  of harmful content}.
\newblock In \emph{Proceedings of the Fourth Workshop on Online Abuse and
  Harms}, pages 125--137, Online. Association for Computational Linguistics.

\bibitem[{Barth and {de Jong}(2017)}]{BARTH20171038}
Susanne Barth and Menno~D.T. {de Jong}. 2017.
\newblock \href {https://doi.org/https://doi.org/10.1016/j.tele.2017.04.013}
  {The privacy paradox – investigating discrepancies between expressed
  privacy concerns and actual online behavior – a systematic literature
  review}.
\newblock \emph{Telematics and Informatics}, 34(7):1038--1058.

\bibitem[{Basile et~al.(2017)Basile, Dwyer, Medvedeva, Rawee, Haagsma, and
  Nissim}]{DBLP:conf/clef/BasileDMRHN17a}
Angelo Basile, Gareth Dwyer, Maria Medvedeva, Josine Rawee, Hessel Haagsma, and
  Malvina Nissim. 2017.
\newblock \href {https://doi.org/10.1007/978-3-319-98932-7\_14} {Simply the
  best: Minimalist system trumps complex models in author profiling}.
\newblock In \emph{Experimental {IR} Meets Multilinguality, Multimodality, and
  Interaction - 9th International Conference of the {CLEF} Association, {CLEF}
  2018, Avignon, France, September 10-14, 2018, Proceedings}, volume 11018 of
  \emph{Lecture Notes in Computer Science}, pages 143--156. Springer.

\bibitem[{Bayer et~al.(2021)Bayer, Kaufhold, and
  Reuter}]{DBLP:journals/corr/abs-2107-03158}
Markus Bayer, Marc{-}Andr{\'{e}} Kaufhold, and Christian Reuter. 2021.
\newblock \href {http://arxiv.org/abs/2107.03158} {A survey on data
  augmentation for text classification}.
\newblock \emph{CoRR}, abs/2107.03158.

\bibitem[{Bayzick et~al.(2011)Bayzick, Kontostathis, and
  Edwards}]{bayzick2011detecting}
Jennifer Bayzick, April Kontostathis, and Lynne Edwards. 2011.
\newblock \href {https://april-edwards.me/BayzickHonors.pdf} {Detecting the
  presence of cyberbullying using computer software}.
\newblock In \emph{Proceedings of the ACM WebSci Conference, Koblenz, Germany,
  2011}.

\bibitem[{Belinkov and Bisk(2018)}]{DBLP:conf/iclr/BelinkovB18}
Yonatan Belinkov and Yonatan Bisk. 2018.
\newblock \href {https://openreview.net/forum?id=BJ8vJebC-} {Synthetic and
  natural noise both break neural machine translation}.
\newblock In \emph{6th International Conference on Learning Representations,
  {ICLR} 2018, Vancouver, BC, Canada, April 30 - May 3, 2018, Conference Track
  Proceedings}. OpenReview.net.

\bibitem[{Beller et~al.(2014)Beller, Knowles, Harman, Bergsma, Mitchell, and
  Van~Durme}]{beller-etal-2014-im}
Charley Beller, Rebecca Knowles, Craig Harman, Shane Bergsma, Margaret
  Mitchell, and Benjamin Van~Durme. 2014.
\newblock \href {https://doi.org/10.3115/v1/P14-2030} {{I}{'}m a belieber:
  Social roles via self-identification and conceptual attributes}.
\newblock In \emph{Proceedings of the 52nd Annual Meeting of the Association
  for Computational Linguistics (Volume 2: Short Papers)}, pages 181--186,
  Baltimore, Maryland. Association for Computational Linguistics.

\bibitem[{Bender and Friedman(2018)}]{bender-friedman-2018-data}
Emily~M. Bender and Batya Friedman. 2018.
\newblock \href {https://doi.org/10.1162/tacl_a_00041} {Data statements for
  natural language processing: Toward mitigating system bias and enabling
  better science}.
\newblock \emph{Transactions of the Association for Computational Linguistics},
  6:587--604.

\bibitem[{Bender et~al.(2021)Bender, Gebru, McMillan{-}Major, and
  Shmitchell}]{DBLP:conf/fat/BenderGMS21}
Emily~M. Bender, Timnit Gebru, Angelina McMillan{-}Major, and Shmargaret
  Shmitchell. 2021.
\newblock \href {https://doi.org/10.1145/3442188.3445922} {On the dangers of
  stochastic parrots: Can language models be too big?}
\newblock In \emph{FAccT '21: 2021 {ACM} Conference on Fairness,
  Accountability, and Transparency, Virtual Event / Toronto, Canada, March
  3-10, 2021}, pages 610--623. {ACM}.

\bibitem[{Ben{\'{\i}}tez et~al.(1997)Ben{\'{\i}}tez, Castro, and
  Requena}]{DBLP:journals/tnn/BenitezCR97}
Jos{\'{e}}~Manuel Ben{\'{\i}}tez, Juan~Luis Castro, and Ignacio Requena. 1997.
\newblock \href {https://doi.org/10.1109/72.623216} {Are artificial neural
  networks black boxes?}
\newblock \emph{{IEEE} Trans. Neural Networks}, 8(5):1156--1164.

\bibitem[{Beran and Li(2008)}]{Beran2008}
Tanya Beran and Qing Li. 2008.
\newblock \href {https://doi.org/https://doi.org/10.21913/JSW.v1i2.172} {The
  relationship between cyberbullying and school bullying}.
\newblock \emph{The Journal of Student Wellbeing}, 1(2):16--33.

\bibitem[{Bessière et~al.(2008)Bessière, Kiesler, Kraut, and
  Boneva}]{doi:10.1080/13691180701858851}
Katherine Bessière, Sara Kiesler, Robert Kraut, and Bonka~S. Boneva. 2008.
\newblock \href {https://doi.org/10.1080/13691180701858851} {Effects of
  internet use and social resources on changes in depression}.
\newblock \emph{Information, Communication \& Society}, 11(1):47--70.

\bibitem[{Bevendorff et~al.(2020)Bevendorff, Wenzel, Potthast, Hagen, and
  Stein}]{DBLP:journals/it/BevendorffWPH020}
Janek Bevendorff, Tobias Wenzel, Martin Potthast, Matthias Hagen, and Benno
  Stein. 2020.
\newblock \href {https://doi.org/10.1515/itit-2019-0046} {On divergence-based
  author obfuscation: An attack on the state of the art in statistical
  authorship verification}.
\newblock \emph{it Inf. Technol.}, 62(2):99--115.

\bibitem[{Biber(1995)}]{biber-1995-dimensions}
Douglas Biber. 1995.
\newblock \href {https://doi.org/10.1017/CBO9780511519871} {\emph{Dimensions of
  Register Variation: A Cross-Linguistic Comparison}}.
\newblock Cambridge University Press.

\bibitem[{Binns(2013)}]{uclan8378}
Amy Binns. 2013.
\newblock \href {http://clok.uclan.ac.uk/8378/} {Facebook's ugly sisters:
  Anonymity and abuse on formspring and ask.fm}.
\newblock \emph{Media Education Research Journal}.

\bibitem[{Birhane et~al.(2021)Birhane, Prabhu, and
  Kahembwe}]{DBLP:journals/corr/abs-2110-01963}
Abeba Birhane, Vinay~Uday Prabhu, and Emmanuel Kahembwe. 2021.
\newblock \href {http://arxiv.org/abs/2110.01963} {Multimodal datasets:
  misogyny, pornography, and malignant stereotypes}.
\newblock \emph{CoRR}, abs/2110.01963.

\bibitem[{Blitzer et~al.(2007)Blitzer, Dredze, and
  Pereira}]{blitzer-etal-2007-biographies}
John Blitzer, Mark Dredze, and Fernando Pereira. 2007.
\newblock \href {https://aclanthology.org/P07-1056} {Biographies, {B}ollywood,
  boom-boxes and blenders: Domain adaptation for sentiment classification}.
\newblock In \emph{Proceedings of the 45th Annual Meeting of the Association of
  Computational Linguistics}, pages 440--447, Prague, Czech Republic.
  Association for Computational Linguistics.

\bibitem[{Bo et~al.(2021)Bo, Ding, Fung, and Iqbal}]{bo-etal-2021-er}
Haohan Bo, Steven H.~H. Ding, Benjamin C.~M. Fung, and Farkhund Iqbal. 2021.
\newblock \href {https://doi.org/10.18653/v1/2021.naacl-main.314} {{ER}-{AE}:
  Differentially private text generation for authorship anonymization}.
\newblock In \emph{Proceedings of the 2021 Conference of the North American
  Chapter of the Association for Computational Linguistics: Human Language
  Technologies}, pages 3997--4007, Online. Association for Computational
  Linguistics.

\bibitem[{Bogdanova et~al.(2014)Bogdanova, Rosso, and
  Solorio}]{DBLP:journals/csl/BogdanovaRS14}
Dasha Bogdanova, Paolo Rosso, and Thamar Solorio. 2014.
\newblock \href {https://doi.org/10.1016/j.csl.2013.04.007} {Exploring
  high-level features for detecting cyberpedophilia}.
\newblock \emph{Comput. Speech Lang.}, 28(1):108--120.

\bibitem[{Bojanowski et~al.(2016)Bojanowski, Grave, Joulin, and
  Mikolov}]{bojanowski2016enriching}
Piotr Bojanowski, Edouard Grave, Armand Joulin, and Tomas Mikolov. 2016.
\newblock Enriching word vectors with subword information.
\newblock \emph{arXiv preprint arXiv:1607.04606}.

\bibitem[{Bojanowski et~al.(2017)Bojanowski, Grave, Joulin, and
  Mikolov}]{bojanowski-etal-2017-enriching}
Piotr Bojanowski, Edouard Grave, Armand Joulin, and Tomas Mikolov. 2017.
\newblock \href {https://doi.org/10.1162/tacl_a_00051} {Enriching word vectors
  with subword information}.
\newblock \emph{Transactions of the Association for Computational Linguistics},
  5:135--146.

\bibitem[{Brainerd(1974)}]{10.3138/j.ctt15jjcf0}
Barron Brainerd. 1974.
\newblock \href {http://www.jstor.org/stable/10.3138/j.ctt15jjcf0}
  {\emph{Weighting Evidence in Language and Literature}}.
\newblock University of Toronto Press.

\bibitem[{Brashier and Schacter(2020)}]{doi:10.1177/0963721420915872}
Nadia~M. Brashier and Daniel~L. Schacter. 2020.
\newblock \href {https://doi.org/10.1177/0963721420915872} {Aging in an era of
  fake news}.
\newblock \emph{Current Directions in Psychological Science}, 29(3):316--323.
\newblock PMID: 32968336.

\bibitem[{Brassard-Gourdeau and
  Khoury(2019)}]{brassard-gourdeau-khoury-2019-subversive}
Eloi Brassard-Gourdeau and Richard Khoury. 2019.
\newblock \href {https://doi.org/10.18653/v1/W19-3501} {Subversive toxicity
  detection using sentiment information}.
\newblock In \emph{Proceedings of the Third Workshop on Abusive Language
  Online}, pages 1--10, Florence, Italy. Association for Computational
  Linguistics.

\bibitem[{Brennan et~al.(2012)Brennan, Afroz, and
  Greenstadt}]{DBLP:journals/tissec/BrennanAG12}
Michael Brennan, Sadia Afroz, and Rachel Greenstadt. 2012.
\newblock \href {https://doi.org/10.1145/2382448.2382450} {Adversarial
  stylometry: Circumventing authorship recognition to preserve privacy and
  anonymity}.
\newblock \emph{{ACM} Trans. Inf. Syst. Secur.}, 15(3):12:1--12:22.

\bibitem[{Bretschneider et~al.(2014)Bretschneider, W{\"{o}}hner, and
  Peters}]{DBLP:conf/icis/BretschneiderWP14}
Uwe Bretschneider, Thomas W{\"{o}}hner, and Ralf Peters. 2014.
\newblock \href
  {http://aisel.aisnet.org/icis2014/proceedings/ConferenceTheme/2} {Detecting
  online harassment in social networks}.
\newblock In \emph{Proceedings of the International Conference on Information
  Systems - Building a Better World through Information Systems, {ICIS} 2014,
  Auckland, New Zealand, December 14-17, 2014}. Association for Information
  Systems.

\bibitem[{Brinegar(1963)}]{10.2307/2282956}
Claude~S. Brinegar. 1963.
\newblock Mark twain and the quintus curtius snodgrass letters: A statistical
  test of authorship.
\newblock \emph{Journal of the American Statistical Association},
  58(301):85--96.

\bibitem[{{Broadcasting Standards Authority}(2013)}]{broadc}
{Broadcasting Standards Authority}. 2013.
\newblock {What not to swear: The acceptability of words in broadcasting}.
\newblock
  \url{http://bsa.govt.nz/images/assets/Research/Acceptability_of_Words_2013_WEB.pdf}.
\newblock Last accessed 15/03/2016.

\bibitem[{Brown et~al.(2020)Brown, Mann, Ryder, Subbiah, Kaplan, Dhariwal,
  Neelakantan, Shyam, Sastry, Askell, Agarwal, Herbert{-}Voss, Krueger,
  Henighan, Child, Ramesh, Ziegler, Wu, Winter, Hesse, Chen, Sigler, Litwin,
  Gray, Chess, Clark, Berner, McCandlish, Radford, Sutskever, and
  Amodei}]{DBLP:conf/nips/BrownMRSKDNSSAA20}
Tom~B. Brown, Benjamin Mann, Nick Ryder, Melanie Subbiah, Jared Kaplan,
  Prafulla Dhariwal, Arvind Neelakantan, Pranav Shyam, Girish Sastry, Amanda
  Askell, Sandhini Agarwal, Ariel Herbert{-}Voss, Gretchen Krueger, Tom
  Henighan, Rewon Child, Aditya Ramesh, Daniel~M. Ziegler, Jeffrey Wu, Clemens
  Winter, Christopher Hesse, Mark Chen, Eric Sigler, Mateusz Litwin, Scott
  Gray, Benjamin Chess, Jack Clark, Christopher Berner, Sam McCandlish, Alec
  Radford, Ilya Sutskever, and Dario Amodei. 2020.
\newblock \href
  {https://proceedings.neurips.cc/paper/2020/hash/1457c0d6bfcb4967418bfb8ac142f64a-Abstract.html}
  {Language models are few-shot learners}.
\newblock In \emph{Advances in Neural Information Processing Systems 33: Annual
  Conference on Neural Information Processing Systems 2020, NeurIPS 2020,
  December 6-12, 2020, virtual}.

\bibitem[{Burger et~al.(2011)Burger, Henderson, Kim, and
  Zarrella}]{burger-etal-2011-discriminating}
John~D. Burger, John Henderson, George Kim, and Guido Zarrella. 2011.
\newblock \href {https://aclanthology.org/D11-1120} {Discriminating gender on
  {T}witter}.
\newblock In \emph{Proceedings of the 2011 Conference on Empirical Methods in
  Natural Language Processing}, pages 1301--1309, Edinburgh, Scotland, UK.
  Association for Computational Linguistics.

\bibitem[{Burtenshaw and Kestemont(2021)}]{burtenshaw-kestemont-2021-dutch}
Ben Burtenshaw and Mike Kestemont. 2021.
\newblock \href {https://aclanthology.org/2021.bucc-1.10} {A {D}utch dataset
  for cross-lingual multilabel toxicity detection}.
\newblock In \emph{Proceedings of the 14th Workshop on Building and Using
  Comparable Corpora (BUCC 2021)}, pages 75--79, Online (Virtual Mode). INCOMA
  Ltd.

\bibitem[{Cai and Guo(2019)}]{DBLP:conf/vl/CaiG19}
Carrie~J. Cai and Philip~J. Guo. 2019.
\newblock \href {https://doi.org/10.1109/VLHCC.2019.8818751} {Software
  developers learning machine learning: Motivations, hurdles, and desires}.
\newblock In \emph{2019 {IEEE} Symposium on Visual Languages and Human-Centric
  Computing, {VL/HCC} 2019, Memphis, Tennessee, USA, October 14-18, 2019},
  pages 25--34. {IEEE} Computer Society.

\bibitem[{Caliskan and Greenstadt(2012)}]{DBLP:conf/semco/CaliskanG12}
Aylin Caliskan and Rachel Greenstadt. 2012.
\newblock \href {https://doi.org/10.1109/ICSC.2012.46} {Translate once,
  translate twice, translate thrice and attribute: Identifying authors and
  machine translation tools in translated text}.
\newblock In \emph{Sixth {IEEE} International Conference on Semantic Computing,
  {ICSC} 2012, Palermo, Italy, September 19-21, 2012}, pages 121--125. {IEEE}
  Computer Society.

\bibitem[{Caliskan et~al.(2018)Caliskan, Yamaguchi, Dauber, Harang, Rieck,
  Greenstadt, and Narayanan}]{caliskan2015coding}
Aylin Caliskan, Fabian Yamaguchi, Edwin Dauber, Richard~E. Harang, Konrad
  Rieck, Rachel Greenstadt, and Arvind Narayanan. 2018.
\newblock When coding style survives compilation: De-anonymizing programmers
  from executable binaries.
\newblock In \emph{25th Annual Network and Distributed System Security
  Symposium, {NDSS} 2018, San Diego, California, USA, February 18-21, 2018}.
  The Internet Society.

\bibitem[{Canfora et~al.(2018)Canfora, Sorbo, Emanuele, Forootani, and
  Visaggio}]{DBLP:conf/IEEEares/CanforaSEFV18}
Gerardo Canfora, Andrea~Di Sorbo, Enrico Emanuele, Sara Forootani, and
  Corrado~Aaron Visaggio. 2018.
\newblock \href {https://doi.org/10.1145/3230833.3230845} {A nlp-based solution
  to prevent from privacy leaks in social network posts}.
\newblock In \emph{Proceedings of the 13th International Conference on
  Availability, Reliability and Security, {ARES} 2018, Hamburg, Germany, August
  27-30, 2018}, pages 36:1--36:6. {ACM}.

\bibitem[{Carlini et~al.(2021)Carlini, Tram{\`{e}}r, Wallace, Jagielski,
  Herbert{-}Voss, Lee, Roberts, Brown, Song, Erlingsson, Oprea, and
  Raffel}]{DBLP:journals/corr/abs-2012-07805}
Nicholas Carlini, Florian Tram{\`{e}}r, Eric Wallace, Matthew Jagielski, Ariel
  Herbert{-}Voss, Katherine Lee, Adam Roberts, Tom~B. Brown, Dawn Song,
  {\'{U}}lfar Erlingsson, Alina Oprea, and Colin Raffel. 2021.
\newblock \href
  {https://www.usenix.org/conference/usenixsecurity21/presentation/carlini-extracting}
  {Extracting training data from large language models}.
\newblock In \emph{30th {USENIX} Security Symposium, {USENIX} Security 2021,
  August 11-13, 2021}, pages 2633--2650. {USENIX} Association.

\bibitem[{Carlson et~al.(2017)Carlson, Riddell, and
  Rockmore}]{DBLP:journals/corr/abs-1711-04731}
Keith Carlson, Allen Riddell, and Daniel~N. Rockmore. 2017.
\newblock \href {http://arxiv.org/abs/1711.04731} {Zero-shot style transfer in
  text using recurrent neural networks}.
\newblock \emph{CoRR}, abs/1711.04731.

\bibitem[{Carton et~al.(2020)Carton, Mei, and
  Resnick}]{DBLP:conf/icwsm/CartonMR20}
Samuel Carton, Qiaozhu Mei, and Paul Resnick. 2020.
\newblock \href {https://aaai.org/ojs/index.php/ICWSM/article/view/7282}
  {Feature-based explanations don't help people detect misclassifications of
  online toxicity}.
\newblock In \emph{Proceedings of the Fourteenth International {AAAI}
  Conference on Web and Social Media, {ICWSM} 2020, Held Virtually, Original
  Venue: Atlanta, Georgia, USA, June 8-11, 2020}, pages 95--106. {AAAI} Press.

\bibitem[{Caselli et~al.(2020)Caselli, Basile, Mitrovic, and
  Granitzer}]{DBLP:journals/corr/abs-2010-12472}
Tommaso Caselli, Valerio Basile, Jelena Mitrovic, and Michael Granitzer. 2020.
\newblock \href {http://arxiv.org/abs/2010.12472} {Hatebert: Retraining {BERT}
  for abusive language detection in english}.
\newblock \emph{CoRR}, abs/2010.12472.

\bibitem[{Caselli et~al.(2021)Caselli, Basile, Mitrovi{\'c}, and
  Granitzer}]{caselli-etal-2021-hatebert}
Tommaso Caselli, Valerio Basile, Jelena Mitrovi{\'c}, and Michael Granitzer.
  2021.
\newblock \href {https://doi.org/10.18653/v1/2021.woah-1.3} {{H}ate{BERT}:
  Retraining {BERT} for abusive language detection in {E}nglish}.
\newblock In \emph{Proceedings of the 5th Workshop on Online Abuse and Harms
  (WOAH 2021)}, pages 17--25, Online. Association for Computational
  Linguistics.

\bibitem[{Caswell et~al.(2019)Caswell, Chelba, and
  Grangier}]{caswell-etal-2019-tagged}
Isaac Caswell, Ciprian Chelba, and David Grangier. 2019.
\newblock \href {https://doi.org/10.18653/v1/W19-5206} {Tagged
  back-translation}.
\newblock In \emph{Proceedings of the Fourth Conference on Machine Translation
  (Volume 1: Research Papers)}, pages 53--63, Florence, Italy. Association for
  Computational Linguistics.

\bibitem[{Cer et~al.(2018)Cer, Yang, Kong, Hua, Limtiaco, St.~John, Constant,
  Guajardo-Cespedes, Yuan, Tar, Strope, and Kurzweil}]{cer-etal-2018-universal}
Daniel Cer, Yinfei Yang, Sheng-yi Kong, Nan Hua, Nicole Limtiaco, Rhomni
  St.~John, Noah Constant, Mario Guajardo-Cespedes, Steve Yuan, Chris Tar,
  Brian Strope, and Ray Kurzweil. 2018.
\newblock \href {https://doi.org/10.18653/v1/D18-2029} {Universal sentence
  encoder for {E}nglish}.
\newblock In \emph{Proceedings of the 2018 Conference on Empirical Methods in
  Natural Language Processing: System Demonstrations}, pages 169--174,
  Brussels, Belgium. Association for Computational Linguistics.

\bibitem[{Chan and Ng(2006)}]{chan-ng-2006-estimating}
Yee~Seng Chan and Hwee~Tou Ng. 2006.
\newblock \href {https://doi.org/10.3115/1220175.1220187} {Estimating class
  priors in domain adaptation for word sense disambiguation}.
\newblock In \emph{Proceedings of the 21st International Conference on
  Computational Linguistics and 44th Annual Meeting of the Association for
  Computational Linguistics}, pages 89--96, Sydney, Australia. Association for
  Computational Linguistics.

\bibitem[{Chandar et~al.(2014)Chandar, Lauly, Larochelle, Khapra, Ravindran,
  Raykar, and Saha}]{DBLP:conf/nips/PLLKRRS14}
A.~P.~Sarath Chandar, Stanislas Lauly, Hugo Larochelle, Mitesh~M. Khapra,
  Balaraman Ravindran, Vikas~C. Raykar, and Amrita Saha. 2014.
\newblock \href
  {https://proceedings.neurips.cc/paper/2014/hash/2bcab9d935d219641434683dd9d18a03-Abstract.html}
  {An autoencoder approach to learning bilingual word representations}.
\newblock In \emph{Advances in Neural Information Processing Systems 27: Annual
  Conference on Neural Information Processing Systems 2014, December 8-13 2014,
  Montreal, Quebec, Canada}, pages 1853--1861.

\bibitem[{Chapman et~al.(2011)Chapman, Nadkarni, Hirschman, D'Avolio, Savova,
  and Uzuner}]{10.1136/amiajnl-2011-000465}
Wendy~W Chapman, Prakash~M Nadkarni, Lynette Hirschman, Leonard~W D'Avolio,
  Guergana~K Savova, and Ozlem Uzuner. 2011.
\newblock \href {https://doi.org/10.1136/amiajnl-2011-000465} {{Overcoming
  barriers to NLP for clinical text: the role of shared tasks and the need for
  additional creative solutions}}.
\newblock \emph{Journal of the American Medical Informatics Association},
  18(5):540--543.

\bibitem[{Chauhan et~al.(2021)Chauhan, Bhukar, and
  Kaul}]{DBLP:journals/corr/abs-2106-07047}
Jatin Chauhan, Karan Bhukar, and Manohar Kaul. 2021.
\newblock \href {http://arxiv.org/abs/2106.07047} {Target model agnostic
  adversarial attacks with query budgets on language understanding models}.
\newblock \emph{CoRR}, abs/2106.07047.

\bibitem[{Chen and Dolan(2011)}]{chen-dolan-2011-collecting}
David Chen and William Dolan. 2011.
\newblock \href {https://aclanthology.org/P11-1020} {Collecting highly parallel
  data for paraphrase evaluation}.
\newblock In \emph{Proceedings of the 49th Annual Meeting of the Association
  for Computational Linguistics: Human Language Technologies}, pages 190--200,
  Portland, Oregon, USA. Association for Computational Linguistics.

\bibitem[{Chen et~al.(2012)Chen, Xu, Weinberger, and
  Sha}]{DBLP:conf/icml/ChenXWS12}
Minmin Chen, Zhixiang~Eddie Xu, Kilian~Q. Weinberger, and Fei Sha. 2012.
\newblock \href {http://icml.cc/2012/papers/416.pdf} {Marginalized denoising
  autoencoders for domain adaptation}.
\newblock In \emph{Proceedings of the 29th International Conference on Machine
  Learning, {ICML} 2012, Edinburgh, Scotland, UK, June 26 - July 1, 2012}.
  icml.cc / Omnipress.

\bibitem[{Chen et~al.(2017)Chen, Liao, Chuang, Hsu, Fu, and Sun}]{chen2017show}
Tseng-Hung Chen, Yuan-Hong Liao, Ching-Yao Chuang, Wan-Ting Hsu, Jianlong Fu,
  and Min Sun. 2017.
\newblock Show, adapt and tell: Adversarial training of cross-domain image
  captioner.
\newblock In \emph{The IEEE International Conference on Computer Vision
  (ICCV)}, volume~2.

\bibitem[{Cheng et~al.(2020{\natexlab{a}})Cheng, Guo, Candan, and
  Liu}]{DBLP:conf/sdm/ChengGCL20}
Lu~Cheng, Ruocheng Guo, K.~Sel{\c{c}}uk Candan, and Huan Liu.
  2020{\natexlab{a}}.
\newblock \href {https://doi.org/10.1137/1.9781611976236.54} {Representation
  learning for imbalanced cross-domain classification}.
\newblock In \emph{Proceedings of the 2020 {SIAM} International Conference on
  Data Mining, {SDM} 2020, Cincinnati, Ohio, USA, May 7-9, 2020}, pages
  478--486. {SIAM}.

\bibitem[{Cheng et~al.(2019)Cheng, Guo, Silva, Hall, and
  Liu}]{DBLP:conf/sdm/ChengGSHL19}
Lu~Cheng, Ruocheng Guo, Yasin~N. Silva, Deborah~L. Hall, and Huan Liu. 2019.
\newblock \href {https://doi.org/10.1137/1.9781611975673.27} {Hierarchical
  attention networks for cyberbullying detection on the instagram social
  network}.
\newblock In \emph{Proceedings of the 2019 {SIAM} International Conference on
  Data Mining, {SDM} 2019, Calgary, Alberta, Canada, May 2-4, 2019}, pages
  235--243. {SIAM}.

\bibitem[{Cheng et~al.(2020{\natexlab{b}})Cheng, Yi, Chen, Zhang, and
  Hsieh}]{DBLP:conf/aaai/ChengYCZH20}
Minhao Cheng, Jinfeng Yi, Pin{-}Yu Chen, Huan Zhang, and Cho{-}Jui Hsieh.
  2020{\natexlab{b}}.
\newblock \href {https://ojs.aaai.org/index.php/AAAI/article/view/5767}
  {Seq2sick: Evaluating the robustness of sequence-to-sequence models with
  adversarial examples}.
\newblock In \emph{The Thirty-Fourth {AAAI} Conference on Artificial
  Intelligence, {AAAI} 2020, The Thirty-Second Innovative Applications of
  Artificial Intelligence Conference, {IAAI} 2020, The Tenth {AAAI} Symposium
  on Educational Advances in Artificial Intelligence, {EAAI} 2020, New York,
  NY, USA, February 7-12, 2020}, pages 3601--3608. {AAAI} Press.

\bibitem[{Chollet et~al.(2015)}]{chollet2015keras}
Fran\c{c}ois Chollet et~al. 2015.
\newblock Keras.
\newblock \url{https://keras.io}.

\bibitem[{Choudhury et~al.(2013)Choudhury, Gamon, Counts, and
  Horvitz}]{DBLP:conf/icwsm/ChoudhuryGCH13}
Munmun~De Choudhury, Michael Gamon, Scott Counts, and Eric Horvitz. 2013.
\newblock \href
  {http://www.aaai.org/ocs/index.php/ICWSM/ICWSM13/paper/view/6124} {Predicting
  depression via social media}.
\newblock In \emph{Proceedings of the Seventh International Conference on
  Weblogs and Social Media, {ICWSM} 2013, Cambridge, Massachusetts, USA, July
  8-11, 2013}. The {AAAI} Press.

\bibitem[{Collobert et~al.(2011)Collobert, Weston, Bottou, Karlen, Kavukcuoglu,
  and Kuksa}]{DBLP:journals/jmlr/CollobertWBKKK11}
Ronan Collobert, Jason Weston, L{\'{e}}on Bottou, Michael Karlen, Koray
  Kavukcuoglu, and Pavel~P. Kuksa. 2011.
\newblock \href {https://dl.acm.org/doi/10.5555/1953048.2078186} {Natural
  language processing (almost) from scratch}.
\newblock \emph{J. Mach. Learn. Res.}, 12:2493--2537.

\bibitem[{Constantiou and
  Kallinikos(2015)}]{DBLP:journals/jitech/ConstantiouK15}
Ioanna~D. Constantiou and Jannis Kallinikos. 2015.
\newblock \href {https://doi.org/10.1057/jit.2014.17} {New games, new rules:
  big data and the changing context of strategy}.
\newblock \emph{J. Inf. Technol.}, 30(1):44--57.

\bibitem[{Coppersmith et~al.(2015)Coppersmith, Dredze, Harman, and
  Hollingshead}]{coppersmith2015}
Glen Coppersmith, Mark Dredze, Craig Harman, and Kristy Hollingshead. 2015.
\newblock From adhd to sad: Analyzing the language of mental health on twitter
  through self-reported diagnoses.
\newblock \emph{NAACL HLT 2015}, page~1.

\bibitem[{Cortes and Vapnik(1995)}]{DBLP:journals/ml/CortesV95}
Corinna Cortes and Vladimir Vapnik. 1995.
\newblock \href {https://doi.org/10.1007/BF00994018} {Support-vector networks}.
\newblock \emph{Mach. Learn.}, 20(3):273--297.

\bibitem[{Dadvar(2014)}]{Dadvar2014}
Maral Dadvar. 2014.
\newblock \href {https://doi.org/10.3990/1.9789036537391} {\emph{Experts and
  Machines United Against Cyberbullying}}.
\newblock Ph.D. thesis, University of Twente, Netherlands.

\bibitem[{Dadvar et~al.(2014)Dadvar, Trieschnigg, and
  de~Jong}]{DBLP:conf/ai/DadvarTJ14}
Maral Dadvar, Dolf Trieschnigg, and Franciska de~Jong. 2014.
\newblock \href {https://doi.org/10.1007/978-3-319-06483-3\_25} {Experts and
  machines against bullies: {A} hybrid approach to detect cyberbullies}.
\newblock In \emph{Advances in Artificial Intelligence - 27th Canadian
  Conference on Artificial Intelligence, Canadian {AI} 2014, Montr{\'{e}}al,
  QC, Canada, May 6-9, 2014. Proceedings}, volume 8436 of \emph{Lecture Notes
  in Computer Science}, pages 275--281. Springer.

\bibitem[{Daelemans(2013)}]{DBLP:conf/cicling/Daelemans13}
Walter Daelemans. 2013.
\newblock \href {https://doi.org/10.1007/978-3-642-37256-8\_37} {Explanation in
  computational stylometry}.
\newblock In \emph{Computational Linguistics and Intelligent Text Processing -
  14th International Conference, CICLing 2013, Samos, Greece, March 24-30,
  2013, Proceedings, Part {II}}, volume 7817 of \emph{Lecture Notes in Computer
  Science}, pages 451--462. Springer.

\bibitem[{Daum{\'e}~III(2007)}]{daume-iii-2007-frustratingly}
Hal Daum{\'e}~III. 2007.
\newblock \href {https://aclanthology.org/P07-1033} {Frustratingly easy domain
  adaptation}.
\newblock In \emph{Proceedings of the 45th Annual Meeting of the Association of
  Computational Linguistics}, pages 256--263, Prague, Czech Republic.
  Association for Computational Linguistics.

\bibitem[{Daum{\'e}~III et~al.(2010)Daum{\'e}~III, Kumar, and
  Saha}]{daume-iii-etal-2010-frustratingly}
Hal Daum{\'e}~III, Abhishek Kumar, and Avishek Saha. 2010.
\newblock \href {https://aclanthology.org/W10-2608} {Frustratingly easy
  semi-supervised domain adaptation}.
\newblock In \emph{Proceedings of the 2010 Workshop on Domain Adaptation for
  Natural Language Processing}, pages 53--59, Uppsala, Sweden. Association for
  Computational Linguistics.

\bibitem[{Dechand et~al.(2019)Dechand, Naiakshina, Danilova, and
  Smith}]{DBLP:conf/eurosp/DechandND019}
Sergej Dechand, Alena Naiakshina, Anastasia Danilova, and Matthew Smith. 2019.
\newblock \href {https://doi.org/10.1109/EuroSP.2019.00037} {In encryption we
  don't trust: The effect of end-to-end encryption to the masses on user
  perception}.
\newblock In \emph{{IEEE} European Symposium on Security and Privacy,
  EuroS{\&}P 2019, Stockholm, Sweden, June 17-19, 2019}, pages 401--415.
  {IEEE}.

\bibitem[{Dehue et~al.(2008)Dehue, Bolman, and V{\"o}llink}]{Dehue2008-rn}
Francine Dehue, Catherine Bolman, and Trijntje V{\"o}llink. 2008.
\newblock \href {https://doi.org/10.1089/cpb.2007.0008} {Cyberbullying:
  youngsters' experiences and parental perception}.
\newblock \emph{Cyberpsychol Behav}, 11(2):217--223.

\bibitem[{Denkowski and Lavie(2011)}]{denkowski-lavie-2011-meteor}
Michael Denkowski and Alon Lavie. 2011.
\newblock \href {https://aclanthology.org/W11-2107} {Meteor 1.3: Automatic
  metric for reliable optimization and evaluation of machine translation
  systems}.
\newblock In \emph{Proceedings of the Sixth Workshop on Statistical Machine
  Translation}, pages 85--91, Edinburgh, Scotland. Association for
  Computational Linguistics.

\bibitem[{Devlin et~al.(2019{\natexlab{a}})Devlin, Chang, Lee, and
  Toutanova}]{devlin-etal-2019-bert}
Jacob Devlin, Ming-Wei Chang, Kenton Lee, and Kristina Toutanova.
  2019{\natexlab{a}}.
\newblock \href {https://doi.org/10.18653/v1/N19-1423} {{BERT}: Pre-training of
  deep bidirectional transformers for language understanding}.
\newblock In \emph{Proceedings of the 2019 Conference of the North {A}merican
  Chapter of the Association for Computational Linguistics: Human Language
  Technologies, Volume 1 (Long and Short Papers)}, pages 4171--4186,
  Minneapolis, Minnesota. Association for Computational Linguistics.

\bibitem[{Devlin et~al.(2019{\natexlab{b}})Devlin, Chang, Lee, and
  Toutanova}]{DBLP:conf/naacl/DevlinCLT19}
Jacob Devlin, Ming{-}Wei Chang, Kenton Lee, and Kristina Toutanova.
  2019{\natexlab{b}}.
\newblock \href {https://doi.org/10.18653/v1/n19-1423} {{BERT:} pre-training of
  deep bidirectional transformers for language understanding}.
\newblock In \emph{Proceedings of the 2019 Conference of the North American
  Chapter of the Association for Computational Linguistics: Human Language
  Technologies, {NAACL-HLT} 2019, Minneapolis, MN, USA, June 2-7, 2019, Volume
  1 (Long and Short Papers)}, pages 4171--4186. Association for Computational
  Linguistics.

\bibitem[{Dinakar et~al.(2011)Dinakar, Reichart, and
  Lieberman}]{DBLP:conf/icwsm/DinakarRL11}
Karthik Dinakar, Roi Reichart, and Henry Lieberman. 2011.
\newblock \href
  {http://www.aaai.org/ocs/index.php/ICWSM/ICWSM11/paper/view/3841} {Modeling
  the detection of textual cyberbullying}.
\newblock In \emph{The Social Mobile Web, Papers from the 2011 {ICWSM}
  Workshop, Barcelona, Catalonia, Spain, July 21, 2011}, volume {WS-11-02} of
  \emph{{AAAI} Technical Report}. {AAAI}.

\bibitem[{Dong et~al.(2021)Dong, Luu, Ji, and Liu}]{DBLP:conf/iclr/DongLJ021}
Xinshuai Dong, Anh~Tuan Luu, Rongrong Ji, and Hong Liu. 2021.
\newblock \href {https://openreview.net/forum?id=ks5nebunVn\_} {Towards
  robustness against natural language word substitutions}.
\newblock In \emph{9th International Conference on Learning Representations,
  {ICLR} 2021, Virtual Event, Austria, May 3-7, 2021}. OpenReview.net.

\bibitem[{Dong et~al.(2014)Dong, Yang, Tang, Yang, and
  Chawla}]{DBLP:conf/kdd/DongYTYC14}
Yuxiao Dong, Yang Yang, Jie Tang, Yang Yang, and Nitesh~V. Chawla. 2014.
\newblock \href {https://doi.org/10.1145/2623330.2623703} {Inferring user
  demographics and social strategies in mobile social networks}.
\newblock In \emph{The 20th {ACM} {SIGKDD} International Conference on
  Knowledge Discovery and Data Mining, {KDD} '14, New York, NY, {USA} - August
  24 - 27, 2014}, pages 15--24. {ACM}.

\bibitem[{Duggan(2015)}]{Duggan2015}
Maeve Duggan. 2015.
\newblock \href
  {https://www.pewresearch.org/wp-content/uploads/sites/9/2015/08/Social-Media-Update-2015-FINAL2.pdf}
  {Mobile messaging and social media 2015}.
\newblock \emph{Pew Research Center}.

\bibitem[{Dyvik(2004)}]{dyvik-2004-translations}
Helge Dyvik. 2004.
\newblock \href {https://doi.org/https://doi.org/10.1163/9789004333710_019}
  {\emph{Translations as semantic mirrors: from parallel corpus to wordnet}},
  pages 309 -- 326. Brill, Leiden, The Netherlands.

\bibitem[{Ebrahimi et~al.(2018{\natexlab{a}})Ebrahimi, Lowd, and
  Dou}]{ebrahimi-etal-2018-adversarial}
Javid Ebrahimi, Daniel Lowd, and Dejing Dou. 2018{\natexlab{a}}.
\newblock \href {https://aclanthology.org/C18-1055} {On adversarial examples
  for character-level neural machine translation}.
\newblock In \emph{Proceedings of the 27th International Conference on
  Computational Linguistics}, pages 653--663, Santa Fe, New Mexico, USA.
  Association for Computational Linguistics.

\bibitem[{Ebrahimi et~al.(2018{\natexlab{b}})Ebrahimi, Rao, Lowd, and
  Dou}]{ebrahimi-etal-2018-hotflip}
Javid Ebrahimi, Anyi Rao, Daniel Lowd, and Dejing Dou. 2018{\natexlab{b}}.
\newblock \href {https://doi.org/10.18653/v1/P18-2006} {{H}ot{F}lip: White-box
  adversarial examples for text classification}.
\newblock In \emph{Proceedings of the 56th Annual Meeting of the Association
  for Computational Linguistics (Volume 2: Short Papers)}, pages 31--36,
  Melbourne, Australia. Association for Computational Linguistics.

\bibitem[{Ebrahimi et~al.(2016)Ebrahimi, Suen, and
  Ormandjieva}]{DBLP:journals/di/EbrahimiSO16}
Mohammad~Reza Ebrahimi, Ching~Y. Suen, and Olga Ormandjieva. 2016.
\newblock \href {https://doi.org/10.1016/j.diin.2016.07.001} {Detecting
  predatory conversations in social media by deep convolutional neural
  networks}.
\newblock \emph{Digit. Investig.}, 18:33--49.

\bibitem[{Edwards and Storkey(2016)}]{DBLP:journals/corr/EdwardsS15}
Harrison Edwards and Amos~J. Storkey. 2016.
\newblock \href {http://arxiv.org/abs/1511.05897} {Censoring representations
  with an adversary}.
\newblock In \emph{4th International Conference on Learning Representations,
  {ICLR} 2016, San Juan, Puerto Rico, May 2-4, 2016, Conference Track
  Proceedings}.

\bibitem[{Eger et~al.(2019)Eger, {\c{S}}ahin, R{\"u}ckl{\'e}, Lee, Schulz,
  Mesgar, Swarnkar, Simpson, and Gurevych}]{eger-etal-2019-text}
Steffen Eger, G{\"o}zde~G{\"u}l {\c{S}}ahin, Andreas R{\"u}ckl{\'e}, Ji-Ung
  Lee, Claudia Schulz, Mohsen Mesgar, Krishnkant Swarnkar, Edwin Simpson, and
  Iryna Gurevych. 2019.
\newblock \href {https://doi.org/10.18653/v1/N19-1165} {Text processing like
  humans do: Visually attacking and shielding {NLP} systems}.
\newblock In \emph{Proceedings of the 2019 Conference of the North {A}merican
  Chapter of the Association for Computational Linguistics: Human Language
  Technologies, Volume 1 (Long and Short Papers)}, pages 1634--1647,
  Minneapolis, Minnesota. Association for Computational Linguistics.

\bibitem[{Ehni(2008)}]{Ehni2008}
Hans-J{\"o}rg Ehni. 2008.
\newblock \href {https://doi.org/10.1007/s00005-008-0020-7} {Dual use and the
  ethical responsibility of scientists}.
\newblock \emph{Archivum Immunologiae et Therapiae Experimentalis}, 56(3):147.

\bibitem[{Eisenstein et~al.(2011)Eisenstein, Smith, and
  Xing}]{eisenstein-etal-2011-discovering}
Jacob Eisenstein, Noah~A. Smith, and Eric~P. Xing. 2011.
\newblock \href {https://aclanthology.org/P11-1137} {Discovering
  sociolinguistic associations with structured sparsity}.
\newblock In \emph{Proceedings of the 49th Annual Meeting of the Association
  for Computational Linguistics: Human Language Technologies}, pages
  1365--1374, Portland, Oregon, USA. Association for Computational Linguistics.

\bibitem[{Elsafoury et~al.(2021{\natexlab{a}})Elsafoury, Katsigiannis, Pervez,
  and Ramzan}]{DBLP:journals/access/ElsafouryKPR21}
Fatma Elsafoury, Stamos Katsigiannis, Zeeshan Pervez, and Naeem Ramzan.
  2021{\natexlab{a}}.
\newblock \href {https://doi.org/10.1109/ACCESS.2021.3098979} {When the
  timeline meets the pipeline: {A} survey on automated cyberbullying
  detection}.
\newblock \emph{{IEEE} Access}, 9:103541--103563.

\bibitem[{Elsafoury et~al.(2021{\natexlab{b}})Elsafoury, Katsigiannis, Wilson,
  and Ramzan}]{DBLP:conf/sigir/ElsafouryKWR21}
Fatma Elsafoury, Stamos Katsigiannis, Steven~R. Wilson, and Naeem Ramzan.
  2021{\natexlab{b}}.
\newblock \href {https://doi.org/10.1145/3404835.3463029} {Does {BERT} pay
  attention to cyberbullying?}
\newblock In \emph{{SIGIR} '21: The 44th International {ACM} {SIGIR} Conference
  on Research and Development in Information Retrieval, Virtual Event, Canada,
  July 11-15, 2021}, pages 1900--1904. {ACM}.

\bibitem[{Elueze and Quan-Haase(2018)}]{doi:10.1177/0002764218787026}
Isioma Elueze and Anabel Quan-Haase. 2018.
\newblock \href {https://doi.org/10.1177/0002764218787026} {Privacy attitudes
  and concerns in the digital lives of older adults: Westin’s privacy
  attitude typology revisited}.
\newblock \emph{American Behavioral Scientist}, 62(10):1372--1391.

\bibitem[{Emmery et~al.(2017)Emmery, Chrupa{\l}a, and
  Daelemans}]{emmery-etal-2017-simple}
Chris Emmery, Grzegorz Chrupa{\l}a, and Walter Daelemans. 2017.
\newblock \href {https://doi.org/10.18653/v1/W17-4407} {Simple queries as
  distant labels for predicting gender on {T}witter}.
\newblock In \emph{Proceedings of the 3rd Workshop on Noisy User-generated
  Text}, pages 50--55, Copenhagen, Denmark. Association for Computational
  Linguistics.

\bibitem[{Emmery et~al.(2021{\natexlab{a}})Emmery, K{\'a}d{\'a}r, and
  Chrupa{\l}a}]{emmery-etal-2021-adversarial}
Chris Emmery, {\'A}kos K{\'a}d{\'a}r, and Grzegorz Chrupa{\l}a.
  2021{\natexlab{a}}.
\newblock \href {https://aclanthology.org/2021.eacl-main.203} {Adversarial
  stylometry in the wild: {T}ransferable lexical substitution attacks on author
  profiling}.
\newblock In \emph{Proceedings of the 16th Conference of the European Chapter
  of the Association for Computational Linguistics: Main Volume}, pages
  2388--2402, Online. Association for Computational Linguistics.

\bibitem[{Emmery et~al.(2022)Emmery, K{\'a}d{\'a}r, Chrupa{\l}a, and
  Daelemans}]{emmery-etal-2022-cyberbullying}
Chris Emmery, {\'A}kos K{\'a}d{\'a}r, Grzegorz Chrupa{\l}a, and Walter
  Daelemans. 2022.
\newblock \href {https://aclanthology.org/2022.lrec-1.319} {Cyberbullying
  classifiers are sensitive to model-agnostic perturbations}.
\newblock In \emph{Proceedings of the Thirteenth Language Resources and
  Evaluation Conference}, pages 2976--2988, Marseille, France. European
  Language Resources Association.

\bibitem[{Emmery et~al.(2019)Emmery, K{\'a}d{\'a}r, Wiltshire, and
  Hendrickson}]{emmery-2019-towards}
Chris Emmery, {\'A}kos K{\'a}d{\'a}r, Travis~J. Wiltshire, and Andrew~T.
  Hendrickson. 2019.
\newblock \href {https://doi.org/10.1007/s42113-019-00055-w} {Towards
  replication in computational cognitive modeling: a machine learning
  perspective}.
\newblock \emph{Computational Brain {\&} Behavior}, 2(3):242--246.

\bibitem[{Emmery et~al.(2018)Emmery, Manjavacas~Arevalo, and
  Chrupa{\l}a}]{emmery-etal-2018-style}
Chris Emmery, Enrique Manjavacas~Arevalo, and Grzegorz Chrupa{\l}a. 2018.
\newblock \href {https://aclanthology.org/C18-1084} {Style obfuscation by
  invariance}.
\newblock In \emph{Proceedings of the 27th International Conference on
  Computational Linguistics}, pages 984--996, Santa Fe, New Mexico, USA.
  Association for Computational Linguistics.

\bibitem[{Emmery et~al.(2021{\natexlab{b}})Emmery, Verhoeven, De~Pauw, Jacobs,
  Van~Hee, Lefever, Desmet, Hoste, and Daelemans}]{emmery-etal-2021-current}
Chris Emmery, Ben Verhoeven, Guy De~Pauw, Gilles Jacobs, Cynthia Van~Hee, Els
  Lefever, Bart Desmet, V{\'e}ronique Hoste, and Walter Daelemans.
  2021{\natexlab{b}}.
\newblock \href {https://doi.org/10.1007/s10579-020-09509-1} {Current
  limitations in cyberbullying detection: On evaluation criteria,
  reproducibility, and data scarcity}.
\newblock \emph{Language Resources and Evaluation}, 55(3):597--633.

\bibitem[{Escalante et~al.(2017)Escalante, Villatoro{-}Tello, Villarreal,
  L{\'{o}}pez{-}Monroy, Montes{-}y{-}G{\'{o}}mez, and
  Pineda}]{DBLP:journals/eswa/EscalanteVGLMP17}
Hugo~Jair Escalante, Esa{\'{u}} Villatoro{-}Tello, Sara Elena~Garza Villarreal,
  Adri{\'{a}}n~Pastor L{\'{o}}pez{-}Monroy, Manuel Montes{-}y{-}G{\'{o}}mez,
  and Luis~Villase{\~{n}}or Pineda. 2017.
\newblock \href {https://doi.org/10.1016/j.eswa.2017.07.040} {Early detection
  of deception and aggressiveness using profile-based representations}.
\newblock \emph{Expert Syst. Appl.}, 89:99--111.

\bibitem[{Evans(2017)}]{evans2017dynamics}
David~S Evans. 2017.
\newblock \href {http://dx.doi.org/10.2139/ssrn.3009438} {Why the dynamics of
  competition for online platforms leads to sleepless nights but not sleepy
  monopolies}.
\newblock \emph{Available at SSRN 3009438}.

\bibitem[{Fan et~al.(2008)Fan, Chang, Hsieh, Wang, and
  Lin}]{DBLP:journals/jmlr/FanCHWL08}
Rong{-}En Fan, Kai{-}Wei Chang, Cho{-}Jui Hsieh, Xiang{-}Rui Wang, and
  Chih{-}Jen Lin. 2008.
\newblock \href {https://dl.acm.org/doi/10.5555/1390681.1442794} {{LIBLINEAR:}
  {A} library for large linear classification}.
\newblock \emph{J. Mach. Learn. Res.}, 9:1871--1874.

\bibitem[{Feng et~al.(2020)Feng, He, Liu, Wang, and
  Choo}]{DBLP:journals/tifs/FengHLWC20}
Qi~Feng, Debiao He, Zhe Liu, Huaqun Wang, and Kim{-}Kwang~Raymond Choo. 2020.
\newblock \href {https://doi.org/10.1109/TIFS.2020.2997134} {Securenlp: {A}
  system for multi-party privacy-preserving natural language processing}.
\newblock \emph{{IEEE} Trans. Inf. Forensics Secur.}, 15:3709--3721.

\bibitem[{Feng et~al.(2012)Feng, Banerjee, and Choi}]{feng-etal-2012-syntactic}
Song Feng, Ritwik Banerjee, and Yejin Choi. 2012.
\newblock \href {https://aclanthology.org/P12-2034} {Syntactic stylometry for
  deception detection}.
\newblock In \emph{Proceedings of the 50th Annual Meeting of the Association
  for Computational Linguistics (Volume 2: Short Papers)}, pages 171--175, Jeju
  Island, Korea. Association for Computational Linguistics.

\bibitem[{Feng et~al.(2021)Feng, Gangal, Wei, Chandar, Vosoughi, Mitamura, and
  Hovy}]{feng-etal-2021-survey}
Steven Feng, Varun Gangal, Jason Wei, Sarath Chandar, Soroush Vosoughi, Teruko
  Mitamura, and Eduard Hovy. 2021.
\newblock \href {https://doi.org/10.18653/v1/2021.findings-acl.84} {A survey of
  data augmentation approaches for {NLP}}.
\newblock In \emph{Findings of the Association for Computational Linguistics:
  ACL-IJCNLP 2021}, pages 968--988, Online. Association for Computational
  Linguistics.

\bibitem[{Fernandes et~al.(2019)Fernandes, Dras, and
  McIver}]{DBLP:conf/post/FernandesDM19}
Natasha Fernandes, Mark Dras, and Annabelle McIver. 2019.
\newblock \href {https://doi.org/10.1007/978-3-030-17138-4\_6} {Generalised
  differential privacy for text document processing}.
\newblock In \emph{Principles of Security and Trust - 8th International
  Conference, {POST} 2019, Held as Part of the European Joint Conferences on
  Theory and Practice of Software, {ETAPS} 2019, Prague, Czech Republic, April
  6-11, 2019, Proceedings}, volume 11426 of \emph{Lecture Notes in Computer
  Science}, pages 123--148. Springer.

\bibitem[{Feyisetan et~al.(2020)Feyisetan, Ghanavati, and
  Thaine}]{DBLP:conf/wsdm/FeyisetanGT20}
Oluwaseyi Feyisetan, Sepideh Ghanavati, and Patricia Thaine. 2020.
\newblock \href {https://doi.org/10.1145/3336191.3371881} {Workshop on privacy
  in {NLP} (privatenlp 2020)}.
\newblock In \emph{{WSDM} '20: The Thirteenth {ACM} International Conference on
  Web Search and Data Mining, Houston, TX, USA, February 3-7, 2020}, pages
  903--904. {ACM}.

\bibitem[{Ficler and Goldberg(2017)}]{ficler2017controlling}
Jessica Ficler and Yoav Goldberg. 2017.
\newblock Controlling linguistic style aspects in neural language generation.
\newblock \emph{arXiv preprint arXiv:1707.02633}.

\bibitem[{Finkel and Manning(2009)}]{finkel-manning-2009-hierarchical}
Jenny~Rose Finkel and Christopher~D. Manning. 2009.
\newblock \href {https://aclanthology.org/N09-1068} {Hierarchical {B}ayesian
  domain adaptation}.
\newblock In \emph{Proceedings of Human Language Technologies: The 2009 Annual
  Conference of the North {A}merican Chapter of the Association for
  Computational Linguistics}, pages 602--610, Boulder, Colorado. Association
  for Computational Linguistics.

\bibitem[{Fornaciari and Poesio(2014)}]{fornaciari-poesio-2014-identifying}
Tommaso Fornaciari and Massimo Poesio. 2014.
\newblock \href {https://doi.org/10.3115/v1/E14-1030} {Identifying fake
  {A}mazon reviews as learning from crowds}.
\newblock In \emph{Proceedings of the 14th Conference of the {E}uropean Chapter
  of the Association for Computational Linguistics}, pages 279--287,
  Gothenburg, Sweden. Association for Computational Linguistics.

\bibitem[{Fortuna and Nunes(2018)}]{DBLP:journals/csur/FortunaN18}
Paula Fortuna and S{\'{e}}rgio Nunes. 2018.
\newblock \href {https://doi.org/10.1145/3232676} {A survey on automatic
  detection of hate speech in text}.
\newblock \emph{{ACM} Comput. Surv.}, 51(4):85:1--85:30.

\bibitem[{Fortuna et~al.(2021)Fortuna, {Soler Company}, and
  Wanner}]{DBLP:journals/ipm/FortunaCW21}
Paula Fortuna, Juan {Soler Company}, and Leo Wanner. 2021.
\newblock \href {https://doi.org/10.1016/j.ipm.2021.102524} {How well do hate
  speech, toxicity, abusive and offensive language classification models
  generalize across datasets?}
\newblock \emph{Inf. Process. Manag.}, 58(3):102524.

\bibitem[{Founta et~al.(2018)Founta, Djouvas, Chatzakou, Leontiadis, Blackburn,
  Stringhini, Vakali, Sirivianos, and
  Kourtellis}]{DBLP:conf/icwsm/FountaDCLBSVSK18}
Antigoni{-}Maria Founta, Constantinos Djouvas, Despoina Chatzakou, Ilias
  Leontiadis, Jeremy Blackburn, Gianluca Stringhini, Athena Vakali, Michael
  Sirivianos, and Nicolas Kourtellis. 2018.
\newblock \href {https://aaai.org/ocs/index.php/ICWSM/ICWSM18/paper/view/17909}
  {Large scale crowdsourcing and characterization of twitter abusive behavior}.
\newblock In \emph{Proceedings of the Twelfth International Conference on Web
  and Social Media, {ICWSM} 2018, Stanford, California, USA, June 25-28, 2018},
  pages 491--500. {AAAI} Press.

\bibitem[{Ganin and Lempitsky(2015)}]{DBLP:conf/icml/GaninL15}
Yaroslav Ganin and Victor~S. Lempitsky. 2015.
\newblock \href {http://proceedings.mlr.press/v37/ganin15.html} {Unsupervised
  domain adaptation by backpropagation}.
\newblock In \emph{Proceedings of the 32nd International Conference on Machine
  Learning, {ICML} 2015, Lille, France, 6-11 July 2015}, volume~37 of
  \emph{{JMLR} Workshop and Conference Proceedings}, pages 1180--1189.
  JMLR.org.

\bibitem[{Gardner et~al.(2020)Gardner, Artzi, Basmova, Berant, Bogin, Chen,
  Dasigi, Dua, Elazar, Gottumukkala, Gupta, Hajishirzi, Ilharco, Khashabi, Lin,
  Liu, Liu, Mulcaire, Ning, Singh, Smith, Subramanian, Tsarfaty, Wallace,
  Zhang, and Zhou}]{DBLP:journals/corr/abs-2004-02709}
Matt Gardner, Yoav Artzi, Victoria Basmova, Jonathan Berant, Ben Bogin, Sihao
  Chen, Pradeep Dasigi, Dheeru Dua, Yanai Elazar, Ananth Gottumukkala, Nitish
  Gupta, Hanna Hajishirzi, Gabriel Ilharco, Daniel Khashabi, Kevin Lin,
  Jiangming Liu, Nelson~F. Liu, Phoebe Mulcaire, Qiang Ning, Sameer Singh,
  Noah~A. Smith, Sanjay Subramanian, Reut Tsarfaty, Eric Wallace, Ally Zhang,
  and Ben Zhou. 2020.
\newblock \href {http://arxiv.org/abs/2004.02709} {Evaluating {NLP} models via
  contrast sets}.
\newblock \emph{CoRR}, abs/2004.02709.

\bibitem[{Gebru et~al.(2021)Gebru, Morgenstern, Vecchione, Vaughan, Wallach,
  III, and Crawford}]{DBLP:journals/cacm/GebruMVVWDC21}
Timnit Gebru, Jamie Morgenstern, Briana Vecchione, Jennifer~Wortman Vaughan,
  Hanna~M. Wallach, Hal~Daum{\'{e}} III, and Kate Crawford. 2021.
\newblock \href {https://doi.org/10.1145/3458723} {Datasheets for datasets}.
\newblock \emph{Commun. {ACM}}, 64(12):86--92.

\bibitem[{Gehman et~al.(2020)Gehman, Gururangan, Sap, Choi, and
  Smith}]{gehman-etal-2020-realtoxicityprompts}
Samuel Gehman, Suchin Gururangan, Maarten Sap, Yejin Choi, and Noah~A. Smith.
  2020.
\newblock \href {https://doi.org/10.18653/v1/2020.findings-emnlp.301}
  {{R}eal{T}oxicity{P}rompts: Evaluating neural toxic degeneration in language
  models}.
\newblock In \emph{Findings of the Association for Computational Linguistics:
  EMNLP 2020}, pages 3356--3369, Online. Association for Computational
  Linguistics.

\bibitem[{Gencoglu(2021)}]{DBLP:journals/internet/Gencoglu21}
Oguzhan Gencoglu. 2021.
\newblock \href {https://doi.org/10.1109/MIC.2020.3032461} {Cyberbullying
  detection with fairness constraints}.
\newblock \emph{{IEEE} Internet Comput.}, 25(1):20--29.

\bibitem[{Gers et~al.(2002)Gers, Schraudolph, and
  Schmidhuber}]{DBLP:journals/jmlr/GersSS02}
Felix~A. Gers, Nicol~N. Schraudolph, and J{\"{u}}rgen Schmidhuber. 2002.
\newblock \href {http://jmlr.org/papers/v3/gers02a.html} {Learning precise
  timing with {LSTM} recurrent networks}.
\newblock \emph{J. Mach. Learn. Res.}, 3:115--143.

\bibitem[{Glorot et~al.(2011)Glorot, Bordes, and
  Bengio}]{DBLP:conf/icml/GlorotBB11}
Xavier Glorot, Antoine Bordes, and Yoshua Bengio. 2011.
\newblock \href {https://icml.cc/2011/papers/342\_icmlpaper.pdf} {Domain
  adaptation for large-scale sentiment classification: {A} deep learning
  approach}.
\newblock In \emph{Proceedings of the 28th International Conference on Machine
  Learning, {ICML} 2011, Bellevue, Washington, USA, June 28 - July 2, 2011},
  pages 513--520. Omnipress.

\bibitem[{Gonen and Goldberg(2019)}]{gonen-goldberg-2019-lipstick}
Hila Gonen and Yoav Goldberg. 2019.
\newblock \href {https://doi.org/10.18653/v1/N19-1061} {Lipstick on a pig:
  {D}ebiasing methods cover up systematic gender biases in word embeddings but
  do not remove them}.
\newblock In \emph{Proceedings of the 2019 Conference of the North {A}merican
  Chapter of the Association for Computational Linguistics: Human Language
  Technologies, Volume 1 (Long and Short Papers)}, pages 609--614, Minneapolis,
  Minnesota. Association for Computational Linguistics.

\bibitem[{Goodfellow et~al.(2015)Goodfellow, Shlens, and
  Szegedy}]{DBLP:journals/corr/GoodfellowSS14}
Ian~J. Goodfellow, Jonathon Shlens, and Christian Szegedy. 2015.
\newblock \href {http://arxiv.org/abs/1412.6572} {Explaining and harnessing
  adversarial examples}.
\newblock In \emph{3rd International Conference on Learning Representations,
  {ICLR} 2015, San Diego, CA, USA, May 7-9, 2015, Conference Track
  Proceedings}.

\bibitem[{Greevy and Smeaton(2004)}]{DBLP:conf/sigir/GreevyS04}
Edel Greevy and Alan~F. Smeaton. 2004.
\newblock \href {https://doi.org/10.1145/1008992.1009074} {Classifying racist
  texts using a support vector machine}.
\newblock In \emph{{SIGIR} 2004: Proceedings of the 27th Annual International
  {ACM} {SIGIR} Conference on Research and Development in Information
  Retrieval, Sheffield, UK, July 25-29, 2004}, pages 468--469. {ACM}.

\bibitem[{Gui et~al.(2017)Gui, Zhang, Huang, Peng, and
  Huang}]{gui-etal-2017-part}
Tao Gui, Qi~Zhang, Haoran Huang, Minlong Peng, and Xuanjing Huang. 2017.
\newblock \href {https://doi.org/10.18653/v1/D17-1256} {Part-of-speech tagging
  for {T}witter with adversarial neural networks}.
\newblock In \emph{Proceedings of the 2017 Conference on Empirical Methods in
  Natural Language Processing}, pages 2411--2420, Copenhagen, Denmark.
  Association for Computational Linguistics.

\bibitem[{Gunasekara and Nejadgholi(2018)}]{gunasekara-nejadgholi-2018-review}
Isuru Gunasekara and Isar Nejadgholi. 2018.
\newblock \href {https://doi.org/10.18653/v1/W18-5103} {A review of standard
  text classification practices for multi-label toxicity identification of
  online content}.
\newblock In \emph{Proceedings of the 2nd Workshop on Abusive Language Online
  ({ALW}2)}, pages 21--25, Brussels, Belgium. Association for Computational
  Linguistics.

\bibitem[{Gundersen and Kjensmo(2018)}]{Gundersen_Kjensmo_2018}
Odd~Erik Gundersen and Sigbjørn Kjensmo. 2018.
\newblock \href {https://doi.org/10.1609/aaai.v32i1.11503} {State of the art:
  Reproducibility in artificial intelligence}.
\newblock \emph{Proceedings of the AAAI Conference on Artificial Intelligence},
  32(1).

\bibitem[{Guzman{-}Silverio et~al.(2020)Guzman{-}Silverio, Balderas{-}Paredes,
  and L{\'{o}}pez{-}Monroy}]{DBLP:conf/sepln/Guzman-Silverio20}
Mario Guzman{-}Silverio, {\'{A}}ngel Balderas{-}Paredes, and
  Adri{\'{a}}n~Pastor L{\'{o}}pez{-}Monroy. 2020.
\newblock \href {http://ceur-ws.org/Vol-2664/mexa3t\_paper9.pdf} {Transformers
  and data augmentation for aggressiveness detection in mexican spanish}.
\newblock In \emph{Proceedings of the Iberian Languages Evaluation Forum
  (IberLEF 2020) co-located with 36th Conference of the Spanish Society for
  Natural Language Processing {(SEPLN} 2020), M{\'{a}}laga, Spain, September
  23th, 2020}, volume 2664 of \emph{{CEUR} Workshop Proceedings}, pages
  293--302. CEUR-WS.org.

\bibitem[{Hagen et~al.(2017)Hagen, Potthast, and
  Stein}]{DBLP:conf/clef/HagenPS17}
Matthias Hagen, Martin Potthast, and Benno Stein. 2017.
\newblock \href {http://ceur-ws.org/Vol-1866/invited\_paper\_4.pdf} {Overview
  of the author obfuscation task at {PAN} 2017: Safety evaluation revisited}.
\newblock In \emph{Working Notes of {CLEF} 2017 - Conference and Labs of the
  Evaluation Forum, Dublin, Ireland, September 11-14, 2017}, volume 1866 of
  \emph{{CEUR} Workshop Proceedings}. CEUR-WS.org.

\bibitem[{Han and Tsvetkov(2020)}]{han-tsvetkov-2020-fortifying}
Xiaochuang Han and Yulia Tsvetkov. 2020.
\newblock \href {https://doi.org/10.18653/v1/2020.emnlp-main.622} {Fortifying
  toxic speech detectors against veiled toxicity}.
\newblock In \emph{Proceedings of the 2020 Conference on Empirical Methods in
  Natural Language Processing (EMNLP)}, pages 7732--7739, Online. Association
  for Computational Linguistics.

\bibitem[{Hanu et~al.(2020)Hanu, Unitary, and team}]{Detoxify}
Laura Hanu, Unitary, and team. 2020.
\newblock \href {https://github.com/unitaryai/detoxify} {Detoxify}.
\newblock GitHub.

\bibitem[{Haralick et~al.(1973)Haralick, Shanmugam, and
  Dinstein}]{DBLP:journals/tsmc/HaralickSD73}
Robert~M. Haralick, K.~Sam Shanmugam, and Its'hak Dinstein. 1973.
\newblock \href {https://doi.org/10.1109/TSMC.1973.4309314} {Textural features
  for image classification}.
\newblock \emph{{IEEE} Trans. Syst. Man Cybern.}, 3(6):610--621.

\bibitem[{Hargittai(2010)}]{https://doi.org/10.1111/j.1475-682X.2009.00317.x}
Eszter Hargittai. 2010.
\newblock \href
  {https://doi.org/https://doi.org/10.1111/j.1475-682X.2009.00317.x} {Digital
  na(t)ives? variation in internet skills and uses among members of the “net
  generation”*}.
\newblock \emph{Sociological Inquiry}, 80(1):92--113.

\bibitem[{Haroon et~al.(2021)Haroon, Zaffar, Srinivasan, and
  Shafiq}]{DBLP:journals/corr/abs-2109-07028}
Muhammad Haroon, Muhammad~Fareed Zaffar, Padmini Srinivasan, and Zubair Shafiq.
  2021.
\newblock \href {http://arxiv.org/abs/2109.07028} {Avengers ensemble! improving
  transferability of authorship obfuscation}.
\newblock \emph{CoRR}, abs/2109.07028.

\bibitem[{He et~al.(2018)He, Cai, and Yu}]{DBLP:journals/tvt/HeCY18}
Zaobo He, Zhipeng Cai, and Jiguo Yu. 2018.
\newblock \href {https://doi.org/10.1109/TVT.2017.2738018} {Latent-data privacy
  preserving with customized data utility for social network data}.
\newblock \emph{{IEEE} Trans. Veh. Technol.}, 67(1):665--673.

\bibitem[{Hee et~al.(2015)Hee, Lefever, Verhoeven, Mennes, Desmet, Pauw,
  Daelemans, and Hoste}]{DBLP:conf/ranlp/HeeLVMDPDH15}
Cynthia~Van Hee, Els Lefever, Ben Verhoeven, Julie Mennes, Bart Desmet, Guy~De
  Pauw, Walter Daelemans, and V{\'{e}}ronique Hoste. 2015.
\newblock \href {https://aclanthology.org/R15-1086/} {Detection and
  fine-grained classification of cyberbullying events}.
\newblock In \emph{Recent Advances in Natural Language Processing, {RANLP}
  2015, 7-9 September, 2015, Hissar, Bulgaria}, pages 672--680. {RANLP} 2015
  Organising Committee / {ACL}.

\bibitem[{Henderson and Brunskill(2018)}]{DBLP:journals/corr/abs-1812-01074}
Peter Henderson and Emma Brunskill. 2018.
\newblock \href {http://arxiv.org/abs/1812.01074} {Distilling information from
  a flood: {A} possibility for the use of meta-analysis and systematic review
  in machine learning research}.
\newblock \emph{CoRR}, abs/1812.01074.

\bibitem[{Hill et~al.(2015)Hill, Reichart, and
  Korhonen}]{hill-etal-2015-simlex}
Felix Hill, Roi Reichart, and Anna Korhonen. 2015.
\newblock \href {https://doi.org/10.1162/COLI_a_00237} {{S}im{L}ex-999:
  Evaluating semantic models with (genuine) similarity estimation}.
\newblock \emph{Computational Linguistics}, 41(4):665--695.

\bibitem[{Hochreiter and Schmidhuber(1997)}]{DBLP:journals/neco/HochreiterS97}
Sepp Hochreiter and J{\"{u}}rgen Schmidhuber. 1997.
\newblock \href {https://doi.org/10.1162/neco.1997.9.8.1735} {Long short-term
  memory}.
\newblock \emph{Neural Comput.}, 9(8):1735--1780.

\bibitem[{Hollingsworth(2012)}]{DBLP:conf/tsd/Hollingsworth12}
Charles Hollingsworth. 2012.
\newblock \href {https://doi.org/10.1007/978-3-642-32790-2\_38} {Using
  dependency-based annotations for authorship identification}.
\newblock In \emph{Text, Speech and Dialogue - 15th International Conference,
  {TSD} 2012, Brno, Czech Republic, September 3-7, 2012. Proceedings}, volume
  7499 of \emph{Lecture Notes in Computer Science}, pages 314--319. Springer.

\bibitem[{Holmes and Forsyth(1995)}]{10.1093/llc/10.2.111}
D.~I. Holmes and R.~S. Forsyth. 1995.
\newblock \href {https://doi.org/10.1093/llc/10.2.111} {{The Federalist
  Revisited: New Directions in Authorship Attribution}}.
\newblock \emph{Literary and Linguistic Computing}, 10(2):111--127.

\bibitem[{Holmes(1998)}]{10.1093/llc/13.3.111}
David~I. Holmes. 1998.
\newblock \href {https://doi.org/10.1093/llc/13.3.111} {{The Evolution of
  Stylometry in Humanities Scholarship}}.
\newblock \emph{Literary and Linguistic Computing}, 13(3):111--117.

\bibitem[{Honnibal et~al.(2020)Honnibal, Montani, Van~Landeghem, and
  Adriane}]{honnibal2020spacy}
Matthew Honnibal, Ines Montani, Sofie Van~Landeghem, and Boyd Adriane. 2020.
\newblock \href {https://doi.org/10.5281/zenodo.1212303} {spacy:
  Industrial-strength natural language processing in python}.

\bibitem[{Hosseini et~al.(2017)Hosseini, Kannan, Zhang, and
  Poovendran}]{DBLP:journals/corr/HosseiniKZP17}
Hossein Hosseini, Sreeram Kannan, Baosen Zhang, and Radha Poovendran. 2017.
\newblock \href {http://arxiv.org/abs/1702.08138} {Deceiving google's
  perspective {API} built for detecting toxic comments}.
\newblock \emph{CoRR}, abs/1702.08138.

\bibitem[{Hosseinmardi et~al.(2015)Hosseinmardi, Mattson, Rafiq, Han, Lv, and
  Mishra}]{DBLP:journals/corr/HosseinmardiMRH15a}
Homa Hosseinmardi, Sabrina~Arredondo Mattson, Rahat~Ibn Rafiq, Richard Han, Qin
  Lv, and Shivakant Mishra. 2015.
\newblock \href {http://arxiv.org/abs/1508.06257} {Prediction of cyberbullying
  incidents on the instagram social network}.
\newblock \emph{CoRR}, abs/1508.06257.

\bibitem[{Hovy and Spruit(2016)}]{hovy-spruit-2016-social}
Dirk Hovy and Shannon~L. Spruit. 2016.
\newblock \href {https://doi.org/10.18653/v1/P16-2096} {The social impact of
  natural language processing}.
\newblock In \emph{Proceedings of the 54th Annual Meeting of the Association
  for Computational Linguistics (Volume 2: Short Papers)}, pages 591--598,
  Berlin, Germany. Association for Computational Linguistics.

\bibitem[{Howcroft et~al.(2020)Howcroft, Belz, Clinciu, Gkatzia, Hasan,
  Mahamood, Mille, van Miltenburg, Santhanam, and
  Rieser}]{howcroft-etal-2020-twenty}
David~M. Howcroft, Anya Belz, Miruna-Adriana Clinciu, Dimitra Gkatzia, Sadid~A.
  Hasan, Saad Mahamood, Simon Mille, Emiel van Miltenburg, Sashank Santhanam,
  and Verena Rieser. 2020.
\newblock \href {https://aclanthology.org/2020.inlg-1.23} {Twenty years of
  confusion in human evaluation: {NLG} needs evaluation sheets and standardised
  definitions}.
\newblock In \emph{Proceedings of the 13th International Conference on Natural
  Language Generation}, pages 169--182, Dublin, Ireland. Association for
  Computational Linguistics.

\bibitem[{Huang et~al.(2020)Huang, Xing, Dernoncourt, and
  Paul}]{huang-etal-2020-multilingual}
Xiaolei Huang, Linzi Xing, Franck Dernoncourt, and Michael~J. Paul. 2020.
\newblock \href {https://aclanthology.org/2020.lrec-1.180} {Multilingual
  {T}witter corpus and baselines for evaluating demographic bias in hate speech
  recognition}.
\newblock In \emph{Proceedings of the 12th Language Resources and Evaluation
  Conference}, pages 1440--1448, Marseille, France. European Language Resources
  Association.

\bibitem[{Hutson(2018)}]{doi:10.1126/science.359.6377.725}
Matthew Hutson. 2018.
\newblock \href {https://doi.org/10.1126/science.359.6377.725} {Artificial
  intelligence faces reproducibility crisis}.
\newblock \emph{Science}, 359(6377):725--726.

\bibitem[{Ibrahim et~al.(2018)Ibrahim, Torki, and
  El{-}Makky}]{DBLP:conf/icmla/IbrahimTE18}
Mai Ibrahim, Marwan Torki, and Nagwa~M. El{-}Makky. 2018.
\newblock \href {https://doi.org/10.1109/ICMLA.2018.00141} {Imbalanced toxic
  comments classification using data augmentation and deep learning}.
\newblock In \emph{17th {IEEE} International Conference on Machine Learning and
  Applications, {ICMLA} 2018, Orlando, FL, USA, December 17-20, 2018}, pages
  875--878. {IEEE}.

\bibitem[{Ide et~al.(2002)Ide, Erjavec, and Tufis}]{ide-etal-2002-sense}
Nancy Ide, Tomaz Erjavec, and Dan Tufis. 2002.
\newblock \href {https://doi.org/10.3115/1118675.1118683} {Sense discrimination
  with parallel corpora}.
\newblock In \emph{Proceedings of the {ACL}-02 Workshop on Word Sense
  Disambiguation: Recent Successes and Future Directions}, pages 61--66.
  Association for Computational Linguistics.

\bibitem[{Ignatov et~al.(2019)Ignatov, Timofte, Kulik, Yang, Wang, Baum, Wu,
  Xu, and Gool}]{DBLP:conf/iccvw/IgnatovTKYWBWXG19}
Andrey Ignatov, Radu Timofte, Andrei Kulik, Seungsoo Yang, Ke~Wang, Felix Baum,
  Max Wu, Lirong Xu, and Luc~Van Gool. 2019.
\newblock \href {https://doi.org/10.1109/ICCVW.2019.00447} {{AI} benchmark: All
  about deep learning on smartphones in 2019}.
\newblock In \emph{2019 {IEEE/CVF} International Conference on Computer Vision
  Workshops, {ICCV} Workshops 2019, Seoul, Korea (South), October 27-28, 2019},
  pages 3617--3635. {IEEE}.

\bibitem[{Ippolito et~al.(2020)Ippolito, Duckworth, Callison-Burch, and
  Eck}]{ippolito-etal-2020-automatic}
Daphne Ippolito, Daniel Duckworth, Chris Callison-Burch, and Douglas Eck. 2020.
\newblock \href {https://doi.org/10.18653/v1/2020.acl-main.164} {Automatic
  detection of generated text is easiest when humans are fooled}.
\newblock In \emph{Proceedings of the 58th Annual Meeting of the Association
  for Computational Linguistics}, pages 1808--1822, Online. Association for
  Computational Linguistics.

\bibitem[{Isaak and Hanna(2018)}]{DBLP:journals/computer/IsaakH18}
Jim Isaak and Mina~J. Hanna. 2018.
\newblock \href {https://doi.org/10.1109/MC.2018.3191268} {User data privacy:
  Facebook, cambridge analytica, and privacy protection}.
\newblock \emph{Computer}, 51(8):56--59.

\bibitem[{Jahan and Oussalah(2021)}]{DBLP:journals/corr/abs-2106-00742}
Md~Saroar Jahan and Mourad Oussalah. 2021.
\newblock \href {http://arxiv.org/abs/2106.00742} {A systematic review of hate
  speech automatic detection using natural language processing}.
\newblock \emph{CoRR}, abs/2106.00742.

\bibitem[{Jia and Liang(2017)}]{jia-liang-2017-adversarial}
Robin Jia and Percy Liang. 2017.
\newblock \href {https://doi.org/10.18653/v1/D17-1215} {Adversarial examples
  for evaluating reading comprehension systems}.
\newblock In \emph{Proceedings of the 2017 Conference on Empirical Methods in
  Natural Language Processing}, pages 2021--2031, Copenhagen, Denmark.
  Association for Computational Linguistics.

\bibitem[{Jiang and Zhai(2007)}]{DBLP:conf/cikm/JiangZ07}
Jing Jiang and ChengXiang Zhai. 2007.
\newblock \href {https://doi.org/10.1145/1321440.1321498} {A two-stage approach
  to domain adaptation for statistical classifiers}.
\newblock In \emph{Proceedings of the Sixteenth {ACM} Conference on Information
  and Knowledge Management, {CIKM} 2007, Lisbon, Portugal, November 6-10,
  2007}, pages 401--410. {ACM}.

\bibitem[{Jin et~al.(2020)Jin, Jin, Zhou, and
  Szolovits}]{DBLP:conf/aaai/JinJZS20}
Di~Jin, Zhijing Jin, Joey~Tianyi Zhou, and Peter Szolovits. 2020.
\newblock Is {BERT} really robust? {A} strong baseline for natural language
  attack on text classification and entailment.
\newblock In \emph{The Thirty-Fourth {AAAI} Conference on Artificial
  Intelligence, {AAAI} 2020, The Thirty-Second Innovative Applications of
  Artificial Intelligence Conference, {IAAI} 2020, The Tenth {AAAI} Symposium
  on Educational Advances in Artificial Intelligence, {EAAI} 2020, New York,
  NY, USA, February 7-12, 2020}, pages 8018--8025. {AAAI} Press.

\bibitem[{Jo and Gebru(2020)}]{DBLP:conf/fat/JoG20}
Eun~Seo Jo and Timnit Gebru. 2020.
\newblock \href {https://doi.org/10.1145/3351095.3372829} {Lessons from
  archives: strategies for collecting sociocultural data in machine learning}.
\newblock In \emph{FAT* '20: Conference on Fairness, Accountability, and
  Transparency, Barcelona, Spain, January 27-30, 2020}, pages 306--316. {ACM}.

\bibitem[{John et~al.(2019)John, Mou, Bahuleyan, and
  Vechtomova}]{john-etal-2019-disentangled}
Vineet John, Lili Mou, Hareesh Bahuleyan, and Olga Vechtomova. 2019.
\newblock \href {https://doi.org/10.18653/v1/P19-1041} {Disentangled
  representation learning for non-parallel text style transfer}.
\newblock In \emph{Proceedings of the 57th Annual Meeting of the Association
  for Computational Linguistics}, pages 424--434, Florence, Italy. Association
  for Computational Linguistics.

\bibitem[{Johnson et~al.(2017)Johnson, Schuster, Le, Krikun, Wu, Chen, Thorat,
  Vi{\'{e}}gas, Wattenberg, Corrado, Hughes, and
  Dean}]{DBLP:journals/tacl/JohnsonSLKWCTVW17}
Melvin Johnson, Mike Schuster, Quoc~V. Le, Maxim Krikun, Yonghui Wu, Zhifeng
  Chen, Nikhil Thorat, Fernanda~B. Vi{\'{e}}gas, Martin Wattenberg, Greg
  Corrado, Macduff Hughes, and Jeffrey Dean. 2017.
\newblock \href {https://transacl.org/ojs/index.php/tacl/article/view/1081}
  {Google's multilingual neural machine translation system: Enabling zero-shot
  translation}.
\newblock \emph{Trans. Assoc. Comput. Linguistics}, 5:339--351.

\bibitem[{Joo and Hwang(2019)}]{DBLP:conf/clef/JooH19}
Youngjun Joo and Inchon Hwang. 2019.
\newblock \href {http://ceur-ws.org/Vol-2380/paper\_174.pdf} {Author profiling
  on social media: An ensemble learning approach using various features}.
\newblock In \emph{Working Notes of {CLEF} 2019 - Conference and Labs of the
  Evaluation Forum, Lugano, Switzerland, September 9-12, 2019}, volume 2380 of
  \emph{{CEUR} Workshop Proceedings}. CEUR-WS.org.

\bibitem[{Joulin et~al.(2017)Joulin, Grave, Bojanowski, and
  Mikolov}]{joulin-etal-2017-bag}
Armand Joulin, Edouard Grave, Piotr Bojanowski, and Tomas Mikolov. 2017.
\newblock \href {https://aclanthology.org/E17-2068} {Bag of tricks for
  efficient text classification}.
\newblock In \emph{Proceedings of the 15th Conference of the {E}uropean Chapter
  of the Association for Computational Linguistics: Volume 2, Short Papers},
  pages 427--431, Valencia, Spain. Association for Computational Linguistics.

\bibitem[{Jungiewicz and Smywinski-Pohl(2019)}]{Jungiewicz_Smywinski-Pohl_2019}
Michał Jungiewicz and Aleksander Smywinski-Pohl. 2019.
\newblock \href {https://doi.org/10.7494/csci.2019.20.1.3023} {Towards textual
  data augmentation for neural networks: synonyms and maximum loss}.
\newblock \emph{Computer Science}, 20(1).

\bibitem[{Juola and Vescovi(2011)}]{DBLP:conf/ifip11-9/JuolaV11}
Patrick Juola and Darren Vescovi. 2011.
\newblock \href {https://doi.org/10.1007/978-3-642-24212-0\_9} {Analyzing
  stylometric approaches to author obfuscation}.
\newblock In \emph{Advances in Digital Forensics {VII} - 7th {IFIP} {WG} 11.9
  International Conference on Digital Forensics, Orlando, FL, USA, January 31 -
  February 2, 2011, Revised Selected Papers}, volume 361 of \emph{{IFIP}
  Advances in Information and Communication Technology}, pages 115--125.
  Springer.

\bibitem[{Jurowetzki et~al.(2021)Jurowetzki, Hain, Mateos{-}Garcia, and
  Stathoulopoulos}]{DBLP:journals/corr/abs-2102-01648}
Roman Jurowetzki, Daniel~S. Hain, Juan Mateos{-}Garcia, and Konstantinos
  Stathoulopoulos. 2021.
\newblock \href {http://arxiv.org/abs/2102.01648} {The privatization of {AI}
  research(-ers): Causes and potential consequences - from university-industry
  interaction to public research brain-drain?}
\newblock \emph{CoRR}, abs/2102.01648.

\bibitem[{Kaati et~al.(2015)Kaati, Omer, Prucha, and
  Shrestha}]{DBLP:conf/icdm/KaatiOPS15}
Lisa Kaati, Enghin Omer, Nico Prucha, and Amendra Shrestha. 2015.
\newblock \href {https://doi.org/10.1109/ICDMW.2015.9} {Detecting multipliers
  of jihadism on twitter}.
\newblock In \emph{{IEEE} International Conference on Data Mining Workshop,
  {ICDMW} 2015, Atlantic City, NJ, USA, November 14-17, 2015}, pages 954--960.
  {IEEE} Computer Society.

\bibitem[{Kabbara and Cheung(2016)}]{kabbara-cheung-2016-stylistic}
Jad Kabbara and Jackie Chi~Kit Cheung. 2016.
\newblock \href {https://doi.org/10.18653/v1/W16-6010} {Stylistic transfer in
  natural language generation systems using recurrent neural networks}.
\newblock In \emph{Proceedings of the Workshop on Uphill Battles in Language
  Processing: Scaling Early Achievements to Robust Methods}, pages 43--47,
  Austin, TX. Association for Computational Linguistics.

\bibitem[{Kacmarcik and Gamon(2006)}]{kacmarcik-gamon-2006-obfuscating}
Gary Kacmarcik and Michael Gamon. 2006.
\newblock \href {https://aclanthology.org/P06-2058} {Obfuscating document
  stylometry to preserve author anonymity}.
\newblock In \emph{Proceedings of the {COLING}/{ACL} 2006 Main Conference
  Poster Sessions}, pages 444--451, Sydney, Australia. Association for
  Computational Linguistics.

\bibitem[{K{\'a}d{\'a}r et~al.(2017)K{\'a}d{\'a}r, Chrupa{\l}a, and
  Alishahi}]{kadar-etal-2017-representation}
{\'A}kos K{\'a}d{\'a}r, Grzegorz Chrupa{\l}a, and Afra Alishahi. 2017.
\newblock \href {https://doi.org/10.1162/COLI_a_00300} {Representation of
  linguistic form and function in recurrent neural networks}.
\newblock \emph{Computational Linguistics}, 43(4):761--780.

\bibitem[{K{\'{a}}d{\'{a}}r et~al.(2017)K{\'{a}}d{\'{a}}r, Chrupa{\l}a, and
  Alishahi}]{DBLP:journals/coling/KadarCA17}
{\'{A}}kos K{\'{a}}d{\'{a}}r, Grzegorz Chrupa{\l}a, and Afra Alishahi. 2017.
\newblock \href {https://doi.org/10.1162/COLI\_a\_00300} {Representation of
  linguistic form and function in recurrent neural networks}.
\newblock \emph{Comput. Linguistics}, 43(4).

\bibitem[{Kann et~al.(2014)Kann, Kinchen, Shanklin, Flint, Kawkins, Harris,
  Lowry, Olsen, McManus, Chyen, Whittle, Taylor, Demissie, Brener, Thornton,
  Moore, Zaza, and {Centers for Disease Control and Prevention
  (CDC)}}]{Kann2014-mp}
Laura Kann, Steve Kinchen, Shari~L Shanklin, Katherine~H Flint, Joseph Kawkins,
  William~A Harris, Richard Lowry, Emily~O'malley Olsen, Tim McManus, David
  Chyen, Lisa Whittle, Eboni Taylor, Zewditu Demissie, Nancy Brener, Jemekia
  Thornton, John Moore, Stephanie Zaza, and {Centers for Disease Control and
  Prevention (CDC)}. 2014.
\newblock Youth risk behavior surveillance--united states, 2013.
\newblock \emph{MMWR Suppl}, 63(4):1--168.

\bibitem[{Karadzhov et~al.(2017)Karadzhov, Mihaylova, Kiprov, Georgiev,
  Koychev, and Nakov}]{10.1007/978-3-319-65813-1_18}
Georgi Karadzhov, Tsvetomila Mihaylova, Yasen Kiprov, Georgi Georgiev, Ivan
  Koychev, and Preslav Nakov. 2017.
\newblock The case for being average: A mediocrity approach to style masking
  and author obfuscation.
\newblock In \emph{Experimental IR Meets Multilinguality, Multimodality, and
  Interaction}, pages 173--185, Cham. Springer International Publishing.

\bibitem[{Kehr et~al.(2015)Kehr, Kowatsch, Wentzel, and
  Fleisch}]{DBLP:journals/isj/KehrKWF15}
Flavius Kehr, Tobias Kowatsch, Daniel Wentzel, and Elgar Fleisch. 2015.
\newblock \href {https://doi.org/10.1111/isj.12062} {Blissfully ignorant: the
  effects of general privacy concerns, general institutional trust, and affect
  in the privacy calculus}.
\newblock \emph{Inf. Syst. J.}, 25(6):607--635.

\bibitem[{Keller et~al.(2021)Keller, Mackensen, and
  Eger}]{keller-etal-2021-bert}
Yannik Keller, Jan Mackensen, and Steffen Eger. 2021.
\newblock \href {https://doi.org/10.18653/v1/2021.findings-acl.141}
  {{BERT}-defense: A probabilistic model based on {BERT} to combat cognitively
  inspired orthographic adversarial attacks}.
\newblock In \emph{Findings of the Association for Computational Linguistics:
  ACL-IJCNLP 2021}, pages 1616--1629, Online. Association for Computational
  Linguistics.

\bibitem[{Kelly et~al.(2006)Kelly, Kantor, Morse, Scholtz, and
  Sun}]{kelly-etal-2006-user}
Diane Kelly, Paul Kantor, Emile Morse, Jean Scholtz, and Ying Sun. 2006.
\newblock \href {https://aclanthology.org/W06-3007} {User-centered evaluation
  of interactive question answering systems}.
\newblock In \emph{Proceedings of the Interactive Question Answering Workshop
  at {HLT}-{NAACL} 2006}, pages 49--56, New York, NY, USA. Association for
  Computational Linguistics.

\bibitem[{Kim et~al.(2017)Kim, Kim, Sarikaya, and
  Fosler-Lussier}]{kim-etal-2017-cross}
Joo-Kyung Kim, Young-Bum Kim, Ruhi Sarikaya, and Eric Fosler-Lussier. 2017.
\newblock \href {https://doi.org/10.18653/v1/D17-1302} {Cross-lingual transfer
  learning for {POS} tagging without cross-lingual resources}.
\newblock In \emph{Proceedings of the 2017 Conference on Empirical Methods in
  Natural Language Processing}, pages 2832--2838, Copenhagen, Denmark.
  Association for Computational Linguistics.

\bibitem[{Kim et~al.(2021)Kim, Razi, Stringhini, Wisniewski, and
  De~Choudhury}]{10.1145/3476066}
Seunghyun Kim, Afsaneh Razi, Gianluca Stringhini, Pamela~J. Wisniewski, and
  Munmun De~Choudhury. 2021.
\newblock \href {https://doi.org/10.1145/3476066} {A human-centered systematic
  literature review of cyberbullying detection algorithms}.
\newblock \emph{Proc. ACM Hum.-Comput. Interact.}, 5(CSCW2).

\bibitem[{Kim(2014)}]{kim-2014-convolutional}
Yoon Kim. 2014.
\newblock \href {https://doi.org/10.3115/v1/D14-1181} {Convolutional neural
  networks for sentence classification}.
\newblock In \emph{Proceedings of the 2014 Conference on Empirical Methods in
  Natural Language Processing ({EMNLP})}, pages 1746--1751, Doha, Qatar.
  Association for Computational Linguistics.

\bibitem[{Kingma and Ba(2015)}]{DBLP:journals/corr/KingmaB14}
Diederik~P. Kingma and Jimmy Ba. 2015.
\newblock \href {http://arxiv.org/abs/1412.6980} {Adam: {A} method for
  stochastic optimization}.
\newblock In \emph{3rd International Conference on Learning Representations,
  {ICLR} 2015, San Diego, CA, USA, May 7-9, 2015, Conference Track
  Proceedings}.

\bibitem[{Kiritchenko and Nejadgholi(2020)}]{DBLP:journals/corr/abs-2010-14952}
Svetlana Kiritchenko and Isar Nejadgholi. 2020.
\newblock \href {http://arxiv.org/abs/2010.14952} {Towards ethics by design in
  online abusive content detection}.
\newblock \emph{CoRR}, abs/2010.14952.

\bibitem[{Kleinberg et~al.(2022)Kleinberg, Davies, and
  Mozes}]{DBLP:journals/corr/abs-2208-13081}
Bennett Kleinberg, Toby Davies, and Maximilian Mozes. 2022.
\newblock \href {https://doi.org/10.48550/arXiv.2208.13081} {Textwash -
  automated open-source text anonymisation}.
\newblock \emph{CoRR}, abs/2208.13081.

\bibitem[{Kontostathis et~al.(2010)Kontostathis, Edwards, and
  Leatherman}]{doi:https://doi.org/10.1002/9780470689646.ch8}
April Kontostathis, Lynne Edwards, and Amanda Leatherman. 2010.
\newblock \href {https://doi.org/https://doi.org/10.1002/9780470689646.ch8}
  {\emph{Text Mining and Cybercrime}}, chapter~8. John Wiley \& Sons, Ltd.

\bibitem[{Kontostathis et~al.(2013)Kontostathis, Reynolds, Garron, and
  Edwards}]{DBLP:conf/websci/KontostathisRGE13}
April Kontostathis, Kelly Reynolds, Andy Garron, and Lynne Edwards. 2013.
\newblock \href {https://doi.org/10.1145/2464464.2464499} {Detecting
  cyberbullying: query terms and techniques}.
\newblock In \emph{Web Science 2013 (co-located with ECRC), WebSci '13, Paris,
  France, May 2-4, 2013}, pages 195--204. {ACM}.

\bibitem[{Koolen and van
  Cranenburgh(2017)}]{koolen-van-cranenburgh-2017-stereotypes}
Corina Koolen and Andreas van Cranenburgh. 2017.
\newblock \href {https://doi.org/10.18653/v1/W17-1602} {These are not the
  stereotypes you are looking for: Bias and fairness in authorial gender
  attribution}.
\newblock In \emph{Proceedings of the First {ACL} Workshop on Ethics in Natural
  Language Processing}, pages 12--22, Valencia, Spain. Association for
  Computational Linguistics.

\bibitem[{Koppel et~al.(2009)Koppel, Akiva, Alshech, and
  Bar}]{DBLP:conf/isi/KoppelAAB09}
Moshe Koppel, Navot Akiva, Eli Alshech, and Kfir Bar. 2009.
\newblock \href {https://doi.org/10.1109/ISI.2009.5137294} {Automatically
  classifying documents by ideological and organizational affiliation}.
\newblock In \emph{{IEEE} International Conference on Intelligence and Security
  Informatics, {ISI} 2009, Dallas, Texas, USA, June 8-11, 2009, Proceedings},
  pages 176--178. {IEEE}.

\bibitem[{Koppel et~al.(2002)Koppel, Argamon, and
  Shimoni}]{DBLP:journals/lalc/KoppelAS02}
Moshe Koppel, Shlomo Argamon, and Anat~Rachel Shimoni. 2002.
\newblock \href {https://doi.org/10.1093/llc/17.4.401} {Automatically
  categorizing written texts by author gender}.
\newblock \emph{Lit. Linguistic Comput.}, 17(4):401--412.

\bibitem[{Koppel and Schler(2004)}]{DBLP:conf/icml/KoppelS04}
Moshe Koppel and Jonathan Schler. 2004.
\newblock \href {https://doi.org/10.1145/1015330.1015448} {Authorship
  verification as a one-class classification problem}.
\newblock In \emph{Machine Learning, Proceedings of the Twenty-first
  International Conference {(ICML} 2004), Banff, Alberta, Canada, July 4-8,
  2004}, volume~69 of \emph{{ACM} International Conference Proceeding Series}.
  {ACM}.

\bibitem[{Kowalski et~al.(2012)Kowalski, Limber, and
  Agatston}]{kowalski-2012-cyber}
Robin Kowalski, Susan Limber, and Patricia Agatston. 2012.
\newblock \href {https://doi.org/10.1002/9780470694176} {Cyberbullying:
  Bullying in the digital age}.
\newblock \emph{American Journal of Psychiatry - AMER J PSYCHIAT}, 165:296.

\bibitem[{Kumar et~al.(2020)Kumar, Choudhary, and
  Cho}]{DBLP:journals/corr/abs-2003-02245}
Varun Kumar, Ashutosh Choudhary, and Eunah Cho. 2020.
\newblock \href {http://arxiv.org/abs/2003.02245} {Data augmentation using
  pre-trained transformer models}.
\newblock \emph{CoRR}, abs/2003.02245.

\bibitem[{Kurakin et~al.(2017{\natexlab{a}})Kurakin, Goodfellow, and
  Bengio}]{DBLP:conf/iclr/KurakinGB17a}
Alexey Kurakin, Ian~J. Goodfellow, and Samy Bengio. 2017{\natexlab{a}}.
\newblock \href {https://openreview.net/forum?id=HJGU3Rodl} {Adversarial
  examples in the physical world}.
\newblock In \emph{5th International Conference on Learning Representations,
  {ICLR} 2017, Toulon, France, April 24-26, 2017, Workshop Track Proceedings}.
  OpenReview.net.

\bibitem[{Kurakin et~al.(2017{\natexlab{b}})Kurakin, Goodfellow, and
  Bengio}]{DBLP:conf/iclr/KurakinGB17}
Alexey Kurakin, Ian~J. Goodfellow, and Samy Bengio. 2017{\natexlab{b}}.
\newblock \href {https://openreview.net/forum?id=BJm4T4Kgx} {Adversarial
  machine learning at scale}.
\newblock In \emph{5th International Conference on Learning Representations,
  {ICLR} 2017, Toulon, France, April 24-26, 2017, Conference Track
  Proceedings}. OpenReview.net.

\bibitem[{Kurita et~al.(2019)Kurita, Belova, and
  Anastasopoulos}]{DBLP:journals/corr/abs-1912-06872}
Keita Kurita, Anna Belova, and Antonios Anastasopoulos. 2019.
\newblock \href {http://arxiv.org/abs/1912.06872} {Towards robust toxic content
  classification}.
\newblock \emph{CoRR}, abs/1912.06872.

\bibitem[{Kusner et~al.(2015)Kusner, Sun, Kolkin, and
  Weinberger}]{kusner2015word}
Matt Kusner, Yu~Sun, Nicholas Kolkin, and Kilian Weinberger. 2015.
\newblock From word embeddings to document distances.
\newblock In \emph{International Conference on Machine Learning}, pages
  957--966.

\bibitem[{Larochelle and Khoury(2020)}]{DBLP:conf/asunam/LarochelleK20}
Marc{-}Andr{\'{e}} Larochelle and Richard Khoury. 2020.
\newblock \href {https://doi.org/10.1109/ASONAM49781.2020.9381476}
  {Generalisation of cyberbullying detection}.
\newblock In \emph{{IEEE/ACM} International Conference on Advances in Social
  Networks Analysis and Mining, {ASONAM} 2020, The Hague, Netherlands, December
  7-10, 2020}, pages 296--300. {IEEE}.

\bibitem[{Lavie and Denkowski(2009)}]{DBLP:journals/mt/LavieD09}
Alon Lavie and Michael~J. Denkowski. 2009.
\newblock \href {https://doi.org/10.1007/s10590-009-9059-4} {The meteor metric
  for automatic evaluation of machine translation}.
\newblock \emph{Mach. Transl.}, 23(2-3):105--115.

\bibitem[{Leidner and Plachouras(2017)}]{leidner-plachouras-2017-ethical}
Jochen~L. Leidner and Vassilis Plachouras. 2017.
\newblock \href {https://doi.org/10.18653/v1/W17-1604} {Ethical by design:
  Ethics best practices for natural language processing}.
\newblock In \emph{Proceedings of the First {ACL} Workshop on Ethics in Natural
  Language Processing}, pages 30--40, Valencia, Spain. Association for
  Computational Linguistics.

\bibitem[{Lemley et~al.(2017)Lemley, Bazrafkan, and
  Corcoran}]{DBLP:journals/cem/LemleyBC17}
Joe Lemley, Shabab Bazrafkan, and Peter Corcoran. 2017.
\newblock \href {https://doi.org/10.1109/MCE.2016.2640698} {Deep learning for
  consumer devices and services: Pushing the limits for machine learning,
  artificial intelligence, and computer vision}.
\newblock \emph{{IEEE} Consumer Electron. Mag.}, 6(2):48--56.

\bibitem[{Lenhart et~al.(2011)Lenhart, Madden, Smith, Purcell, Zickuhr, and
  Rainie}]{Lenhart2011}
Amanda Lenhart, Mary Madden, Aaron Smith, Kristen Purcell, Kathryn Zickuhr, and
  Lee Rainie. 2011.
\newblock \href
  {https://www.pewresearch.org/internet/wp-content/uploads/sites/9/media/Files/Reports/2011/PIP_Teens_Kindness_Cruelty_SNS_Report_Nov_2011_FINAL_110711.pdf}
  {Teens, kindness and cruelty on social network sites: How american teens
  navigate the new world of}.
\newblock \emph{Pew Research Center}.

\bibitem[{Lewis et~al.(2020)Lewis, Liu, Goyal, Ghazvininejad, Mohamed, Levy,
  Stoyanov, and Zettlemoyer}]{lewis-etal-2020-bart}
Mike Lewis, Yinhan Liu, Naman Goyal, Marjan Ghazvininejad, Abdelrahman Mohamed,
  Omer Levy, Veselin Stoyanov, and Luke Zettlemoyer. 2020.
\newblock \href {https://doi.org/10.18653/v1/2020.acl-main.703} {{BART}:
  Denoising sequence-to-sequence pre-training for natural language generation,
  translation, and comprehension}.
\newblock In \emph{Proceedings of the 58th Annual Meeting of the Association
  for Computational Linguistics}, pages 7871--7880, Online. Association for
  Computational Linguistics.

\bibitem[{Li et~al.(2020)Li, Shengshuo, Liu, Wu, Zhou, and
  Steinert{-}Threlkeld}]{DBLP:conf/blackboxnlp/LiSLWZS20}
Chuanrong Li, Lin Shengshuo, Zeyu Liu, Xinyi Wu, Xuhui Zhou, and Shane
  Steinert{-}Threlkeld. 2020.
\newblock \href {https://doi.org/10.18653/v1/2020.blackboxnlp-1.12}
  {Linguistically-informed transformations {(LIT):} {A} method for
  automatically generating contrast sets}.
\newblock In \emph{Proceedings of the Third BlackboxNLP Workshop on Analyzing
  and Interpreting Neural Networks for NLP, BlackboxNLP@EMNLP 2020, Online,
  November 2020}, pages 126--135. Association for Computational Linguistics.

\bibitem[{Li et~al.(2019)Li, Ji, Du, Li, and Wang}]{DBLP:conf/ndss/LiJDLW19}
Jinfeng Li, Shouling Ji, Tianyu Du, Bo~Li, and Ting Wang. 2019.
\newblock \href
  {https://www.ndss-symposium.org/ndss-paper/textbugger-generating-adversarial-text-against-real-world-applications/}
  {Textbugger: Generating adversarial text against real-world applications}.
\newblock In \emph{26th Annual Network and Distributed System Security
  Symposium, {NDSS} 2019, San Diego, California, USA, February 24-27, 2019}.
  The Internet Society.

\bibitem[{Li et~al.(2015)Li, Luong, and Jurafsky}]{DBLP:conf/acl/LiLJ15}
Jiwei Li, Minh{-}Thang Luong, and Dan Jurafsky. 2015.
\newblock \href {https://doi.org/10.3115/v1/p15-1107} {A hierarchical neural
  autoencoder for paragraphs and documents}.
\newblock In \emph{Proceedings of the 53rd Annual Meeting of the Association
  for Computational Linguistics and the 7th International Joint Conference on
  Natural Language Processing of the Asian Federation of Natural Language
  Processing, {ACL} 2015, July 26-31, 2015, Beijing, China, Volume 1: Long
  Papers}, pages 1106--1115. The Association for Computer Linguistics.

\bibitem[{Li et~al.(2016)Li, Monroe, and Jurafsky}]{DBLP:journals/corr/LiMJ16a}
Jiwei Li, Will Monroe, and Dan Jurafsky. 2016.
\newblock \href {http://arxiv.org/abs/1612.08220} {Understanding neural
  networks through representation erasure}.
\newblock \emph{CoRR}, abs/1612.08220.

\bibitem[{Li et~al.(2012)Li, Ju, Zhou, and Li}]{li-etal-2012-active-learning}
Shoushan Li, Shengfeng Ju, Guodong Zhou, and Xiaojun Li. 2012.
\newblock \href {https://aclanthology.org/D12-1013} {Active learning for
  imbalanced sentiment classification}.
\newblock In \emph{Proceedings of the 2012 Joint Conference on Empirical
  Methods in Natural Language Processing and Computational Natural Language
  Learning}, pages 139--148, Jeju Island, Korea. Association for Computational
  Linguistics.

\bibitem[{Li et~al.(2018)Li, Baldwin, and Cohn}]{li-etal-2018-towards}
Yitong Li, Timothy Baldwin, and Trevor Cohn. 2018.
\newblock \href {https://doi.org/10.18653/v1/P18-2005} {Towards robust and
  privacy-preserving text representations}.
\newblock In \emph{Proceedings of the 56th Annual Meeting of the Association
  for Computational Linguistics (Volume 2: Short Papers)}, pages 25--30,
  Melbourne, Australia. Association for Computational Linguistics.

\bibitem[{Liang et~al.(2018{\natexlab{a}})Liang, Li, Su, Bian, Li, and
  Shi}]{DBLP:conf/ijcai/0002LSBLS18}
Bin Liang, Hongcheng Li, Miaoqiang Su, Pan Bian, Xirong Li, and Wenchang Shi.
  2018{\natexlab{a}}.
\newblock \href {https://doi.org/10.24963/ijcai.2018/585} {Deep text
  classification can be fooled}.
\newblock In \emph{Proceedings of the Twenty-Seventh International Joint
  Conference on Artificial Intelligence, {IJCAI} 2018, July 13-19, 2018,
  Stockholm, Sweden}, pages 4208--4215. ijcai.org.

\bibitem[{Liang et~al.(2018{\natexlab{b}})Liang, Cai, Yu, Han, and
  Li}]{DBLP:journals/network/LiangCYHL18}
Yi~Liang, Zhipeng Cai, Jiguo Yu, Qilong Han, and Yingshu Li.
  2018{\natexlab{b}}.
\newblock \href {https://doi.org/10.1109/MNET.2018.1700349} {Deep learning
  based inference of private information using embedded sensors in smart
  devices}.
\newblock \emph{{IEEE} Netw.}, 32(4):8--14.

\bibitem[{Lipton and Steinhardt(2019)}]{DBLP:journals/cacm/LiptonS19}
Zachary~C. Lipton and Jacob Steinhardt. 2019.
\newblock \href {https://doi.org/10.1145/3316774} {Research for practice:
  troubling trends in machine-learning scholarship}.
\newblock \emph{Commun. {ACM}}, 62(6):45--53.

\bibitem[{Liu et~al.(2017)Liu, Qiu, and Huang}]{liu-etal-2017-adversarial}
Pengfei Liu, Xipeng Qiu, and Xuanjing Huang. 2017.
\newblock \href {https://doi.org/10.18653/v1/P17-1001} {Adversarial multi-task
  learning for text classification}.
\newblock In \emph{Proceedings of the 55th Annual Meeting of the Association
  for Computational Linguistics (Volume 1: Long Papers)}, pages 1--10,
  Vancouver, Canada. Association for Computational Linguistics.

\bibitem[{Liu et~al.(2022)Liu, Lu, Chen, and
  Tang}]{DBLP:journals/taslp/LiuLCT22}
Shengcai Liu, Ning Lu, Cheng Chen, and Ke~Tang. 2022.
\newblock \href {https://doi.org/10.1109/TASLP.2021.3130970} {Efficient
  combinatorial optimization for word-level adversarial textual attack}.
\newblock \emph{{IEEE} {ACM} Trans. Audio Speech Lang. Process.}, 30:98--111.

\bibitem[{Livingstone et~al.(2011)Livingstone, Haddon, G{\"{o}}rzig, and
  {\'{O}}lafsson}]{Livingstone2011}
Sonia Livingstone, Leslie Haddon, Anke G{\"{o}}rzig, and Kjartan
  {\'{O}}lafsson. 2011.
\newblock \href {https://doi.org/293} {{EU Kids Online II: Final Report 2011}}.
\newblock Technical report, EU Kids Online.

\bibitem[{Lowe(1999)}]{DBLP:conf/iccv/Lowe99}
David~G. Lowe. 1999.
\newblock \href {https://doi.org/10.1109/ICCV.1999.790410} {Object recognition
  from local scale-invariant features}.
\newblock In \emph{Proceedings of the International Conference on Computer
  Vision, Kerkyra, Corfu, Greece, September 20-25, 1999}, pages 1150--1157.
  {IEEE} Computer Society.

\bibitem[{Lutz and Hoffmann(2017)}]{doi:10.1080/1369118X.2017.1293129}
Christoph Lutz and Christian~Pieter Hoffmann. 2017.
\newblock \href {https://doi.org/10.1080/1369118X.2017.1293129} {The dark side
  of online participation: exploring non-, passive and negative participation}.
\newblock \emph{Information, Communication \& Society}, 20(6):876--897.

\bibitem[{Lutz et~al.(2020)Lutz, Hoffmann, and
  Ranzini}]{DBLP:journals/nms/LutzHR20}
Christoph Lutz, Christian~Pieter Hoffmann, and Giulia Ranzini. 2020.
\newblock \href {https://doi.org/10.1177/1461444820912544} {Data capitalism and
  the user: An exploration of privacy cynicism in germany}.
\newblock \emph{New Media Soc.}, 22(7).

\bibitem[{Lyu et~al.(2020)Lyu, He, and Li}]{lyu-etal-2020-differentially}
Lingjuan Lyu, Xuanli He, and Yitong Li. 2020.
\newblock \href {https://doi.org/10.18653/v1/2020.findings-emnlp.213}
  {Differentially private representation for {NLP}: Formal guarantee and an
  empirical study on privacy and fairness}.
\newblock In \emph{Findings of the Association for Computational Linguistics:
  EMNLP 2020}, pages 2355--2365, Online. Association for Computational
  Linguistics.

\bibitem[{Ma(2014)}]{ma-2014-automatic}
Jianqiang Ma. 2014.
\newblock \href
  {http://www.lrec-conf.org/proceedings/lrec2014/pdf/1158_Paper.pdf} {Automatic
  refinement of syntactic categories in {C}hinese word structures}.
\newblock In \emph{Proceedings of the Ninth International Conference on
  Language Resources and Evaluation ({LREC}'14)}, pages 4087--4092, Reykjavik,
  Iceland. European Language Resources Association (ELRA).

\bibitem[{Madaan et~al.(2020)Madaan, Setlur, Parekh, Poczos, Neubig, Yang,
  Salakhutdinov, Black, and Prabhumoye}]{madaan-etal-2020-politeness}
Aman Madaan, Amrith Setlur, Tanmay Parekh, Barnabas Poczos, Graham Neubig,
  Yiming Yang, Ruslan Salakhutdinov, Alan~W Black, and Shrimai Prabhumoye.
  2020.
\newblock \href {https://doi.org/10.18653/v1/2020.acl-main.169} {Politeness
  transfer: A tag and generate approach}.
\newblock In \emph{Proceedings of the 58th Annual Meeting of the Association
  for Computational Linguistics}, pages 1869--1881, Online. Association for
  Computational Linguistics.

\bibitem[{Madden et~al.(2013)Madden, Lenhart, and Cortesi}]{Madden2013}
Mary Madden, Amanda Lenhart, and Sandra Cortesi. 2013.
\newblock \href {http://www.lateledipenelope.it/public/52dff2e35b812.pdf}
  {{Teens, social media, and privacy}}.
\newblock Technical report, Pew Internet.

\bibitem[{Madnani and Loukina(2020)}]{madnani-loukina-2020-user}
Nitin Madnani and Anastassia Loukina. 2020.
\newblock \href {https://doi.org/10.18653/v1/2020.nlposs-1.20} {User-centered
  {\&} robust {NLP} {OSS}: Lessons learned from developing {\&} maintaining
  {RSMT}ool}.
\newblock In \emph{Proceedings of Second Workshop for NLP Open Source Software
  (NLP-OSS)}, pages 141--146, Online. Association for Computational
  Linguistics.

\bibitem[{Madry et~al.(2018)Madry, Makelov, Schmidt, Tsipras, and
  Vladu}]{DBLP:conf/iclr/MadryMSTV18}
Aleksander Madry, Aleksandar Makelov, Ludwig Schmidt, Dimitris Tsipras, and
  Adrian Vladu. 2018.
\newblock \href {https://openreview.net/forum?id=rJzIBfZAb} {Towards deep
  learning models resistant to adversarial attacks}.
\newblock In \emph{6th International Conference on Learning Representations,
  {ICLR} 2018, Vancouver, BC, Canada, April 30 - May 3, 2018, Conference Track
  Proceedings}. OpenReview.net.

\bibitem[{Madukwe et~al.(2020)Madukwe, Gao, and Xue}]{madukwe-etal-2020-data}
Kosisochukwu Madukwe, Xiaoying Gao, and Bing Xue. 2020.
\newblock \href {https://doi.org/10.18653/v1/2020.alw-1.18} {In data we trust:
  A critical analysis of hate speech detection datasets}.
\newblock In \emph{Proceedings of the Fourth Workshop on Online Abuse and
  Harms}, pages 150--161, Online. Association for Computational Linguistics.

\bibitem[{Mahmood et~al.(2020)Mahmood, Shafiq, and
  Srinivasan}]{mahmood-etal-2020-girl}
Asad Mahmood, Zubair Shafiq, and Padmini Srinivasan. 2020.
\newblock \href {https://doi.org/10.18653/v1/2020.acl-main.203} {A girl has a
  name: Detecting authorship obfuscation}.
\newblock In \emph{Proceedings of the 58th Annual Meeting of the Association
  for Computational Linguistics}, pages 2235--2245, Online. Association for
  Computational Linguistics.

\bibitem[{Manheim and Kaplan(2019)}]{manheim2019artificial}
Karl~M. Manheim and Lyric Kaplan. 2019.
\newblock \href {https://ssrn.com/abstract=3273016} {Artificial intelligence:
  Risks to privacy and democracy}.
\newblock \emph{Yale Journal of Law and Technology 106}.

\bibitem[{Mansoorizadeh et~al.(2016)Mansoorizadeh, Rahgooy, Aminian, and
  Eskandari}]{DBLP:conf/clef/MansoorizadehRA16}
Muharram Mansoorizadeh, Taher Rahgooy, Mohammad Aminian, and Mehdy Eskandari.
  2016.
\newblock \href {http://ceur-ws.org/Vol-1609/16090939.pdf} {Author obfuscation
  using wordnet and language models}.
\newblock In \emph{Working Notes of {CLEF} 2016 - Conference and Labs of the
  Evaluation forum, {\'{E}}vora, Portugal, 5-8 September, 2016}, volume 1609 of
  \emph{{CEUR} Workshop Proceedings}, pages 939--946. CEUR-WS.org.

\bibitem[{Markov et~al.(2021)Markov, Ljube{\v{s}}i{\'c}, Fi{\v{s}}er, and
  Daelemans}]{markov-etal-2021-exploring}
Ilia Markov, Nikola Ljube{\v{s}}i{\'c}, Darja Fi{\v{s}}er, and Walter
  Daelemans. 2021.
\newblock \href {https://aclanthology.org/2021.wassa-1.16} {Exploring
  stylometric and emotion-based features for multilingual cross-domain hate
  speech detection}.
\newblock In \emph{Proceedings of the Eleventh Workshop on Computational
  Approaches to Subjectivity, Sentiment and Social Media Analysis}, pages
  149--159, Online. Association for Computational Linguistics.

\bibitem[{Mason(2008)}]{https://doi.org/10.1002/pits.20301}
Kimberly~L. Mason. 2008.
\newblock \href {https://doi.org/https://doi.org/10.1002/pits.20301}
  {Cyberbullying: A preliminary assessment for school personnel}.
\newblock \emph{Psychology in the Schools}, 45(4):323--348.

\bibitem[{Mateo and Yus(2013)}]{jbp:/content/journals/10.1075/jlac.1.1.05mat}
José Mateo and Francisco~Ramos Yus. 2013.
\newblock \href {https://doi.org/https://doi.org/10.1075/jlac.1.1.05mat}
  {Towards a cross-cultural pragmatic taxonomy of insults}.
\newblock \emph{Journal of Language Aggression and Conflict}, 1(1):87--114.

\bibitem[{Mathai et~al.(2020)Mathai, Khare, Tamilselvam, and
  Mani}]{DBLP:journals/corr/abs-2011-03901}
Alex Mathai, Shreya Khare, Srikanth Tamilselvam, and Senthil Mani. 2020.
\newblock \href {http://arxiv.org/abs/2011.03901} {Adversarial black-box
  attacks on text classifiers using multi-objective genetic optimization guided
  by deep networks}.
\newblock \emph{CoRR}, abs/2011.03901.

\bibitem[{Matthews and Merriam(1993)}]{10.1093/llc/8.4.203}
Robert A.~J. Matthews and Thomas V.~N. Merriam. 1993.
\newblock \href {https://doi.org/10.1093/llc/8.4.203} {{Neural Computation in
  Stylometry I: An Application to the Works of Shakespeare and Fletcher}}.
\newblock \emph{Literary and Linguistic Computing}, 8(4):203--209.

\bibitem[{McClosky et~al.(2006)McClosky, Charniak, and
  Johnson}]{mcclosky-etal-2006-reranking}
David McClosky, Eugene Charniak, and Mark Johnson. 2006.
\newblock \href {https://doi.org/10.3115/1220175.1220218} {Reranking and
  self-training for parser adaptation}.
\newblock In \emph{Proceedings of the 21st International Conference on
  Computational Linguistics and 44th Annual Meeting of the Association for
  Computational Linguistics}, pages 337--344, Sydney, Australia. Association
  for Computational Linguistics.

\bibitem[{Melis et~al.(2018)Melis, Dyer, and
  Blunsom}]{DBLP:conf/iclr/MelisDB18}
G{\'{a}}bor Melis, Chris Dyer, and Phil Blunsom. 2018.
\newblock \href {https://openreview.net/forum?id=ByJHuTgA-} {On the state of
  the art of evaluation in neural language models}.
\newblock In \emph{6th International Conference on Learning Representations,
  {ICLR} 2018, Vancouver, BC, Canada, April 30 - May 3, 2018, Conference Track
  Proceedings}. OpenReview.net.

\bibitem[{Merriam and Matthews(1994)}]{10.1093/llc/9.1.1}
Thomas V.~N. Merriam and Robert A.~J. Matthews. 1994.
\newblock \href {https://doi.org/10.1093/llc/9.1.1} {{Neural Computation in
  Stylometry II: An Application to the Works of Shakespeare and Marlowe}}.
\newblock \emph{Literary and Linguistic Computing}, 9(1):1--6.

\bibitem[{Mikolov et~al.(2013{\natexlab{a}})Mikolov, Chen, Corrado, and
  Dean}]{DBLP:journals/corr/abs-1301-3781}
Tom{\'{a}}s Mikolov, Kai Chen, Greg Corrado, and Jeffrey Dean.
  2013{\natexlab{a}}.
\newblock \href {http://arxiv.org/abs/1301.3781} {Efficient estimation of word
  representations in vector space}.
\newblock In \emph{1st International Conference on Learning Representations,
  {ICLR} 2013, Scottsdale, Arizona, USA, May 2-4, 2013, Workshop Track
  Proceedings}.

\bibitem[{Mikolov et~al.(2013{\natexlab{b}})Mikolov, Sutskever, Chen, Corrado,
  and Dean}]{DBLP:conf/nips/MikolovSCCD13}
Tom{\'{a}}s Mikolov, Ilya Sutskever, Kai Chen, Gregory~S. Corrado, and Jeffrey
  Dean. 2013{\natexlab{b}}.
\newblock \href
  {https://proceedings.neurips.cc/paper/2013/hash/9aa42b31882ec039965f3c4923ce901b-Abstract.html}
  {Distributed representations of words and phrases and their
  compositionality}.
\newblock In \emph{Advances in Neural Information Processing Systems 26: 27th
  Annual Conference on Neural Information Processing Systems 2013. Proceedings
  of a meeting held December 5-8, 2013, Lake Tahoe, Nevada, United States},
  pages 3111--3119.

\bibitem[{Miller(1995)}]{DBLP:journals/cacm/Miller95}
George~A. Miller. 1995.
\newblock \href {https://doi.org/10.1145/219717.219748} {Wordnet: {A} lexical
  database for english}.
\newblock \emph{Commun. {ACM}}, 38(11):39--41.

\bibitem[{Millwood-Hargreave(2000)}]{Millwood2000}
Andrea Millwood-Hargreave. 2000.
\newblock Delete expletives? research undertaken jointly by the advertising
  standards authority, british broadcasting corporation, broadcasting standards
  commission and the independent television commission.
\newblock \emph{ASA, BBC, BSC and ITC, London}.

\bibitem[{Mishra et~al.(2019)Mishra, Yannakoudakis, and
  Shutova}]{DBLP:journals/corr/abs-1908-06024}
Pushkar Mishra, Helen Yannakoudakis, and Ekaterina Shutova. 2019.
\newblock \href {http://arxiv.org/abs/1908.06024} {Tackling online abuse: {A}
  survey of automated abuse detection methods}.
\newblock \emph{CoRR}, abs/1908.06024.

\bibitem[{Mislove et~al.(2010)Mislove, Viswanath, Gummadi, and
  Druschel}]{DBLP:conf/wsdm/MisloveVGD10}
Alan Mislove, Bimal Viswanath, P.~Krishna Gummadi, and Peter Druschel. 2010.
\newblock \href {https://doi.org/10.1145/1718487.1718519} {You are who you
  know: inferring user profiles in online social networks}.
\newblock In \emph{Proceedings of the Third International Conference on Web
  Search and Web Data Mining, {WSDM} 2010, New York, NY, USA, February 4-6,
  2010}, pages 251--260. {ACM}.

\bibitem[{Mitchell et~al.(2019)Mitchell, Wu, Zaldivar, Barnes, Vasserman,
  Hutchinson, Spitzer, Raji, and Gebru}]{DBLP:conf/fat/MitchellWZBVHSR19}
Margaret Mitchell, Simone Wu, Andrew Zaldivar, Parker Barnes, Lucy Vasserman,
  Ben Hutchinson, Elena Spitzer, Inioluwa~Deborah Raji, and Timnit Gebru. 2019.
\newblock \href {https://doi.org/10.1145/3287560.3287596} {Model cards for
  model reporting}.
\newblock In \emph{Proceedings of the Conference on Fairness, Accountability,
  and Transparency, FAT* 2019, Atlanta, GA, USA, January 29-31, 2019}, pages
  220--229. {ACM}.

\bibitem[{Mladenovic et~al.(2021)Mladenovic, Osmjanski, and {Vujicic
  Stankovic}}]{DBLP:journals/csur/MladenovicOS21}
Miljana Mladenovic, Vera Osmjanski, and Stasa {Vujicic Stankovic}. 2021.
\newblock \href {https://doi.org/10.1145/3424246} {Cyber-aggression,
  cyberbullying, and cyber-grooming: {A} survey and research challenges}.
\newblock \emph{{ACM} Comput. Surv.}, 54(1):1:1--1:42.

\bibitem[{Morris et~al.(2020)Morris, Lifland, Yoo, Grigsby, Jin, and
  Qi}]{DBLP:conf/emnlp/MorrisLYGJQ20}
John~X. Morris, Eli Lifland, Jin~Yong Yoo, Jake Grigsby, Di~Jin, and Yanjun Qi.
  2020.
\newblock \href {https://doi.org/10.18653/v1/2020.emnlp-demos.16} {Textattack:
  {A} framework for adversarial attacks, data augmentation, and adversarial
  training in {NLP}}.
\newblock In \emph{Proceedings of the 2020 Conference on Empirical Methods in
  Natural Language Processing: System Demonstrations, {EMNLP} 2020 - Demos,
  Online, November 16-20, 2020}, pages 119--126. Association for Computational
  Linguistics.

\bibitem[{Morrison(2009)}]{DBLP:journals/nms/Morrison09}
Aim{\'{e}}e~Hope Morrison. 2009.
\newblock \href {https://doi.org/10.1177/1461444808100161} {An impossible
  future: John perry barlow's 'declaration of the independence of cyberspace'}.
\newblock \emph{New Media Soc.}, 11(1-2):53--71.

\bibitem[{Morton(1965)}]{10.2307/2344178}
A.~Q. Morton. 1965.
\newblock \href {http://www.jstor.org/stable/2344178} {The authorship of greek
  prose}.
\newblock \emph{Journal of the Royal Statistical Society. Series A (General)},
  128(2):169--233.

\bibitem[{Mosteller and Wallace(1963)}]{doi:10.1080/01621459.1963.10500849}
Frederick Mosteller and David~L. Wallace. 1963.
\newblock \href {https://doi.org/10.1080/01621459.1963.10500849} {Inference in
  an authorship problem}.
\newblock \emph{Journal of the American Statistical Association},
  58(302):275--309.

\bibitem[{Moura et~al.(2020)Moura, Castro, Hardaker, Wullink, and
  Hesselman}]{DBLP:conf/imc/MouraCHWH20}
Giovane C.~M. Moura, Sebastian Castro, Wes Hardaker, Maarten Wullink, and
  Cristian Hesselman. 2020.
\newblock \href {https://doi.org/10.1145/3419394.3423625} {Clouding up the
  internet: how centralized is {DNS} traffic becoming?}
\newblock In \emph{{IMC} '20: {ACM} Internet Measurement Conference, Virtual
  Event, USA, October 27-29, 2020}, pages 42--49. {ACM}.

\bibitem[{Mozes et~al.(2021{\natexlab{a}})Mozes, Bartolo, Stenetorp, Kleinberg,
  and Griffin}]{mozes-etal-2021-contrasting}
Maximilian Mozes, Max Bartolo, Pontus Stenetorp, Bennett Kleinberg, and Lewis
  Griffin. 2021{\natexlab{a}}.
\newblock \href {https://doi.org/10.18653/v1/2021.emnlp-main.651} {Contrasting
  human- and machine-generated word-level adversarial examples for text
  classification}.
\newblock In \emph{Proceedings of the 2021 Conference on Empirical Methods in
  Natural Language Processing}, pages 8258--8270, Online and Punta Cana,
  Dominican Republic. Association for Computational Linguistics.

\bibitem[{Mozes et~al.(2021{\natexlab{b}})Mozes, Stenetorp, Kleinberg, and
  Griffin}]{mozes-etal-2021-frequency}
Maximilian Mozes, Pontus Stenetorp, Bennett Kleinberg, and Lewis Griffin.
  2021{\natexlab{b}}.
\newblock \href {https://aclanthology.org/2021.eacl-main.13} {Frequency-guided
  word substitutions for detecting textual adversarial examples}.
\newblock In \emph{Proceedings of the 16th Conference of the European Chapter
  of the Association for Computational Linguistics: Main Volume}, pages
  171--186, Online. Association for Computational Linguistics.

\bibitem[{Mrk{\v{s}}i{\'c} et~al.(2016)Mrk{\v{s}}i{\'c}, {\'O}~S{\'e}aghdha,
  Thomson, Ga{\v{s}}i{\'c}, Rojas-Barahona, Su, Vandyke, Wen, and
  Young}]{mrksic-etal-2016-counter}
Nikola Mrk{\v{s}}i{\'c}, Diarmuid {\'O}~S{\'e}aghdha, Blaise Thomson, Milica
  Ga{\v{s}}i{\'c}, Lina~M. Rojas-Barahona, Pei-Hao Su, David Vandyke,
  Tsung-Hsien Wen, and Steve Young. 2016.
\newblock \href {https://doi.org/10.18653/v1/N16-1018} {Counter-fitting word
  vectors to linguistic constraints}.
\newblock In \emph{Proceedings of the 2016 Conference of the North {A}merican
  Chapter of the Association for Computational Linguistics: Human Language
  Technologies}, pages 142--148, San Diego, California. Association for
  Computational Linguistics.

\bibitem[{Munaf{\`o} et~al.(2017)Munaf{\`o}, Nosek, Bishop, Button, Chambers,
  Percie~du Sert, Simonsohn, Wagenmakers, Ware, and Ioannidis}]{Munafo2017}
Marcus~R. Munaf{\`o}, Brian~A. Nosek, Dorothy V.~M. Bishop, Katherine~S.
  Button, Christopher~D. Chambers, Nathalie Percie~du Sert, Uri Simonsohn,
  Eric-Jan Wagenmakers, Jennifer~J. Ware, and John P.~A. Ioannidis. 2017.
\newblock \href {https://doi.org/10.1038/s41562-016-0021} {A manifesto for
  reproducible science}.
\newblock \emph{Nature Human Behaviour}, 1(1):0021.

\bibitem[{Muneer and Fati(2020)}]{DBLP:journals/fi/MuneerF20}
Amgad Muneer and Suliman~Mohamed Fati. 2020.
\newblock \href {https://doi.org/10.3390/fi12110187} {A comparative analysis of
  machine learning techniques for cyberbullying detection on twitter}.
\newblock \emph{Future Internet}, 12(11):187.

\bibitem[{Murray and Durrell(1999)}]{murray1999inferring}
Dan Murray and Kevan Durrell. 1999.
\newblock Inferring demographic attributes of anonymous internet users.
\newblock In \emph{International Workshop on Web Usage Analysis and User
  Profiling}, pages 7--20. Springer.

\bibitem[{Nahar et~al.(2013)Nahar, Li, and Pang}]{Nahar2013}
Vinita Nahar, Xue Li, and Chaoyi Pang. 2013.
\newblock \href
  {http://ijircce.com/admin/main/storage/app/pdf/leAaQQOYNEXmpP0qNswfGGUsBmYthZaZX5eFQxpg.pdf}
  {{An Effective Approach for Cyberbullying Detection}}.
\newblock \emph{Communications in Information Science and Management
  Engineering}, 3(May):238--247.

\bibitem[{Nakov et~al.(2021)Nakov, Nayak, Dent, Bhatawdekar, Sarwar, Hardalov,
  Dinkov, Zlatkova, Bouchard, and
  Augenstein}]{DBLP:journals/corr/abs-2103-00153}
Preslav Nakov, Vibha Nayak, Kyle Dent, Ameya Bhatawdekar, Sheikh~Muhammad
  Sarwar, Momchil Hardalov, Yoan Dinkov, Dimitrina Zlatkova, Guillaume
  Bouchard, and Isabelle Augenstein. 2021.
\newblock \href {http://arxiv.org/abs/2103.00153} {Detecting abusive language
  on online platforms: {A} critical analysis}.
\newblock \emph{CoRR}, abs/2103.00153.

\bibitem[{Nam and Kim(2018)}]{DBLP:conf/nips/NamK18}
Hyeonseob Nam and Hyo{-}Eun Kim. 2018.
\newblock \href
  {https://proceedings.neurips.cc/paper/2018/hash/018b59ce1fd616d874afad0f44ba338d-Abstract.html}
  {Batch-instance normalization for adaptively style-invariant neural
  networks}.
\newblock In \emph{Advances in Neural Information Processing Systems 31: Annual
  Conference on Neural Information Processing Systems 2018, NeurIPS 2018,
  December 3-8, 2018, Montr{\'{e}}al, Canada}, pages 2563--2572.

\bibitem[{Neal et~al.(2018)Neal, Sundararajan, Fatima, Yan, Xiang, and
  Woodard}]{DBLP:journals/csur/NealSFYXW17}
Tempestt~J. Neal, Kalaivani Sundararajan, Aneez Fatima, Yiming Yan, Yingfei
  Xiang, and Damon~L. Woodard. 2018.
\newblock \href {https://doi.org/10.1145/3132039} {Surveying stylometry
  techniques and applications}.
\newblock \emph{{ACM} Comput. Surv.}, 50(6):86:1--86:36.

\bibitem[{Nejadgholi and Kiritchenko(2020)}]{nejadgholi-kiritchenko-2020-cross}
Isar Nejadgholi and Svetlana Kiritchenko. 2020.
\newblock \href {https://doi.org/10.18653/v1/2020.alw-1.20} {On cross-dataset
  generalization in automatic detection of online abuse}.
\newblock In \emph{Proceedings of the Fourth Workshop on Online Abuse and
  Harms}, pages 173--183, Online. Association for Computational Linguistics.

\bibitem[{Nitta et~al.(2013)Nitta, Masui, Ptaszynski, Kimura, Rzepka, and
  Araki}]{nitta-etal-2013-detecting}
Taisei Nitta, Fumito Masui, Michal Ptaszynski, Yasutomo Kimura, Rafal Rzepka,
  and Kenji Araki. 2013.
\newblock \href {https://aclanthology.org/I13-1066} {Detecting cyberbullying
  entries on informal school websites based on category relevance
  maximization}.
\newblock In \emph{Proceedings of the Sixth International Joint Conference on
  Natural Language Processing}, pages 579--586, Nagoya, Japan. Asian Federation
  of Natural Language Processing.

\bibitem[{{Noswearing.com}(2016)}]{noswear}
{Noswearing.com}. 2016.
\newblock {Bad Word List \& Swear Filter}.
\newblock \url{http://www.noswearing.com}.
\newblock Last accessed 15/03/2016.

\bibitem[{Nothdurft et~al.(2015)Nothdurft, Behnke, Bercher, Biundo, and
  Minker}]{nothdurft-etal-2015-interplay}
Florian Nothdurft, Gregor Behnke, Pascal Bercher, Susanne Biundo, and Wolfgang
  Minker. 2015.
\newblock \href {https://doi.org/10.18653/v1/W15-4646} {The interplay of
  user-centered dialog systems and {AI} planning}.
\newblock In \emph{Proceedings of the 16th Annual Meeting of the Special
  Interest Group on Discourse and Dialogue}, pages 344--353, Prague, Czech
  Republic. Association for Computational Linguistics.

\bibitem[{Novikova et~al.(2017)Novikova, Du{\v{s}}ek, Cercas~Curry, and
  Rieser}]{novikova-etal-2017-need}
Jekaterina Novikova, Ond{\v{r}}ej Du{\v{s}}ek, Amanda Cercas~Curry, and Verena
  Rieser. 2017.
\newblock \href {https://doi.org/10.18653/v1/D17-1238} {Why we need new
  evaluation metrics for {NLG}}.
\newblock In \emph{Proceedings of the 2017 Conference on Empirical Methods in
  Natural Language Processing}, pages 2241--2252, Copenhagen, Denmark.
  Association for Computational Linguistics.

\bibitem[{Ott et~al.(2011)Ott, Choi, Cardie, and
  Hancock}]{ott-etal-2011-finding}
Myle Ott, Yejin Choi, Claire Cardie, and Jeffrey~T. Hancock. 2011.
\newblock \href {https://aclanthology.org/P11-1032} {Finding deceptive opinion
  spam by any stretch of the imagination}.
\newblock In \emph{Proceedings of the 49th Annual Meeting of the Association
  for Computational Linguistics: Human Language Technologies}, pages 309--319,
  Portland, Oregon, USA. Association for Computational Linguistics.

\bibitem[{Otterbacher(2010)}]{DBLP:conf/cikm/Otterbacher10}
Jahna Otterbacher. 2010.
\newblock \href {https://doi.org/10.1145/1871437.1871487} {Inferring gender of
  movie reviewers: exploiting writing style, content and metadata}.
\newblock In \emph{Proceedings of the 19th {ACM} Conference on Information and
  Knowledge Management, {CIKM} 2010, Toronto, Ontario, Canada, October 26-30,
  2010}, pages 369--378. {ACM}.

\bibitem[{Ousidhoum et~al.(2021)Ousidhoum, Zhao, Fang, Song, and
  Yeung}]{DBLP:conf/acl/OusidhoumZFSY20}
Nedjma Ousidhoum, Xinran Zhao, Tianqing Fang, Yangqiu Song, and Dit{-}Yan
  Yeung. 2021.
\newblock \href {https://doi.org/10.18653/v1/2021.acl-long.329} {Probing toxic
  content in large pre-trained language models}.
\newblock In \emph{Proceedings of the 59th Annual Meeting of the Association
  for Computational Linguistics and the 11th International Joint Conference on
  Natural Language Processing, {ACL/IJCNLP} 2021, (Volume 1: Long Papers),
  Virtual Event, August 1-6, 2021}, pages 4262--4274. Association for
  Computational Linguistics.

\bibitem[{Pan et~al.(2021)Pan, Hang, Sil, Potdar, and
  Yu}]{DBLP:journals/corr/abs-2107-10137}
Lin Pan, Chung{-}Wei Hang, Avirup Sil, Saloni Potdar, and Mo~Yu. 2021.
\newblock \href {http://arxiv.org/abs/2107.10137} {Improved text classification
  via contrastive adversarial training}.
\newblock \emph{CoRR}, abs/2107.10137.

\bibitem[{Pan et~al.(2010)Pan, Ni, Sun, Yang, and
  Chen}]{DBLP:conf/www/PanNSYC10}
Sinno~Jialin Pan, Xiaochuan Ni, Jian{-}Tao Sun, Qiang Yang, and Zheng Chen.
  2010.
\newblock \href {https://doi.org/10.1145/1772690.1772767} {Cross-domain
  sentiment classification via spectral feature alignment}.
\newblock In \emph{Proceedings of the 19th International Conference on World
  Wide Web, {WWW} 2010, Raleigh, North Carolina, USA, April 26-30, 2010}, pages
  751--760. {ACM}.

\bibitem[{Pan et~al.(2020)Pan, Zhang, Ji, and Yang}]{DBLP:conf/sp/PanZJY20}
Xudong Pan, Mi~Zhang, Shouling Ji, and Min Yang. 2020.
\newblock \href {https://doi.org/10.1109/SP40000.2020.00095} {Privacy risks of
  general-purpose language models}.
\newblock In \emph{2020 {IEEE} Symposium on Security and Privacy, {SP} 2020,
  San Francisco, CA, USA, May 18-21, 2020}, pages 1314--1331. {IEEE}.

\bibitem[{Papernot et~al.(2016{\natexlab{a}})Papernot, McDaniel, and
  Goodfellow}]{DBLP:journals/corr/PapernotMG16}
Nicolas Papernot, Patrick~D. McDaniel, and Ian~J. Goodfellow.
  2016{\natexlab{a}}.
\newblock \href {http://arxiv.org/abs/1605.07277} {Transferability in machine
  learning: from phenomena to black-box attacks using adversarial samples}.
\newblock \emph{CoRR}, abs/1605.07277.

\bibitem[{Papernot et~al.(2016{\natexlab{b}})Papernot, McDaniel, Jha,
  Fredrikson, Celik, and Swami}]{DBLP:conf/eurosp/PapernotMJFCS16}
Nicolas Papernot, Patrick~D. McDaniel, Somesh Jha, Matt Fredrikson, Z.~Berkay
  Celik, and Ananthram Swami. 2016{\natexlab{b}}.
\newblock \href {https://doi.org/10.1109/EuroSP.2016.36} {The limitations of
  deep learning in adversarial settings}.
\newblock In \emph{{IEEE} European Symposium on Security and Privacy,
  EuroS{\&}P 2016, Saarbr{\"{u}}cken, Germany, March 21-24, 2016}, pages
  372--387. {IEEE}.

\bibitem[{Papernot et~al.(2016{\natexlab{c}})Papernot, McDaniel, Swami, and
  Harang}]{DBLP:conf/milcom/PapernotMSH16}
Nicolas Papernot, Patrick~D. McDaniel, Ananthram Swami, and Richard~E. Harang.
  2016{\natexlab{c}}.
\newblock \href {https://doi.org/10.1109/MILCOM.2016.7795300} {Crafting
  adversarial input sequences for recurrent neural networks}.
\newblock In \emph{2016 {IEEE} Military Communications Conference, {MILCOM}
  2016, Baltimore, MD, USA, November 1-3, 2016}, pages 49--54. {IEEE}.

\bibitem[{Papineni et~al.(2002)Papineni, Roukos, Ward, and
  Zhu}]{papineni-etal-2002-bleu}
Kishore Papineni, Salim Roukos, Todd Ward, and Wei-Jing Zhu. 2002.
\newblock \href {https://doi.org/10.3115/1073083.1073135} {{B}leu: a method for
  automatic evaluation of machine translation}.
\newblock In \emph{Proceedings of the 40th Annual Meeting of the Association
  for Computational Linguistics}, pages 311--318, Philadelphia, Pennsylvania,
  USA. Association for Computational Linguistics.

\bibitem[{Pardo et~al.(2016)Pardo, Rosso, Verhoeven, Daelemans, Potthast, and
  Stein}]{DBLP:conf/clef/PardoRVDPS16}
Francisco Manuel~Rangel Pardo, Paolo Rosso, Ben Verhoeven, Walter Daelemans,
  Martin Potthast, and Benno Stein. 2016.
\newblock \href {http://ceur-ws.org/Vol-1609/16090750.pdf} {Overview of the 4th
  author profiling task at {PAN} 2016: Cross-genre evaluations}.
\newblock In \emph{Working Notes of {CLEF} 2016 - Conference and Labs of the
  Evaluation forum, {\'{E}}vora, Portugal, 5-8 September, 2016}, volume 1609 of
  \emph{{CEUR} Workshop Proceedings}, pages 750--784. CEUR-WS.org.

\bibitem[{Passali et~al.(2021)Passali, Gidiotis, Chatzikyriakidis, and
  Tsoumakas}]{passali-etal-2021-towards}
Tatiana Passali, Alexios Gidiotis, Efstathios Chatzikyriakidis, and Grigorios
  Tsoumakas. 2021.
\newblock \href {https://aclanthology.org/2021.hcinlp-1.4} {Towards
  human-centered summarization: A case study on financial news}.
\newblock In \emph{Proceedings of the First Workshop on Bridging
  Human{--}Computer Interaction and Natural Language Processing}, pages 21--27,
  Online. Association for Computational Linguistics.

\bibitem[{Paszke et~al.(2019)Paszke, Gross, Massa, Lerer, Bradbury, Chanan,
  Killeen, Lin, Gimelshein, Antiga, Desmaison, K{\"{o}}pf, Yang, DeVito,
  Raison, Tejani, Chilamkurthy, Steiner, Fang, Bai, and
  Chintala}]{DBLP:conf/nips/PaszkeGMLBCKLGA19}
Adam Paszke, Sam Gross, Francisco Massa, Adam Lerer, James Bradbury, Gregory
  Chanan, Trevor Killeen, Zeming Lin, Natalia Gimelshein, Luca Antiga, Alban
  Desmaison, Andreas K{\"{o}}pf, Edward~Z. Yang, Zachary DeVito, Martin Raison,
  Alykhan Tejani, Sasank Chilamkurthy, Benoit Steiner, Lu~Fang, Junjie Bai, and
  Soumith Chintala. 2019.
\newblock \href
  {https://proceedings.neurips.cc/paper/2019/hash/bdbca288fee7f92f2bfa9f7012727740-Abstract.html}
  {Pytorch: An imperative style, high-performance deep learning library}.
\newblock In \emph{Advances in Neural Information Processing Systems 32: Annual
  Conference on Neural Information Processing Systems 2019, NeurIPS 2019,
  December 8-14, 2019, Vancouver, BC, Canada}, pages 8024--8035.

\bibitem[{Paul and Saha(2020)}]{paul_cyberbert_2020}
Sayanta Paul and Sriparna Saha. 2020.
\newblock \href {https://doi.org/10.1007/s00530-020-00710-4} {{CyberBERT}:
  {BERT} for cyberbullying identification}.
\newblock \emph{Multimedia Systems}.
\newblock 00000.

\bibitem[{Pavlopoulos et~al.(2020)Pavlopoulos, Sorensen, Dixon, Thain, and
  Androutsopoulos}]{pavlopoulos-etal-2020-toxicity}
John Pavlopoulos, Jeffrey Sorensen, Lucas Dixon, Nithum Thain, and Ion
  Androutsopoulos. 2020.
\newblock \href {https://doi.org/10.18653/v1/2020.acl-main.396} {Toxicity
  detection: Does context really matter?}
\newblock In \emph{Proceedings of the 58th Annual Meeting of the Association
  for Computational Linguistics}, pages 4296--4305, Online. Association for
  Computational Linguistics.

\bibitem[{Pedregosa et~al.(2011)Pedregosa, Varoquaux, Gramfort, Michel,
  Thirion, Grisel, Blondel, Prettenhofer, Weiss, Dubourg, VanderPlas, Passos,
  Cournapeau, Brucher, Perrot, and
  Duchesnay}]{DBLP:journals/jmlr/PedregosaVGMTGBPWDVPCBPD11}
Fabian Pedregosa, Ga{\"{e}}l Varoquaux, Alexandre Gramfort, Vincent Michel,
  Bertrand Thirion, Olivier Grisel, Mathieu Blondel, Peter Prettenhofer, Ron
  Weiss, Vincent Dubourg, Jake VanderPlas, Alexandre Passos, David Cournapeau,
  Matthieu Brucher, Matthieu Perrot, and Edouard Duchesnay. 2011.
\newblock \href {https://dl.acm.org/doi/10.5555/1953048.2078195} {Scikit-learn:
  Machine learning in python}.
\newblock \emph{J. Mach. Learn. Res.}, 12:2825--2830.

\bibitem[{Peersman et~al.(2011)Peersman, Daelemans, and
  Vaerenbergh}]{DBLP:conf/cikm/PeersmanDV11}
Claudia Peersman, Walter Daelemans, and Leona~Van Vaerenbergh. 2011.
\newblock \href {https://doi.org/10.1145/2065023.2065035} {Predicting age and
  gender in online social networks}.
\newblock In \emph{Proceedings of the 3rd International {CIKM} Workshop on
  Search and Mining User-Generated Contents, {SMUC} 2011, Glasgow, United
  Kingdom, October 28, 2011}, pages 37--44. {ACM}.

\bibitem[{Pennacchiotti and Popescu(2011)}]{DBLP:conf/kdd/PennacchiottiP11}
Marco Pennacchiotti and Ana{-}Maria Popescu. 2011.
\newblock \href {https://doi.org/10.1145/2020408.2020477} {Democrats,
  republicans and starbucks afficionados: user classification in twitter}.
\newblock In \emph{Proceedings of the 17th {ACM} {SIGKDD} International
  Conference on Knowledge Discovery and Data Mining, San Diego, CA, USA, August
  21-24, 2011}, pages 430--438. {ACM}.

\bibitem[{Pennington et~al.(2014)Pennington, Socher, and
  Manning}]{pennington-etal-2014-glove}
Jeffrey Pennington, Richard Socher, and Christopher Manning. 2014.
\newblock \href {https://doi.org/10.3115/v1/D14-1162} {{G}lo{V}e: Global
  vectors for word representation}.
\newblock In \emph{Proceedings of the 2014 Conference on Empirical Methods in
  Natural Language Processing ({EMNLP})}, pages 1532--1543, Doha, Qatar.
  Association for Computational Linguistics.

\bibitem[{Pineau et~al.(2018)Pineau, Fried, Ke, and
  Larochelle}]{pineau2018iclr}
Joelle Pineau, G~Fried, RN~Ke, and H~Larochelle. 2018.
\newblock Iclr 2018 reproducibility challenge.
\newblock In \emph{ICML workshop on Reproducibility in Machine Learning}.

\bibitem[{Plank and Hovy(2015)}]{plank-hovy-2015-personality}
Barbara Plank and Dirk Hovy. 2015.
\newblock \href {https://doi.org/10.18653/v1/W15-2913} {Personality traits on
  {T}witter{---}or{---}{H}ow to get 1,500 personality tests in a week}.
\newblock In \emph{Proceedings of the 6th Workshop on Computational Approaches
  to Subjectivity, Sentiment and Social Media Analysis}, pages 92--98, Lisboa,
  Portugal. Association for Computational Linguistics.

\bibitem[{Potthast et~al.(2016)Potthast, Hagen, and
  Stein}]{DBLP:conf/clef/PotthastHS16}
Martin Potthast, Matthias Hagen, and Benno Stein. 2016.
\newblock \href {http://ceur-ws.org/Vol-1609/16090716.pdf} {Author obfuscation:
  Attacking the state of the art in authorship verification}.
\newblock In \emph{Working Notes of {CLEF} 2016 - Conference and Labs of the
  Evaluation forum, {\'{E}}vora, Portugal, 5-8 September, 2016}, volume 1609 of
  \emph{{CEUR} Workshop Proceedings}, pages 716--749. CEUR-WS.org.

\bibitem[{Potthast et~al.(2019)Potthast, Rosso, Stamatatos, and
  Stein}]{10.1007/978-3-030-15719-7_39}
Martin Potthast, Paolo Rosso, Efstathios Stamatatos, and Benno Stein. 2019.
\newblock A decade of shared tasks in digital text forensics at pan.
\newblock In \emph{Advances in Information Retrieval}, pages 291--300, Cham.
  Springer International Publishing.

\bibitem[{Potthast et~al.(2018)Potthast, Schremmer, Hagen, and
  Stein}]{DBLP:conf/clef/PotthastSHS18}
Martin Potthast, Felix Schremmer, Matthias Hagen, and Benno Stein. 2018.
\newblock \href {http://ceur-ws.org/Vol-2125/invited\_paper\_16.pdf} {Overview
  of the author obfuscation task at {PAN} 2018: {A} new approach to measuring
  safety}.
\newblock In \emph{Working Notes of {CLEF} 2018 - Conference and Labs of the
  Evaluation Forum, Avignon, France, September 10-14, 2018}, volume 2125 of
  \emph{{CEUR} Workshop Proceedings}. CEUR-WS.org.

\bibitem[{Privitera and Campbell(2009)}]{Privitera2009-iu}
Carmel Privitera and Marilyn~Anne Campbell. 2009.
\newblock \href {https://doi.org/https://doi.org/10.1089/cpb.2009.0025}
  {Cyberbullying: the new face of workplace bullying?}
\newblock \emph{Cyberpsychol Behav}, 12(4):395--400.

\bibitem[{Pryzant et~al.(2017)Pryzant, Chung, and
  Jurafsky}]{DBLP:conf/sigir/PryzantCJ17}
Reid Pryzant, Youngjoo Chung, and Dan Jurafsky. 2017.
\newblock \href {http://ceur-ws.org/Vol-2311/paper\_3.pdf} {Predicting sales
  from the language of product descriptions}.
\newblock In \emph{Proceedings of the {SIGIR} 2017 Workshop On eCommerce
  co-located with the 40th International {ACM} {SIGIR} Conference on Research
  and Development in Information Retrieval, eCOM@SIGIR 2017, Tokyo, Japan,
  August 11, 2017}, volume 2311 of \emph{{CEUR} Workshop Proceedings}.
  CEUR-WS.org.

\bibitem[{Quteineh et~al.(2020)Quteineh, Samothrakis, and
  Sutcliffe}]{quteineh-etal-2020-textual}
Husam Quteineh, Spyridon Samothrakis, and Richard Sutcliffe. 2020.
\newblock \href {https://doi.org/10.18653/v1/2020.emnlp-main.600} {Textual data
  augmentation for efficient active learning on tiny datasets}.
\newblock In \emph{Proceedings of the 2020 Conference on Empirical Methods in
  Natural Language Processing (EMNLP)}, pages 7400--7410, Online. Association
  for Computational Linguistics.

\bibitem[{Radfar et~al.(2020)Radfar, Shivaram, and
  Culotta}]{DBLP:conf/icwsm/RadfarSC20}
Bahar Radfar, Karthik Shivaram, and Aron Culotta. 2020.
\newblock \href {https://aaai.org/ojs/index.php/ICWSM/article/view/7366}
  {Characterizing variation in toxic language by social context}.
\newblock In \emph{Proceedings of the Fourteenth International {AAAI}
  Conference on Web and Social Media, {ICWSM} 2020, Held Virtually, Original
  Venue: Atlanta, Georgia, USA, June 8-11, 2020}, pages 959--963. {AAAI} Press.

\bibitem[{Radford et~al.(2018)Radford, Narasimhan, Salimans, and
  Sutskever}]{radford2018improving}
Alec Radford, Karthik Narasimhan, Tim Salimans, and Ilya Sutskever. 2018.
\newblock \href
  {https://cdn.openai.com/research-covers/language-unsupervised/language_understanding_paper.pdf}
  {Improving language understanding by generative pre-training}.
\newblock \emph{OpenAI blog}.

\bibitem[{Radford et~al.(2019)Radford, Wu, Child, Luan, Amodei, Sutskever
  et~al.}]{radford2019language}
Alec Radford, Jeffrey Wu, Rewon Child, David Luan, Dario Amodei, Ilya
  Sutskever, et~al. 2019.
\newblock \href
  {https://cdn.openai.com/better-language-models/language_models_are_unsupervised_multitask_learners.pdf}
  {Language models are unsupervised multitask learners}.
\newblock \emph{OpenAI blog}, 1(8):9.

\bibitem[{Rao et~al.(2010)Rao, Yarowsky, Shreevats, and
  Gupta}]{DBLP:conf/cikm/RaoYSG10}
Delip Rao, David Yarowsky, Abhishek Shreevats, and Manaswi Gupta. 2010.
\newblock \href {https://doi.org/10.1145/1871985.1871993} {Classifying latent
  user attributes in twitter}.
\newblock In \emph{Proceedings of the 2nd international workshop on Search and
  mining user-generated contents, SMUC@CIKM 2010, Toronto, ON, Canada, October
  30, 2010}, pages 37--44. {ACM}.

\bibitem[{Rao and Rohatgi(2000)}]{DBLP:conf/uss/RaoR00}
Josyula~R. Rao and Pankaj Rohatgi. 2000.
\newblock \href
  {https://www.usenix.org/conference/9th-usenix-security-symposium/can-pseudonymity-really-guarantee-privacy}
  {Can pseudonymity really guarantee privacy?}
\newblock In \emph{9th {USENIX} Security Symposium, Denver, Colorado, USA,
  August 14-17, 2000}. {USENIX} Association.

\bibitem[{Raskin et~al.(2002)Raskin, Nirenburg, Atallah, Hempelmann, and
  Triezenberg}]{raskin-etal-2002-nlp}
Victor Raskin, Sergei Nirenburg, Mikhail~J. Atallah, Christian~F. Hempelmann,
  and Katrina~E. Triezenberg. 2002.
\newblock \href {https://aclanthology.org/W02-1303} {Why {NLP} should move into
  {IAS}}.
\newblock In \emph{{COLING}-02: A Roadmap for Computational Linguistics}.

\bibitem[{Rayner et~al.(2006)Rayner, White, Johnson, and
  Liversedge}]{Rayner2006-de}
Keith Rayner, Sarah~J White, Rebecca~L Johnson, and Simon~P Liversedge. 2006.
\newblock \href {https://doi.org/10.1111/j.1467-9280.2006.01684.x} {Raeding
  wrods with jubmled lettres: there is a cost}.
\newblock \emph{Psychol Sci"}, 17(3):192--193.

\bibitem[{Recht et~al.(2019)Recht, Roelofs, Schmidt, and
  Shankar}]{DBLP:conf/icml/RechtRSS19}
Benjamin Recht, Rebecca Roelofs, Ludwig Schmidt, and Vaishaal Shankar. 2019.
\newblock \href {http://proceedings.mlr.press/v97/recht19a.html} {Do imagenet
  classifiers generalize to imagenet?}
\newblock In \emph{Proceedings of the 36th International Conference on Machine
  Learning, {ICML} 2019, 9-15 June 2019, Long Beach, California, {USA}},
  volume~97 of \emph{Proceedings of Machine Learning Research}, pages
  5389--5400. {PMLR}.

\bibitem[{Reddy and Knight(2016)}]{reddy-knight-2016-obfuscating}
Sravana Reddy and Kevin Knight. 2016.
\newblock \href {https://doi.org/10.18653/v1/W16-5603} {Obfuscating gender in
  social media writing}.
\newblock In \emph{Proceedings of the First Workshop on {NLP} and Computational
  Social Science}, pages 17--26, Austin, Texas. Association for Computational
  Linguistics.

\bibitem[{{\v R}eh{\r u}{\v r}ek and Sojka(2010)}]{rehurek_lrec}
Radim {\v R}eh{\r u}{\v r}ek and Petr Sojka. 2010.
\newblock \href {http://is.muni.cz/publication/884893/en} {{Software Framework
  for Topic Modelling with Large Corpora}}.
\newblock In \emph{{Proceedings of the LREC 2010 Workshop on New Challenges for
  NLP Frameworks}}, pages 45--50, Valletta, Malta. ELRA.

\bibitem[{Reynolds et~al.(2011{\natexlab{a}})Reynolds, Kontostathis, and
  Edwards}]{6147681}
Kelly Reynolds, April Kontostathis, and Lynne Edwards. 2011{\natexlab{a}}.
\newblock \href {https://doi.org/10.1109/ICMLA.2011.152} {Using machine
  learning to detect cyberbullying}.
\newblock In \emph{Proceedings of the 10th International Conference on Machine
  Learning and Applications and Workshops}, volume~2, pages 241--244.

\bibitem[{Reynolds et~al.(2011{\natexlab{b}})Reynolds, Kontostathis, and
  Edwards}]{DBLP:conf/icmla/ReynoldsKE11}
Kelly Reynolds, April Kontostathis, and Lynne Edwards. 2011{\natexlab{b}}.
\newblock \href {https://doi.org/10.1109/ICMLA.2011.152} {Using machine
  learning to detect cyberbullying}.
\newblock In \emph{10th International Conference on Machine Learning and
  Applications and Workshops, {ICMLA} 2011, Honolulu, Hawaii, USA, December
  18-21, 2011. Volume 2: Special Sessions and Workshop}, pages 241--244. {IEEE}
  Computer Society.

\bibitem[{Rieser and Lemon(2008)}]{rieser-lemon-2008-automatic}
Verena Rieser and Oliver Lemon. 2008.
\newblock \href
  {http://www.lrec-conf.org/proceedings/lrec2008/pdf/592_paper.pdf} {Automatic
  learning and evaluation of user-centered objective functions for dialogue
  system optimisation}.
\newblock In \emph{Proceedings of the Sixth International Conference on
  Language Resources and Evaluation ({LREC}'08)}, Marrakech, Morocco. European
  Language Resources Association (ELRA).

\bibitem[{Robers et~al.(2015)Robers, Zhang, and Morgan}]{Morgan2015}
Simone Robers, Anlan Zhang, and Rachel~E Morgan. 2015.
\newblock Indicators of school crime and safety: 2014.
\newblock \emph{National Center for Education Statistics, U.S. Department of
  Education, and Bureau of Justice Statistics, Office of Justice Programs, U.S.
  Department of Justice}.
\newblock NCES 2015-072/NCJ 248036.

\bibitem[{Rosa et~al.(2019{\natexlab{a}})Rosa, Pereira, Ribeiro, Ferreira,
  Carvalho, Oliveira, Coheur, Paulino, {Veiga Simão}, and
  Trancoso}]{ROSA2019333}
H.~Rosa, N.~Pereira, R.~Ribeiro, P.C. Ferreira, J.P. Carvalho, S.~Oliveira,
  L.~Coheur, P.~Paulino, A.M. {Veiga Simão}, and I.~Trancoso.
  2019{\natexlab{a}}.
\newblock \href {https://doi.org/https://doi.org/10.1016/j.chb.2018.12.021}
  {Automatic cyberbullying detection: A systematic review}.
\newblock \emph{Computers in Human Behavior}, 93:333--345.

\bibitem[{Rosa et~al.(2018{\natexlab{a}})Rosa, Carvalho, Calado, Martins,
  Ribeiro, and Coheur}]{DBLP:conf/fuzzIEEE/RosaCCMRC18}
Hugo Rosa, Jo{\~{a}}o~Paulo Carvalho, P{\'{a}}vel Calado, Bruno Martins,
  Ricardo Ribeiro, and Lu{\'{\i}}sa Coheur. 2018{\natexlab{a}}.
\newblock \href {https://doi.org/10.1109/FUZZ-IEEE.2018.8491557} {Using fuzzy
  fingerprints for cyberbullying detection in social networks}.
\newblock In \emph{2018 {IEEE} International Conference on Fuzzy Systems,
  {FUZZ-IEEE} 2018, Rio de Janeiro, Brazil, July 8-13, 2018}, pages 1--7.
  {IEEE}.

\bibitem[{Rosa et~al.(2018{\natexlab{b}})Rosa, de~Matos, Ribeiro, Coheur, and
  Carvalho}]{DBLP:conf/ijcnn/RosaM0CC18}
Hugo Rosa, David~Martins de~Matos, Ricardo Ribeiro, Lu{\'{\i}}sa Coheur, and
  Jo{\~{a}}o~Paulo Carvalho. 2018{\natexlab{b}}.
\newblock \href {https://doi.org/10.1109/IJCNN.2018.8489211} {A "deeper" look
  at detecting cyberbullying in social networks}.
\newblock In \emph{2018 International Joint Conference on Neural Networks,
  {IJCNN} 2018, Rio de Janeiro, Brazil, July 8-13, 2018}, pages 1--8. {IEEE}.

\bibitem[{Rosa et~al.(2019{\natexlab{b}})Rosa, Pereira, Ribeiro, Ferreira,
  Carvalho, Oliveira, Coheur, Paulino, Sim{\~{a}}o, and
  Trancoso}]{DBLP:journals/chb/RosaPRFCOCPST19}
Hugo Rosa, N{\'{a}}dia~Salgado Pereira, Ricardo Ribeiro, Paula~Costa Ferreira,
  Jo{\~{a}}o~Paulo Carvalho, Sofia Oliveira, Lu{\'{\i}}sa Coheur, Paula
  Paulino, Ana Margarida~Veiga Sim{\~{a}}o, and Isabel Trancoso.
  2019{\natexlab{b}}.
\newblock \href {https://doi.org/10.1016/j.chb.2018.12.021} {Automatic
  cyberbullying detection: {A} systematic review}.
\newblock \emph{Comput. Hum. Behav.}, 93:333--345.

\bibitem[{Rosenthal et~al.(2021)Rosenthal, Atanasova, Karadzhov, Zampieri, and
  Nakov}]{rosenthal-etal-2021-solid}
Sara Rosenthal, Pepa Atanasova, Georgi Karadzhov, Marcos Zampieri, and Preslav
  Nakov. 2021.
\newblock \href {https://doi.org/10.18653/v1/2021.findings-acl.80} {{SOLID}: A
  large-scale semi-supervised dataset for offensive language identification}.
\newblock In \emph{Findings of the Association for Computational Linguistics:
  ACL-IJCNLP 2021}, pages 915--928, Online. Association for Computational
  Linguistics.

\bibitem[{Roth et~al.(2021)Roth, Gao, Abuadbba, Nepal, and
  Liu}]{DBLP:journals/corr/abs-2103-00676}
Tom Roth, Yansong Gao, Alsharif Abuadbba, Surya Nepal, and Wei Liu. 2021.
\newblock \href {http://arxiv.org/abs/2103.00676} {Token-modification
  adversarial attacks for natural language processing: {A} survey}.
\newblock \emph{CoRR}, abs/2103.00676.

\bibitem[{Saedi and Dras(2020)}]{saedi-dras-2020-large}
Chakaveh Saedi and Mark Dras. 2020.
\newblock \href {https://aclanthology.org/2020.starsem-1.19} {Large scale
  author obfuscation using {S}iamese variational auto-encoder: The {S}iam{AO}
  system}.
\newblock In \emph{Proceedings of the Ninth Joint Conference on Lexical and
  Computational Semantics}, pages 179--189, Barcelona, Spain (Online).
  Association for Computational Linguistics.

\bibitem[{Salawu et~al.(2020)Salawu, He, and
  Lumsden}]{DBLP:journals/taffco/SalawuHL20}
Semiu Salawu, Yulan He, and Joanna Lumsden. 2020.
\newblock \href {https://doi.org/10.1109/TAFFC.2017.2761757} {Approaches to
  automated detection of cyberbullying: {A} survey}.
\newblock \emph{{IEEE} Trans. Affect. Comput.}, 11(1):3--24.

\bibitem[{Salminen et~al.(2020)Salminen, Hopf, Chowdhury, Jung, Almerekhi, and
  Jansen}]{DBLP:journals/hcis/SalminenHCJAJ20}
Joni Salminen, Maximilian Hopf, Shammur~A. Chowdhury, Soon{-}Gyo Jung, Hind
  Almerekhi, and Bernard~J. Jansen. 2020.
\newblock \href {https://doi.org/10.1186/s13673-019-0205-6} {Developing an
  online hate classifier for multiple social media platforms}.
\newblock \emph{Hum. centric Comput. Inf. Sci.}, 10:1.

\bibitem[{Samek et~al.(2017)Samek, Binder, Montavon, Lapuschkin, and
  M{\"{u}}ller}]{DBLP:journals/tnn/SamekBMLM17}
Wojciech Samek, Alexander Binder, Gr{\'{e}}goire Montavon, Sebastian
  Lapuschkin, and Klaus{-}Robert M{\"{u}}ller. 2017.
\newblock \href {https://doi.org/10.1109/TNNLS.2016.2599820} {Evaluating the
  visualization of what a deep neural network has learned}.
\newblock \emph{{IEEE} Trans. Neural Networks Learn. Syst.}, 28(11):2660--2673.

\bibitem[{Sanh et~al.(2019)Sanh, Debut, Chaumond, and
  Wolf}]{DBLP:journals/corr/abs-1910-01108}
Victor Sanh, Lysandre Debut, Julien Chaumond, and Thomas Wolf. 2019.
\newblock \href {http://arxiv.org/abs/1910.01108} {Distilbert, a distilled
  version of {BERT:} smaller, faster, cheaper and lighter}.
\newblock \emph{CoRR}, abs/1910.01108.

\bibitem[{Sanh et~al.(2021)Sanh, Webson, Raffel, Bach, Sutawika, Alyafeai,
  Chaffin, Stiegler, Scao, Raja, Dey, Bari, Xu, Thakker, Sharma, Szczechla,
  Kim, Chhablani, Nayak, Datta, Chang, Jiang, Wang, Manica, Shen, Yong, Pandey,
  Bawden, Wang, Neeraj, Rozen, Sharma, Santilli, F{\'{e}}vry, Fries, Teehan,
  Biderman, Gao, Bers, Wolf, and Rush}]{DBLP:journals/corr/abs-2110-08207}
Victor Sanh, Albert Webson, Colin Raffel, Stephen~H. Bach, Lintang Sutawika,
  Zaid Alyafeai, Antoine Chaffin, Arnaud Stiegler, Teven~Le Scao, Arun Raja,
  Manan Dey, M.~Saiful Bari, Canwen Xu, Urmish Thakker, Shanya Sharma, Eliza
  Szczechla, Taewoon Kim, Gunjan Chhablani, Nihal~V. Nayak, Debajyoti Datta,
  Jonathan Chang, Mike~Tian{-}Jian Jiang, Han Wang, Matteo Manica, Sheng Shen,
  Zheng~Xin Yong, Harshit Pandey, Rachel Bawden, Thomas Wang, Trishala Neeraj,
  Jos Rozen, Abheesht Sharma, Andrea Santilli, Thibault F{\'{e}}vry, Jason~Alan
  Fries, Ryan Teehan, Stella Biderman, Leo Gao, Tali Bers, Thomas Wolf, and
  Alexander~M. Rush. 2021.
\newblock \href {http://arxiv.org/abs/2110.08207} {Multitask prompted training
  enables zero-shot task generalization}.
\newblock \emph{CoRR}, abs/2110.08207.

\bibitem[{Sap et~al.(2019)Sap, Card, Gabriel, Choi, and
  Smith}]{sap-etal-2019-risk}
Maarten Sap, Dallas Card, Saadia Gabriel, Yejin Choi, and Noah~A. Smith. 2019.
\newblock \href {https://doi.org/10.18653/v1/P19-1163} {The risk of racial bias
  in hate speech detection}.
\newblock In \emph{Proceedings of the 57th Annual Meeting of the Association
  for Computational Linguistics}, pages 1668--1678, Florence, Italy.
  Association for Computational Linguistics.

\bibitem[{Sap et~al.(2014)Sap, Park, Eichstaedt, Kern, Stillwell, Kosinski,
  Ungar, and Schwartz}]{sap-etal-2014-developing}
Maarten Sap, Gregory Park, Johannes Eichstaedt, Margaret Kern, David Stillwell,
  Michal Kosinski, Lyle Ungar, and Hansen~Andrew Schwartz. 2014.
\newblock \href {https://doi.org/10.3115/v1/D14-1121} {Developing age and
  gender predictive lexica over social media}.
\newblock In \emph{Proceedings of the 2014 Conference on Empirical Methods in
  Natural Language Processing ({EMNLP})}, pages 1146--1151, Doha, Qatar.
  Association for Computational Linguistics.

\bibitem[{Sch{\"a}fer and Bildhauer(2012)}]{schafer-bildhauer-2012-building}
Roland Sch{\"a}fer and Felix Bildhauer. 2012.
\newblock \href
  {http://www.lrec-conf.org/proceedings/lrec2012/pdf/834_Paper.pdf} {Building
  large corpora from the web using a new efficient tool chain}.
\newblock In \emph{Proceedings of the Eighth International Conference on
  Language Resources and Evaluation ({LREC}'12)}, pages 486--493, Istanbul,
  Turkey. European Language Resources Association (ELRA).

\bibitem[{Schaltenbrand et~al.(1996)Schaltenbrand, Lengelle, Toussaint,
  Luthringer, Carelli, Jacqmin, Lainey, Muzet, and
  Macher}]{10.1093/sleep/19.1.26}
N.~Schaltenbrand, R.~Lengelle, M.~Toussaint, R.~Luthringer, G.~Carelli,
  A.~Jacqmin, E.~Lainey, A.~Muzet, and J.~P. Macher. 1996.
\newblock \href {https://doi.org/10.1093/sleep/19.1.26} {{Sleep Stage Scoring
  Using the Neural Network Model: Comparison Between Visual and Automatic
  Analysis in Normal Subjects and Patients}}.
\newblock \emph{Sleep}, 19(1):26--35.

\bibitem[{Schler et~al.(2006)Schler, Koppel, Argamon, and
  Pennebaker}]{DBLP:conf/aaaiss/SchlerKAP06}
Jonathan Schler, Moshe Koppel, Shlomo Argamon, and James~W. Pennebaker. 2006.
\newblock \href
  {http://www.aaai.org/Library/Symposia/Spring/2006/ss06-03-039.php} {Effects
  of age and gender on blogging}.
\newblock In \emph{Computational Approaches to Analyzing Weblogs, Papers from
  the 2006 {AAAI} Spring Symposium, Technical Report SS-06-03, Stanford,
  California, USA, March 27-29, 2006}, pages 199--205. {AAAI}.

\bibitem[{Schmidhuber(2015)}]{SCHMIDHUBER201585}
Jürgen Schmidhuber. 2015.
\newblock \href {https://doi.org/https://doi.org/10.1016/j.neunet.2014.09.003}
  {Deep learning in neural networks: An overview}.
\newblock \emph{Neural Networks}, 61:85--117.

\bibitem[{Schnabel and Sch{\"u}tze(2014)}]{schnabel-schutze-2014-flors}
Tobias Schnabel and Hinrich Sch{\"u}tze. 2014.
\newblock \href {https://doi.org/10.1162/tacl_a_00162} {{FLORS}: Fast and
  simple domain adaptation for part-of-speech tagging}.
\newblock \emph{Transactions of the Association for Computational Linguistics},
  2:15--26.

\bibitem[{Schneier(2010)}]{DBLP:journals/ieeesp/Schneier10}
Bruce Schneier. 2010.
\newblock \href {https://doi.org/10.1109/MSP.2010.47} {Security and function
  creep}.
\newblock \emph{{IEEE} Secur. Priv.}, 8(1):88.

\bibitem[{Schuff et~al.(2020)Schuff, Adel, and Vu}]{schuff-etal-2020-f1}
Hendrik Schuff, Heike Adel, and Ngoc~Thang Vu. 2020.
\newblock \href {https://doi.org/10.18653/v1/2020.emnlp-main.575} {{F}1 is
  {N}ot {E}nough! {M}odels and {E}valuation {T}owards {U}ser-{C}entered
  {E}xplainable {Q}uestion {A}nswering}.
\newblock In \emph{Proceedings of the 2020 Conference on Empirical Methods in
  Natural Language Processing (EMNLP)}, pages 7076--7095, Online. Association
  for Computational Linguistics.

\bibitem[{Schuster and Paliwal(1997)}]{DBLP:journals/tsp/SchusterP97}
Mike Schuster and Kuldip~K. Paliwal. 1997.
\newblock \href {https://doi.org/10.1109/78.650093} {Bidirectional recurrent
  neural networks}.
\newblock \emph{{IEEE} Trans. Signal Process.}, 45(11):2673--2681.

\bibitem[{Sculley et~al.(2018)Sculley, Snoek, Wiltschko, and
  Rahimi}]{sculley2018winner}
D.~Sculley, Jasper Snoek, Alex Wiltschko, and Ali Rahimi. 2018.
\newblock \href {https://openreview.net/forum?id=rJWF0Fywf} {Winner's curse? on
  pace, progress, and empirical rigor}.

\bibitem[{Sculley et~al.(2019)Sculley, Snoek, and
  Wiltschko}]{DBLP:journals/corr/abs-1901-06246}
D.~Sculley, Jasper Snoek, and Alexander~B. Wiltschko. 2019.
\newblock \href {http://arxiv.org/abs/1901.06246} {Avoiding a tragedy of the
  commons in the peer review process}.
\newblock \emph{CoRR}, abs/1901.06246.

\bibitem[{Sellam et~al.(2020)Sellam, Das, and Parikh}]{sellam-etal-2020-bleurt}
Thibault Sellam, Dipanjan Das, and Ankur Parikh. 2020.
\newblock \href {https://doi.org/10.18653/v1/2020.acl-main.704} {{BLEURT}:
  Learning robust metrics for text generation}.
\newblock In \emph{Proceedings of the 58th Annual Meeting of the Association
  for Computational Linguistics}, pages 7881--7892, Online. Association for
  Computational Linguistics.

\bibitem[{Sen and Can(2021)}]{DBLP:journals/corr/abs-2107-03072}
Sevil Sen and Burcu Can. 2021.
\newblock \href {http://arxiv.org/abs/2107.03072} {Android security using {NLP}
  techniques: {A} review}.
\newblock \emph{CoRR}, abs/2107.03072.

\bibitem[{Shetty et~al.(2018)Shetty, Schiele, and
  Fritz}]{DBLP:conf/uss/ShettySF18}
Rakshith Shetty, Bernt Schiele, and Mario Fritz. 2018.
\newblock \href
  {https://www.usenix.org/conference/usenixsecurity18/presentation/shetty}
  {{A4NT:} author attribute anonymity by adversarial training of neural machine
  translation}.
\newblock In \emph{27th {USENIX} Security Symposium, {USENIX} Security 2018,
  Baltimore, MD, USA, August 15-17, 2018}, pages 1633--1650. {USENIX}
  Association.

\bibitem[{Shokri et~al.(2017)Shokri, Stronati, Song, and
  Shmatikov}]{DBLP:conf/sp/ShokriSSS17}
Reza Shokri, Marco Stronati, Congzheng Song, and Vitaly Shmatikov. 2017.
\newblock \href {https://doi.org/10.1109/SP.2017.41} {Membership inference
  attacks against machine learning models}.
\newblock In \emph{2017 {IEEE} Symposium on Security and Privacy, {SP} 2017,
  San Jose, CA, USA, May 22-26, 2017}, pages 3--18. {IEEE} Computer Society.

\bibitem[{Shropshire(2018)}]{shropshire-2018-natural}
Jordan Shropshire. 2018.
\newblock \href {https://aisel.aisnet.org/wisp2018/26} {Natural language
  processing as a weapon}.
\newblock In \emph{Proceedings of the 13th Pre-ICIS Workshop on Information
  Security and Privacy}, volume~1.

\bibitem[{Si et~al.(2021)Si, Zhang, Qi, Liu, Wang, Liu, and
  Sun}]{DBLP:conf/acl/SiZQLWLS21}
Chenglei Si, Zhengyan Zhang, Fanchao Qi, Zhiyuan Liu, Yasheng Wang, Qun Liu,
  and Maosong Sun. 2021.
\newblock \href {https://doi.org/10.18653/v1/2021.findings-acl.137} {Better
  robustness by more coverage: Adversarial and mixup data augmentation for
  robust finetuning}.
\newblock In \emph{Findings of the Association for Computational Linguistics:
  {ACL/IJCNLP} 2021, Online Event, August 1-6, 2021}, volume {ACL/IJCNLP} 2021
  of \emph{Findings of {ACL}}, pages 1569--1576. Association for Computational
  Linguistics.

\bibitem[{Smith and Anderson(2018)}]{Smith2018}
Aaron Smith and Monica Anderson. 2018.
\newblock \href
  {https://www.pewresearch.org/internet/wp-content/uploads/sites/9/2018/02/PI_2018.03.01_Social-Media_FINAL.pdf}
  {Social media use in 2018}.
\newblock \emph{Pew Research Center}.

\bibitem[{Smith(2012)}]{DBLP:journals/corr/abs-1207-0245}
Noah~A. Smith. 2012.
\newblock \href {http://arxiv.org/abs/1207.0245} {Adversarial evaluation for
  models of natural language}.
\newblock \emph{CoRR}, abs/1207.0245.

\bibitem[{Snover et~al.(2009)Snover, Madnani, Dorr, and
  Schwartz}]{snover-etal-2009-fluency}
Matthew Snover, Nitin Madnani, Bonnie Dorr, and Richard Schwartz. 2009.
\newblock \href {https://aclanthology.org/W09-0441} {Fluency, adequacy, or
  {HTER}? {E}xploring different human judgments with a tunable {MT} metric}.
\newblock In \emph{Proceedings of the Fourth Workshop on Statistical Machine
  Translation}, pages 259--268, Athens, Greece. Association for Computational
  Linguistics.

\bibitem[{Song et~al.(2017)Song, Ristenpart, and
  Shmatikov}]{DBLP:conf/ccs/SongRS17}
Congzheng Song, Thomas Ristenpart, and Vitaly Shmatikov. 2017.
\newblock \href {https://doi.org/10.1145/3133956.3134077} {Machine learning
  models that remember too much}.
\newblock In \emph{Proceedings of the 2017 {ACM} {SIGSAC} Conference on
  Computer and Communications Security, {CCS} 2017, Dallas, TX, USA, October 30
  - November 03, 2017}, pages 587--601. {ACM}.

\bibitem[{Song et~al.(2021)Song, Yu, Peng, and
  Narasimhan}]{song-etal-2021-universal}
Liwei Song, Xinwei Yu, Hsuan-Tung Peng, and Karthik Narasimhan. 2021.
\newblock \href {https://doi.org/10.18653/v1/2021.naacl-main.291} {Universal
  adversarial attacks with natural triggers for text classification}.
\newblock In \emph{Proceedings of the 2021 Conference of the North American
  Chapter of the Association for Computational Linguistics: Human Language
  Technologies}, pages 3724--3733, Online. Association for Computational
  Linguistics.

\bibitem[{Squicciarini et~al.(2015)Squicciarini, Rajtmajer, Liu, and
  Griffin}]{DBLP:conf/asunam/SquicciariniRLG15}
Anna~Cinzia Squicciarini, Sarah~Michele Rajtmajer, Y.~Liu, and Christopher
  Griffin. 2015.
\newblock \href {https://doi.org/10.1145/2808797.2809398} {Identification and
  characterization of cyberbullying dynamics in an online social network}.
\newblock In \emph{Proceedings of the 2015 {IEEE/ACM} International Conference
  on Advances in Social Networks Analysis and Mining, {ASONAM} 2015, Paris,
  France, August 25 - 28, 2015}, pages 280--285. {ACM}.

\bibitem[{Srivastava et~al.(1996)Srivastava, Chandrakasan, and
  Brodersen}]{DBLP:journals/tvlsi/SrivastavaCB96}
Mani~B. Srivastava, Anantha~P. Chandrakasan, and Robert~W. Brodersen. 1996.
\newblock \href {https://doi.org/10.1109/92.486080} {Predictive system shutdown
  and other architectural techniques for energy efficient programmable
  computation}.
\newblock \emph{{IEEE} Trans. Very Large Scale Integr. Syst.}, 4(1):42--55.

\bibitem[{Srivastava et~al.(2014)Srivastava, Hinton, Krizhevsky, Sutskever, and
  Salakhutdinov}]{DBLP:journals/jmlr/SrivastavaHKSS14}
Nitish Srivastava, Geoffrey~E. Hinton, Alex Krizhevsky, Ilya Sutskever, and
  Ruslan Salakhutdinov. 2014.
\newblock \href {http://dl.acm.org/citation.cfm?id=2670313} {Dropout: a simple
  way to prevent neural networks from overfitting}.
\newblock \emph{J. Mach. Learn. Res.}, 15(1):1929--1958.

\bibitem[{Stamatatos et~al.(1999)Stamatatos, Fakotakis, and
  Kokkinakis}]{stamatatos-etal-1999-automatic}
E.~Stamatatos, N.~Fakotakis, and G.~Kokkinakis. 1999.
\newblock \href {https://aclanthology.org/E99-1021} {Automatic authorship
  attribution}.
\newblock In \emph{Ninth Conference of the {E}uropean Chapter of the
  Association for Computational Linguistics}, pages 158--164, Bergen, Norway.
  Association for Computational Linguistics.

\bibitem[{Stamatatos et~al.(2000)Stamatatos, Fakotakis, and
  Kokkinakis}]{stamatatos-etal-2000-automatic}
Efstathios Stamatatos, Nikos Fakotakis, and George Kokkinakis. 2000.
\newblock \href {https://aclanthology.org/J00-4001} {Automatic text
  categorization in terms of genre and author}.
\newblock \emph{Computational Linguistics}, 26(4):471--495.

\bibitem[{Steijn and Schouten(2013)}]{Steijn2013-hi}
Wouter M~P Steijn and Alexander~P Schouten. 2013.
\newblock \href {https://doi.org/10.1089/cyber.2012.0392} {Information sharing
  and relationships on social networking sites}.
\newblock \emph{Cyberpsychol Behav Soc Netw}, 16(8):582--587.

\bibitem[{Stolerman et~al.(2014)Stolerman, Overdorf, Afroz, and
  Greenstadt}]{DBLP:conf/ifip11-9/StolermanOAG14}
Ariel Stolerman, Rebekah Overdorf, Sadia Afroz, and Rachel Greenstadt. 2014.
\newblock \href {https://doi.org/10.1007/978-3-662-44952-3\_13} {Breaking the
  closed-world assumption in stylometric authorship attribution}.
\newblock In \emph{Advances in Digital Forensics {X} - 10th {IFIP} {WG} 11.9
  International Conference, Vienna, Austria, January 8-10, 2014, Revised
  Selected Papers}, volume 433 of \emph{{IFIP} Advances in Information and
  Communication Technology}, pages 185--205. Springer.

\bibitem[{Stratford(1996)}]{https://doi.org/10.1002/cbm.123}
Margaret Stratford. 1996.
\newblock \href {https://doi.org/https://doi.org/10.1002/cbm.123} {School
  bullying: insights and perspectives}.
\newblock \emph{Criminal Behaviour and Mental Health}, 6(4):362--363.

\bibitem[{Sui(2015)}]{Sui2015}
Junming Sui. 2015.
\newblock \href {https://research.cs.wisc.edu/bullying/pub/junming-thesis.pdf}
  {\emph{Understanding and fighting bullying with machine learning}}.
\newblock Ph.D. thesis, The University of Wisconsin-Madison.

\bibitem[{Sun et~al.(2019)Sun, Qiu, Xu, and
  Huang}]{10.1007/978-3-030-32381-3_16}
Chi Sun, Xipeng Qiu, Yige Xu, and Xuanjing Huang. 2019.
\newblock How to fine-tune bert for text classification?
\newblock In \emph{Chinese Computational Linguistics}, pages 194--206, Cham.
  Springer International Publishing.

\bibitem[{Sun et~al.(2017)Sun, Li, and Zha}]{DBLP:conf/mswim/SunLZ17}
Mingxuan Sun, Changbin Li, and Hongyuan Zha. 2017.
\newblock \href {https://doi.org/10.1145/3127540.3127566} {Inferring private
  demographics of new users in recommender systems}.
\newblock In \emph{Proceedings of the 20th {ACM} International Conference on
  Modelling, Analysis and Simulation of Wireless and Mobile Systems, MSWiM
  2017, Miami, FL, USA, November 21 - 25, 2017}, pages 237--244. {ACM}.

\bibitem[{Sutskever et~al.(2014)Sutskever, Vinyals, and
  Le}]{DBLP:conf/nips/SutskeverVL14}
Ilya Sutskever, Oriol Vinyals, and Quoc~V. Le. 2014.
\newblock \href
  {https://proceedings.neurips.cc/paper/2014/hash/a14ac55a4f27472c5d894ec1c3c743d2-Abstract.html}
  {Sequence to sequence learning with neural networks}.
\newblock In \emph{Advances in Neural Information Processing Systems 27: Annual
  Conference on Neural Information Processing Systems 2014, December 8-13 2014,
  Montreal, Quebec, Canada}, pages 3104--3112.

\bibitem[{Swamy et~al.(2019)Swamy, Jamatia, and
  Gamb{\"a}ck}]{swamy-etal-2019-studying}
Steve~Durairaj Swamy, Anupam Jamatia, and Bj{\"o}rn Gamb{\"a}ck. 2019.
\newblock \href {https://doi.org/10.18653/v1/K19-1088} {Studying
  generalisability across abusive language detection datasets}.
\newblock In \emph{Proceedings of the 23rd Conference on Computational Natural
  Language Learning (CoNLL)}, pages 940--950, Hong Kong, China. Association for
  Computational Linguistics.

\bibitem[{Szegedy et~al.(2014)Szegedy, Zaremba, Sutskever, Bruna, Erhan,
  Goodfellow, and Fergus}]{DBLP:journals/corr/SzegedyZSBEGF13}
Christian Szegedy, Wojciech Zaremba, Ilya Sutskever, Joan Bruna, Dumitru Erhan,
  Ian~J. Goodfellow, and Rob Fergus. 2014.
\newblock \href {http://arxiv.org/abs/1312.6199} {Intriguing properties of
  neural networks}.
\newblock In \emph{2nd International Conference on Learning Representations,
  {ICLR} 2014, Banff, AB, Canada, April 14-16, 2014, Conference Track
  Proceedings}.

\bibitem[{Tandoc et~al.(2018)Tandoc, Lou, and Min}]{10.1093/jcmc/zmy022}
Jr. Tandoc, Edson~C, Chen Lou, and Velyn Lee~Hui Min. 2018.
\newblock \href {https://doi.org/10.1093/jcmc/zmy022} {{Platform-swinging in a
  poly-social-media context: How and why users navigate multiple social media
  platforms}}.
\newblock \emph{Journal of Computer-Mediated Communication}, 24(1):21--35.

\bibitem[{Tapia{-}T{\'{e}}llez and
  Escalante(2020)}]{DBLP:conf/micai/Tapia-TellezE20}
Jos{\'{e}}~Medardo Tapia{-}T{\'{e}}llez and Hugo~Jair Escalante. 2020.
\newblock \href {https://doi.org/10.1007/978-3-030-60887-3\_22} {Data
  augmentation with transformers for text classification}.
\newblock In \emph{Advances in Computational Intelligence - 19th Mexican
  International Conference on Artificial Intelligence, {MICAI} 2020, Mexico
  City, Mexico, October 12-17, 2020, Proceedings, Part {II}}, volume 12469 of
  \emph{Lecture Notes in Computer Science}, pages 247--259. Springer.

\bibitem[{Thain et~al.(2017)Thain, Dixon, and Wulczyn}]{Thain2017}
Nithum Thain, Lucas Dixon, and Ellery Wulczyn. 2017.
\newblock \href {https://doi.org/10.6084/m9.figshare.4563973.v2} {{Wikipedia
  Talk Labels: Toxicity}}.

\bibitem[{Thi et~al.(2015)Thi, Safavi{-}Naini, and
  Galib}]{DBLP:conf/wistp/ThiSG15}
Hoi~Le Thi, Reihaneh Safavi{-}Naini, and Asadullah~Al Galib. 2015.
\newblock \href {https://doi.org/10.1007/978-3-319-24018-3\_6} {Secure
  obfuscation of authoring style}.
\newblock In \emph{Information Security Theory and Practice - 9th {IFIP} {WG}
  11.2 International Conference, {WISTP} 2015 Heraklion, Crete, Greece, August
  24-25, 2015 Proceedings}, volume 9311 of \emph{Lecture Notes in Computer
  Science}, pages 88--103. Springer.

\bibitem[{Thylstrup and Waseem(2020)}]{thylstrup2020detecting}
Nanna Thylstrup and Zeerak Waseem. 2020.
\newblock Detecting ‘dirt’and ‘toxicity’: Rethinking content moderation
  as pollution behaviour.
\newblock \emph{Available at SSRN 3709719}.

\bibitem[{Tian et~al.(2019)Tian, Ma, Gong, Sengupta, Chen, Pinkerton, and
  Zitnick}]{DBLP:conf/icml/TianMGSCPZ19}
Yuandong Tian, Jerry Ma, Qucheng Gong, Shubho Sengupta, Zhuoyuan Chen, James
  Pinkerton, and Larry Zitnick. 2019.
\newblock \href {http://proceedings.mlr.press/v97/tian19a.html} {{ELF} opengo:
  an analysis and open reimplementation of alphazero}.
\newblock In \emph{Proceedings of the 36th International Conference on Machine
  Learning, {ICML} 2019, 9-15 June 2019, Long Beach, California, {USA}},
  volume~97 of \emph{Proceedings of Machine Learning Research}, pages
  6244--6253. {PMLR}.

\bibitem[{Tian et~al.(2018)Tian, Pei, Jana, and Ray}]{DBLP:conf/icse/TianPJR18}
Yuchi Tian, Kexin Pei, Suman Jana, and Baishakhi Ray. 2018.
\newblock \href {https://doi.org/10.1145/3180155.3180220} {Deeptest: automated
  testing of deep-neural-network-driven autonomous cars}.
\newblock In \emph{Proceedings of the 40th International Conference on Software
  Engineering, {ICSE} 2018, Gothenburg, Sweden, May 27 - June 03, 2018}, pages
  303--314. {ACM}.

\bibitem[{Tomkins et~al.(2018)Tomkins, Getoor, Chen, and
  Zhang}]{DBLP:conf/asunam/TomkinsGCZ18}
Sabina Tomkins, Lise Getoor, Yunfei Chen, and Yi~Zhang. 2018.
\newblock \href {https://doi.org/10.1109/ASONAM.2018.8508294} {A
  socio-linguistic model for cyberbullying detection}.
\newblock In \emph{{IEEE/ACM} 2018 International Conference on Advances in
  Social Networks Analysis and Mining, {ASONAM} 2018, Barcelona, Spain, August
  28-31, 2018}, pages 53--60. {IEEE} Computer Society.

\bibitem[{Tulkens et~al.(2016)Tulkens, Emmery, and
  Daelemans}]{tulkens-etal-2016-evaluating}
St{\'e}phan Tulkens, Chris Emmery, and Walter Daelemans. 2016.
\newblock \href {https://aclanthology.org/L16-1652} {Evaluating unsupervised
  {D}utch word embeddings as a linguistic resource}.
\newblock In \emph{Proceedings of the Tenth International Conference on
  Language Resources and Evaluation ({LREC}'16)}, pages 4130--4136,
  Portoro{\v{z}}, Slovenia. European Language Resources Association (ELRA).

\bibitem[{Turk and Pentland(1991)}]{10.1162/jocn.1991.3.1.71}
Matthew Turk and Alex Pentland. 1991.
\newblock \href {https://doi.org/10.1162/jocn.1991.3.1.71} {{Eigenfaces for
  Recognition}}.
\newblock \emph{Journal of Cognitive Neuroscience}, 3(1):71--86.

\bibitem[{Vaez et~al.(2004)Vaez, Ekberg, and Laflamme}]{Vaez2004-sl}
Marjan Vaez, Kerstin Ekberg, and Lucie Laflamme. 2004.
\newblock \href {https://www.jstor.org/stable/23077597} {Abusive events at work
  among young working adults: Magnitude of the problem and its effect on
  {Self-Rated} health}.
\newblock \emph{Relations Industrielles / Industrial Relations},
  59(3):569--584.

\bibitem[{Valkenburg and Peter(2007)}]{ValkePeter2007fw}
Patti~M. Valkenburg and Jochen Peter. 2007.
\newblock \href {https://doi.org/10.1037/0012-1649.43.2.267} {Preadolescents'
  and adolescents' online communication and their closeness to friends}.
\newblock \emph{Developmental Psychology}, 43(2):267--277.

\bibitem[{Van~den Broeck et~al.(2014)Van~den Broeck, Poels, Vandebosch, and
  Van~Royen}]{Broeck2014}
Evert Van~den Broeck, Karolien Poels, Heidi Vandebosch, and Kathleen Van~Royen.
  2014.
\newblock \href {https://doi.org/10.3233/978-1-61499-401-5-113} {Online
  perspective-taking as an intervention tool against cyberbullying}.
\newblock \emph{Studies in health technology and informatics}, 199:113--7.

\bibitem[{van~der Lee et~al.(2019)van~der Lee, Gatt, van Miltenburg, Wubben,
  and Krahmer}]{van-der-lee-etal-2019-best}
Chris van~der Lee, Albert Gatt, Emiel van Miltenburg, Sander Wubben, and Emiel
  Krahmer. 2019.
\newblock \href {https://doi.org/10.18653/v1/W19-8643} {Best practices for the
  human evaluation of automatically generated text}.
\newblock In \emph{Proceedings of the 12th International Conference on Natural
  Language Generation}, pages 355--368, Tokyo, Japan. Association for
  Computational Linguistics.

\bibitem[{Van~Hee et~al.(2018)Van~Hee, Jacobs, Emmery, Desmet, Lefever,
  Verhoeven, De~Pauw, Daelemans, and Hoste}]{10.1371/journal.pone.0203794}
Cynthia Van~Hee, Gilles Jacobs, Chris Emmery, Bart Desmet, Els Lefever, Ben
  Verhoeven, Guy De~Pauw, Walter Daelemans, and Véronique Hoste. 2018.
\newblock \href {https://doi.org/10.1371/journal.pone.0203794} {Automatic
  detection of cyberbullying in social media text}.
\newblock \emph{PLOS ONE}, 13(10):1--22.

\bibitem[{Van~Hee et~al.(2015{\natexlab{a}})Van~Hee, Lefever, Verhoeven,
  Mennes, Desmet, De~Pauw, Daelemans, and Hoste}]{van-hee-etal-2015-detection}
Cynthia Van~Hee, Els Lefever, Ben Verhoeven, Julie Mennes, Bart Desmet, Guy
  De~Pauw, Walter Daelemans, and Veronique Hoste. 2015{\natexlab{a}}.
\newblock \href {https://aclanthology.org/R15-1086} {Detection and fine-grained
  classification of cyberbullying events}.
\newblock In \emph{Proceedings of the International Conference Recent Advances
  in Natural Language Processing}, pages 672--680, Hissar, Bulgaria. INCOMA
  Ltd. Shoumen, BULGARIA.

\bibitem[{Van~Hee et~al.(2015{\natexlab{b}})Van~Hee, Verhoeven, Lefever,
  De~Pauw, Daelemans, and Hoste}]{VanHee2015guide}
Cynthia Van~Hee, Ben Verhoeven, Els Lefever, Guy De~Pauw, Walter Daelemans, and
  V{\'e}ronique Hoste. 2015{\natexlab{b}}.
\newblock \href
  {https://www.lt3.ugent.be/media/uploads/publications/2015/Guidelines_Cyberbullying_TechnicalReport_2.pdf}
  {Guidelines for the fine-grained analysis of cyberbullying}.
\newblock Technical report, version 1.0. Technical Report LT3 15-01, LT3,
  Language and Translation Technology Team--Ghent University.

\bibitem[{Vaswani et~al.(2017)Vaswani, Shazeer, Parmar, Uszkoreit, Jones,
  Gomez, Kaiser, and Polosukhin}]{DBLP:conf/nips/VaswaniSPUJGKP17}
Ashish Vaswani, Noam Shazeer, Niki Parmar, Jakob Uszkoreit, Llion Jones,
  Aidan~N. Gomez, Lukasz Kaiser, and Illia Polosukhin. 2017.
\newblock \href
  {https://proceedings.neurips.cc/paper/2017/hash/3f5ee243547dee91fbd053c1c4a845aa-Abstract.html}
  {Attention is all you need}.
\newblock In \emph{Advances in Neural Information Processing Systems 30: Annual
  Conference on Neural Information Processing Systems 2017, December 4-9, 2017,
  Long Beach, CA, {USA}}, pages 5998--6008.

\bibitem[{Vidgen and Derczynski(2021)}]{10.1371/journal.pone.0243300}
Bertie Vidgen and Leon Derczynski. 2021.
\newblock \href {https://doi.org/10.1371/journal.pone.0243300} {Directions in
  abusive language training data, a systematic review: Garbage in, garbage
  out}.
\newblock \emph{PLOS ONE}, 15(12):1--32.

\bibitem[{Vilain et~al.(2007)Vilain, Su, and Lubar}]{vilain-etal-2007-entity}
Marc Vilain, Jennifer Su, and Suzi Lubar. 2007.
\newblock \href {https://aclanthology.org/N07-2046} {Entity extraction is a
  boring solved {P}roblem{---}{O}r is it?}
\newblock In \emph{Human Language Technologies 2007: The Conference of the
  North {A}merican Chapter of the Association for Computational Linguistics;
  Companion Volume, Short Papers}, pages 181--184, Rochester, New York.
  Association for Computational Linguistics.

\bibitem[{Vlachos and Riedel(2014)}]{vlachos-riedel-2014-fact}
Andreas Vlachos and Sebastian Riedel. 2014.
\newblock \href {https://doi.org/10.3115/v1/W14-2508} {Fact checking: Task
  definition and dataset construction}.
\newblock In \emph{Proceedings of the {ACL} 2014 Workshop on Language
  Technologies and Computational Social Science}, pages 18--22, Baltimore, MD,
  USA. Association for Computational Linguistics.

\bibitem[{Volkova and Bachrach(2016)}]{volkova-bachrach-2016-inferring}
Svitlana Volkova and Yoram Bachrach. 2016.
\newblock \href {https://doi.org/10.18653/v1/P16-1148} {Inferring perceived
  demographics from user emotional tone and user-environment emotional
  contrast}.
\newblock In \emph{Proceedings of the 54th Annual Meeting of the Association
  for Computational Linguistics (Volume 1: Long Papers)}, pages 1567--1578,
  Berlin, Germany. Association for Computational Linguistics.

\bibitem[{Volkova et~al.(2015)Volkova, Bachrach, Armstrong, and
  Sharma}]{DBLP:conf/aaai/VolkovaBAS15}
Svitlana Volkova, Yoram Bachrach, Michael Armstrong, and Vijay Sharma. 2015.
\newblock \href {http://www.aaai.org/ocs/index.php/AAAI/AAAI15/paper/view/9358}
  {Inferring latent user properties from texts published in social media}.
\newblock In \emph{Proceedings of the Twenty-Ninth {AAAI} Conference on
  Artificial Intelligence, January 25-30, 2015, Austin, Texas, {USA}}, pages
  4296--4297. {AAAI} Press.

\bibitem[{Volkova et~al.(2014)Volkova, Coppersmith, and
  Van~Durme}]{volkova-etal-2014-inferring}
Svitlana Volkova, Glen Coppersmith, and Benjamin Van~Durme. 2014.
\newblock \href {https://doi.org/10.3115/v1/P14-1018} {Inferring user political
  preferences from streaming communications}.
\newblock In \emph{Proceedings of the 52nd Annual Meeting of the Association
  for Computational Linguistics (Volume 1: Long Papers)}, pages 186--196,
  Baltimore, Maryland. Association for Computational Linguistics.

\bibitem[{Wachter and Mittelstadt(2018)}]{Wachter2018ART}
Sandra Wachter and B.~Mittelstadt. 2018.
\newblock A right to reasonable inferences: Re-thinking data protection law in
  the age of big data and ai.
\newblock \emph{Columbia Business Law Review}, 2019:494–620--494–620.

\bibitem[{Wallace et~al.(2019)Wallace, Feng, Kandpal, Gardner, and
  Singh}]{wallace-etal-2019-universal}
Eric Wallace, Shi Feng, Nikhil Kandpal, Matt Gardner, and Sameer Singh. 2019.
\newblock \href {https://doi.org/10.18653/v1/D19-1221} {Universal adversarial
  triggers for attacking and analyzing {NLP}}.
\newblock In \emph{Proceedings of the 2019 Conference on Empirical Methods in
  Natural Language Processing and the 9th International Joint Conference on
  Natural Language Processing (EMNLP-IJCNLP)}, pages 2153--2162, Hong Kong,
  China. Association for Computational Linguistics.

\bibitem[{Wallach(2014)}]{wallach-2014-big}
Hanna Wallach. 2014.
\newblock \href
  {https://www.microsoft.com/en-us/research/publication/big-data-machine-learning-and-the-social-sciences-fairness-accountability-and-transparency/}
  {Big data, machine learning, and the social sciences: Fairness,
  accountability, and transparency}.
\newblock \emph{Medium}.

\bibitem[{Wang et~al.(2016)Wang, Guo, Lan, Xu, and
  Cheng}]{DBLP:conf/wsdm/WangGLXC16}
Pengfei Wang, Jiafeng Guo, Yanyan Lan, Jun Xu, and Xueqi Cheng. 2016.
\newblock \href {https://doi.org/10.1145/2835776.2835783} {Your cart tells you:
  Inferring demographic attributes from purchase data}.
\newblock In \emph{Proceedings of the Ninth {ACM} International Conference on
  Web Search and Data Mining, San Francisco, CA, USA, February 22-25, 2016},
  pages 173--182. {ACM}.

\bibitem[{Wang and Manning(2012)}]{wang-manning-2012-baselines}
Sida Wang and Christopher Manning. 2012.
\newblock \href {https://aclanthology.org/P12-2018} {Baselines and bigrams:
  Simple, good sentiment and topic classification}.
\newblock In \emph{Proceedings of the 50th Annual Meeting of the Association
  for Computational Linguistics (Volume 2: Short Papers)}, pages 90--94, Jeju
  Island, Korea. Association for Computational Linguistics.

\bibitem[{Waseem(2016)}]{waseem-2016-racist}
Zeerak Waseem. 2016.
\newblock \href {https://doi.org/10.18653/v1/W16-5618} {Are you a racist or am
  {I} seeing things? annotator influence on hate speech detection on
  {T}witter}.
\newblock In \emph{Proceedings of the First Workshop on {NLP} and Computational
  Social Science}, pages 138--142, Austin, Texas. Association for Computational
  Linguistics.

\bibitem[{Waseem and Hovy(2016)}]{waseem-hovy-2016-hateful}
Zeerak Waseem and Dirk Hovy. 2016.
\newblock \href {https://doi.org/10.18653/v1/N16-2013} {Hateful symbols or
  hateful people? predictive features for hate speech detection on {T}witter}.
\newblock In \emph{Proceedings of the {NAACL} Student Research Workshop}, pages
  88--93, San Diego, California. Association for Computational Linguistics.

\bibitem[{Wei and Zou(2019)}]{wei-zou-2019-eda}
Jason Wei and Kai Zou. 2019.
\newblock \href {https://doi.org/10.18653/v1/D19-1670} {{EDA}: Easy data
  augmentation techniques for boosting performance on text classification
  tasks}.
\newblock In \emph{Proceedings of the 2019 Conference on Empirical Methods in
  Natural Language Processing and the 9th International Joint Conference on
  Natural Language Processing (EMNLP-IJCNLP)}, pages 6382--6388, Hong Kong,
  China. Association for Computational Linguistics.

\bibitem[{Welbl et~al.(2021)Welbl, Glaese, Uesato, Dathathri, Mellor,
  Hendricks, Anderson, Kohli, Coppin, and
  Huang}]{welbl-etal-2021-challenges-detoxifying}
Johannes Welbl, Amelia Glaese, Jonathan Uesato, Sumanth Dathathri, John Mellor,
  Lisa~Anne Hendricks, Kirsty Anderson, Pushmeet Kohli, Ben Coppin, and Po-Sen
  Huang. 2021.
\newblock \href {https://doi.org/10.18653/v1/2021.findings-emnlp.210}
  {Challenges in detoxifying language models}.
\newblock In \emph{Findings of the Association for Computational Linguistics:
  EMNLP 2021}, pages 2447--2469, Punta Cana, Dominican Republic. Association
  for Computational Linguistics.

\bibitem[{Wiegand et~al.(2019)Wiegand, Ruppenhofer, and
  Kleinbauer}]{wiegand-etal-2019-detection}
Michael Wiegand, Josef Ruppenhofer, and Thomas Kleinbauer. 2019.
\newblock \href {https://doi.org/10.18653/v1/N19-1060} {{D}etection of
  {A}busive {L}anguage: the {P}roblem of {B}iased {D}atasets}.
\newblock In \emph{Proceedings of the 2019 Conference of the North {A}merican
  Chapter of the Association for Computational Linguistics: Human Language
  Technologies, Volume 1 (Long and Short Papers)}, pages 602--608, Minneapolis,
  Minnesota. Association for Computational Linguistics.

\bibitem[{Willard(2007)}]{Willard2007-mm}
Nancy~E Willard. 2007.
\newblock \emph{Cyberbullying and cyberthreats: Responding to the challenge of
  online social aggression, threats, and distress}.
\newblock Cyberbullying and cyberthreats: Responding to the challenge of online
  social aggression, threats, and distress. Research Press, Champaign, IL, US.

\bibitem[{Williams-Jones et~al.(2013)Williams-Jones, Olivier, and
  Smith}]{10.1093/scipol/sct038}
Bryn Williams-Jones, Catherine Olivier, and Elise Smith. 2013.
\newblock \href {https://doi.org/10.1093/scipol/sct038} {{Governing
  ‘dual-use’ research in Canada: A policy review}}.
\newblock \emph{Science and Public Policy}, 41(1):76--93.

\bibitem[{Wolf et~al.(2020)Wolf, Debut, Sanh, Chaumond, Delangue, Moi, Cistac,
  Rault, Louf, Funtowicz, Davison, Shleifer, von Platen, Ma, Jernite, Plu, Xu,
  Le~Scao, Gugger, Drame, Lhoest, and Rush}]{wolf-etal-2020-transformers}
Thomas Wolf, Lysandre Debut, Victor Sanh, Julien Chaumond, Clement Delangue,
  Anthony Moi, Pierric Cistac, Tim Rault, Remi Louf, Morgan Funtowicz, Joe
  Davison, Sam Shleifer, Patrick von Platen, Clara Ma, Yacine Jernite, Julien
  Plu, Canwen Xu, Teven Le~Scao, Sylvain Gugger, Mariama Drame, Quentin Lhoest,
  and Alexander Rush. 2020.
\newblock \href {https://doi.org/10.18653/v1/2020.emnlp-demos.6} {Transformers:
  State-of-the-art natural language processing}.
\newblock In \emph{Proceedings of the 2020 Conference on Empirical Methods in
  Natural Language Processing: System Demonstrations}, pages 38--45, Online.
  Association for Computational Linguistics.

\bibitem[{Wu et~al.(2016)Wu, Schuster, Chen, Le, Norouzi, Macherey, Krikun,
  Cao, Gao, Macherey, Klingner, Shah, Johnson, Liu, Kaiser, Gouws, Kato, Kudo,
  Kazawa, Stevens, Kurian, Patil, Wang, Young, Smith, Riesa, Rudnick, Vinyals,
  Corrado, Hughes, and Dean}]{DBLP:journals/corr/WuSCLNMKCGMKSJL16}
Yonghui Wu, Mike Schuster, Zhifeng Chen, Quoc~V. Le, Mohammad Norouzi, Wolfgang
  Macherey, Maxim Krikun, Yuan Cao, Qin Gao, Klaus Macherey, Jeff Klingner,
  Apurva Shah, Melvin Johnson, Xiaobing Liu, Lukasz Kaiser, Stephan Gouws,
  Yoshikiyo Kato, Taku Kudo, Hideto Kazawa, Keith Stevens, George Kurian,
  Nishant Patil, Wei Wang, Cliff Young, Jason Smith, Jason Riesa, Alex Rudnick,
  Oriol Vinyals, Greg Corrado, Macduff Hughes, and Jeffrey Dean. 2016.
\newblock \href {http://arxiv.org/abs/1609.08144} {Google's neural machine
  translation system: Bridging the gap between human and machine translation}.
\newblock \emph{CoRR}, abs/1609.08144.

\bibitem[{Wu et~al.(2018)Wu, Kambhatla, and Sarkar}]{DBLP:conf/acl-alw/WuKS18}
Zhelun Wu, Nishant Kambhatla, and Anoop Sarkar. 2018.
\newblock \href {https://doi.org/10.18653/v1/w18-5119} {Decipherment for
  adversarial offensive language detection}.
\newblock In \emph{Proceedings of the 2nd Workshop on Abusive Language Online,
  ALW@EMNLP 2018, Brussels, Belgium, October 31, 2018}, pages 149--159.
  Association for Computational Linguistics.

\bibitem[{Wulczyn et~al.(2017)Wulczyn, Thain, and
  Dixon}]{DBLP:conf/www/WulczynTD17}
Ellery Wulczyn, Nithum Thain, and Lucas Dixon. 2017.
\newblock \href {https://doi.org/10.1145/3038912.3052591} {Ex machina: Personal
  attacks seen at scale}.
\newblock In \emph{Proceedings of the 26th International Conference on World
  Wide Web, {WWW} 2017, Perth, Australia, April 3-7, 2017}, pages 1391--1399.
  {ACM}.

\bibitem[{Xenos et~al.(2021)Xenos, Pavlopoulos, Androutsopoulos, Dixon,
  Sorensen, and Laugier}]{DBLP:journals/corr/abs-2111-10223}
Alexandros Xenos, John Pavlopoulos, Ion Androutsopoulos, Lucas Dixon, Jeffrey
  Sorensen, and Leo Laugier. 2021.
\newblock \href {http://arxiv.org/abs/2111.10223} {Toxicity detection can be
  sensitive to the conversational context}.
\newblock \emph{CoRR}, abs/2111.10223.

\bibitem[{Xie et~al.(2017)Xie, Dai, Du, Hovy, and
  Neubig}]{DBLP:conf/nips/XieDDHN17}
Qizhe Xie, Zihang Dai, Yulun Du, Eduard~H. Hovy, and Graham Neubig. 2017.
\newblock \href
  {https://proceedings.neurips.cc/paper/2017/hash/8cb22bdd0b7ba1ab13d742e22eed8da2-Abstract.html}
  {Controllable invariance through adversarial feature learning}.
\newblock In \emph{Advances in Neural Information Processing Systems 30: Annual
  Conference on Neural Information Processing Systems 2017, December 4-9, 2017,
  Long Beach, CA, {USA}}, pages 585--596.

\bibitem[{Xu et~al.(2012)Xu, Jun, Zhu, and Bellmore}]{xu-etal-2012-learning}
Jun-Ming Xu, Kwang-Sung Jun, Xiaojin Zhu, and Amy Bellmore. 2012.
\newblock \href {https://aclanthology.org/N12-1084} {Learning from bullying
  traces in social media}.
\newblock In \emph{Proceedings of the 2012 Conference of the North {A}merican
  Chapter of the Association for Computational Linguistics: Human Language
  Technologies}, pages 656--666, Montr{\'e}al, Canada. Association for
  Computational Linguistics.

\bibitem[{Xu et~al.(2019)Xu, Xu, and Qu}]{xu-etal-2019-alter}
Qiongkai Xu, Chenchen Xu, and Lizhen Qu. 2019.
\newblock \href {https://doi.org/10.18653/v1/D19-3003} {{ALTER}: Auxiliary text
  rewriting tool for natural language generation}.
\newblock In \emph{Proceedings of the 2019 Conference on Empirical Methods in
  Natural Language Processing and the 9th International Joint Conference on
  Natural Language Processing (EMNLP-IJCNLP): System Demonstrations}, pages
  13--18, Hong Kong, China. Association for Computational Linguistics.

\bibitem[{Xu and Yang(2017)}]{xu-yang-2017-cross}
Ruochen Xu and Yiming Yang. 2017.
\newblock \href {https://doi.org/10.18653/v1/P17-1130} {Cross-lingual
  distillation for text classification}.
\newblock In \emph{Proceedings of the 55th Annual Meeting of the Association
  for Computational Linguistics (Volume 1: Long Papers)}, pages 1415--1425,
  Vancouver, Canada. Association for Computational Linguistics.

\bibitem[{Xu et~al.(2020)Xu, Zhong, Jimeno{-}Yepes, and
  Lau}]{DBLP:journals/corr/abs-2001-07820}
Ying Xu, Xu~Zhong, Antonio~Jos{\'{e}} Jimeno{-}Yepes, and Jey~Han Lau. 2020.
\newblock \href {http://arxiv.org/abs/2001.07820} {Elephant in the room: An
  evaluation framework for assessing adversarial examples in {NLP}}.
\newblock \emph{CoRR}, abs/2001.07820.

\bibitem[{Yang et~al.(2019)Yang, Xie, Tan, Xiong, Li, and
  Lin}]{DBLP:journals/corr/abs-1904-06652}
Wei Yang, Yuqing Xie, Luchen Tan, Kun Xiong, Ming Li, and Jimmy Lin. 2019.
\newblock \href {http://arxiv.org/abs/1904.06652} {Data augmentation for {BERT}
  fine-tuning in open-domain question answering}.
\newblock \emph{CoRR}, abs/1904.06652.

\bibitem[{Yates et~al.(2017)Yates, Cohan, and
  Goharian}]{yates-etal-2017-depression}
Andrew Yates, Arman Cohan, and Nazli Goharian. 2017.
\newblock \href {https://doi.org/10.18653/v1/D17-1322} {Depression and
  self-harm risk assessment in online forums}.
\newblock In \emph{Proceedings of the 2017 Conference on Empirical Methods in
  Natural Language Processing}, pages 2968--2978, Copenhagen, Denmark.
  Association for Computational Linguistics.

\bibitem[{Yin et~al.(2009)Yin, Xue, Hong, Davison, Kontostathis, and
  Edwards}]{yin2009detection}
Dawei Yin, Zhenzhen Xue, Liangjie Hong, Brian~D Davison, April Kontostathis,
  and Lynne Edwards. 2009.
\newblock \href
  {http://www.cse.lehigh.edu/~brian/pubs/2009/CAW2/harassment.pdf} {Detection
  of harassment on web 2.0}.
\newblock \emph{Proceedings of the Content Analysis in the WEB 2.0 (CAW 2.0)
  Workshop at WWW2009}, 2:1--7.

\bibitem[{Yoo et~al.(2020)Yoo, Morris, Lifland, and
  Qi}]{yoo-etal-2020-searching}
Jin~Yong Yoo, John Morris, Eli Lifland, and Yanjun Qi. 2020.
\newblock \href {https://doi.org/10.18653/v1/2020.blackboxnlp-1.30} {Searching
  for a search method: Benchmarking search algorithms for generating {NLP}
  adversarial examples}.
\newblock In \emph{Proceedings of the Third BlackboxNLP Workshop on Analyzing
  and Interpreting Neural Networks for NLP}, pages 323--332, Online.
  Association for Computational Linguistics.

\bibitem[{Yoo et~al.(2021)Yoo, Park, Kang, Lee, and
  Park}]{DBLP:journals/corr/abs-2104-08826}
Kang~Min Yoo, Dongju Park, Jaewook Kang, Sang{-}Woo Lee, and Woo{-}Myoung Park.
  2021.
\newblock \href {https://doi.org/10.18653/v1/2021.findings-emnlp.192} {Gpt3mix:
  Leveraging large-scale language models for text augmentation}.
\newblock In \emph{Findings of the Association for Computational Linguistics:
  {EMNLP} 2021, Virtual Event / Punta Cana, Dominican Republic, 16-20 November,
  2021}, pages 2225--2239. Association for Computational Linguistics.

\bibitem[{Zhang et~al.(2020{\natexlab{a}})Zhang, Bengio, Hardt, Mozer, and
  Singer}]{DBLP:conf/iclr/ZhangBHMS20}
Chiyuan Zhang, Samy Bengio, Moritz Hardt, Michael~C. Mozer, and Yoram Singer.
  2020{\natexlab{a}}.
\newblock \href {https://openreview.net/forum?id=B1l6y0VFPr} {Identity crisis:
  Memorization and generalization under extreme overparameterization}.
\newblock In \emph{8th International Conference on Learning Representations,
  {ICLR} 2020, Addis Ababa, Ethiopia, April 26-30, 2020}. OpenReview.net.

\bibitem[{Zhang et~al.(2020{\natexlab{b}})Zhang, Kishore, Wu, Weinberger, and
  Artzi}]{zhang2019bertscore}
Tianyi Zhang, Varsha Kishore, Felix Wu, Kilian~Q. Weinberger, and Yoav Artzi.
  2020{\natexlab{b}}.
\newblock \href {https://openreview.net/forum?id=SkeHuCVFDr} {Bertscore:
  Evaluating text generation with {BERT}}.
\newblock In \emph{8th International Conference on Learning Representations,
  {ICLR} 2020, Addis Ababa, Ethiopia, April 26-30, 2020}. OpenReview.net.

\bibitem[{Zhang et~al.(2020{\natexlab{c}})Zhang, Sheng, Alhazmi, and
  Li}]{DBLP:journals/tist/ZhangSAL20}
Wei~Emma Zhang, Quan~Z. Sheng, Ahoud Alhazmi, and Chenliang Li.
  2020{\natexlab{c}}.
\newblock \href {https://doi.org/10.1145/3374217} {Adversarial attacks on
  deep-learning models in natural language processing: {A} survey}.
\newblock \emph{{ACM} Trans. Intell. Syst. Technol.}, 11(3):24:1--24:41.

\bibitem[{Zhang et~al.(2016)Zhang, Tong, Vishwamitra, Whittaker, Mazer,
  Kowalski, Hu, Luo, Macbeth, and Dillon}]{DBLP:conf/icmla/ZhangTVWMKHLMD16}
Xiang Zhang, Jonathan Tong, Nishant Vishwamitra, Elizabeth Whittaker, Joseph~P.
  Mazer, Robin~M. Kowalski, Hongxin Hu, Feng Luo, Jamie Macbeth, and Edward
  Dillon. 2016.
\newblock \href {https://doi.org/10.1109/ICMLA.2016.0132} {Cyberbullying
  detection with a pronunciation based convolutional neural network}.
\newblock In \emph{15th {IEEE} International Conference on Machine Learning and
  Applications, {ICMLA} 2016, Anaheim, CA, USA, December 18-20, 2016}, pages
  740--745. {IEEE} Computer Society.

\bibitem[{Zhao and Mao(2017)}]{DBLP:journals/taffco/ZhaoM17}
Rui Zhao and Kezhi Mao. 2017.
\newblock \href {https://doi.org/10.1109/TAFFC.2016.2531682} {Cyberbullying
  detection based on semantic-enhanced marginalized denoising auto-encoder}.
\newblock \emph{{IEEE} Trans. Affect. Comput.}, 8(3):328--339.

\bibitem[{Zhao et~al.(2016)Zhao, Zhou, and Mao}]{DBLP:conf/icdcn/ZhaoZM16}
Rui Zhao, Anna Zhou, and Kezhi Mao. 2016.
\newblock \href {https://doi.org/10.1145/2833312.2849567} {Automatic detection
  of cyberbullying on social networks based on bullying features}.
\newblock In \emph{Proceedings of the 17th International Conference on
  Distributed Computing and Networking, Singapore, January 4-7, 2016}, pages
  43:1--43:6. {ACM}.

\bibitem[{Zhao et~al.(2021{\natexlab{a}})Zhao, Ge, Hu, and
  Shi}]{DBLP:journals/corr/abs-2110-08036}
Tengfei Zhao, Zhaocheng Ge, Hanping Hu, and Dingmeng Shi. 2021{\natexlab{a}}.
\newblock \href {http://arxiv.org/abs/2110.08036} {Generating natural language
  adversarial examples through an improved beam search algorithm}.
\newblock \emph{CoRR}, abs/2110.08036.

\bibitem[{Zhao et~al.(2021{\natexlab{b}})Zhao, Zhang, and
  Hopfgartner}]{DBLP:conf/www/ZhaoZH21}
Zhixue Zhao, Ziqi Zhang, and Frank Hopfgartner. 2021{\natexlab{b}}.
\newblock \href {https://doi.org/10.1145/3442442.3452313} {A comparative study
  of using pre-trained language models for toxic comment classification}.
\newblock In \emph{Companion of The Web Conference 2021, Virtual Event /
  Ljubljana, Slovenia, April 19-23, 2021}, pages 500--507. {ACM} / {IW3C2}.

\bibitem[{Zhou et~al.(2015)Zhou, Sun, Liu, and
  Lau}]{DBLP:journals/corr/ZhouSLL15b}
Chunting Zhou, Chonglin Sun, Zhiyuan Liu, and Francis C.~M. Lau. 2015.
\newblock \href {http://arxiv.org/abs/1511.08630} {A {C-LSTM} neural network
  for text classification}.
\newblock \emph{CoRR}, abs/1511.08630.

\bibitem[{Zhou et~al.(2019)Zhou, Ge, Xu, Wei, and Zhou}]{zhou-etal-2019-bert}
Wangchunshu Zhou, Tao Ge, Ke~Xu, Furu Wei, and Ming Zhou. 2019.
\newblock \href {https://doi.org/10.18653/v1/P19-1328} {{BERT}-based lexical
  substitution}.
\newblock In \emph{Proceedings of the 57th Annual Meeting of the Association
  for Computational Linguistics}, pages 3368--3373, Florence, Italy.
  Association for Computational Linguistics.

\bibitem[{Zhou et~al.(2021)Zhou, Sap, Swayamdipta, Choi, and
  Smith}]{zhou-etal-2021-challenges}
Xuhui Zhou, Maarten Sap, Swabha Swayamdipta, Yejin Choi, and Noah Smith. 2021.
\newblock \href {https://aclanthology.org/2021.eacl-main.274} {Challenges in
  automated debiasing for toxic language detection}.
\newblock In \emph{Proceedings of the 16th Conference of the European Chapter
  of the Association for Computational Linguistics: Main Volume}, pages
  3143--3155, Online. Association for Computational Linguistics.

\bibitem[{Zuboff(2015)}]{DBLP:journals/jitech/Zuboff15}
Shoshana Zuboff. 2015.
\newblock \href {https://doi.org/10.1057/jit.2015.5} {Big other: surveillance
  capitalism and the prospects of an information civilization}.
\newblock \emph{J. Inf. Technol.}, 30(1):75--89.

\bibitem[{Ólafsson et~al.(2013)Ólafsson, Livingstone, and
  Haddon}]{RePEc:ehl:lserod:50228}
Kjartan Ólafsson, Sonia Livingstone, and Leslie Haddon. 2013.
\newblock \href {https://ideas.repec.org/p/ehl/lserod/50228.html}
  {{Children’s use of online technologies in Europe: a review of the European
  evidence base}}.
\newblock LSE Research Online Documents on Economics 50228, London School of
  Economics and Political Science, LSE Library.

\end{thebibliography}
\bibliographystyle{acl_natbib}
}

\end{document}